\documentclass[acmtog,authorversion]{acmart}

\usepackage{subcaption}
\usepackage{wrapfig}
\usepackage{makecell}
\usepackage{mathtools}
\usepackage{multirow}
\usepackage[inline]{enumitem}

\def\sharpnet{SharpNet}
\let\method\sharpnet

\let\ingp\instantngp
\def\siren{SIREN}
\def\nhrep{NH\nobreakdashes-Rep}
\def\dann{DANN}

\DeclareMathOperator*{\argmin}{arg\,min}

\DeclarePairedDelimiter{\abs}{\lvert}{\rvert}
\DeclarePairedDelimiter{\norm}{\lVert}{\rVert}
\DeclarePairedDelimiter{\set}{\{}{\}}
\DeclarePairedDelimiterX{\condset}[2]{\{}{\}}{#1\,\delimsize\vert\,\mathopen{}#2}

\allowdisplaybreaks[1]

\newcounter{descriptcount}
\newlist{enumdescriptioninline}{description*}{1}
\setlist[enumdescriptioninline,1]{%
  before={%
    \setcounter{descriptcount}{0}%
    \renewcommand*\thedescriptcount{(\arabic{descriptcount})}%
  },
  font={\bfseries\stepcounter{descriptcount}\thedescriptcount~},
  mode=unboxed,
}

\setcopyright{cc}
\setcctype{by}
\acmJournal{TOG}
\acmYear{2026} \acmVolume{45} \acmNumber{4} \acmArticle{113}
\acmMonth{7} \acmDOI{10.1145/3811330}

\begin{document}
\title{{\sharpnet}: Enhancing MLPs to Represent Functions with Controlled Non-differentiability}

\subtitle{}

\author{Hanting Niu}
\authornote{H. Niu and J. Deng contributed equally to this research.}
\email{niuht@ios.ac.cn}
\orcid{0009-0004-1501-1834}
\affiliation{
  \institution{Key Laboratory of System Software (CAS), Institute of Software, Chinese Academy of Sciences}
  \city{Beijing}
  \country{China}
}
\affiliation{
  \institution{University of Chinese Academy of Sciences}
  \city{Beijing}
  \country{China}
}
\author{Junkai Deng}
\authornotemark[1]
\email{junkai006@e.ntu.edu.sg}
\orcid{0009-0009-7298-1577}
\affiliation{
\institution{College of Computing and Data Science, Nanyang Technological University}
\city{Singapore}
\country{Singapore}
}
\author{Fei Hou}
\authornote{Corresponding authors: F. Hou and Y. He.}
\email{houfei@ios.ac.cn}
\orcid{0000-0001-8226-6635}
\affiliation{
  \institution{Key Laboratory of System Software (CAS), Institute of Software, Chinese Academy of Sciences}
  \city{Beijing}
  \country{China}
}
\affiliation{
  \institution{University of Chinese Academy of Sciences}
  \city{Beijing}
  \country{China}
}
\author{Wencheng Wang}
\email{whn@ios.ac.cn}
\orcid{0000-0001-5094-4606}
\affiliation{
  \institution{Key Laboratory of System Software (CAS), Institute of Software, Chinese Academy of Sciences}
  \city{Beijing}
  \country{China}
}
\affiliation{
  \institution{University of Chinese Academy of Sciences}
  \city{Beijing}
  \country{China}
}
\author{Ying He}
\authornotemark[2]
\email{yhe@ntu.edu.sg}
\orcid{0000-0002-6749-4485}
\affiliation{
\institution{College of Computing and Data Science, Nanyang Technological University}
\city{Singapore}
\country{Singapore}
}

\begin{teaserfigure}
\centering
\begin{subcaptionblock}[b][.5\linewidth][s]{.3\linewidth}
    \centering
    \includegraphics[width=\linewidth]{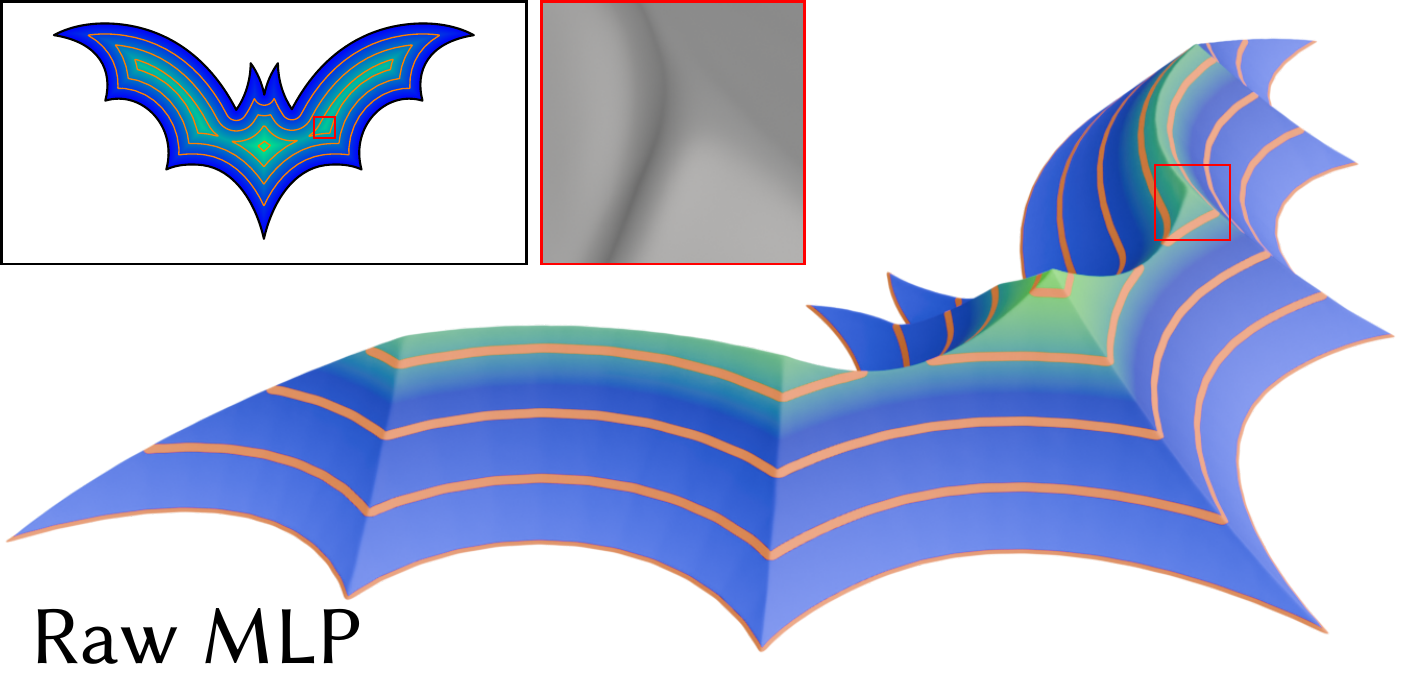}\vfill%
    \includegraphics[width=\linewidth]{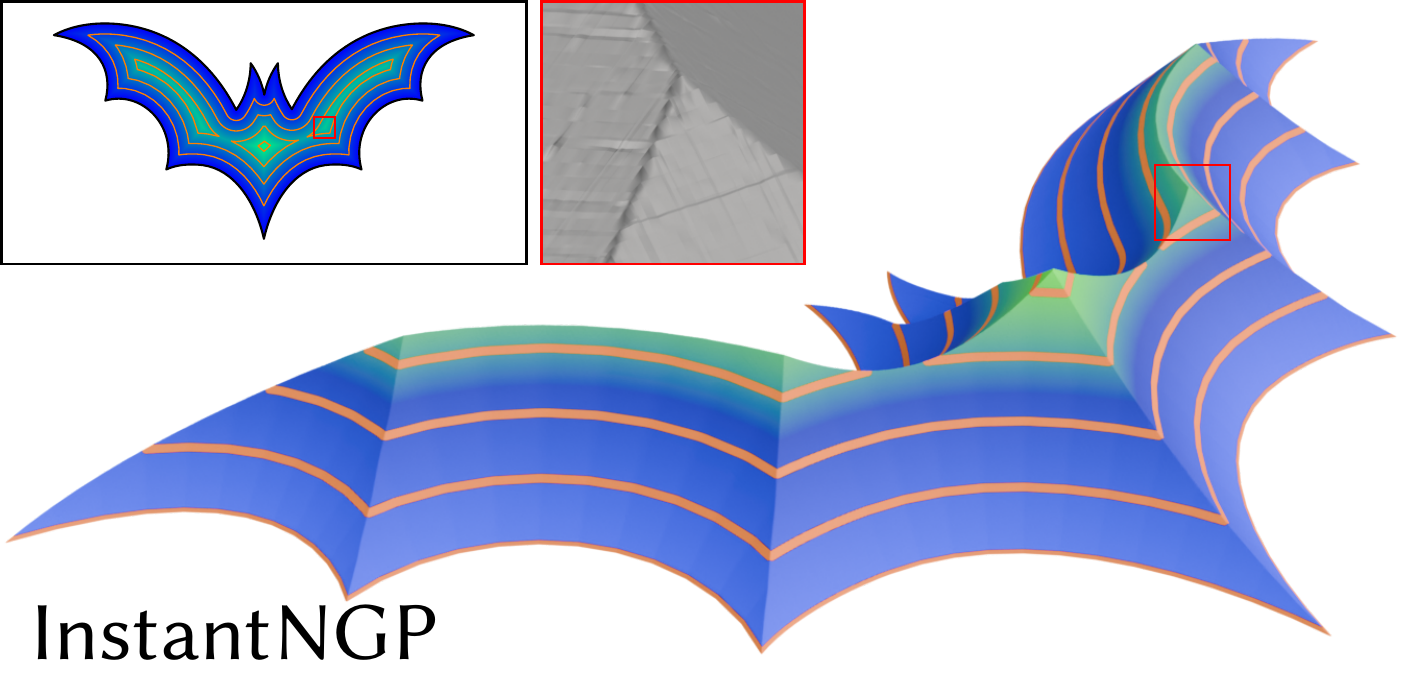}\vfill%
    \includegraphics[width=\linewidth]{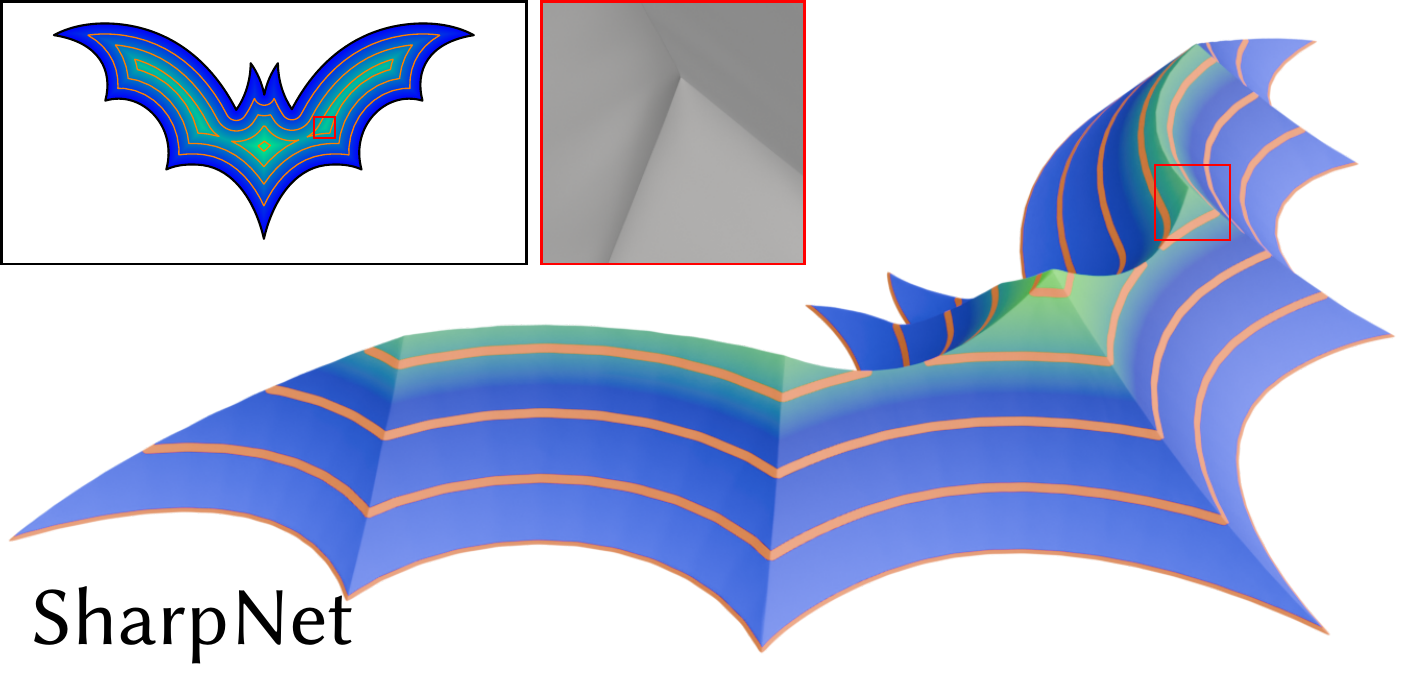}
    \caption{Medial axis}
    \label{fig:teaser_bat}
\end{subcaptionblock}\hfill%
\begin{minipage}[b][.5\linewidth][s]{.3264\linewidth}
\begin{subcaptionblock}{\linewidth}
    \centering
    \includegraphics[width=\linewidth]{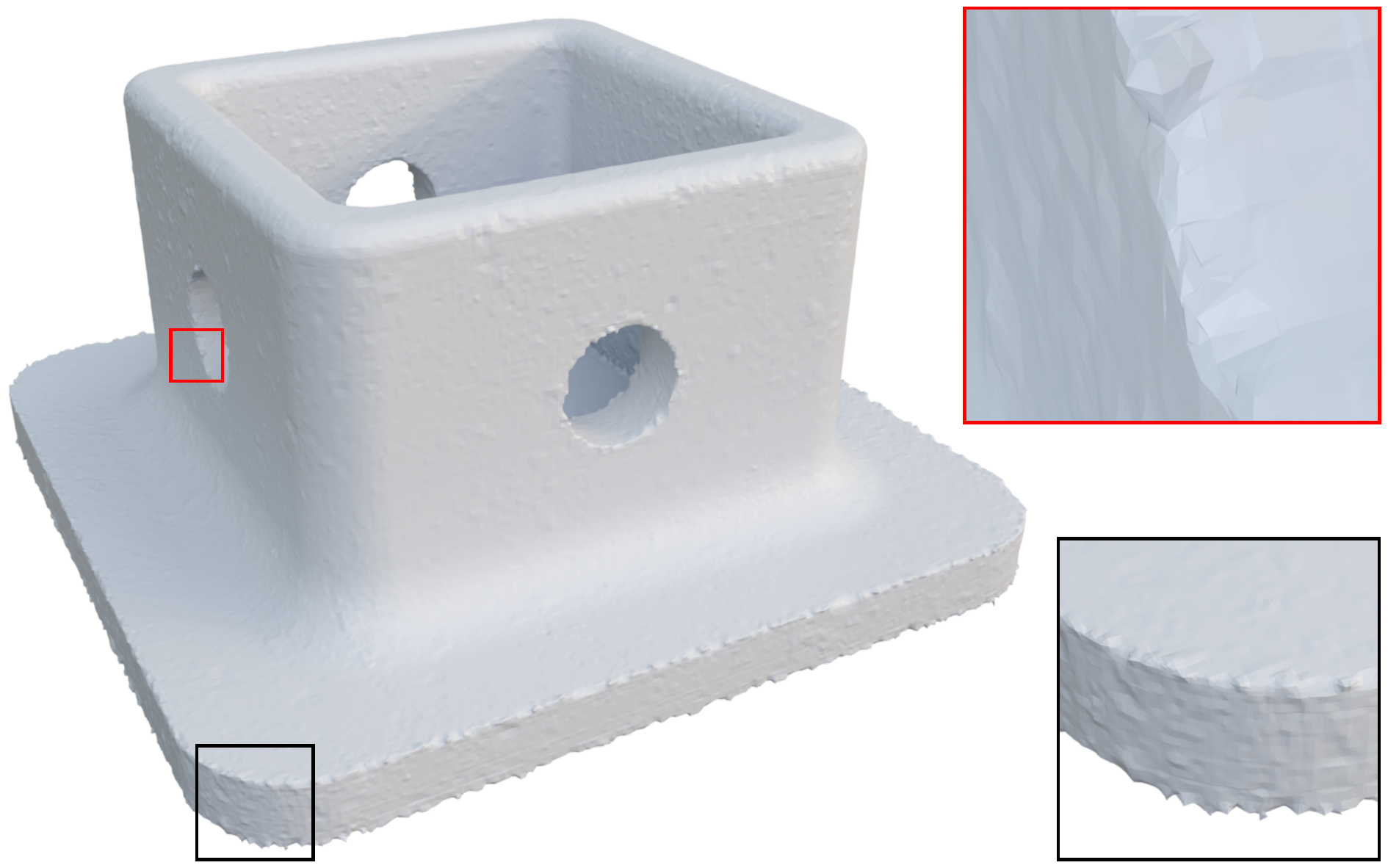}
    \caption{\ingp}
    \label{fig:teaser_ingp-}
\end{subcaptionblock}\vfill%
\begin{subcaptionblock}{\linewidth}
    \centering
    \includegraphics[width=\linewidth]{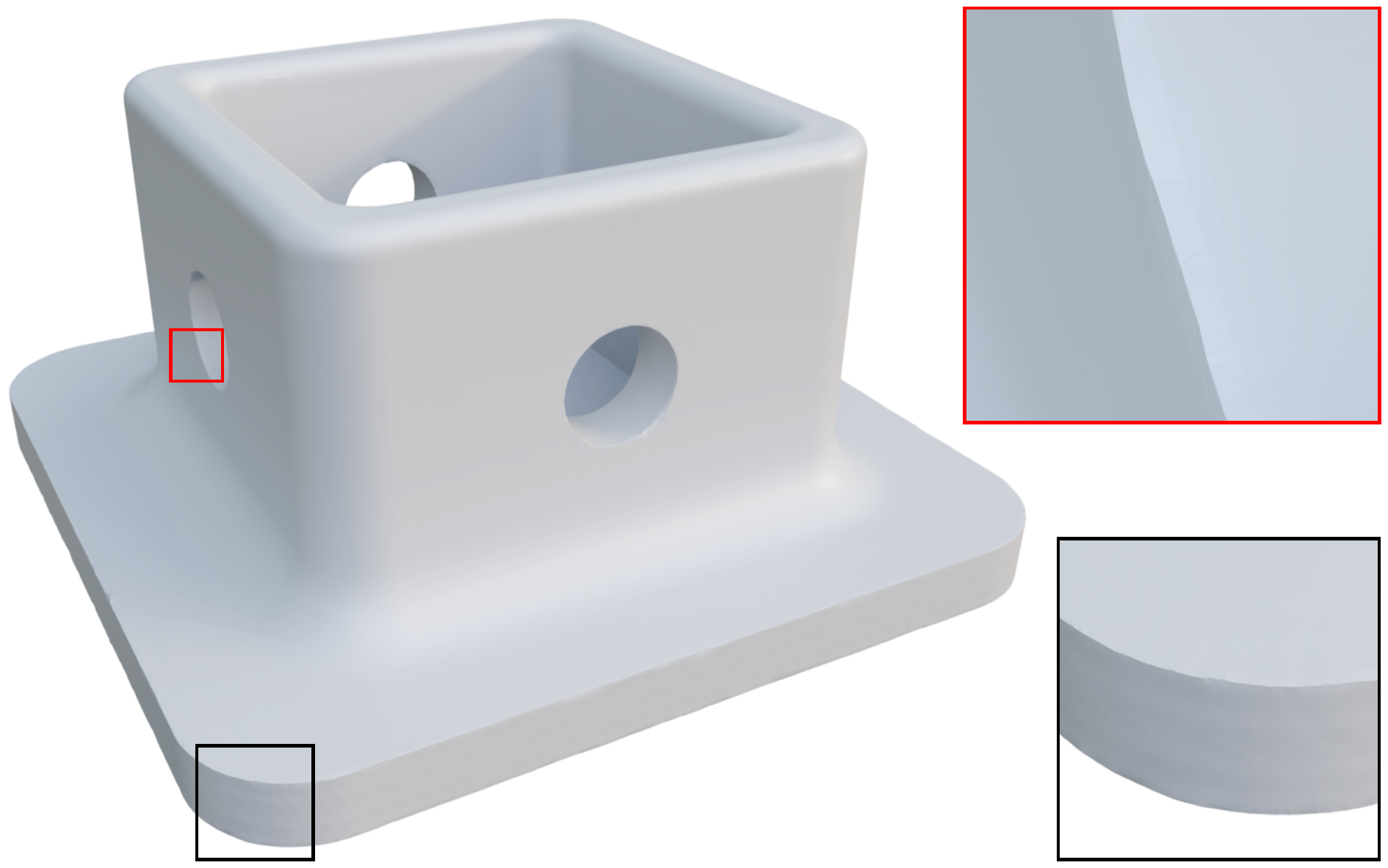}
    \caption{\method}
    \label{fig:teaser_sharp-}
\end{subcaptionblock}
\end{minipage}\hfill%
\begin{minipage}[b][.5\linewidth][s]{.3264\linewidth}
\begin{subcaptionblock}{\linewidth}
    \centering
    \includegraphics[width=\linewidth]{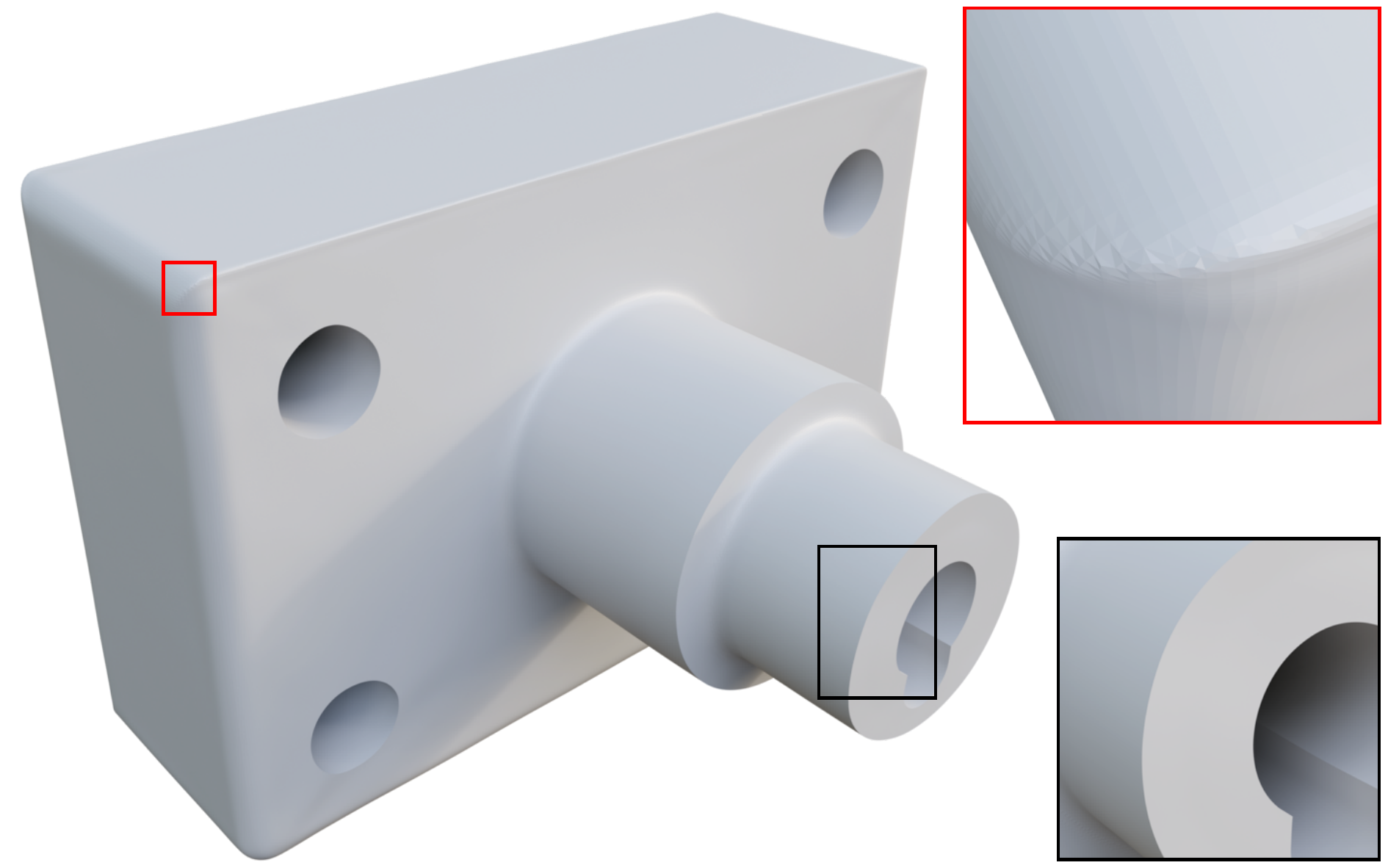}
    \caption{\nhrep}
    \label{fig:teaser_nhrep}
\end{subcaptionblock}\vfill%
\begin{subcaptionblock}{\linewidth}
    \centering
    \includegraphics[width=\linewidth]{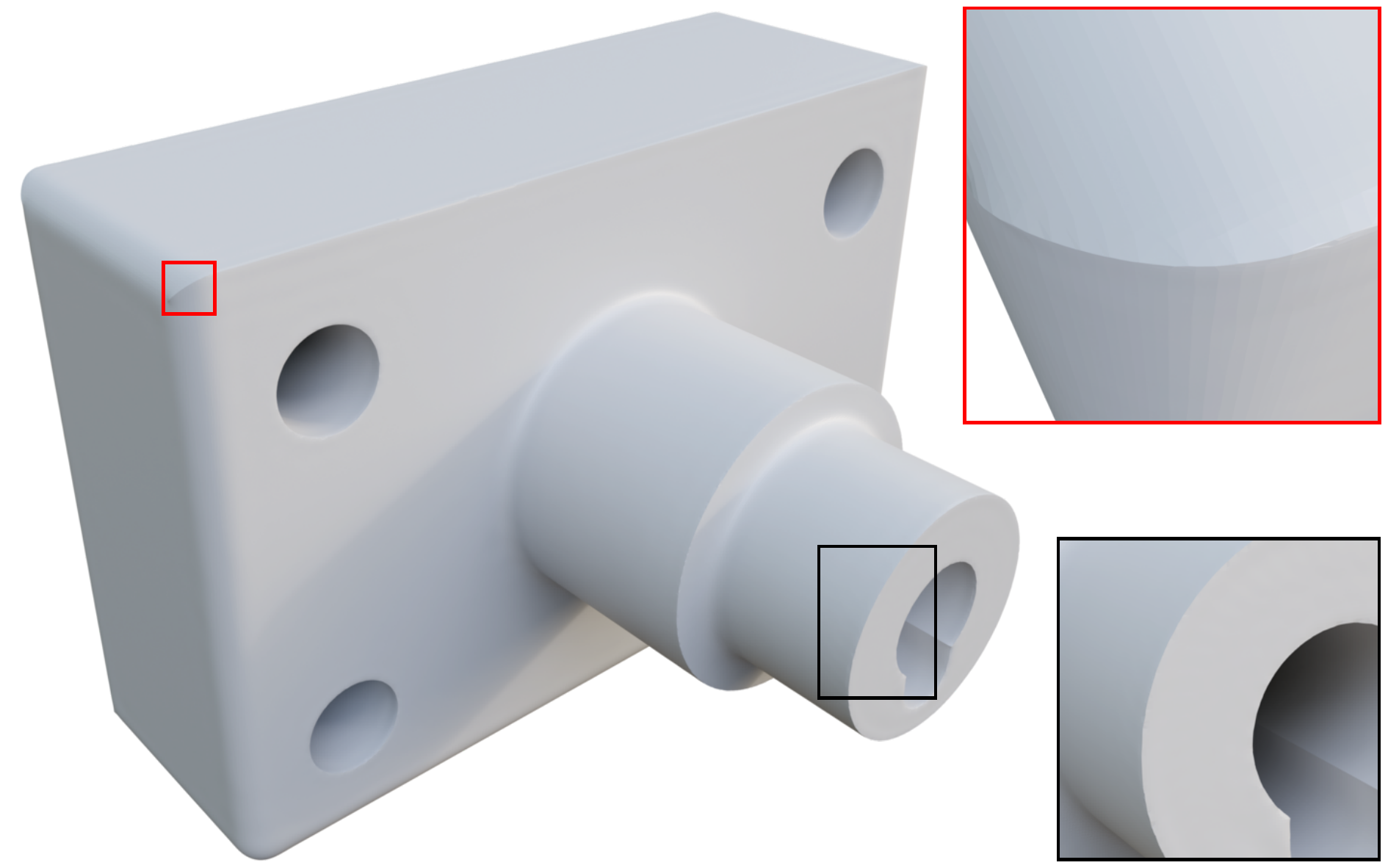}
    \caption{\method}
    \label{fig:teaser_sharp}
\end{subcaptionblock}
\end{minipage}
\caption{MLPs and sinusoidal-activation networks such as SIRENs~\cite{sitzmann2020siren} are effective at representing smooth functions; {\sharpnet} extends these architectures to model functions with sharp ($C^0$) features at user-specified locations.
(\subref{fig:teaser_bat})~Distance field of a 2D closed curve, which is non-differentiable along the medial axis. We visualize the field using both a 2D colormap and a height-field rendering; the height-field view more clearly reveals the non-smooth behavior of the field learned by  {\ingp}~\cite{Muller2022INGP}.
(\subref{fig:teaser_ingp-},\subref{fig:teaser_sharp-})~Reconstruction of a CAD model from a mesh input. {\nhrep}~\cite{Guo2022NH-Rep} constructs the CAD model through patch subdivision and forms sharp edges via patch intersections, but it is unable to represent open (non-closed) sharp edges.
(\subref{fig:teaser_nhrep},\subref{fig:teaser_sharp})~Reconstruction of a CAD model from points with normals as input. {\ingp} introduces notable artifacts, particularly around sharp edges. In contrast, without requiring patch or grid subdivision, {\sharpnet} can generate strictly $C^0$-continuous sharp features, regardless of whether they are closed.}
\label{fig:teaser}
\end{teaserfigure}

\begin{abstract}
Multi-layer perceptrons (MLPs) are a standard tool for learning and function approximation, but they inherently produce globally smooth outputs. Consequently, they struggle to represent functions that are continuous yet intentionally non-differentiable (i.e., functions with prescribed $C^0$ sharp features) without ad hoc post-processing. We present \textbf{\sharpnet}, a modified MLP architecture that encodes user-specified  sharp features by augmenting the network with an auxiliary \emph{feature function} defined as the solution to Poisson's equation with jump Neumann boundary conditions. This feature function is evaluated via an efficient local integral and is fully differentiable with respect to the feature locations, allowing us to jointly optimize both the feature locations and the MLP parameters to recover the target function or geometry. This construction provides precise control over where non-differentiability occurs, enforcing the desired $C^0$ behavior at feature locations while preserving smoothness elsewhere. We validate {\sharpnet} on 2D problems and 3D CAD reconstruction, and compare it with several state-of-the-art baselines. In both settings, {\sharpnet} accurately recovers sharp edges and corners while remaining smooth away from them, whereas existing methods tend to blur gradient discontinuities. Qualitative and quantitative results demonstrate the effectiveness of our approach. Our project page, code and models are publicly available at \href{https://sharpnettech.github.io}{https://sharpnettech.github.io}.
\end{abstract}

%
%

\ccsdesc[500]{Computing methodologies~Mesh geometry models}

%
%

\keywords{MLP, sharp features, Poisson's equation, jump Neumann boundary condition, Green's function, CAD}

\maketitle

\section{Introduction}
\label{sec:introduction}

Multi-layer perceptrons (MLPs) are fundamental building blocks of modern deep learning and are particularly effective at approximating smooth functions. Their expressive power and differentiability have made them a standard choice in geometry and vision~\cite{Mescheder2019,Pang2023}. In 3D shape modeling, MLPs support neural signed distance functions (SDFs) that continuously represent surfaces in space~\cite{DeepSDF,NeuralPull,Wang2022,Li2023}. In novel view synthesis, they serve as the core function approximators in neural radiance fields (NeRFs)~\cite{NeRF} and related methods for multi-view 3D reconstruction~\cite{NeuS}. In these settings, differentiability is central: it enables efficient gradient-based optimization and facilitates the learning of continuous, well-behaved fields.

However, many real-world targets are not globally smooth. In particular, several tasks require the representation of \emph{sharp} geometric features, where the function remains continuous but its derivative changes abruptly. A common example is CAD reconstruction from unoriented point clouds~\cite{Dong2024NeurCADRecon}, where preserving sharp edges and corners is as important as recovering smooth surfaces for maintaining geometric fidelity. From a functional perspective, such structures correspond to $C^0$-continuous functions with gradient discontinuities, which conventional MLP parameterizations struggle to reproduce reliably.

This limitation is closely tied to activation regularity. MLPs with smooth activations (e.g., softplus, sigmoid, or $\tanh$) are $C^\infty$ and therefore cannot represent true non-differentiabilities. ReLU-based MLPs are only $C^0$ and can, in principle, represent gradient discontinuities through their piecewise-linear structure. However, they do not provide precise control over \emph{where} these non-differentiabilities occur, which becomes problematic when sharp features must align with known or learnable geometry.

We introduce \textbf{\sharpnet}, an MLP variant designed to represent functions that are smooth almost everywhere while exhibiting non-differentiability at prescribed locations. {\sharpnet} augments the network input with a geometric prior: an auxiliary \emph{feature function} that encodes the desired set of sharp features. We define this feature function as the solution to Poisson's equation with jump Neumann boundary conditions, yielding a field that is smooth away from the feature set and $C^0$-continuous with a normal-derivative discontinuity along it. This construction draws on classical PDE theory and Green's third identity, leading to a boundary-integral formulation with closed-form expression for numerical evaluation.

Our framework applies both when feature geometry is known and when it must be inferred. If the sharp feature locations are available, the feature function can be treated as fixed during training. When the feature geometry is unknown, the feature locations and the MLP parameters can be optimized jointly in an end-to-end manner because the feature function is  differentiable with respect to the feature geometry.

We validate {\sharpnet} on both 2D and 3D problems. In 2D, we study distance-field fitting and medial-axis learning, where the medial axis coincides with the non-differentiable feature set. In 3D, we evaluate {\sharpnet} on CAD reconstruction under three settings (i.e., meshes, oriented points and unoriented points), encompassing both fixed and learnable sharp features. Across all settings, {\sharpnet} achieves higher accuracy near non-differentiable regions. By contrast, standard MLPs and {\ingp}~\cite{Muller2022INGP} consistently fail to reproduce these sharp structures. NH-Rep~\cite{Guo2022NH-Rep} requires surface partitioning into multiple patches and reconstructs sharp edges through patch intersections, making open (non-closed) sharp edges difficult to handle. As illustrated in Figure~\ref{fig:teaser}, {\sharpnet} accurately reproduces the intended non-differentiable behavior, whereas the baseline architectures struggle substantially.

\section{Related Work}
\label{sec:relatedworks}

\subsection{Improving Representation Capabilities of MLPs}
\label{subsec:mlpcapabilities}

The representational power of neural networks has been studied extensively, most notably through universal approximation theorems~\cite{cybenko1989approximation,hornik1991approximation}. These results establish that MLPs can approximate broad classes of functions, but they say comparatively little about how well, or how efficiently, neural networks capture \emph{non-smooth} structures such as kinks, corners, and discontinuities.

Several works have begun to address this gap. Imaizumi and Fukumizu~\shortcite{Imaizumi2019} analyze approximation error and convergence rates for deep ReLU networks on function classes that include non-smooth targets. While their theory characterizes when accurate approximation is possible, it does not provide a mechanism for explicitly \emph{placing} non-smoothness at prescribed geometric locations, which is critical in many geometry-processing settings. Ismailov~\shortcite{Ismailov2023} further shows that three-layer neural networks can approximate both continuous and discontinuous functions, expanding the theoretical understanding of what shallow networks can represent.

A complementary line of work improves practical representation quality by augmenting MLPs with structured encodings. {\ingp}~\cite{Muller2022INGP} and NGLOD~\cite{Takikawa2021NGLOD} discretize the domain using a grid and learn feature values at grid vertices; features at arbitrary locations are then obtained by (tri)linear interpolation. Such hybrid representations can greatly increase effective capacity and training speed, but their interpolants remain constrained by the underlying cell structure. In particular, because they rely on cellwise interpolation over a fixed grid, they do not provide a principled way to impose \emph{exact} $C^0$ features, i.e., functions that are continuous while showing controlled, geometry-aligned gradient discontinuities, as in our construction.

Most closely related to our goal are discontinuity-aware neural fields for 2D domains~\cite{Belhe2023,Liu2025}. To represent functions with discontinuities at specific locations (e.g., image edges), \citet{Belhe2023} use a fixed triangulation and assign separate feature values to each side of a triangle edge, producing discontinuities through barycentric interpolation. \citet{Liu2025} generalize this idea by making the discontinuity boundaries learnable, enabling more flexible modeling of structured discontinuities. These methods are well suited to  function discontinuities, as commonly encountered in image domains where color values jump across boundaries. In contrast, our setting targets $C^0$ functions with discontinuities in the gradient. When adapted to $C^0$-continuous targets, their barycentric, mesh-based interpolation ties non-smoothness to the entire triangulation. As a result, sharpness is not confined to the desired feature geometry but ``leaks'' onto other mesh boundaries, whereas our \emph{meshless} formulation  localizes $C^0$ features precisely to the specified set.

Recent work in physics simulation has also investigated meshless schemes for representing boundary-aligned discontinuities~\cite{Chang2025LiftingWN, Liu2025PGD}. In particular, \citet{Chang2025LiftingWN} use the generalized winding number (GWN), which is the solution to Poisson's equation with a jump Neumann boundary condition, as an auxiliary spatial lifting to capture $C^{-1}$ behavior. Although their construction also relies on a boundary integral, it is designed primarily for discontinuities in function values. For $C^0$-continuous fields, \citet{Liu2025PGD} propose using the unsigned distance function (UDF) to impose gradient discontinuities along specified boundaries. However, the UDF is also non-differentiable at additional locations, notably along the medial axis of its zero level set, which generally does not coincide with the intended feature set. In contrast, we introduce a fully controllable $C^0$-continuous feature function based on Poisson's equation with a jump Neumann boundary condition that is exactly $C^0$ on the feature surface and $C^\infty$-smooth elsewhere.

\citet{Liu2025KAN} propose Kolmogorov–Arnold Networks (KANs), which replace fixed node-wise activations with learnable edge-wise activation functions. KANs have shown strong empirical performance on function approximation, often matching or surpassing larger MLPs with fewer parameters. That said, they are primarily evaluated on smooth targets, and their behavior in modeling and \emph{controlling} sharp ($C^0$ but non-$C^1$) features remains largely unexplored.

\subsection{Neural CAD Models}
\label{subsec:neuralrepresentationofcadmodels}

Most CAD reconstruction methods aim to recover \emph{explicit} representations defined by analytic primitives and their relationships. Boundary representation (B-rep) models a solid through its bounding surfaces and their adjacency/topology~\cite{liu2024point2cad, li2025deep, shen2025mesh2brep, usama2026nurbgen}, emphasizing accurate surface fitting and consistent patch connectivity. Constructive solid geometry (CSG), in contrast, seeks to infer a shape's construction program as a sequence or tree of Boolean operations~\cite{du2018inversecsg, dupont2025transcad}. Such procedural descriptions are flexible, editable, and closely aligned with manual CAD workflows. 

In addition to fitting analytic primitives, another line of research reconstructs triangular meshes with sharp geometric features directly from raw point clouds. For example, RFEPS~\cite{Xu2022REEPS} employs discrete optimal transport to refine normal distributions around discontinuities, thereby identifying feature lines accurately. A restricted power diagram is then computed to generate a final mesh that preserves the recovered sharp features.

Alongside these explicit approaches, a growing body of work learns \emph{implicit} CAD representations using multi-layer perceptrons, typically in the form of signed distance fields or occupancy fields. Patch-based implicit methods such as {\nhrep}~\cite{Guo2022NH-Rep} and Patch-Grid~\cite{Lin2025PatchGrid} compose multiple learned implicit functions, often one per surface patch, using CSG-style merging to recover sharp edges and corners. However, these approaches typically rely on a given patch decomposition and therefore do not directly address feature discovery from raw, unsegmented inputs.

Other methods incorporate CAD-specific priors while still using standard MLPs. NeurCADRecon~\cite{Dong2024NeurCADRecon} fits an SDF from unoriented point clouds with a developability prior, encouraging near-zero Gaussian curvature to better match piecewise-developable CAD surfaces. NeuVAS~\cite{Wang2025NeuVAS} targets variational surface modeling under sparse geometric control and supports $G^0$ sharp feature curves specified by input curve sketches. Despite these advances, controlling \emph{where} non-differentiability occurs remains challenging for standard MLP parameterizations.

\begin{table*}[!htbp]
\centering
\small
\setlength{\tabcolsep}{4pt}
\begin{tabular}{llcccc}
\toprule
\textbf{Type} & \textbf{Method} & \textbf{Input} & \textbf{Structure} & \textbf{Output} & \textbf{Sharpness} \\
\midrule
\multirow{11}{*}{\textbf{Explicit}} 
& Point2CAD~\cite{liu2024point2cad} & Point cloud & Segmentation + analytic fitting & B-rep & Yes \\
& Split-and-Fit~\cite{Liu2024SplitandFit} & Point cloud & Voronoi & B-rep & Yes\\
& Mesh2Brep~\cite{shen2025mesh2brep} & Mesh & Primitive fitting + constraints & B-rep & Yes \\
& FR-CSG~\cite{Chen2025FRCSG} & Mesh & {Primitive fitting + Optimization} & CSG tree & Yes \\
& D\textsuperscript{2}CSG~\cite{Yu2023D2CSG} & 3D model & Unsupervised learning & CSG tree & Yes\\
& CAPRI-Net~\cite{Yu2022CAPRI-Net} & Point cloud & Self-supervised learning & CSG tree & Yes \\
& CSGNet~\cite{Sharma2022Neural} & 2D/3D model & RNN & CSG program & Yes \\
& \cite{Wu2018Constructing} & Point cloud & Primitive fitting + Optimization & CSG tree & Yes \\
& InverseCSG~\cite{du2018inversecsg} & 3D model & Program synthesis / search & CSG tree & Yes \\
& TransCAD~\cite{dupont2025transcad} & Point cloud & Sequence model & CAD sequence & Yes \\
& NURBGen~\cite{usama2026nurbgen} & Text & LLM-driven NURBS modeling & B-rep & Yes \\
\midrule
\multirow{8}{*}{\textbf{Implicit}} 
& {\siren}~\cite{sitzmann2020siren} & Point cloud & MLP (periodic activations) & SDF & No \\
& NGLOD~\cite{Takikawa2021NGLOD} & SDF & MLP + octree/grid features + interpolation & SDF & Limited \\
& {\ingp}~\cite{Muller2022INGP} & SDF & MLP + hash-grid encoding + interpolation & SDF & Limited \\
& {\nhrep}~\cite{Guo2022NH-Rep} & Subdivided Patches & MLP + neural halfspaces + Boolean tree & SDF & Yes \\
& Patch-Grid~\cite{Lin2025PatchGrid} & Subdivided Patches & MLP + patch SDFs + CSG merge grid & SDF & Yes \\
& NeurCADRecon~\cite{Dong2024NeurCADRecon} & Point cloud & MLP + developability prior & SDF & No \\
& NeuVAS~\cite{Wang2025NeuVAS} & Curve sketches & MLP + variational regularization & SDF & Yes$^\ast$ \\
& {\sharpnet} (ours) & Point cloud & MLP + feature function (PDE) & SDF & Yes \\
\bottomrule
\end{tabular}
\caption{High-level comparison of representative CAD reconstruction and modeling methods. SIREN, NGLOD, and {\ingp} are general-purpose neural field architectures that apply beyond CAD; here we list them specifically in the context of CAD shape representation and reconstruction. ``Limited'' indicates that sharpness is not explicitly localized to the true feature geometry (e.g., it may be grid-/cell-aligned or resolution-limited). $^\ast$NeuVAS models $G^0$ sharp curves specified by input sketches.}
\label{tab:cadcomparison}
\end{table*}

In contrast, we propose an implicit neural representation \emph{explicitly} designed to model $C^0$ sharp features at prescribed (or learned) locations while remaining smooth elsewhere. Table~\ref{tab:cadcomparison} summarizes representative CAD modeling and reconstruction approaches.

\section{SharpNet}
\label{sec:theory}

Let \(\Omega \subset \mathbb{R}^d\) denote the spatial domain, and let
\(\Phi_\theta\colon \Omega \times \mathbb{R}^n \to \mathbb{R}^m\)
be an MLP with parameters \(\theta\).
Unless stated otherwise, we assume smooth activation functions, so $\Phi_\theta$ is $C^\infty$ with respect to its inputs. The spatial input dimension $d$, the auxiliary feature dimension $n$, and the output dimension $m$ are arbitrary. In this paper, we focus on 2D and 3D settings with scalar outputs, i.e., $d \in \set{2,3}$ and 
$m = 1$. For convenience, we consider the case where the auxiliary feature dimension is set to \(n = 1\).

Let $M \subset \Omega$ be a (possibly non-manifold) curve or surface on which the represented function is required to remain continuous but non-differentiable. To encode the location of such sharp features, we introduce an auxiliary \emph{feature function} $f\colon\Omega \rightarrow \mathbb{R}$ that is non-differentiable exactly on $M$ and smooth elsewhere. We concatenate $f$ with the spatial coordinates $\mathbf{x}$ and denote the resulting $f$-augmented MLP by $\Phi_\theta(\mathbf{x}, f(\mathbf{x}))$, which we call {\sharpnet}.

\subsection{Propagation of Non-differentiability under Composition}
\label{subsec:transitivity}

Our approach relies on the fact that non-differentiability in the input map $\mathbf{x}\mapsto f(\mathbf{x})$ can be transferred to the composed field $\mathbf{x}\mapsto \Phi_\theta(\mathbf{x},f(\mathbf{x}))$, provided the network is locally sensitive to the feature channel. This observation is formalized below.

\begin{theorem}
\label{th:transitivity}
Assume $M$ is codimension-one in $\Omega$. Let $f$ be differentiable on $\Omega\setminus M$ and non-differentiable on $M$. Assume $\Phi_\theta(\mathbf{x},u)$ is differentiable with respect to $(\mathbf{x},u)$, and that $\partial \Phi_\theta/\partial u$ does not vanish on $M$ (i.e., the network locally depends on the feature channel). Then the composed function $\mathbf{x}\mapsto \Phi_\theta(\mathbf{x},f(\mathbf{x}))$ is differentiable on $\Omega\setminus M$ and non-differentiable on $M$.
\end{theorem}

The proof is presented in the supplementary material. Theorem~\ref{th:transitivity} demonstrates that non-differentiability does not propagate through composition \emph{unconditionally}. Two common failure cases are:
\begin{enumdescriptioninline}
    \item [Feature insensitivity.] If the composed model does not depend on the feature channel (e.g., the weights associated with $f$ vanish), then the output reduces to a function of $\mathbf{x}$ alone, which is equivalent to replacing $f$ with a constant zero function. In that case, the composite function remains differentiable everywhere.
    \item [Local flattening.] Even when the model depends on $f$, the non-differentiability of $f$ can be canceled if the outer function has zero partial derivative with respect to $f$ at the corresponding non-differentiable point. For example, $f(t)=\abs{t}$ is not differentiable at $t=0$, but $g(f(t)) = f(t)^2 = t^2$ is differentiable because $g'(0)=0$.
\end{enumdescriptioninline}

Therefore, as long as the network remains sensitive to the feature channel along \(M\), the composed field \(\Phi_\theta(\mathbf{x}, f(\mathbf{x}))\) inherits the non-differentiability of $f$. In practice, the condition \(\partial \Phi / \partial f \neq 0\) on (or almost everywhere on) \(M\) is mild, so {\sharpnet} can effectively represent non-differentiable features.

\subsection{Feature Function}
\label{subsec:featurefunction}

\subsubsection{Definition}

To define the feature function $f$, we seek a function that is twice differentiable in $\Omega \setminus M$ and continuous across a codimension-one set $M$, and whose normal directional derivative exhibits a jump discontinuity on $M$. We therefore consider Poisson's equation:
\begin{equation}
    \label{eqn:poisson}
   \nabla^{2}f(\mathbf{x})=h(\mathbf{x}), \quad \mathbf{x} \in \Omega \setminus M
\end{equation}
subject to the following conditions on $M$:
{\setlength{\leftmargini}{1.2em}
\begin{itemize}
    \item Jump Neumann boundary condition:%
    \begin{equation}
    \label{eqn:jumpboundary}
    \quad \frac{\partial f(\mathbf{x})}{\partial\mathbf{n}_{\mathbf{x}}^+}-\frac{\partial f(\mathbf{x})}{\partial\mathbf{n}_{\mathbf{x}}^-}=g(\mathbf{x}), \quad \mathbf{x} \in M \setminus \partial M
    \end{equation}
    \item Continuity condition: \begin{equation}
    \label{eqn:continuity}
    \quad \lim_{\mathbf{y}\rightarrow \mathbf{x}}f(\mathbf{y})=f(\mathbf{x}), \quad \mathbf{x} \in M 
    \end{equation}
\end{itemize}}
Here, $h(\mathbf{x})$ is continuous on $\Omega \setminus M$ and $\partial M$ denotes the boundary of $M$. The vector $\mathbf{n}_{\mathbf{x}}$ denotes a unit normal at a point $\mathbf{x} \in M$, and the one-sided normal derivative of $f(\mathbf{x})$ in the direction of $\mathbf{n}_{\mathbf{x}}$ is defined as:
\begin{equation*}
\frac{\partial f(\mathbf{x})}{\partial \mathbf{n}_{\mathbf{x}}^\pm} = \lim_{\epsilon \to 0^{\pm}} \frac{f(\mathbf{x}+\epsilon\mathbf{n}_{\mathbf{x}})-f(\mathbf{x})}{\epsilon}.
\end{equation*}
Because $M$ is not necessarily a manifold, the normal vector $\mathbf{n}_{\mathbf{x}}$ is defined only at manifold points $\mathbf{x}\in M$. In particular, it is not well defined at non-manifold points, such as Y-junctions in 2D or points on non-manifold edges in 3D. For the function $g$, we require only that $g(\mathbf{x})\neq 0$ for $\mathbf{x}\in M\setminus \partial M$, which ensures a nontrivial jump in the normal derivative along the feature set.

We summarize the resulting regularity of $f$ in the following theorem:
\begin{theorem}\label{th:conditions}
    Equations~\eqref{eqn:poisson}--\eqref{eqn:continuity} are the necessary and sufficient conditions for a function $f(\mathbf{x})$ to be $C^0$-continuous across $M$ and twice differentiable on $\Omega\setminus M$.
\end{theorem}

\subsubsection{Construction}
A solution to Poisson's equation~\eqref{eqn:poisson} can be expressed in boundary-integral form via Green's third identity~\cite{Costabel1987}:
\begin{equation}
    \label{eqn:green_integral}
    f(\mathbf{x})=\int_{\Omega}h(\mathbf{y})G_{\mathbf{x}}(\mathbf{y})\,\mathrm{d} V_\mathbf{y} + \int_{M} g(\mathbf{y})G_{\mathbf{x}}(\mathbf{y})\,\mathrm{d} S_\mathbf{y},
\end{equation}
where $\mathrm{d}V$ and $\mathrm{d}S$ denote the volume and surface (area) elements, respectively. The function $G_{\mathbf{x}}(\mathbf{y})$ is the Green's function of the Laplacian evaluated at the observation point $\bf x$. It takes the following forms:
{\setlength{\leftmargini}{1.2em}
\begin{itemize}
    \item In 2D: \begin{equation*}
    G_{\mathbf{x}}(\mathbf{y})=\frac{1}{2\pi}\ln\norm{\mathbf{x}-\mathbf{y}}, 
\end{equation*}
\item In 3D: \begin{equation*}
G_{\mathbf{x}}(\mathbf{y})=-\frac{1}{4\pi \norm{\mathbf{x} - \mathbf{y}}}. 
\end{equation*}
\end{itemize}}
To simplify computation of the feature function $f$, we consider the special case $h(\mathbf{x})=0$ and $g(\mathbf{x})=1$. Under this choice, Equation~\eqref{eqn:green_integral} reduces to:
\begin{equation}
\label{eqn:feature}
    f(\mathbf{x})=\int_{M} G(\mathbf{x},\mathbf{y})\,\mathrm{d} S_\mathbf{y}.
\end{equation}
In both 2D and 3D, the integrals of the Green's function over line segments and triangular faces admit closed-form expressions, which are detailed in the supplementary material. Therefore, when $M$ is represented as a polyline (in 2D) or a triangular mesh (in 3D), the feature function $f$ can be computed by summing the contributions of individual line segments or triangles.

\subsubsection{Acceleration}
Equation~\eqref{eqn:feature} integrates over all feature curves $M$ (in 2D) or all feature surfaces $M$ (in 3D), and is therefore expensive to evaluate. We partition $M$ into a set of disjoint subregions $M_i$ such that $\bigcup_i M_i=M$,
\begin{equation*}
    f(\mathbf{x})=\int_{M} G(\mathbf{x},\mathbf{y})\,\mathrm{d} S_\mathbf{y}=\sum_{M_i}\int_{M_i} G(\mathbf{x},\mathbf{y})\,\mathrm{d} S_\mathbf{y}.
\end{equation*}
Each partial integral $f_i(\mathbf{x})=\int_{M_i} G(\mathbf{x},\mathbf{y})\,\mathrm{d} S_\mathbf{y}$ is $C^0$-continuous on $M_i$ and $C^\infty$-smooth on $\Omega\setminus M_i$. Let $\phi_i(\mathbf{x})$ be a $C^\infty$ function, and define
\begin{equation*}    \mathfrak{f}_i(\mathbf{x})=\phi_i(\mathbf{x})\int_{M_i} G(\mathbf{x},\mathbf{y})\,\mathrm{d} S_\mathbf{y}.
\end{equation*}
Then, $\mathfrak{f}_i(\mathbf{x})$ is likewise $C^0$-continuous on $M_i$ and $C^\infty$-smooth on $\Omega\setminus M_i$. We define the localized feature function
\begin{equation}
\label{eqn:feature_local}
    \mathfrak{f}(\mathbf{x})=\sum_i \mathfrak{f}_i(\mathbf{x})=\sum_i \phi_i(\mathbf{x})\int_{M_i} G(\mathbf{x},\mathbf{y})\,\mathrm{d} S_\mathbf{y}.
\end{equation}
In the remainder of the paper, $\mathfrak{f}$ denotes the localized feature function in Equation~\eqref{eqn:feature_local}.

By construction, $\mathfrak{f}(\mathbf{x})$ is $C^0$-continuous on $M$ and $C^\infty$-smooth on $\Omega\setminus M$.
Equation~\eqref{eqn:feature_local} therefore suggests that the global integral in Equation~\eqref{eqn:feature} can be reduced to a local one, provided that $\phi_i(\mathbf{x})$ satisfies the following properties:
\begin{enumerate*}
    \item $\phi_i(\mathbf{x})$ is $C^\infty$-continuous;
    \item $\phi_i(\mathbf{x})$ is compactly supported, i.e., $\phi_i(\mathbf{x})$ is nonzero only in a bounded set.
\end{enumerate*}

A mollifier is a smooth, compactly supported function. A standard example is
\begin{center}
\begin{minipage}{\linewidth}
\begin{minipage}{0.6\linewidth}
\begin{equation*}
    \varphi(r) = \begin{cases}
    \frac{1}{I_n}e^{-\frac{1}{1-r^2}} & \abs{r}<1 \\
    0  &  \abs{r} \geq 1
    \end{cases}
\end{equation*}
\end{minipage}%
\hfill%
\begin{minipage}{0.35\linewidth}
    \includegraphics[width=\linewidth]{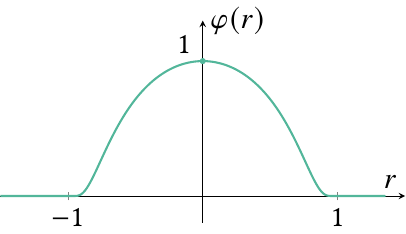}
\end{minipage}
\end{minipage}
\end{center}
where $I_n$ is set to $1/e$ in our experiments. A key property of $\varphi$ is that it is infinitely differentiable on $\mathbb{R}$.

Let $d(\mathbf{x}, M_i)$ denote a smooth distance surrogate from $\mathbf{x}$ to the local feature element $M_i$. Typically, $M_i$ is a line segment in 2D and a triangular face in 3D. Let $c(M_i)$ denote the center of $M_i$, defined as the midpoint of its two endpoints in 2D or the average of its three vertices in 3D. We define $d(\mathbf{x}, M_i)=\norm{\mathbf{x}-c(M_i)}^2$, where the squared form ensures smoothness. Let $r = d(\mathbf{x}, M_i)$ and define $\phi_i(\mathbf{x}) = \varphi(d(\mathbf{x}, M_i))$. Because both $\varphi(r)$ and $d(\mathbf{x})$ are smooth, $\phi_i(\mathbf{x})$ is $C^\infty$-continuous and therefore satisfies the two required conditions.

\subsection{Network Design}
\label{subsec:design}

Define \(
D\partial_{\mathbf{n}}\Phi=\frac{\partial}{\partial\mathbf{n}^+_\mathbf{x}}\Phi(\mathbf{x},\mathfrak{f}(\mathbf{x}))-\frac{\partial}{\partial\mathbf{n}^-_\mathbf{x}}\Phi(\mathbf{x},\mathfrak{f}(\mathbf{x}))\) as the jump in  directional derivatives across \(M\). An important property is that 
\(D\partial_{\mathbf{n}}\Phi\) does not depend on the direction in which \(M\) is crossed; equivalently, it is unchanged if \(\mathbf{n}\) is replaced by \(-\mathbf{n}\). Indeed,
\begin{align*}
\frac{\partial}{\partial(-\mathbf{n})^+_\mathbf{x}}\Phi(\mathbf{x},\mathfrak{f}(\mathbf{x}))=
-\frac{\partial}{\partial\mathbf{n}^-_\mathbf{x}}\Phi(\mathbf{x},\mathfrak{f}(\mathbf{x})),
\\
\frac{\partial}{\partial(-\mathbf{n})^-_\mathbf{x}}\Phi(\mathbf{x},\mathfrak{f}(\mathbf{x}))=
-\frac{\partial}{\partial\mathbf{n}^+_\mathbf{x}}\Phi(\mathbf{x},\mathfrak{f}(\mathbf{x})),
\end{align*}
and therefore
\begin{equation*}
D\partial_{-\mathbf{n}}\Phi=
\frac{\partial}{\partial(-\mathbf{n})^+_\mathbf{x}}\Phi-\frac{\partial}{\partial(-\mathbf{n})^-_\mathbf{x}}\Phi=
-\frac{\partial}{\partial\mathbf{n}^-_\mathbf{x}}\Phi+\frac{\partial}{\partial\mathbf{n}^+_\mathbf{x}}\Phi=
D\partial_{\mathbf{n}}\Phi.
\end{equation*}

\begin{figure}[!htbp]
\centering
\begin{subcaptionblock}{.23\linewidth}
    \centering
    \raisebox{2pt}{\includegraphics[width=\linewidth]{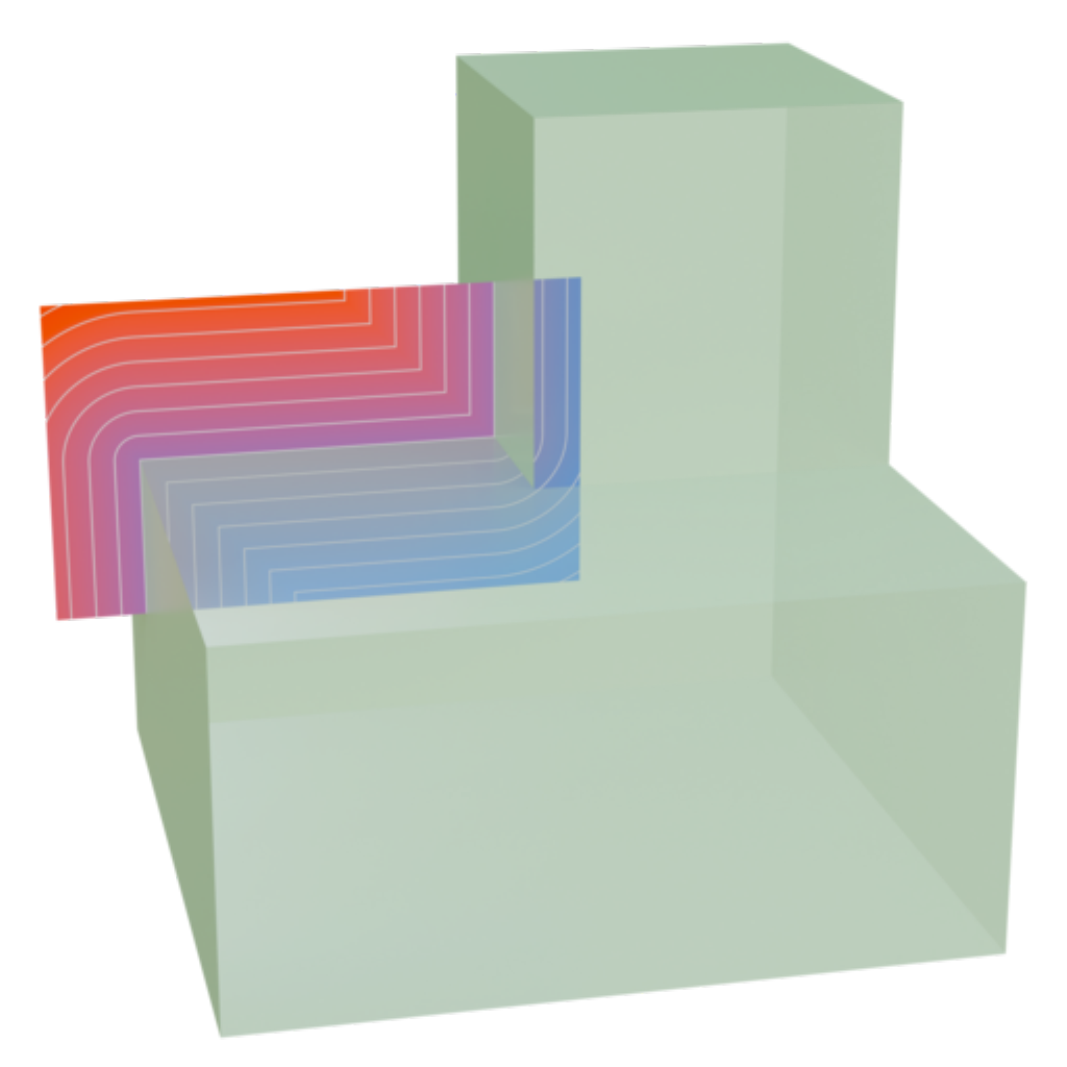}}
    \caption{}
    \label{fig:JumpEdge(cube)}
\end{subcaptionblock}\hfill%
\begin{subcaptionblock}{.37\linewidth}
    \centering
    \raisebox{3pt}{\includegraphics[width=\linewidth]{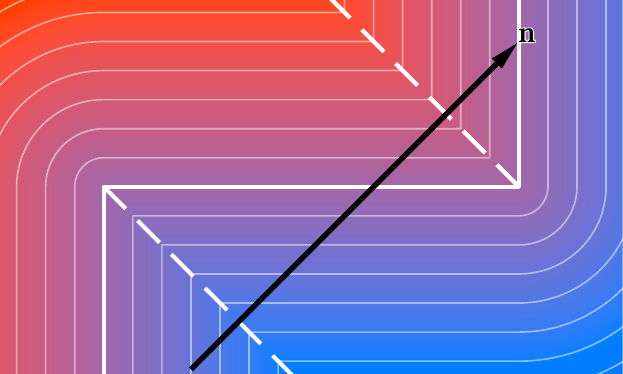}}
    \caption{}
    \label{fig:JumpEdge(face)}
\end{subcaptionblock}\hfill%
\begin{subcaptionblock}{.37\linewidth}
    \centering
    \raisebox{9pt}{\includegraphics[width=\linewidth]{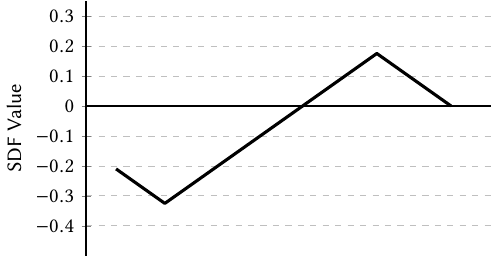}}
    \caption{}
    \label{fig:JumpEdge(line)}
\end{subcaptionblock}
\caption{
The cuboid-like shape in (\subref{fig:JumpEdge(cube)})~features several sharp edges with right dihedral angles. We consider a cross-section that passes through both convex and concave edges. In (\subref{fig:JumpEdge(face)}), the SDF on this cross-section is visualized, and we select a line segment perpendicular to the medial axis; the SDF values along this segment are plotted in (\subref{fig:JumpEdge(line)}). From the SDF profile along this line, we see that medial-axis points corresponding to locally convex surface regions induce a concave SDF curve, which manifests as a positive jump in the directional derivative. In contrast, medial-axis points within locally concave surface regions give rise to a convex SDF curve, leading to a negative jump in the directional derivative.
}
\label{fig:JumpEdge}
\end{figure}

Using the example in Figure~\ref{fig:JumpEdge}, we see that the sign of $D\partial\Phi$ is determined by whether $\Phi$ is locally convex or locally concave. When $\Phi$ is locally concave, $D\partial\Phi$ is positive; when $\Phi$ is locally convex, $D\partial\Phi$ is negative.

\begin{figure}[!htbp]
\centering
\begin{subcaptionblock}{.4\linewidth}
    \centering
    \includegraphics[width=\linewidth]{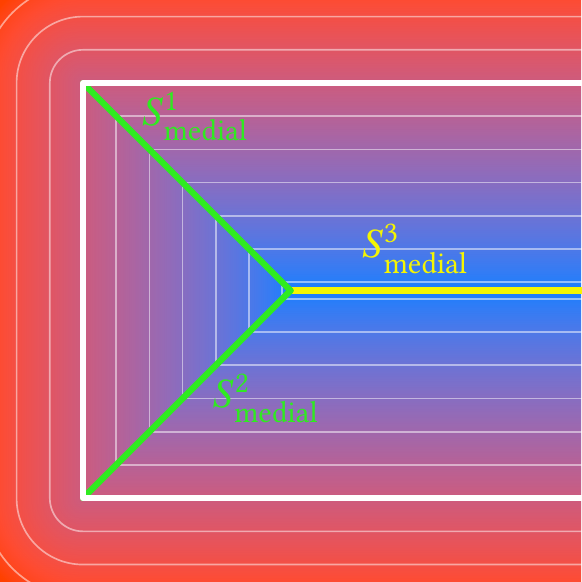}
    \caption{}
    \label{fig:JumpVal3d (jump_rect)}
\end{subcaptionblock}\hfill%
\begin{subcaptionblock}{.5\linewidth}
    \centering
    \includegraphics[width=.84\linewidth]{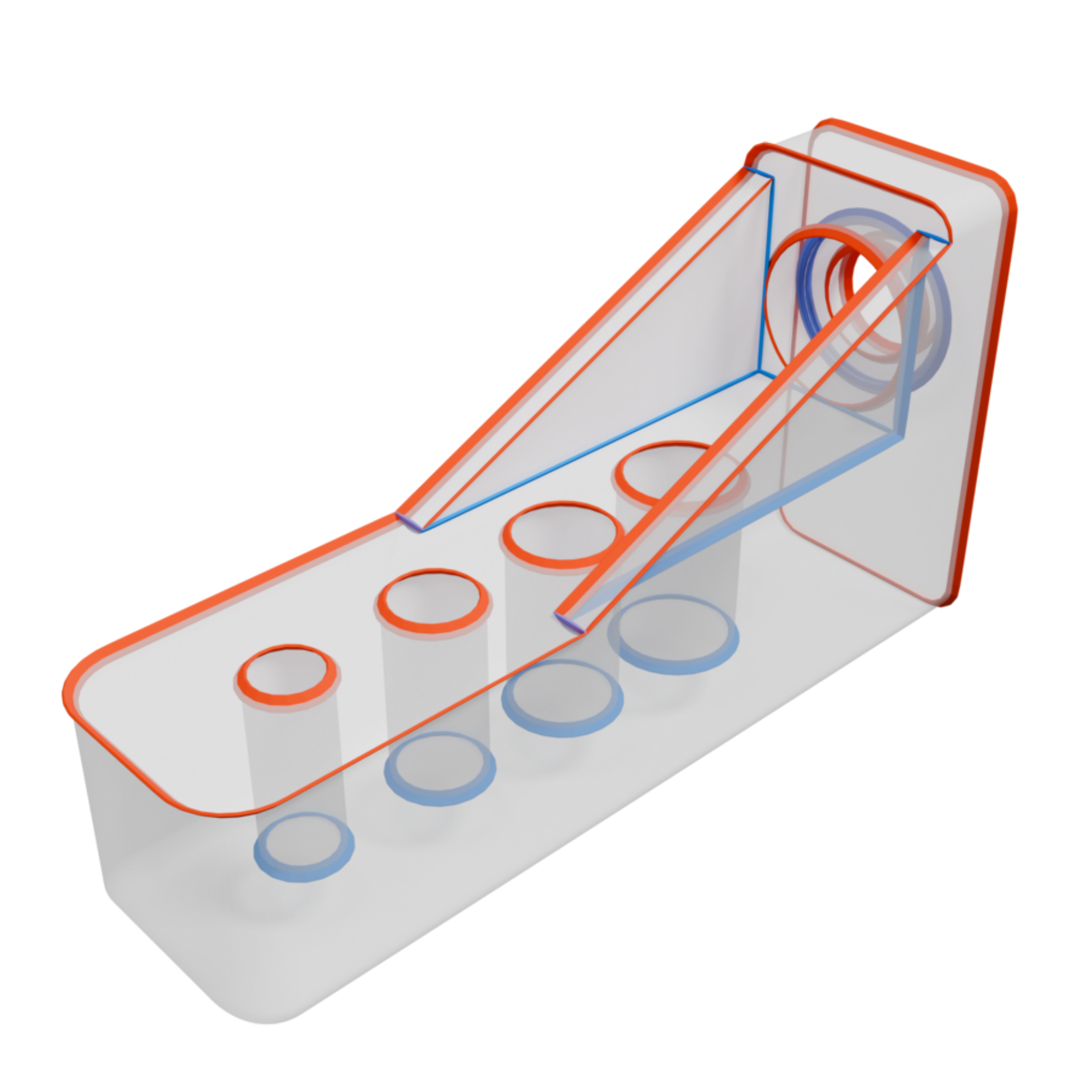}\hfill%
    \raisebox{6pt}{\includegraphics[width=.15\linewidth]{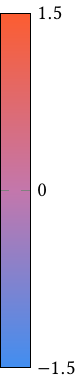}}
    \caption{}
    \label{fig:JumpVal3d (feature meet)}
\end{subcaptionblock}
\caption{
For a watertight surface \(S\) in either 2D or 3D, the SDF may exhibit directional derivative discontinuities across the medial axis or medial surface \(S_\mathrm{medial}\) at junctions. If we define the sign convention of the SDF such that it is positive outside \(S\) and negative inside, then convex portions of \(S\) correspond to positive directional derivative jumps on the interior medial axis, while concave portions of \(S\) correspond to negative directional derivative jumps on the exterior medial axis.
(\subref{fig:JumpVal3d (jump_rect)})~Medial axes carrying different directional derivative jump values may converge at the same point. The branches $\smash{S^1_{\mathrm{medial}}}$ and $\smash{S^2_{\mathrm{medial}}}$ have the same jump magnitude of \(\sqrt{2}\), whereas the jump increases to \(2\) suddenly along $\smash{S^3_{\mathrm{medial}}}$. 
(\subref{fig:JumpVal3d (feature meet)})~In 3D, positive and negative directional derivative jumps can meet at junction points whose local neighborhoods resemble saddle surfaces.
For a CAD model, by coloring the medial surface near sharp edges according to the directional derivative jump values, multiple saddle-like junctions can be observed.
}
\label{fig:JumpVal3d}
\end{figure}

Even when the local shape remains either concave or convex, so that the sign of $D\partial\Phi$ does not change, its magnitude can still jump abruptly at junctions, as illustrated in Figure~\ref{fig:JumpVal3d}(\subref{fig:JumpVal3d (jump_rect)}). At corners of a 3D model, concave and convex sharp edges may meet, causing the convex and concave SDF behaviors to intersect and thereby producing a sign change in $D\partial\Phi$, as shown in Figure~\ref{fig:JumpVal3d}(\subref{fig:JumpVal3d (feature meet)}). At such locations, $D\partial\Phi$ is discontinuous itself.

\begin{figure}[!htbp]
\centering
\includegraphics[width=\linewidth]{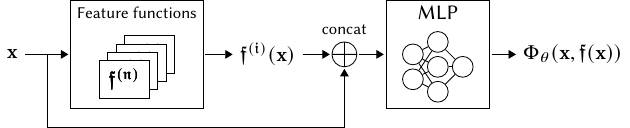}
\caption{Architecture of {\method}. The input coordinates are mapped by feature functions $\mathfrak{f}^{(1)},\dotsc,\mathfrak{f}^{(\mathfrak{n})}$, which are then concatenated with the original input coordinates and sent into an MLP that produces the final value.}
\label{fig:architecture}
\Description[]{The architecture of our method, showing the flow of data.}
\end{figure}

To enable {\sharpnet} to represent discontinuous $D\partial\Phi$, we partition $M$ into disjoint subsets $\set{M^{(i)}}_{i=1}^{\mathfrak{n}}$ such that $M=\bigcup_{i=1}^{\mathfrak{n}} M^{(i)}$. Each $M^{(i)}$ need not be connected. For every subset $M^{(i)}$, we compute a corresponding feature function $\mathfrak{f}^{(i)}(\mathbf{x})$.
We then redefine {\sharpnet} as $\Phi_{\theta}(\mathbf{x},\mathfrak{f}^{(1)}(\mathbf{x}),\dotsc,\mathfrak{f}^{(\mathfrak{n})}(\mathbf{x}))$, which allows the network to represent piecewise-varying directional-derivative jumps across the boundaries of the subsets $M^{(i)}$. 
The resulting network architecture is shown in Figure~\ref{fig:architecture}.

Let $\Phi_{\mathfrak{f}^{(i)}}$ denote the partial derivative of $\Phi_{\theta}(\mathbf{x},\mathfrak{f}^{(1)}(\mathbf{x}),\dotsc,\mathfrak{f}^{(\mathfrak{n})}(\mathbf{x}))$ with respect to $\mathfrak{f}^{(i)}(\mathbf{x})$.
By the chain rule, we can deduce that $D\partial_{\mathbf{n}}\Phi=\Phi_{\mathfrak{f}^{(1)}}D\partial_{\mathbf{n}}\mathfrak{f}^{(1)}(\mathbf{x})+\dotsb+\Phi_{\mathfrak{f}^{(\mathfrak{n})}}D\partial_{\mathbf{n}}\mathfrak{f}^{(\mathfrak{n})}(\mathbf{x})$. When $\mathbf{x} \in M^{(i)}$, only the $i$-th term contributes because $D\partial_{\mathbf{n}}\mathfrak{f}^{(i)}(\mathbf{x}) = 0$ for $\mathbf{x} \notin M^{(i)}$. Consequently, $D\partial_{\mathbf{n}}\Phi=\Phi_{\mathfrak{f}^{(i)}}D\partial_{\mathbf{n}}\mathfrak{f}^{(i)}(\mathbf{x})$, $\mathbf{x}\in M^{(i)}$. Therefore, $D\partial_{\mathbf{n}}\Phi$ may be discontinuous across the boundaries of the subsets $M^{(i)}$, while remaining continuous within each subset if $\Phi_{\mathfrak{f}^{(i)}}$ is appropriately constrained.

For the 2D applications, the feature set is composed of line segments connected at vertices and can therefore be viewed topologically as an undirected graph. We assume that discontinuities in the directional-derivative jump can occur only at junctions, i.e., at vertices with degree greater than or equal to three. Under this assumption,  partitioning \(M\) can be reformulated as an edge-coloring problem: assign colors to the edges of \(M\) such that at all edges incident to a junction have distinct colors. The goal is to use as few colors as possible; in practice, only a small number is needed to keep the feature dimensionality manageable.

We use a similar partitioning strategy in the 3D applications. For CAD signed-distance learning, we construct strip-like surfaces around the model's sharp edges. These strips approximate the local medial surface near the boundary and define the feature surface \(M\), so that the associated feature function exhibits directional-derivative jumps near the sharp edges. Applying the same edge-coloring procedure to the undirected graph of sharp edges yields a partition of the sharp-edge set, and each strip element in \(M\) inherits the color of its corresponding sharp edge. This induces the required partition of \(M\).

\paragraph{Remark}
In this work, we use Poisson's equation with jump Neumann boundary conditions to construct the feature function. Although other functions can also be $C^0$-continuous on user-specified sets, they are not necessarily suitable for our purposes. A commonly alternative is the unsigned distance function~\cite{Liu2025PGD}, whose zero level set is the user-specified set $M$. In theory, the UDF is $C^0$-continuous and non-differentiable on $M$. However, it also non-differentiable at additional locations in the domain, specifically along the medial axis of $M$. \citet{Liu2025PGD} mitigate this issue by applying a smooth clamping function to the UDF $d(\mathbf{p})$,
\begin{displaymath}
    \norm{d(\mathbf{p})}_{\mathrm{SC}} =
    \begin{cases}
        \norm{d(\mathbf{p})}_{2} - \frac{1}{2s} \norm{d(\mathbf{p})}^{2}_{2}, & \text{if } d(\mathbf{p}) < s,\\
        \frac{1}{2}s, & \text{otherwise}.
    \end{cases}
    \label{eqn:smooth_clamping}
\end{displaymath}
Although $\norm{d(\mathbf{p})}_{\mathrm{SC}}$ eliminates the medial axis when $d(\mathbf{p}) \geq s$, the medial axis remains when $d(\mathbf{p})<s$. Thus, the unwanted extra $C^0$ loci are not fully removed.

In practice, the medial-axis problem becomes even more severe. To evaluate the UDF, the feature surface $M$ must first be discretized into a triangular mesh, and the UDF at a point $\mathbf{p}$ is approximated by its distance to the nearest triangle face. This procedure generates a large number of medial axes, because every angular bisector plane formed by the convex angle between a pair of adjacent triangle faces constitutes a medial axis.

These additional discontinuities violate the requirement that the feature function be differentiable on $\Omega \setminus M$, so the UDF does not satisfy Equation~\eqref{eqn:poisson}. Consequently, it is not a suitable choice for our feature function. In addition, the smooth clamping function $\norm{d(\mathbf{p})}_{\mathrm{SC}}$ is only $C^1$-continuous at $d(\mathbf{p}) = s$. In \cite{Liu2025PGD}, the feature surface $M$ is not learnable, likely because computing the closest triangle face is not a differentiable operation. By contrast, our feature function $\mathfrak{f}(\mathbf{x})$ is $C^0$-continuous on $M$, $C^\infty$-smooth elsewhere, and differentiable with respect to $M$, which makes the feature surface itself learnable.

\section{2D Applications}
\label{sec:evaluation}

To evaluate {\method} in 2D applications, we design two experiments that target complementary aspects of performance. The first focuses on fitting accuracy, measuring how well {\method} approximates geodesic distance fields when the feature curves are known. The second examines feature learnability, testing the model's ability to recover medial axes when the feature curves are unknown.

\paragraph{Implementation and baselines.} We instantiate {\method} as $\Phi_{\theta}(\mathbf{x},\mathfrak{f}(\mathbf{x}))$ using a standard MLP with Softplus activation, and compare it against three baselines: a standard MLP without the {\method} module, {\method} with ReLU activation, and {\ingp}~\cite{Muller2022INGP}. In all MLP-based implementations, the network has four hidden layers with 256 neurons per layer. For Softplus, we use $\beta=100$. 

For {\ingp} in the geodesic-distance experiments, we adopt the default configuration except that we reduce the number of levels from 16 to 8. We also remap the input coordinates to $[0.0, 0.9]^2$ before encoding. In all experiments other than {\ingp}, we follow \cite{NeRF} and apply positional encoding to the input coordinates, using 4 frequencies in the geodesic experiments and 8 frequencies in the medial-axis experiments.

\paragraph{Loss functions.} We optimize the parameters $\theta$ of the neural network $\Phi_\theta(\mathbf{x},\mathfrak{f}(\mathbf{x}))$ by minimizing the following loss function,
\begin{equation}
\label{eqn:2Dloss}
\mathcal{L}= \mathcal{L}_F + \alpha \cdot \mathcal{L}_\mathcal{R},
\end{equation}
where $\alpha$ is set to 0.5 in all experiments and  controls the relative weight of the regularization term. The term $\mathcal{L}_F$ measures the mean absolute error between the network prediction and the ground-truth field values $F(\mathbf{x}_i)$ across all $k$ sampled points $\set{\mathbf{x}_i}_{i=1}^k$:
\begin{equation*}
\label{eqn:2D_loss}
    \mathcal{L}_F(\theta) = \frac{1}{k}\sum_{i=1}^k \abs{ F(\mathbf{x}_i) - \Phi_\theta(\mathbf{x}_i, \mathfrak{f}(\mathbf{x}_i)) }.
\end{equation*}
The regularization term $\mathcal{L}_\mathcal{R}$ is designed to encourage uniform polyline segments in \(M\) and to prevent folding. Let $\set{\mathbf{a}_i}_{i=1}^m$ denote the set of degree-2 vertices of \(M\), and let $\mathbf{a}_i^l$ and $\mathbf{a}_i^r$ be the two neighboring vertices of each $\mathbf{a}_i$. We define
\begin{equation*}
\begin{aligned}
\mathbf{v}_i \coloneq \frac{\mathbf{a}^l_i+\mathbf{a}^r_i}{2}-\mathbf{a}_i, &&
\mathbf{u}_i \coloneq \frac{\mathbf{a}^r_i-\mathbf{a}^l_i}{\norm{\mathbf{a}^r_i-\mathbf{a}^l_i}},
\end{aligned}
\end{equation*} where $\mathbf{v}_i$ is the differential coordinate~\cite{Sorkine2006} of the curve $\mathbf{a}_i^l\mathbf{a}_i\mathbf{a}_i^r$. The length of $\mathbf{v}_i$ approximates the curvature at $\mathbf{a}_i$, and the direction of $\mathbf{v}_i$ approximates the normal of $\mathbf{a}_i$. $\mathbf{u}_i$ is the unit vector that points from $\mathbf{a}_i^l$ to $\mathbf{a}_i^r$. The loss of regularization \(\mathcal{L}_\mathcal{R}\) is defined by
\begin{equation*}
    \mathcal{L}_\mathcal{R}(M) = \frac{\lambda}{m} \cdot \sum^m_{i=1}\norm{\mathbf{v}_i}+ \frac{1-\lambda}{m} \cdot \sum^m_{i=1}\abs{\mathbf{v}_i \cdot \mathbf{u}_i}.
\end{equation*}
The first term encourages uniform vertex spacing and overall straightness, while the second term preserves an even vertex distribution but allows moderate bending. We choose $\lambda = 0.3$ to balance these two effects, keeping the $M_i$ uniform while still allowing for slight curvature.

\subsection{Fitting Geodesic Distances}
\label{subsec:geodesic}
\begin{figure*}
\captionsetup[subfigure]{justification=centering}
\centering
\begin{subcaptionblock}{.1557\textwidth}
    \centering
    \includegraphics[width=\textwidth]{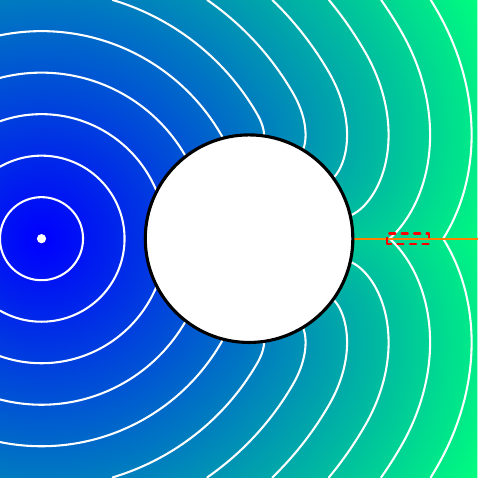}
    \caption{Ground truth\\\phantom{($\varepsilon=0.00\times{10}^{-0}$)}}
\end{subcaptionblock}\hfill%
\begin{subcaptionblock}{.205\textwidth}
    \centering
    \includegraphics[width=\textwidth, page=1]{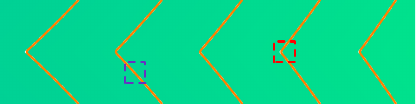}\\[.01\textwidth]
    \includegraphics[width=.49\textwidth, page=2]{img/geodesic/pe.pdf}\hfill%
    \includegraphics[width=.49\textwidth, page=3]{img/geodesic/pe.pdf}
    \caption{Standard MLP\\($\varepsilon=3.13\times{10}^{-5}$)}
    \label{fig:geodesic (raw mlp)}
\end{subcaptionblock}\hfill%
\begin{subcaptionblock}{.205\textwidth}
    \centering
    \includegraphics[width=\textwidth, page=1]{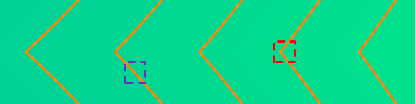}\\[.01\textwidth]
    \includegraphics[width=.49\textwidth, page=2]{img/geodesic/ingp.pdf}\hfill%
    \includegraphics[width=.49\textwidth, page=3]{img/geodesic/ingp.pdf}
    \caption{{\ingp}\\($\varepsilon=3.02\times{10}^{-5}$)}
    \label{fig:geodesic (instantngp)}
\end{subcaptionblock}\hfill%
\begin{subcaptionblock}{.205\textwidth}
    \centering
    \includegraphics[width=\textwidth, page=1]{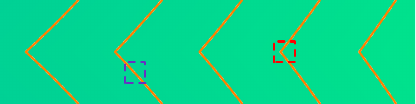}\\[.01\textwidth]
    \includegraphics[width=.49\textwidth, page=2]{img/geodesic/sharp_pe_relu.pdf}\hfill%
    \includegraphics[width=.49\textwidth, page=3]{img/geodesic/sharp_pe_relu.pdf}
    \caption{{\sharpnet} w/ ReLU\\($\varepsilon=1.39\times{10}^{-5}$)}
    \label{fig:geodesic (sharpnet relu)}
\end{subcaptionblock}\hfill%
\begin{subcaptionblock}{.205\textwidth}
    \centering
    \includegraphics[width=\textwidth, page=1]{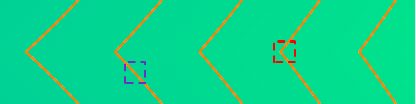}\\[.01\textwidth]
    \includegraphics[width=.49\textwidth, page=2]{img/geodesic/sharp_pe.pdf}\hfill%
    \includegraphics[width=.49\textwidth, page=3]{img/geodesic/sharp_pe.pdf}
    \caption{{\sharpnet} w/ Softplus\\($\varepsilon=0.64\times{10}^{-5}$)}
    \label{fig:geodesic (sharpnet softplus)}
\end{subcaptionblock}\\%
\noindent\rule{\textwidth}{0.4pt}\\[.02\textwidth]%
\begin{subcaptionblock}{.315\textwidth}
    \centering
    \includegraphics[width=\textwidth]{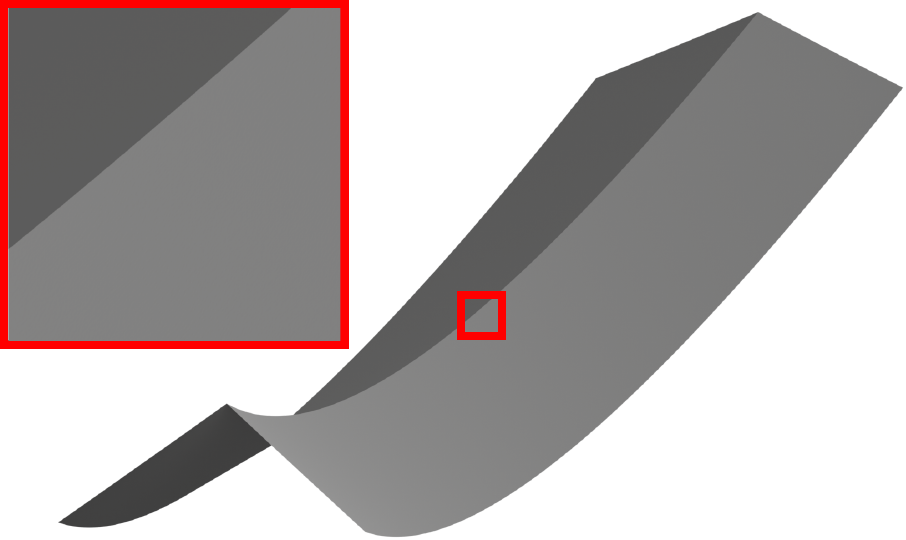}
    \caption{Ground truth\\\phantom{($\varepsilon=0.000$)}}
    \label{fig:geodesic (2) (gt)}
\end{subcaptionblock}\hfill%
\begin{subcaptionblock}{.167\linewidth}
    \centering
    \includegraphics[width=\textwidth]{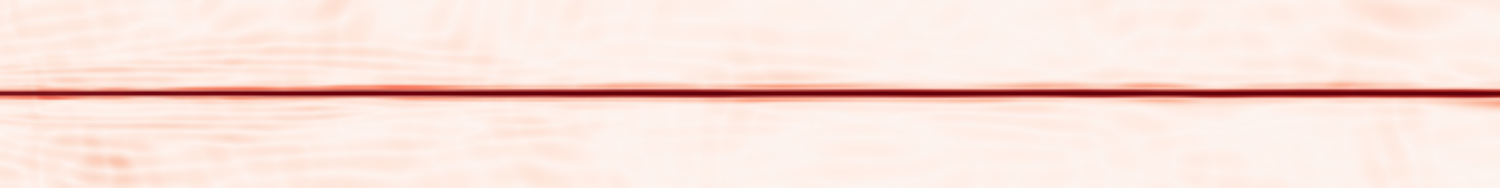}\\
    \includegraphics[width=\textwidth]{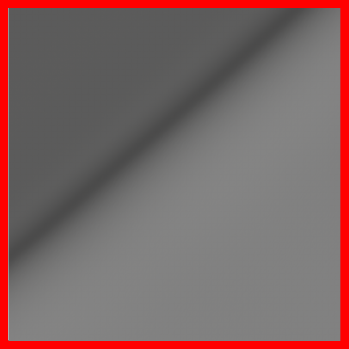}
    \caption{Standard MLP\\($\varepsilon=0.041$)}
    \label{fig:geodesic (2) (raw mlp)}
\end{subcaptionblock}\hfill%
\begin{subcaptionblock}{.167\linewidth}
    \centering
    \includegraphics[width=\textwidth]{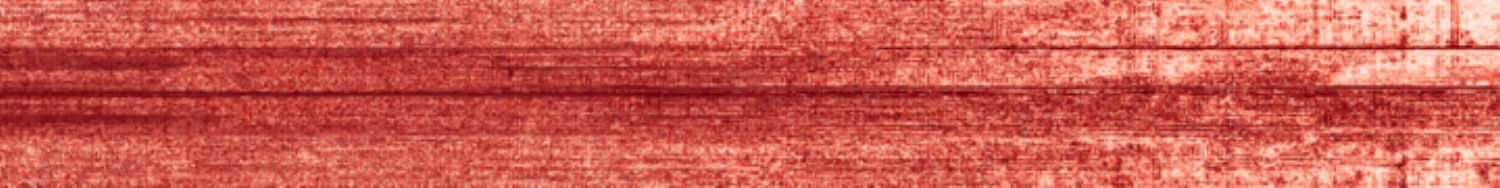}\\
    \includegraphics[width=\textwidth]{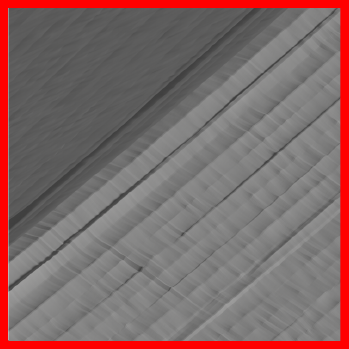}
    \caption{{\ingp}\\($\varepsilon=0.097$)}
    \label{fig:geodesic (2) (instantngp)}
\end{subcaptionblock}\hfill%
\begin{subcaptionblock}{.167\linewidth}
    \centering
    \includegraphics[width=\textwidth]{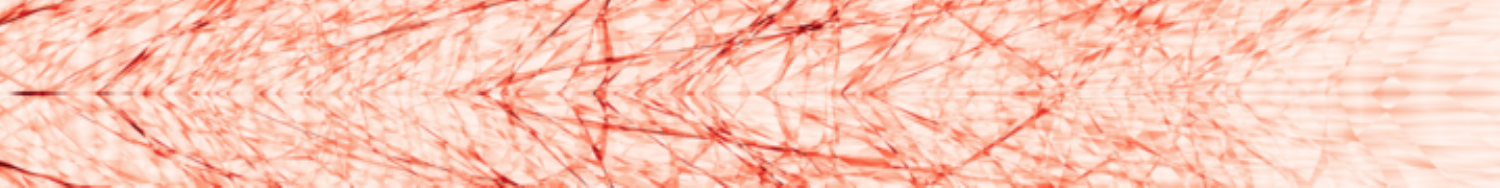}\\
    \includegraphics[width=\textwidth]{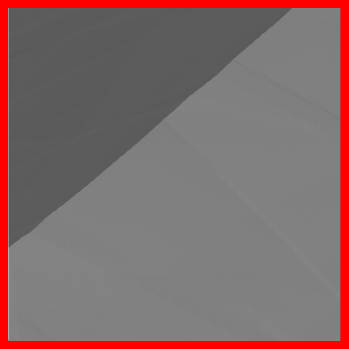}
    \caption{{\sharpnet} w/ ReLU\\($\varepsilon=0.017$)}
    \label{fig:geodesic (2) (sharpnet relu)}
\end{subcaptionblock}\hfill%
\begin{subcaptionblock}{.167\linewidth}
    \centering
    \includegraphics[width=\textwidth]{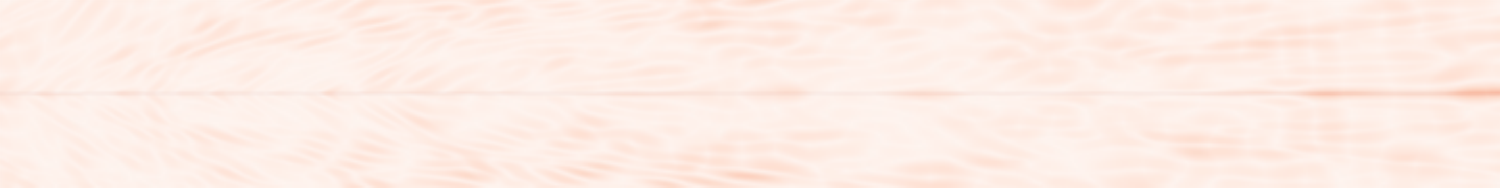}\\
    \includegraphics[width=\textwidth]{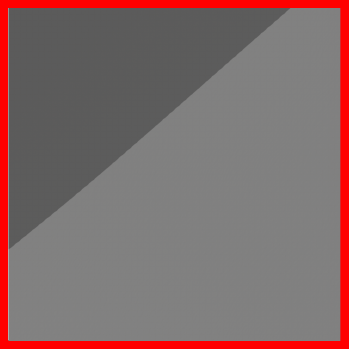}
    \caption{{\sharpnet} w/ Softplus\\($\varepsilon=0.003$)}
    \label{fig:geodesic (2) (sharpnet softplus)}
\end{subcaptionblock}%
\caption{Evaluation of accuracy for training with known feature locations on a 2D geodesic distance field. Top: 2D visualization of the geodesic distance field with iso-distance curves; Bottom: Visualization of the geodesic distance field using height function, which can better reveal errors in the distances. 
}
\label{fig:geodesic-example}
\Description[]{The upper row of figures compare the absolute approximation differences among our method and other baseline methods. The lower row of figures compare the gradient fidelity of the approximation among our method and other baselines.}
\end{figure*}

As illustrated in Figure~\ref{fig:geodesic-example}, we consider a simple 2D domain in which the source point is located at $(0,0)$, and a disk of radius $0.5$ centered at $(1,0)$ is removed. In this configuration, the resulting geodesic distance field is smooth throughout the domain except along a non-differentiable set $M$, which corresponds to the ray $y = 0$ for $x \geq 1.5$. This setup provides a controlled test case in which the feature set $M$ is known a prior. We use it to assess {\method}'s ability to accurately represent geodesic distance fields with sharp transitions.

We evaluate both baseline methods and {\method} variants using neural networks with four layers of width 256. Given the target function $F$, we train {\method} following Equation~\eqref{eqn:2Dloss} without including the regularization term $\mathcal{L}_\mathcal{R}(M)$. A substantial amount of training data is sampled, with $k = 1024^2$. In {\ingp}, we map the coordinates into $[0,0.9]^2$ for hash indexing.

We report our results in Figure~\ref{fig:geodesic-example}. $\varepsilon$ below each figure denotes the error.
In the top row, we evaluate the value accuracy of the learned geodesic distance field. We visualize the predicted distances within a narrow rectangular region ($1.665 \leq x \leq 1.865$, $-0.025 \leq y \leq 0.025$). Since all methods behave similarly away from the sharp feature region $M$, we further restrict the evaluation to a tighter band ($1.55 \leq x \leq 1.95$, $-0.005 \leq y \leq 0.005$) to highlight behavior near the non-differentiable region, and report the absolute error within this band. To aid visual comparison, iso-distance contours of the predicted field (orange solid lines) and the ground truth (white dashed lines) are overlaid. For the value accuracy comparisons,
in Figure~\ref{fig:geodesic-example}(\subref{fig:geodesic (raw mlp)}), the standard MLP oversmooths the sharp corner.
In (\subref{fig:geodesic (instantngp)}), {\ingp} produces jagged artifacts and exhibits undesired sharpness.
In (\subref{fig:geodesic (sharpnet relu)}), {\method} with ReLU improves the result but still introduces spurious sharpness.
In (\subref{fig:geodesic (sharpnet softplus)}), when Softplus activation is used, the result is smooth everywhere except at the intended feature locations.
These artifacts are more pronounced in the bottom row, where we present the learned geodesic field as a height mesh and render the surface of the mesh.

We additionally report the mean absolute error of the gradient of the predicted field and plot the error maps. For the gradient accuracy comparisons,
in (\subref{fig:geodesic (2) (raw mlp)}), the standard MLP remains smooth everywhere, including along the ridge. This produces a prominent dark-red line at the center of the plotted gradient error map.
In (\subref{fig:geodesic (2) (instantngp)}), {\ingp} produces unintended $C^0$ features aligned with the hash-grid structure. This corresponds to grid-like artifacts and creases on the surface. The error of the gradient is very large such that the gradient-error map is nearly saturated (red) across the domain.
In (\subref{fig:geodesic (2) (sharpnet relu)}), for {\sharpnet} with ReLU activation, the non-differentiability of ReLU at zero introduces visible creases that are clearly visible in the rendered mesh as well as in the plotted error map.
Finally, in (\subref{fig:geodesic (2) (sharpnet softplus)}), for {\sharpnet} with Softplus activation, the surface is smooth everywhere except along ridge, resulting in the best overall fitting quality. All images are rendered at high resolution to support close-up inspection. 

\begin{figure*}
\captionsetup[subfigure]{justification=centering}
\centering
\begin{subcaptionblock}{.1015\textwidth}
    \centering
    \includegraphics[width=\textwidth, page=1]{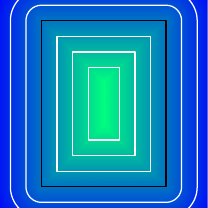}\\
    \includegraphics[width=\textwidth, page=2]{img/medial/rectangle/reference.pdf}
    \caption{GT}
    \label{fig:learningpipeline (reference)}
\end{subcaptionblock}\hfill%
\begin{subcaptionblock}{.205\textwidth}
    \centering
    \includegraphics[width=\textwidth, page=2]{img/medial/rectangle/sharp_pe.pdf}
    \caption{Init medial axis}
    \label{fig:learningpipeline (init axis)}
\end{subcaptionblock}\hfill%
\begin{subcaptionblock}{.205\textwidth}
    \centering
    \includegraphics[width=.48\textwidth, page=4]{img/medial/rectangle/sharp_pe.pdf}\hfill%
    \includegraphics[width=.48\textwidth, page=5]{img/medial/rectangle/sharp_pe.pdf}\\[.02\textwidth]%
    \includegraphics[width=.48\textwidth, page=7]{img/medial/rectangle/sharp_pe.pdf}\hfill%
    \includegraphics[width=.48\textwidth, page=6]{img/medial/rectangle/sharp_pe.pdf}%
    \caption{Medial axis after $i$ iterations}
    \label{fig:learningpipeline (process)}
\end{subcaptionblock}\hfill%
\begin{subcaptionblock}{.260\textwidth}
    \includegraphics[width=\textwidth]{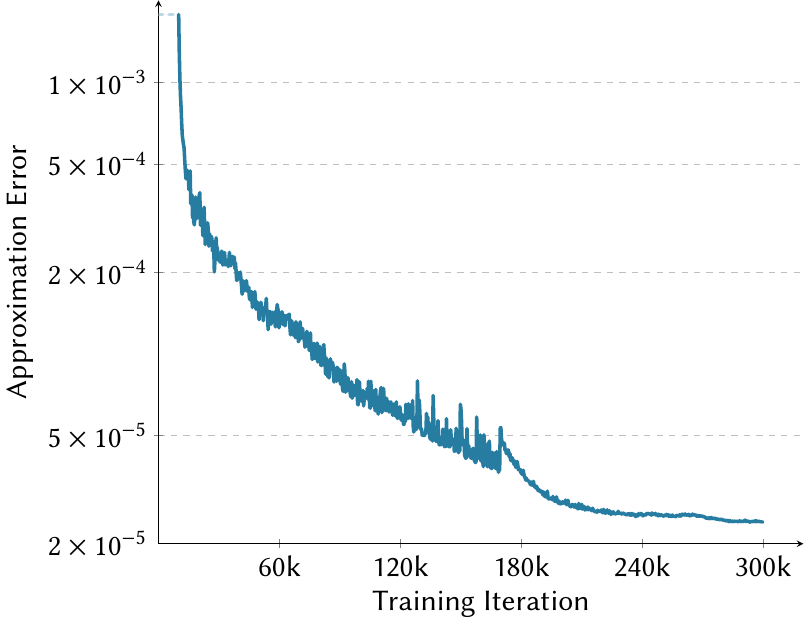}
    \caption{Medial axis error}
    \label{fig:learningpipeline (cd)}
\end{subcaptionblock}\hfill%
\begin{subcaptionblock}{.205\textwidth}
    \centering
    \includegraphics[width=\textwidth, page=3]{img/medial/rectangle/sharp_pe.pdf}
    \caption{Final medial axis}
    \label{fig:learningpipeline (sharp_pe)}
\end{subcaptionblock}
\caption{Evaluation of accuracy for training with inaccurate feature locations on the geodesic distance field of a rectangle.
(\subref{fig:learningpipeline (reference)})~Ground truth geodesic distance field, which is non-differentiable along the medial axis $M$, consisting of 5 segments as shown beneath it.
(\subref{fig:learningpipeline (init axis)})~Initial feature estimate, composed of 94 segments, obtained by subdividing the ground-truth medial axis and perturbing the joint positions to simulate initialization error.
(\subref{fig:learningpipeline (process)})~Close-up views of a Y-junction region, comparing initial feature and the progressively refined feature sets after 100k, 200k, and 300k training iterations (displayed clockwise).
(\subref{fig:learningpipeline (cd)})~Accuracy of the estimated feature set over the course of training. The medial axis is frozen during the first 10k iterations and is marked with dashed line and faint color.
}
\label{fig:learningpipeline}
\end{figure*}

\subsection{Learning Medial Axes}
\label{subsec:medial}

The medial axis of a closed shape $S$ is closely related to geodesic distances, since $S$ defines the source set from which distances are computed. It consists of points in the domain that have more than one closest point on $S$.

Along the medial axis, the distance field is $C^0$-continuous and non-differentiable; therefore, the medial axis naturally serves as the feature set $M$. Given $F$, we jointly optimize the feature curves $M$ and the network $\Phi$ using Equation~(\ref{eqn:2Dloss}), optionally including the regularization term $\mathcal{L}_\mathcal{R}$. Figure~\ref{fig:learningpipeline} illustrates this evaluation on a 2D rectangle example, where we report the medial-axis error. 

In this experiment, we first partition the ground-truth medial axis into 94 segments and perturb the vertex positions to simulate initialization error. We then jointly train the neural network and the medial axis for 300,000 iterations. During the first 10,000 iterations, we freeze the medial axis and optimize only the network, allowing it to fit the distance field without disturbing the coarse medial structure. After this warm-up stage, the medial axis is set free, and all vertices with degree greater than one are optimized. We set the mollifier radius to 0.08, approximately four times the length of each medial-axis segment.

\subsection{Discussion of Alternatives}

Across both the geodesic and medial-axis experiments, {\sharpnet} consistently outperforms the baseline methods,  demonstrating its ability to accurately represent non-differentiable sharp features.

Our feature function $\mathfrak{f}(\mathbf{x})$ is $C^0$-continuous on  $M$ and $C^{\infty}$-smooth elsewhere. This distinguishes our method from closely related approaches such as {\ingp}~\cite{Muller2022INGP} and discontinuity-aware neural networks ({\dann})~\cite{Belhe2023,Liu2025}.

{\ingp}~\cite{Muller2022INGP}, although not  designed specifically for sharp-feature reconstruction, adopts a similar strategy by encoding voxel-corner features in a hash table and performing linear interpolation within each voxel, significantly increasing representational capacity compared to a standard MLP. However, this interpolation enforces $C^0$-continuity across all voxel boundaries. As a result, unintended $C^0$ structures may appear along grid boundaries, while desired non-differentiable features inside a voxel cannot be represented precisely.
{\dann}, introduced by \cite{Belhe2023} and extended by \cite{Liu2025} to incorporate learnable curve features, can be viewed as a two-dimensional, non-uniform, single-resolution variant of {\ingp}. While designed for modeling function discontinuities rather than sharp features, it shares with our method the idea of augmenting neural networks with auxiliary features. In principle, it can be adapted to the $C^0$ setting by assigning identical features on both sides of a discontinuity. However, because it remains interpolation-based, it inherits the same limitation as {\ingp}: it may introduce unwanted sharpness in regions that should remain smooth.

\begin{figure*}
\centering
\setlength{\tabcolsep}{3pt}
\begin{tabular}{ccccccc}
    \begin{minipage}[c][.13\linewidth][c]{.13\linewidth}
        \centering {\sffamily -}
    \end{minipage} &
    \begin{minipage}[c][.13\linewidth][c]{.13\linewidth}
        \centering {\sffamily None}
    \end{minipage} &
    \begin{minipage}[c][.13\linewidth][c]{.13\linewidth}
        \centering \includegraphics[width=\linewidth, page=1]{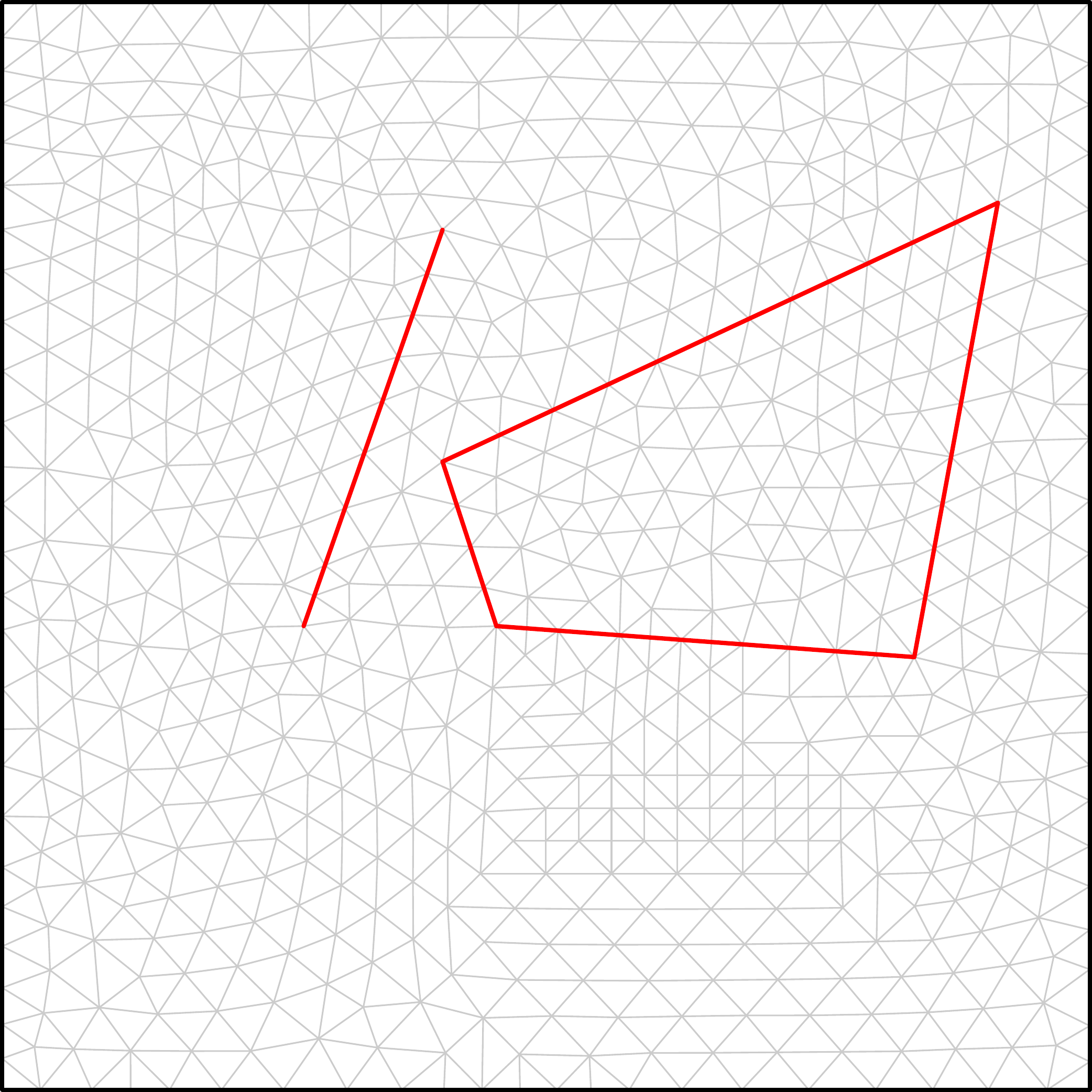}
    \end{minipage} &
    \begin{minipage}[c][.13\linewidth][c]{.13\linewidth}
        \centering \includegraphics[width=\linewidth, page=2]{img/belhe/mesh.pdf}
    \end{minipage} &
    \begin{minipage}[c][.13\linewidth][c]{.13\linewidth}
        \centering {\sffamily Axis-aligned grid}
    \end{minipage} &
    \begin{minipage}[c][.13\linewidth][c]{.13\linewidth}
        \centering \includegraphics[width=\linewidth, page=3]{img/belhe/mesh.pdf}
    \end{minipage} &
    \begin{minipage}[c][.13\linewidth][c]{.13\linewidth}
        \centering \includegraphics[width=\linewidth, page=3]{img/belhe/mesh.pdf}
    \end{minipage} \\[.06\linewidth]
    \includegraphics[width=.13\linewidth, page=3]{img/belhe/reference.pdf} & 
    \includegraphics[width=.13\linewidth, page=3]{img/belhe/pe.pdf} &
    \includegraphics[width=.13\linewidth, page=3]{img/belhe/belhe.pdf} &
    \includegraphics[width=.13\linewidth, page=3]{img/belhe/liu.pdf} &
    \includegraphics[width=.13\linewidth, page=3]{img/belhe/ingp.pdf} &
    \includegraphics[width=.13\linewidth, page=3]{img/belhe/sharp_pe_relu.pdf} &
    \includegraphics[width=.13\linewidth, page=3]{img/belhe/sharp_pe.pdf} \\
    \includegraphics[width=.13\linewidth, page=1]{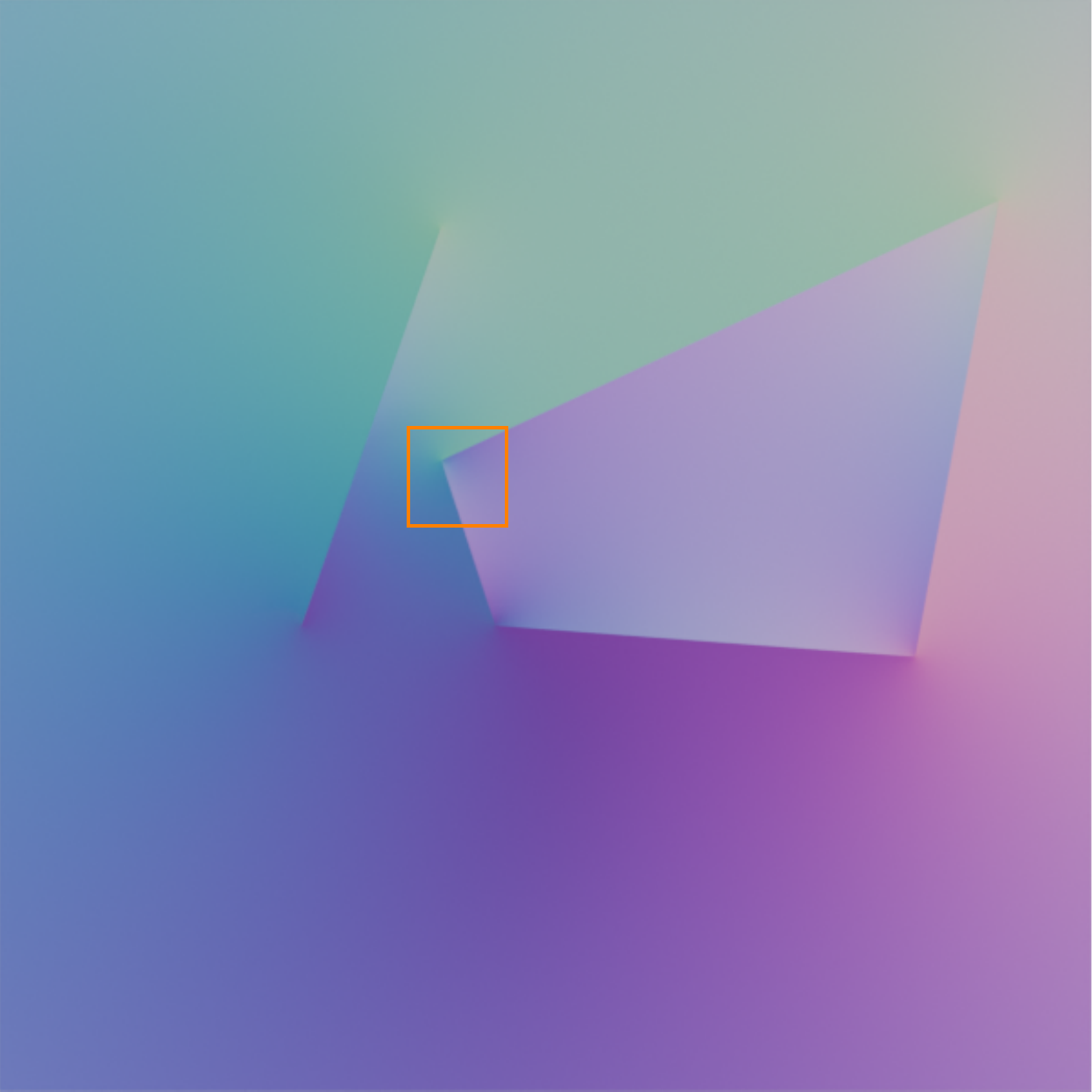} & 
    \includegraphics[width=.13\linewidth, page=1]{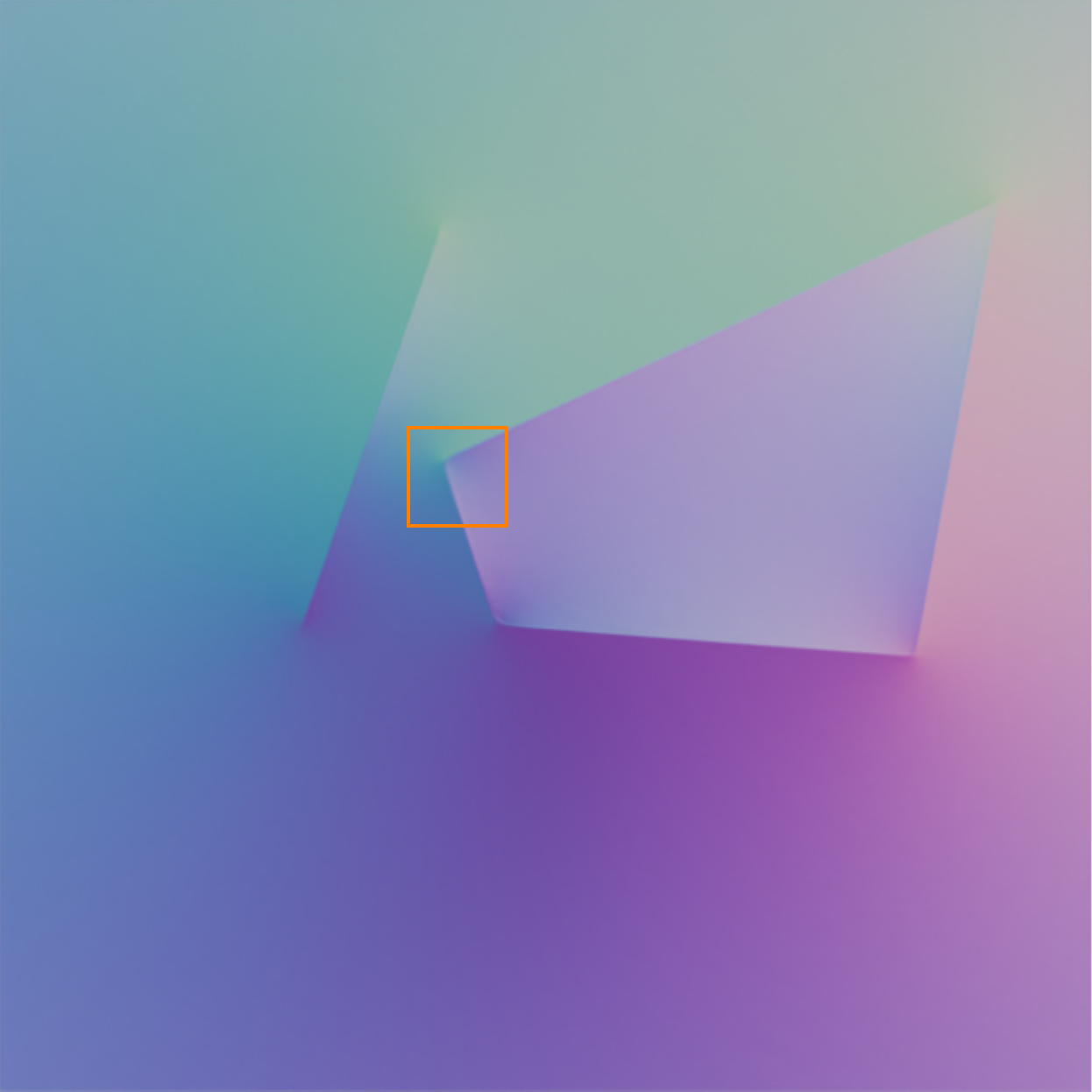} &
    \includegraphics[width=.13\linewidth, page=1]{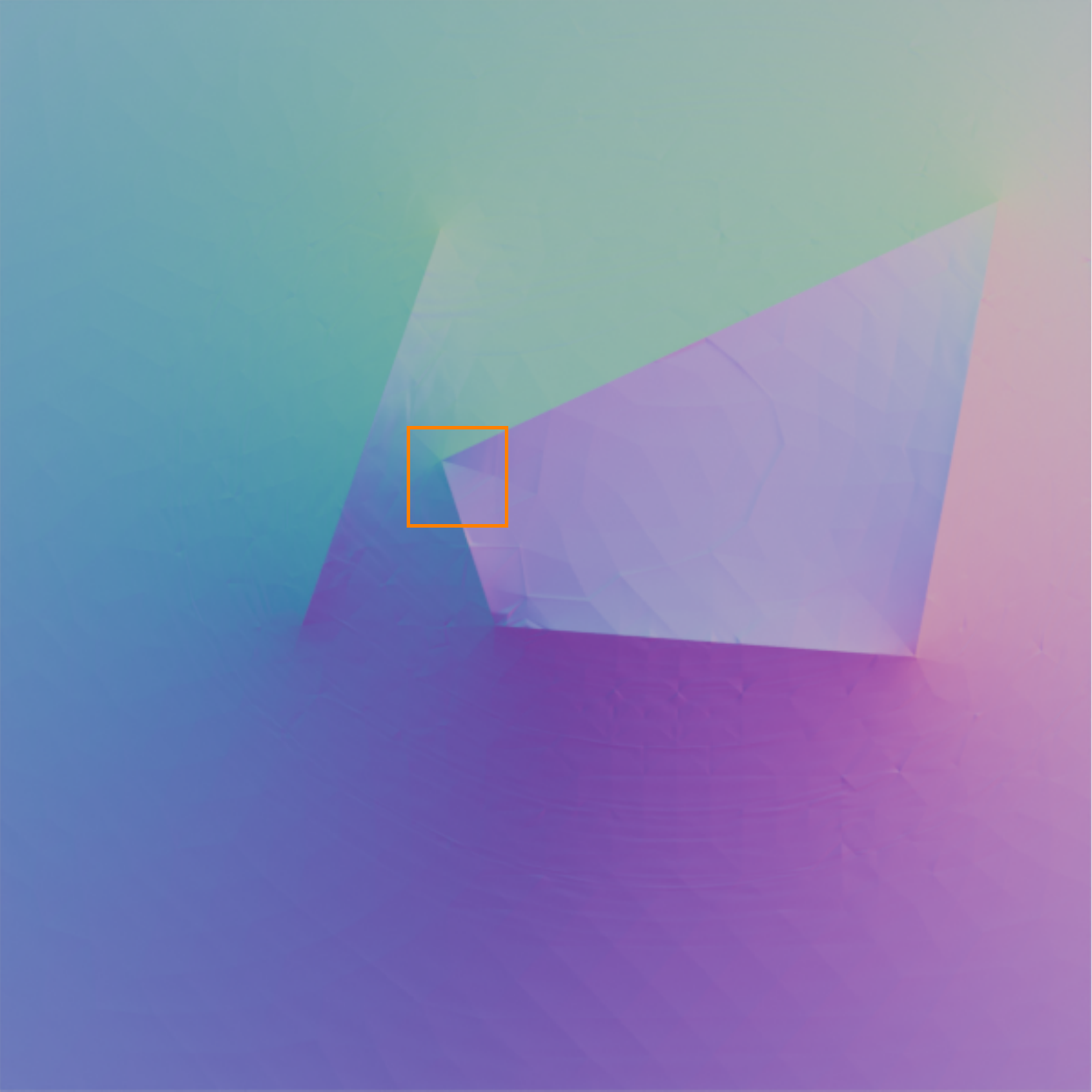} &
    \includegraphics[width=.13\linewidth, page=1]{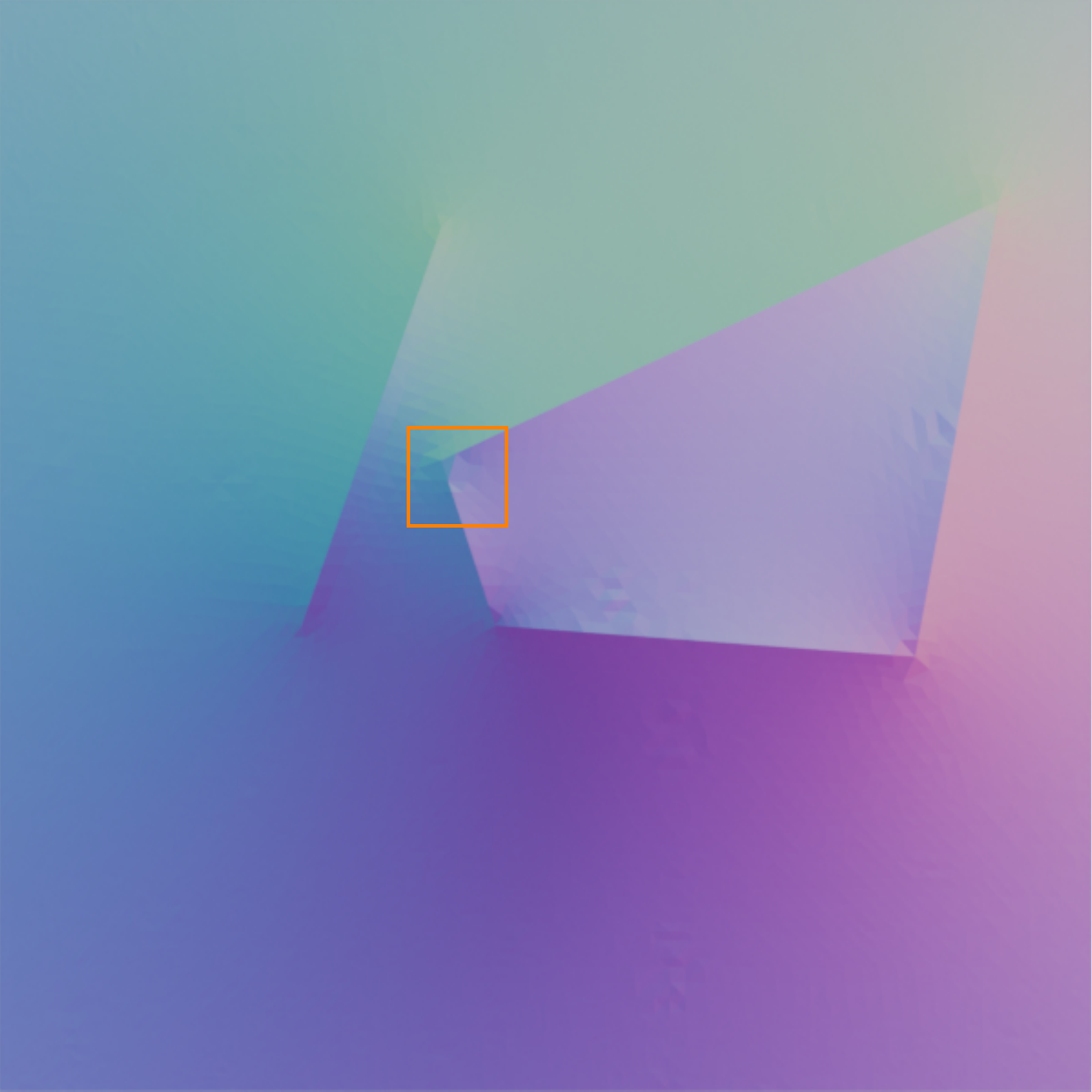} &
    \includegraphics[width=.13\linewidth, page=1]{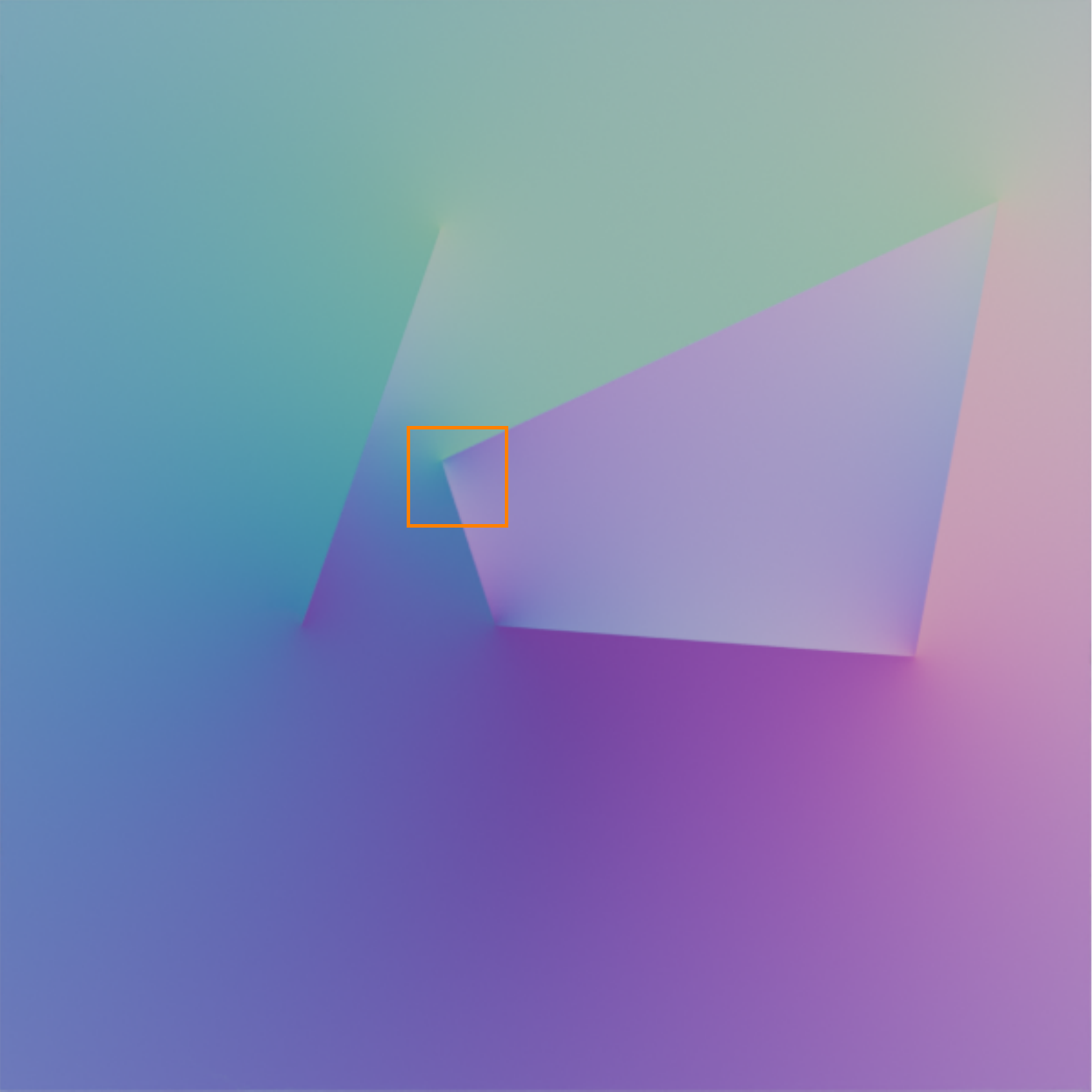} &
    \includegraphics[width=.13\linewidth, page=1]{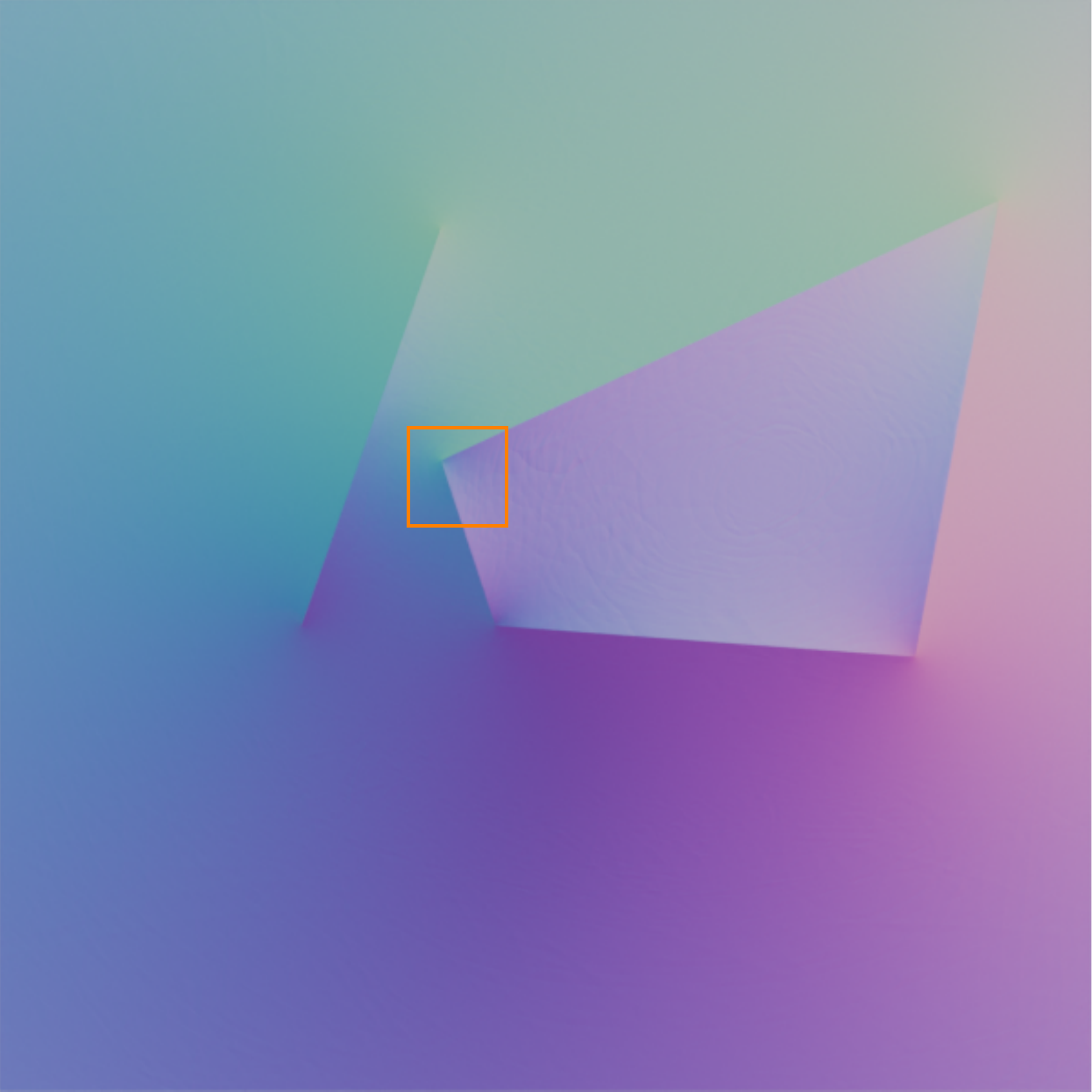} &
    \includegraphics[width=.13\linewidth, page=1]{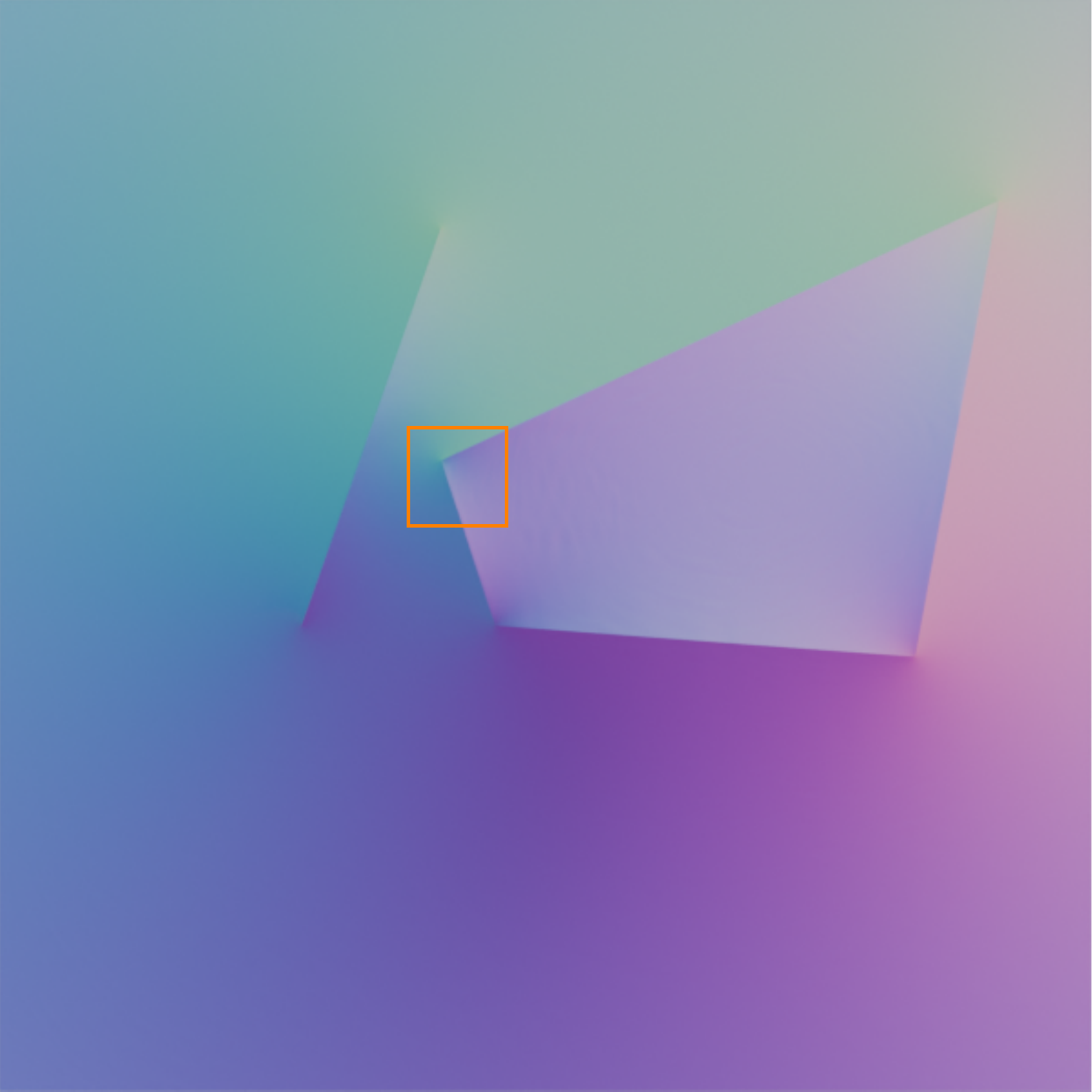} \\
    \includegraphics[width=.13\linewidth, page=2]{img/belhe/reference.pdf} & 
    \includegraphics[width=.13\linewidth, page=2]{img/belhe/pe.pdf} &
    \includegraphics[width=.13\linewidth, page=2]{img/belhe/belhe.pdf} &
    \includegraphics[width=.13\linewidth, page=2]{img/belhe/liu.pdf} &
    \includegraphics[width=.13\linewidth, page=2]{img/belhe/ingp.pdf} &
    \includegraphics[width=.13\linewidth, page=2]{img/belhe/sharp_pe_relu.pdf} &
    \includegraphics[width=.13\linewidth, page=2]{img/belhe/sharp_pe.pdf} \\
    \makecell{\small \sffamily Reference} & {\small \sffamily Raw MLP} & \makecell{\small \sffamily \dann \\ \small \sffamily \cite{Belhe2023}} & {\small \sffamily \cite{Liu2025}} & {\small \sffamily \ingp } & \makecell{\small \sffamily {\sharpnet} \\ \small \sffamily w/ ReLU} & \makecell{\small \sffamily {\sharpnet} \\ \small \sffamily w/ Softplus}
\end{tabular}
\caption{Comparisons against different representations of sharpness strategies. The first row shows the meshing and feature locations. The second to fourth rows display the learned 2.5D models along with locally zoomed-in views.}
\label{fig:belhe_compare}
\end{figure*}

To further illustrate this limitation, in addition to the geodesic experiments in Section~\ref{subsec:geodesic}, we conduct an additional experiment in which we fit a 2D field that is continuous everywhere but non-differentiable along prescribed feature curves. We compare {\sharpnet} (with both Softplus and ReLU activations), {\ingp} with Softplus activation, {\dann} with Softplus activation, and \citet{Liu2025} with their default neural network. The field is visualized as a 3D height surface, with normals used for shading to enhance geometric detail (Figure~\ref{fig:belhe_compare}).

For {\dann}~\cite{Belhe2023} and \citet{Liu2025}, the learned field is non-differentiable not only at the intended feature curves but also at additional locations induced by the mesh structure. {\ingp} exhibits grid-aligned artifacts corresponding to its multi-resolution encoding. {\sharpnet} with ReLU activation introduces visible creases due to the non-differentiability of ReLU itself. In contrast, {\sharpnet} with Softplus activation produces a field that is smooth everywhere except at the specified feature locations, achieving the highest fidelity among all methods.

\section{CAD Applications}
\label{sec:experiments}

We illustrate the practicality of {\sharpnet} using three representative CAD reconstruction tasks. The first examines the direct conversion of explicit CAD meshes into an implicit representation. This scenario emphasizes {\sharpnet}'s capability to accurately encode sharp features when their locations are given. The second addresses a more challenging setting, where CAD models are reconstructed from point clouds augmented with oriented normal constraints. The third and most challenging setting reconstructs CAD models using only point cloud data. In the latter two scenarios, we simultaneously infer both the underlying CAD geometry and its sharp features, demonstrating {\sharpnet}'s strength in sharp feature learning. Depending on the specific input type, we respectively adopt {\nhrep}~\cite{Guo2022NH-Rep}, {\ingp}~\cite{Muller2022INGP}, and NeurCADRecon~\cite{Dong2024NeurCADRecon} as baselines.

\paragraph{Dataset}
We conduct our evaluation on the same collection of 100 models used in \cite{Erler2020}, which are drawn from the ABC dataset~\cite{Koch2019ABC}. The testing models comprise intricate CAD geometries that exhibit a wide range of challenging features, including high-genus surfaces, thin structures, both open and closed sharp edges and curves, as well as other complex and irregular structural characteristics that thoroughly test the robustness of our method.

\paragraph{Metrics}
We use four metrics to evaluate the quality and accuracy of the learned CAD models: Chamfer Distance (CD), Hausdorff Distance (HD), Normal Error (NE), and F-1 Score (FC). Let \(\mathcal{P}_1,\mathcal{P}_2\) be two point sets obtained by sampling from the respective model surfaces, and let \(\norm{\mathcal{P}_i}\) denote the cardinality of \(\mathcal{P}_i\). For any point \(\mathbf{p}\in\mathbb{R}^3\), we define
\begin{equation*}
\mathrm{NN}_{\mathcal{P}_i}(\mathbf{p}) \coloneq \argmin_{\mathbf{q}\in \mathcal{P}_i} \norm{\mathbf{p}-\mathbf{q}} \qquad (i=1,2).
\end{equation*}
For any sample point \(\mathbf{p}\in \mathcal{P}_i\), we denote its associated normal by \(\vec{n}_{\mathbf{p}}\). Let \(r\) be the distance threshold used to determine whether two points are considered a match. We define the metrics above as:

\begin{align*}
\mathrm{CD} =\; &
\frac{\sum_{\mathbf{p}\in\mathcal{P}_1} \norm{\mathbf{p}-\mathrm{NN}_{\mathcal{P}_2}(\mathbf{p})}}
{2\norm{\mathcal{P}_1}}+
\frac{\sum_{\mathbf{p}\in\mathcal{P}_2} \norm{\mathbf{p}-\mathrm{NN}_{\mathcal{P}_1}(\mathbf{p})}}
{2\norm{\mathcal{P}_2}}\,,\\
\mathrm{HD} =\; &
\max\left({\max_{\mathbf{p}\in\mathcal{P}_1} \norm{\mathbf{p}-\mathrm{NN}_{\mathcal{P}_2}(\mathbf{p})}}
,
{\max_{\mathbf{p}\in\mathcal{P}_2} \norm{\mathbf{p}-\mathrm{NN}_{\mathcal{P}_1}(\mathbf{p})}}\right)\,,\\
\mathrm{NE} =\; &
\frac{\sum_{\mathbf{p}\in\mathcal{P}_1} 
\arccos\abs{\vec{n}_{\mathbf{p}}\cdot\vec{n}_{\mathrm{NN}_{\mathcal{P}_2}(\mathbf{p})}}}
{2\norm{\mathcal{P}_1}}+
\frac{\sum_{\mathbf{p}\in\mathcal{P}_2} 
\arccos\abs{\vec{n}_\mathbf{p}\cdot\vec{n}_{\mathrm{NN}_{\mathcal{P}_1}(\mathbf{p})}}}
{2\norm{\mathcal{P}_2}}\,,\\
\mathrm{FC} =\; & \frac{2 R_1 R_2}{R_1+R_2}\,,\;\text{where $R_1$ and $R_2$ denote\,:}\\
& R_1 =
\frac{\norm*{
    \condset{\mathbf{p}\in\mathcal{P}_1}
    {\norm{\mathbf{p}-\mathrm{NN}_{\mathcal{P}_2}(\mathbf{p})}<r}
}}
{\norm{\mathcal{P}_1}}\,,\\
& R_2 =
\frac{\norm*{
    \condset{\mathbf{p}\in\mathcal{P}_2}
    {\norm{\mathbf{p}-\mathrm{NN}_{\mathcal{P}_1}(\mathbf{p})}<r}
}}
{\norm{\mathcal{P}_2}}\,.
\end{align*}

\paragraph{Training details}
For each model, we uniformly sample \(50{,}000\) surface points and rescale them to fit within the bounding box \([-1,1]^3\) centered at \((0,0,0)\). For every sampled point, the distance to its \(k\)-nearest neighbors in the point cloud (\(k=50\)) is used as the standard deviation of a Gaussian distribution from which additional neighbor samples are drawn.
SIREN, {\ingp}, and {\sharpnet} all use a 4-layer 256-width MLP and sine as activation function. In each epoch, we randomly select 20{,}000 points from the 50{,}000 surface samples and draw an additional 20{,}000 near-surface points using Gaussian perturbations around them. We also uniformly sample 10{,}000 points in the extended domain \([-1.1,1.1]^3\) as ambient points. The networks are trained for 15{,}000 epochs using the Adam optimizer~\cite{Kingma2015Adam}, with the specific loss function determined by each experiment. Finally, the mesh is extracted using Dual Contouring~\cite{Ju2022DC}, which preserves sharp features.

\subsection{Implicit Representation from CAD Meshes}
\label{sec:CAD_mesh}
Implicit representation of CAD mesh models has several benefits over explicit representations, such as Boolean operations. Although the representation of the feature surface $M$ is explicit, {\sharpnet} is able to utilize an MLP to learn an implicit distance field to reconstruct CAD models with distinct sharp features. In contrast, other approaches~\cite{Guo2022NH-Rep,Lin2025PatchGrid} have to decompose the model into smaller sub-patches, applying multiple MLPs to represent them. Ultimately, they perform Boolean operations to merge these sub-patches for the reconstruction of sharp features.

We compare our method with {\nhrep}~\cite{Guo2022NH-Rep}, which partitions CAD meshes into sub-patches based on pre-defined input feature curves. Given the mesh as input, we detect sharp feature curves based on the dihedral angles between neighboring faces. For {\sharpnet}, we further define the feature surface \(M\) based on the collection of angular bisector planes along these sharp curves. Detailed information on the extraction of sharp curves and the subsequent generation of the sharp feature surface \(M\) can be found in Appendix~\ref{sec:feature_surface_mesh}.

We learn to approximate a signed offset field (Appendix~\ref{sec:offset}) instead of a strict SDF in these CAD applications, because the feature surface $M$ extends onto both the convex and concave sides of the CAD sharp curves. This ensures that the reconstructed CAD sharp curves intersect with $M$. Although these two formulations coincide for smooth surfaces, they differ when sharp features are present. In particular, offset surfaces preserve the same sharp geometric structures on both sides of the zero-level set, whereas a conventional SDF exhibits ``rounded'' behavior around sharp features when observed from the convex side. Moreover, because the signed offset field also fulfills the Eikonal condition ($\norm{\nabla F} = 1$), it can be trained with standard SDF loss objectives without any modification. For convenience, we still refer to it as SDF.

In this experiment, the features \(M\) remain invariant throughout the training process. We train {\sharpnet} using common SDF learning losses.
\begin{equation}
\label{eqn:3Dloss_mesh}
    \mathcal{L}(\theta) =
        \alpha_{\text{sur}} \cdot \mathcal{L}_{\text{sur}} +
        \alpha_{\text{ext}} \cdot \mathcal{L}_{\text{ext}} +
        \alpha_{\text{ekl}} \cdot \mathcal{L}_{\text{ekl}} +
        \alpha_{\text{nor}} \cdot \mathcal{L}_{\text{nor}},
\end{equation}
where $\mathcal{L}_{\text{sur}}$, $\mathcal{L}_{\text{ext}}$, $\mathcal{L}_{\text{ekl}}$ and $\mathcal{L}_{\text{nor}}$ are
\begin{gather*}
    \mathcal{L}_{\text{sur}}(\theta) = \frac{1}{r}\sum_{i=1}^r \abs*{\Phi_\theta(\mathbf{x}_i, \mathfrak{f}(\mathbf{x}_i))} \\
    \mathcal{L}_{\text{ext}}(\theta) = \frac{1}{s}\sum_{i=1}^s \exp(-\alpha \abs*{\Phi_\theta(\mathbf{x}_i, \mathfrak{f}(\mathbf{x}_i))}) \\
    \mathcal{L}_{\text{nor}}(\theta)=  \frac{1}{r}\sum_{i=1}^r \norm*{ \nabla\Phi_\theta(\mathbf{x}_i, \mathfrak{f}(\mathbf{x}_i))-\mathbf{n}_i}\\
    \mathcal{L}_{\text{ekl}}(\theta) = \frac{1}{t}\sum_{i=1}^t \abs{ 1-\norm{\nabla\Phi_\theta(\mathbf{x}_i, \mathfrak{f}(\mathbf{x}_i))}}.
\end{gather*}
Because {\nhrep} requires oriented normals as constraints, we also incorporate normals. Here, \(\mathbf{n}_i\) denotes the oriented normal associated with the point \(\mathbf{x}_i\). 

At each training epoch, the surface loss \(\mathcal{L}_{\text{sur}}\) and the surface-normal loss \(\mathcal{L}_{\text{nor}}\) are evaluated at surface sampling points. The exterior loss \(\mathcal{L}_{\text{ext}}\), which enforces the network output to stay away from zero far from the surface, is applied to ambient points. For the Eikonal loss \(\mathcal{L}_{\text{ekl}}\), we use both surface samples and near-surface samples as inputs. The corresponding weight parameters are set to \(\alpha_{\text{sur}}=7000\), \(\alpha_{\text{ext}}=600\) and \(\alpha_{\text{ekl}}=50\).

\begin{figure}
    \centering
    \includegraphics[width=0.28\linewidth]{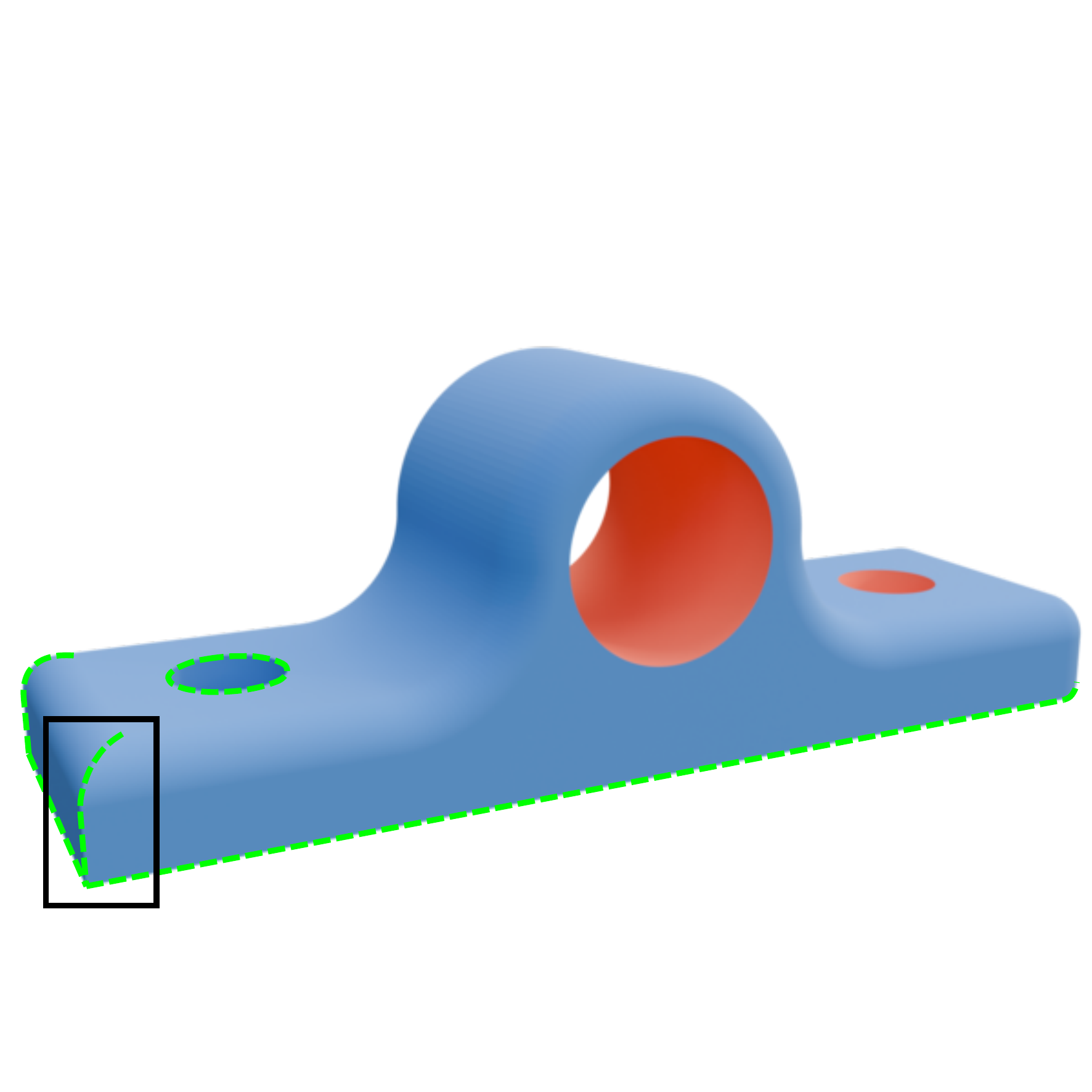}
    \includegraphics[width=0.23\linewidth]{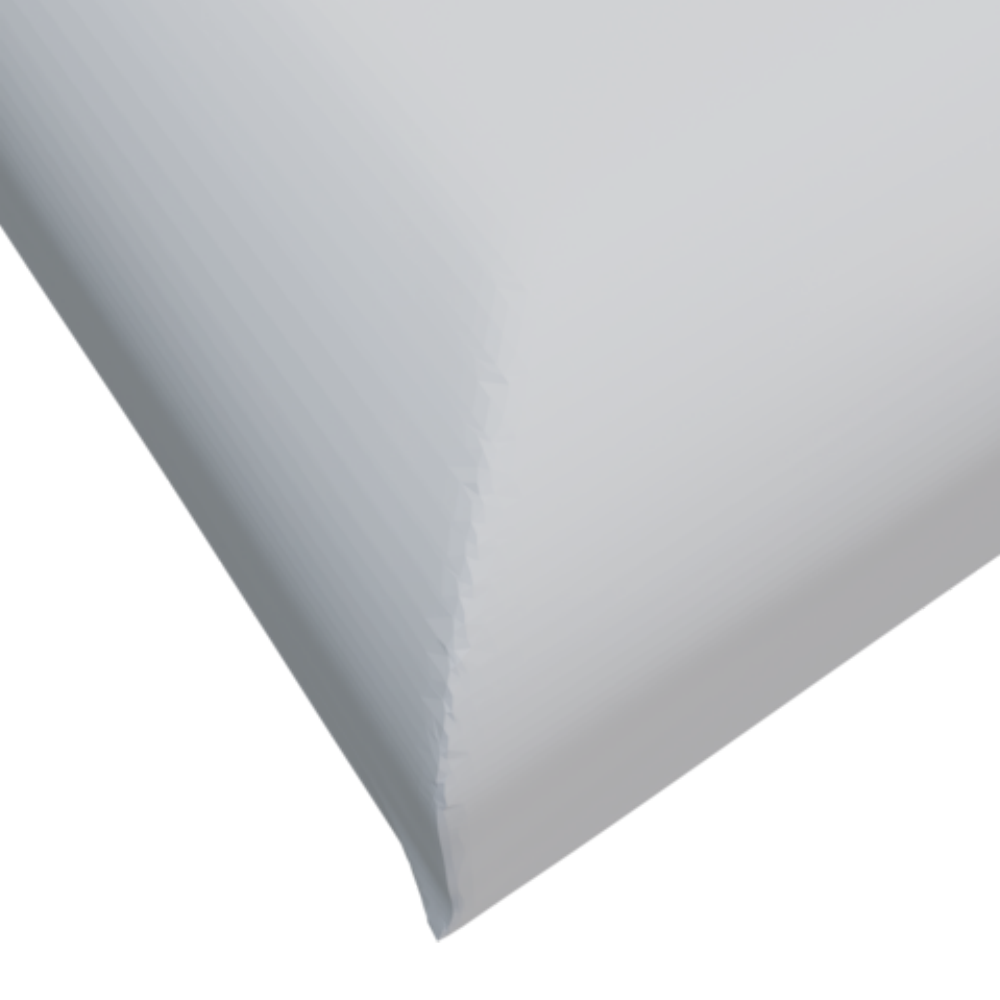}
    \includegraphics[width=0.23\linewidth]{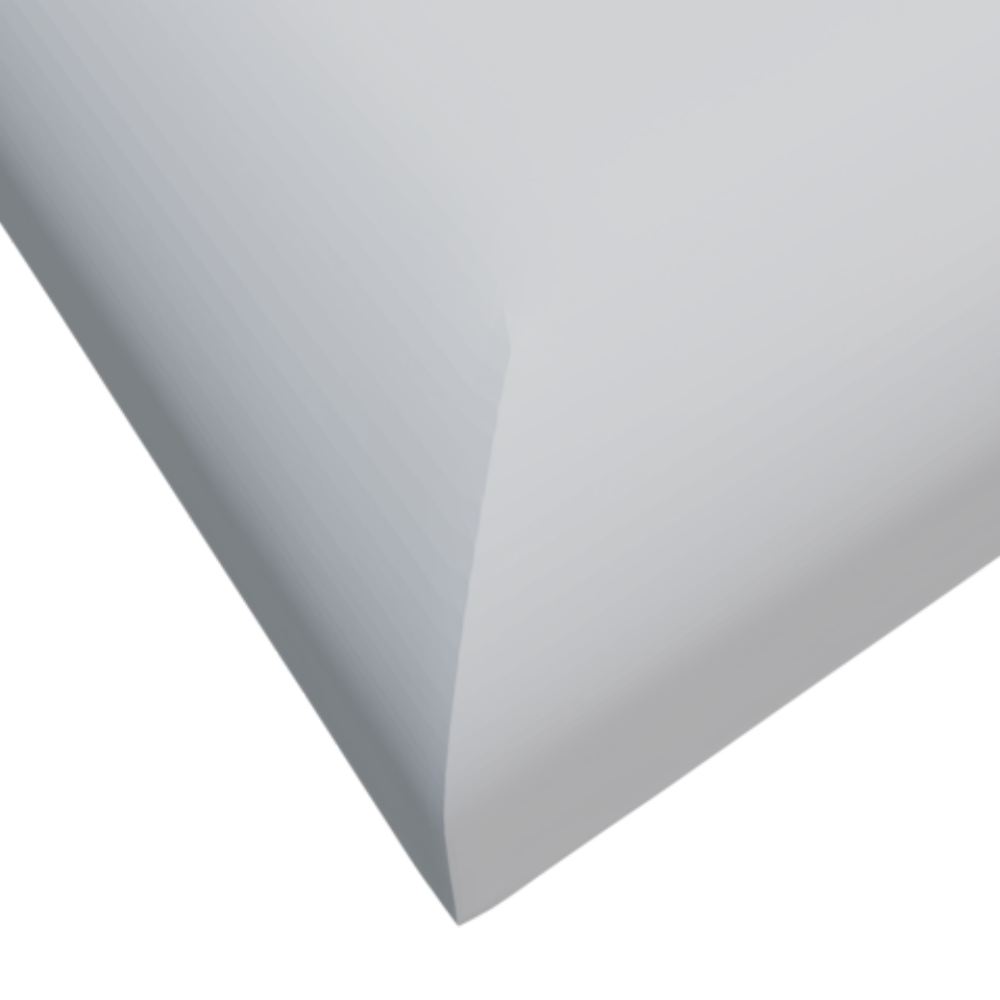}
    \includegraphics[width=0.23\linewidth]{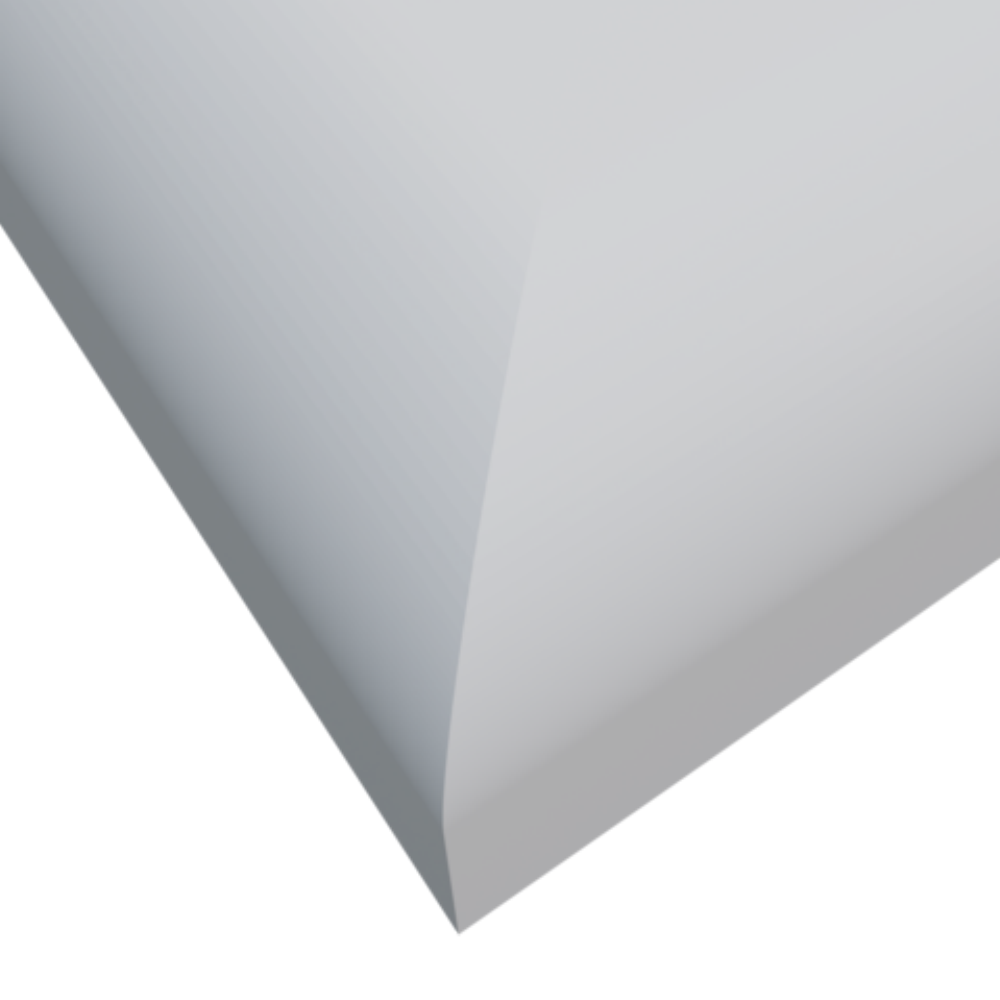}\\
    \includegraphics[width=0.28\linewidth]{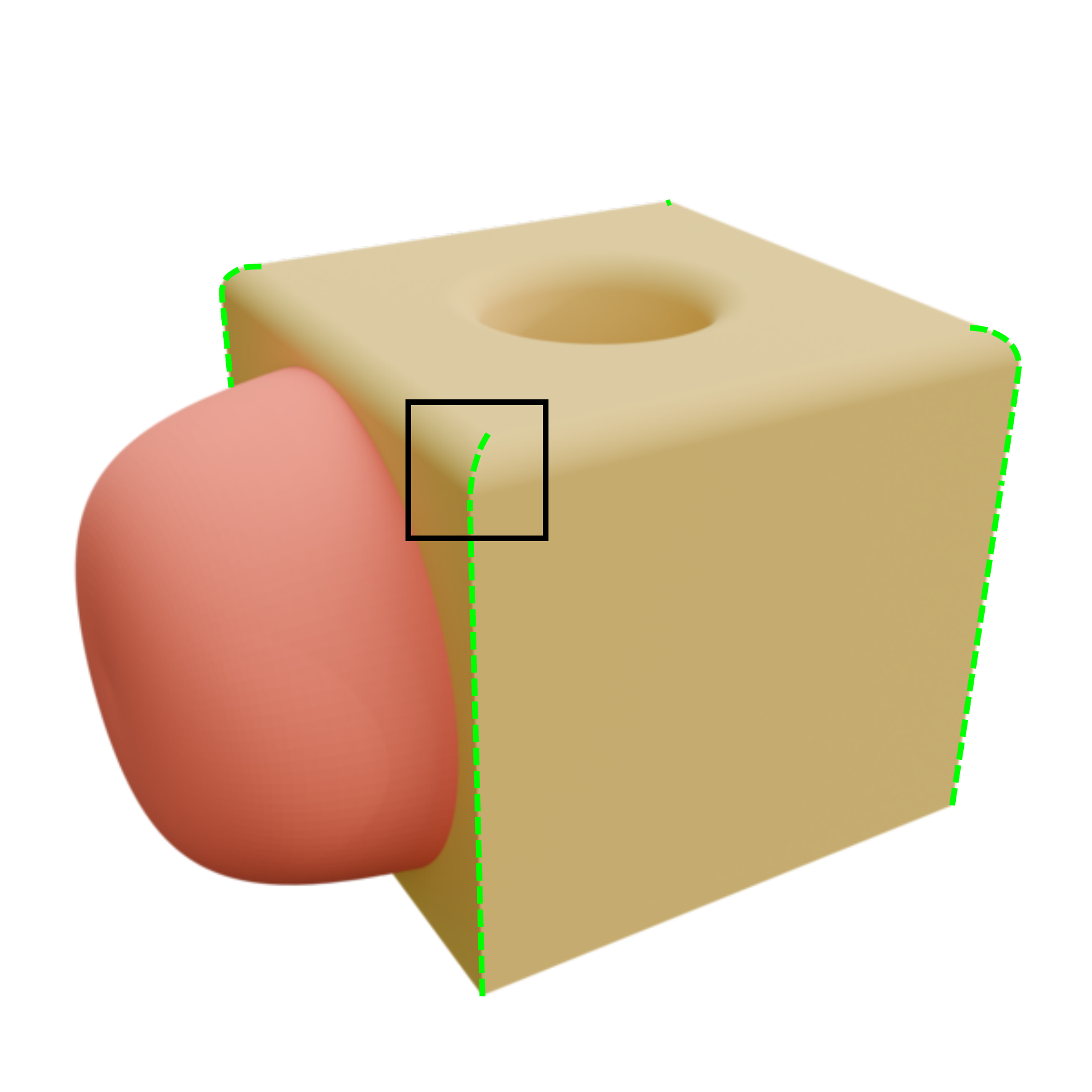}
    \includegraphics[width=0.23\linewidth]{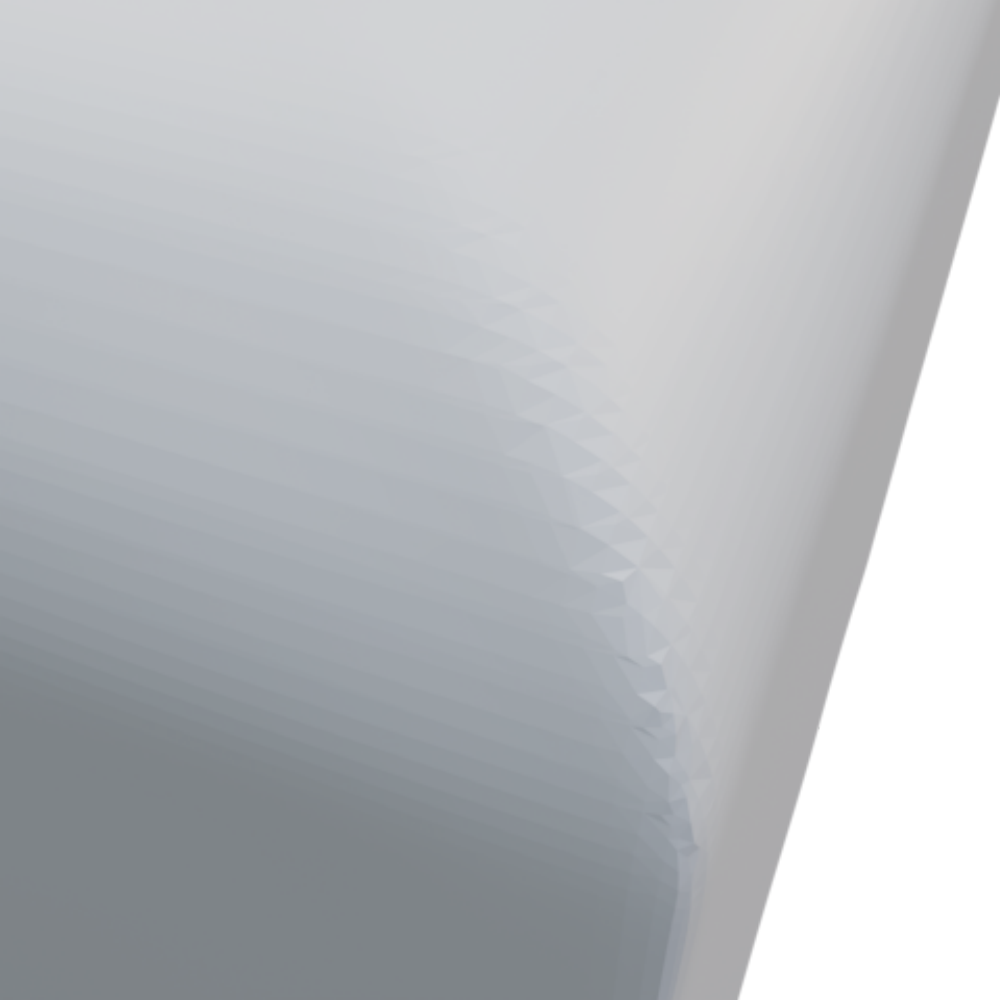}
    \includegraphics[width=0.23\linewidth]{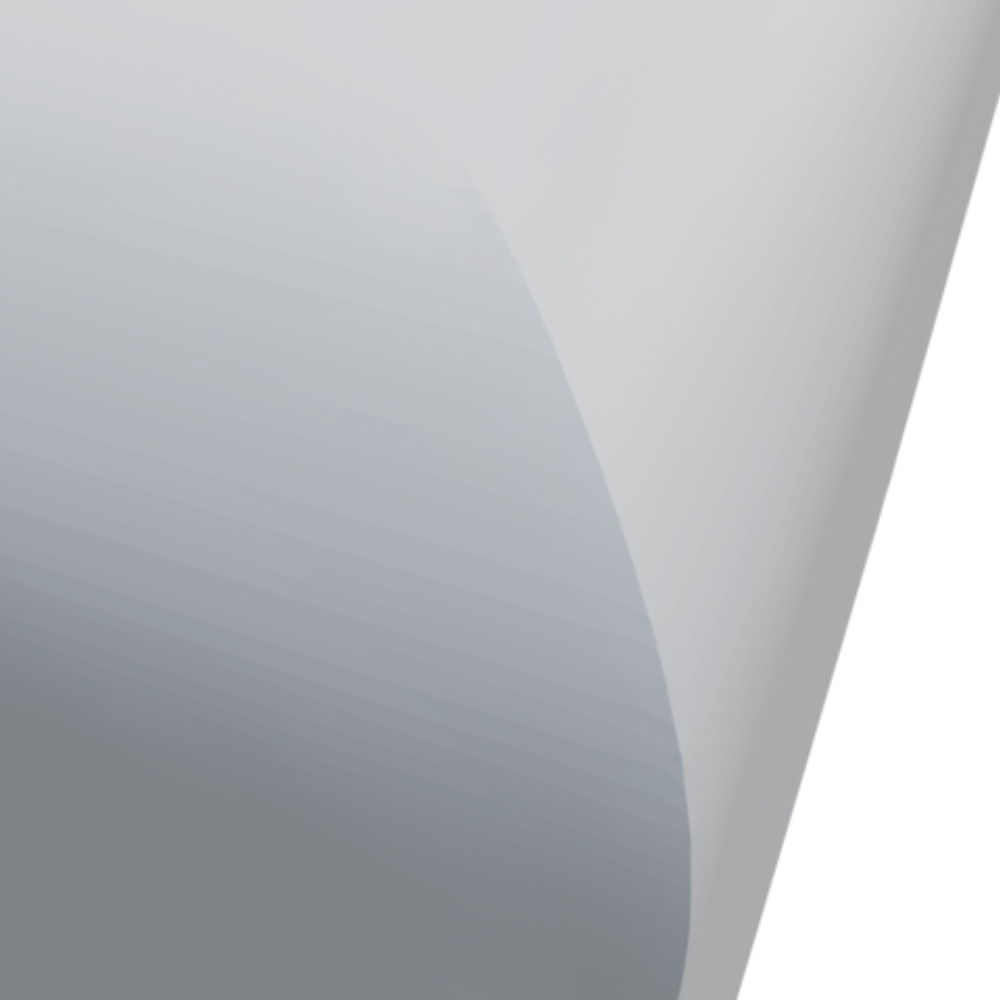}
    \includegraphics[width=0.23\linewidth]{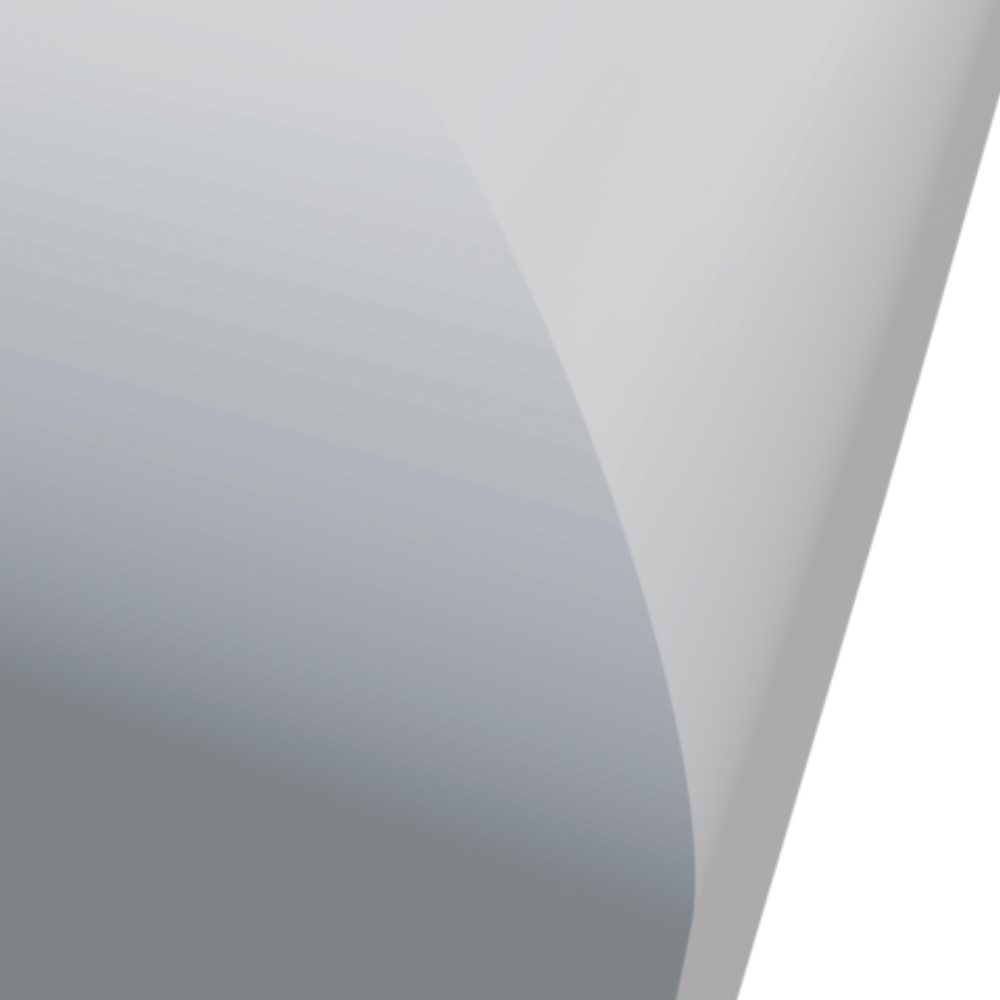}\\
    \makebox[0.28\linewidth]{\sffamily Patches}
    \makebox[0.23\linewidth]{\sffamily {\nhrep}}
    \makebox[0.23\linewidth]{\sffamily {\sharpnet}}
    \makebox[0.23\linewidth]{\sffamily GT}
    \caption{{\nhrep} is unable to reconstruct open sharp feature curves due to its patch subdivision strategy. Dashed green curves are the sharp curves. The reconstruction results of {\nhrep}, {\sharpnet}, and the ground truth for the region indicated illustrate that the sharp curves of {\nhrep} are smoothed.}
    \label{fig:open_feature}
\end{figure}

\begin{figure}
\centering
\newlength{\uniformheight}
\setlength{\uniformheight}{.26\linewidth}
\begin{subcaptionblock}{.26\linewidth}
    \begin{minipage}[c][\uniformheight][c]{\linewidth}
    \includegraphics[width=\linewidth]{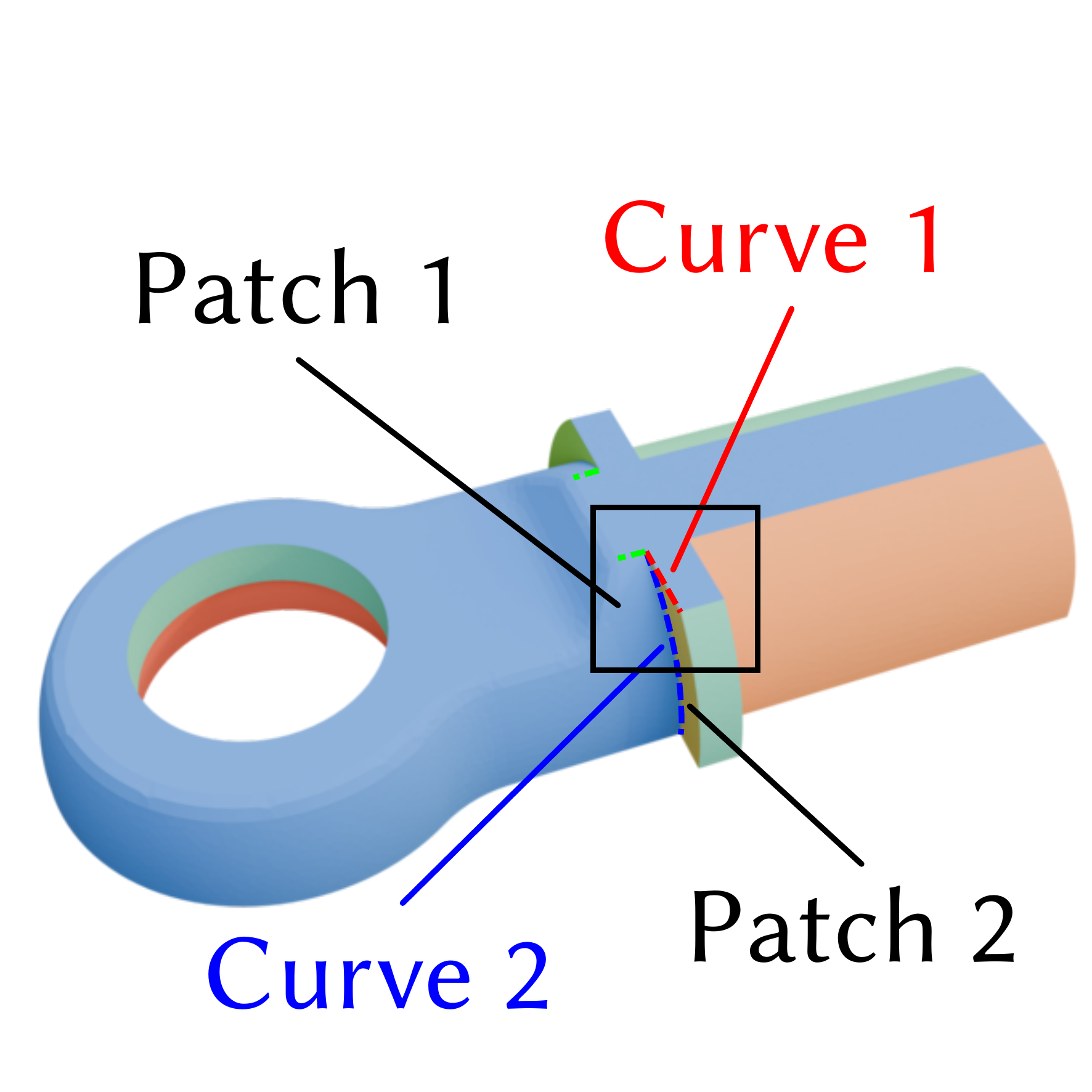}
    \end{minipage}
    \caption{Patches}
    \label{fig:open_feature_fail (patches)}
\end{subcaptionblock}\hfill%
\begin{subcaptionblock}{.23\linewidth}
    \begin{minipage}[c][\uniformheight][c]{\linewidth}
    \centering {\sffamily N.A.}
    \end{minipage}
    \caption{{\nhrep}}
    \label{fig:open_feature_fail (nhrep)}
\end{subcaptionblock}\hfill%
\begin{subcaptionblock}{.24\linewidth}
    \begin{minipage}[c][\uniformheight][c]{\linewidth}
    \includegraphics[width=\linewidth]{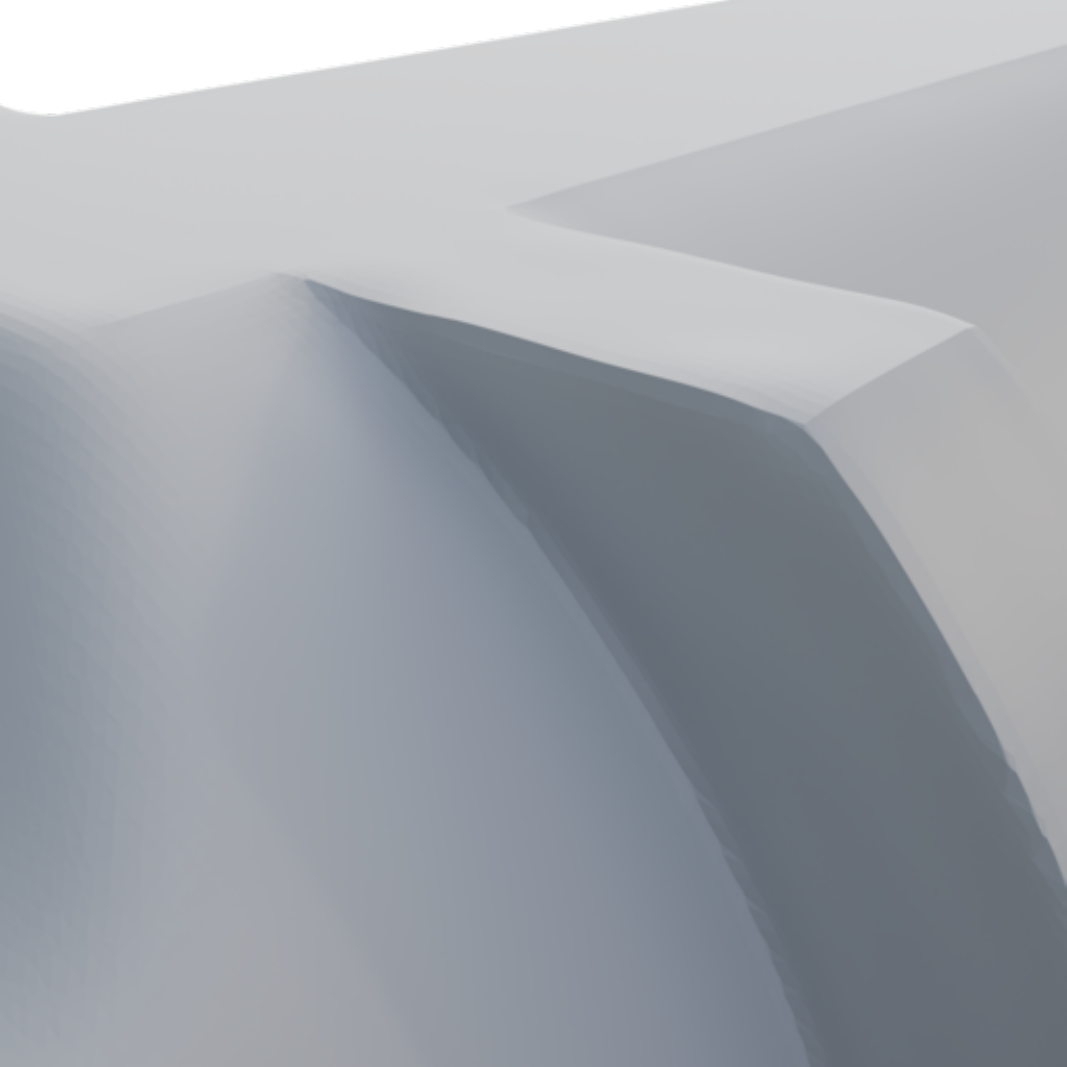}
    \end{minipage}
    \caption{{\sharpnet}}
    \label{fig:open_feature_fail (ours)}
\end{subcaptionblock}\hfill%
\begin{subcaptionblock}{.24\linewidth}
    \begin{minipage}[c][\uniformheight][c]{\linewidth}
    \includegraphics[width=\linewidth]{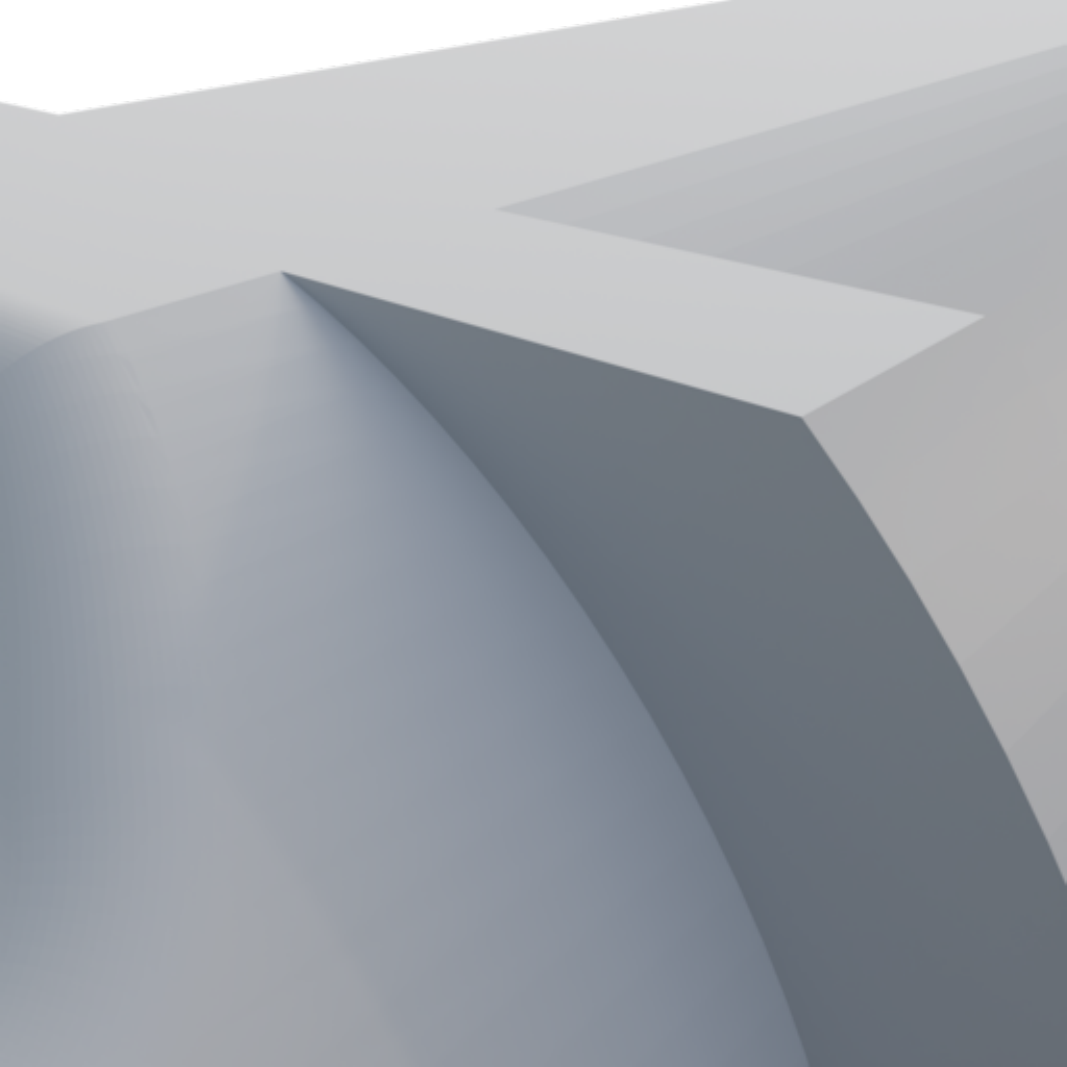}
    \end{minipage}
    \caption{GT}
    \label{fig:open_feature_fail (gt)}
\end{subcaptionblock}
\caption{CSG tree construction of {\nhrep} may fail in the presence of open sharp feature curves.
(\subref{fig:open_feature_fail (patches)})~Due to the presence of open feature curves,  subdivision causes some of them to disappear in {\nhrep}, as indicated in green dashed line. The boundary between Patch~1 and Patch~2 consists of both a convex sharp curve (Curve~1, red dashed line) and a concave sharp curve (Curve~2, blue dashed line), which makes the CSG min/max operation ambiguous.
(\subref{fig:open_feature_fail (nhrep)})~As a result, {\nhrep} does not produce valid outputs.
(\subref{fig:open_feature_fail (ours)})~and~(\subref{fig:open_feature_fail (gt)})~show the reconstruction results of {\sharpnet} and the ground truth for the boxed regions.}
\label{fig:open_feature_fail}
\end{figure}

\paragraph{Non-closed sharp feature curves} {\nhrep}~\cite{Guo2022NH-Rep} relies heavily on the assumption that the feature curves are closed, enabling the model to be decomposed into patches and sharp features to be reconstructed through patch intersections. However, in practice, sharp feature curves are not always closed, as shown in Figure~\ref{fig:open_feature}. When a patch includes such open sharp curves, it is left unsubdivided, which leads to the loss of the associated sharp features. More critically, as shown in Figure~\ref{fig:open_feature_fail}, open curves can generate a patch whose boundary includes a mix of concave and convex segments. In this situation, the min/max evaluation is no longer well-defined, and the construction of the CSG tree breaks down. Consequently, {\nhrep} successfully processes only 65 out of 100 models (with some of the failures arising from other causes).

\begin{figure}
\centering
\setlength{\tabcolsep}{2pt}
\begin{tabular}{ccccc}
    \raisebox{0.11\linewidth}{\rotatebox[origin=c]{90}{\sffamily \scriptsize{Complex Feature}}} &
    \includegraphics[width=0.22\linewidth]{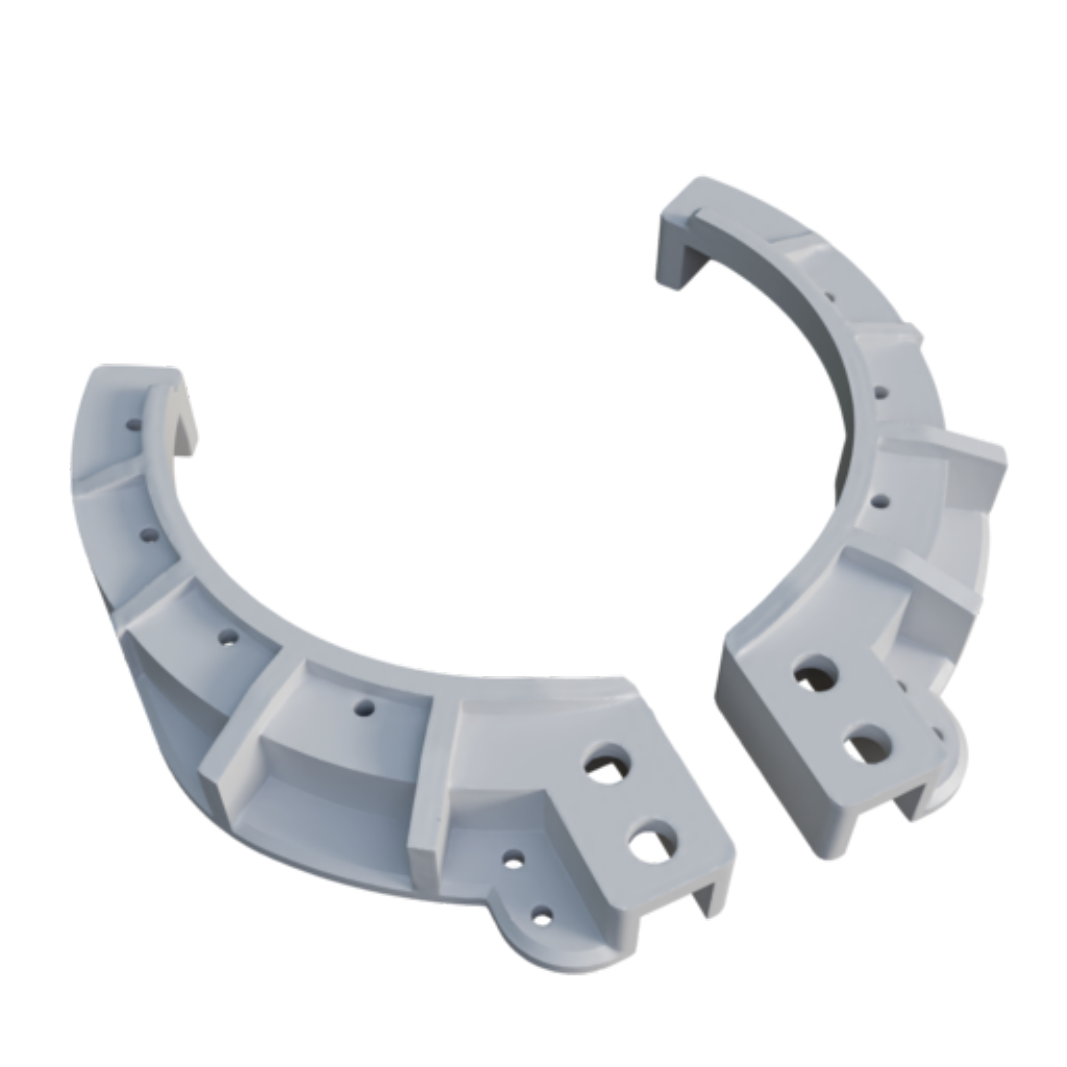} &
    \includegraphics[width=0.22\linewidth]{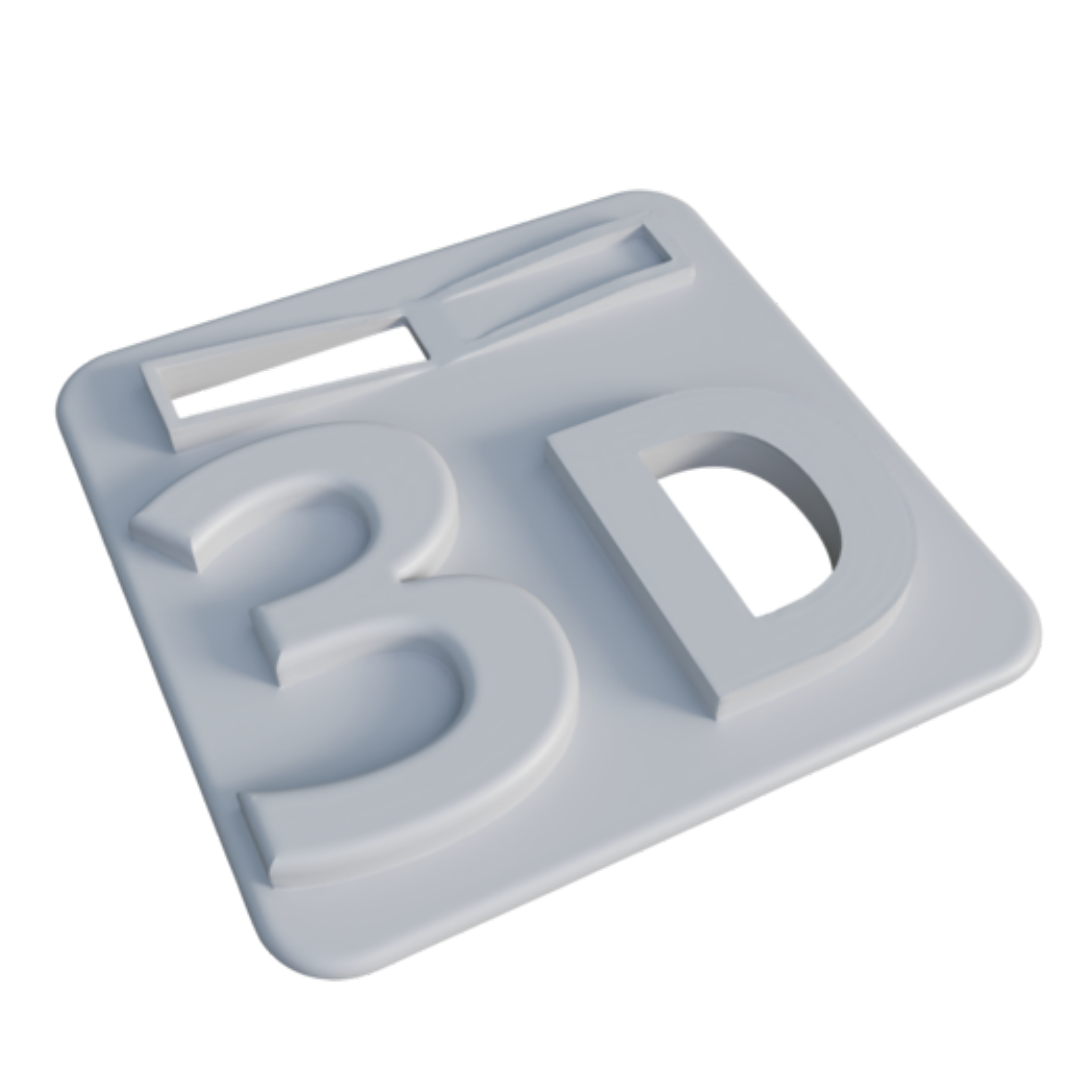} &
    \includegraphics[width=0.22\linewidth]{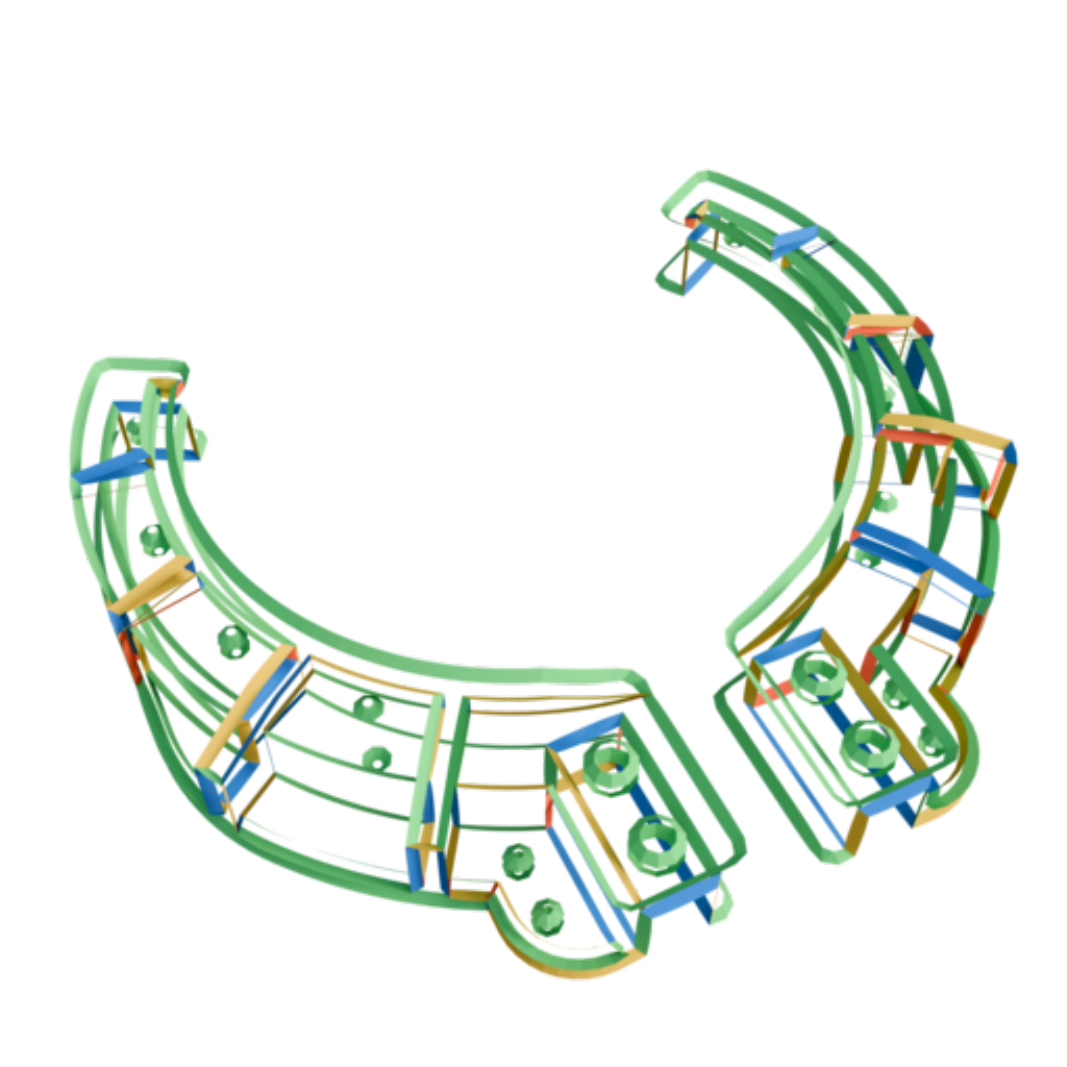} &
    \includegraphics[width=0.22\linewidth]{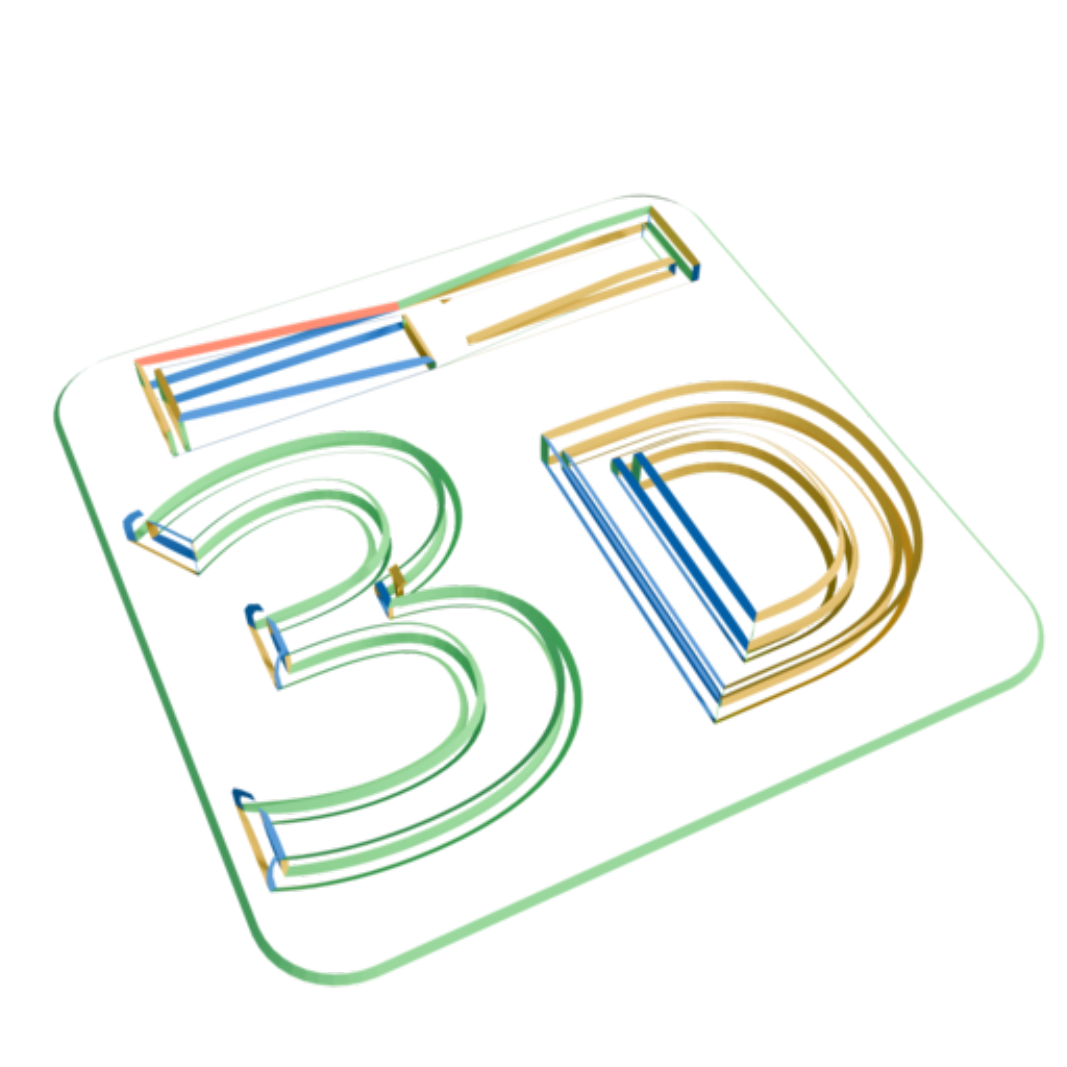}\\
    \raisebox{0.11\linewidth}{\rotatebox[origin=c]{90}{\sffamily\scriptsize{Thin Structure}}} &
    \includegraphics[width=0.22\linewidth]{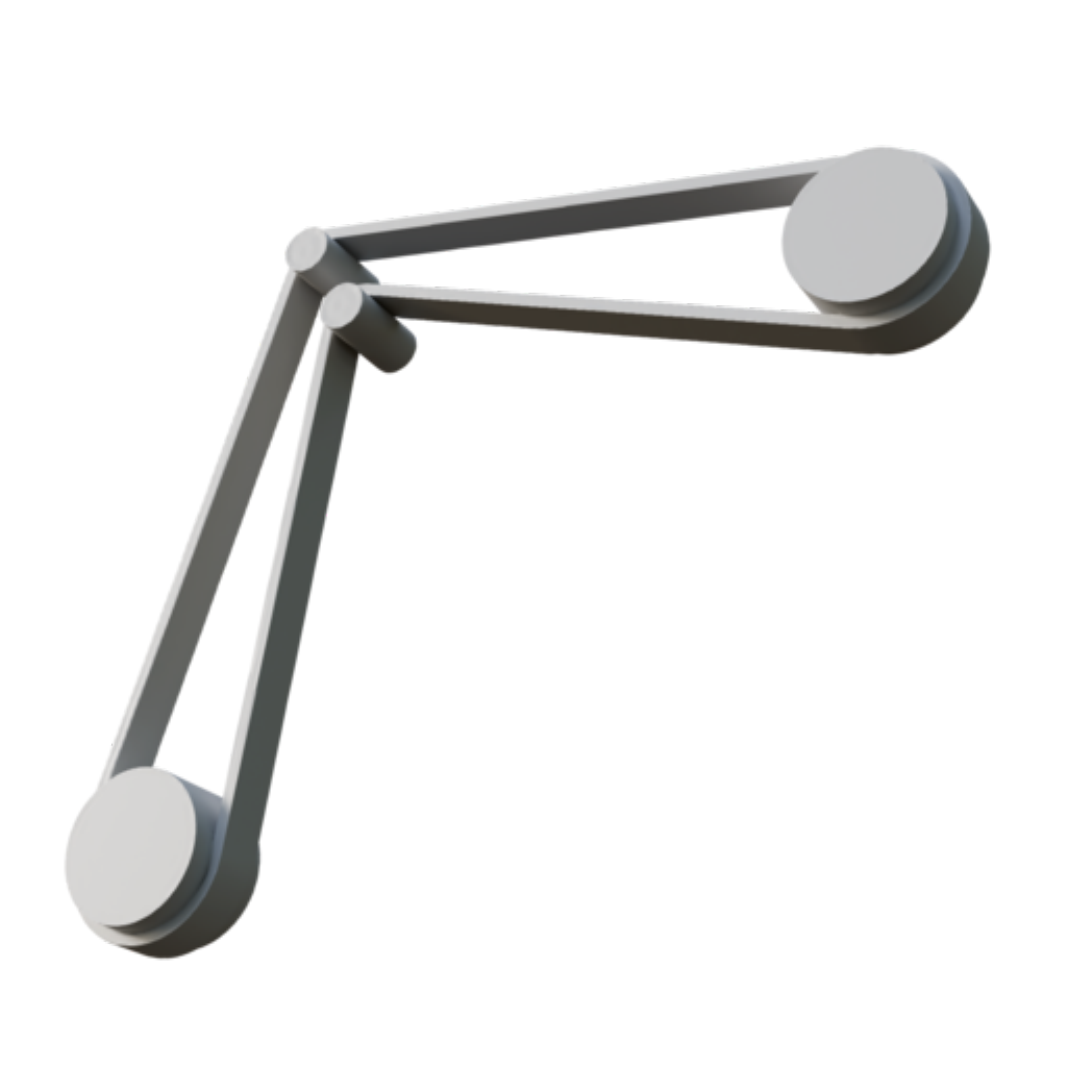} &
    \includegraphics[width=0.22\linewidth]{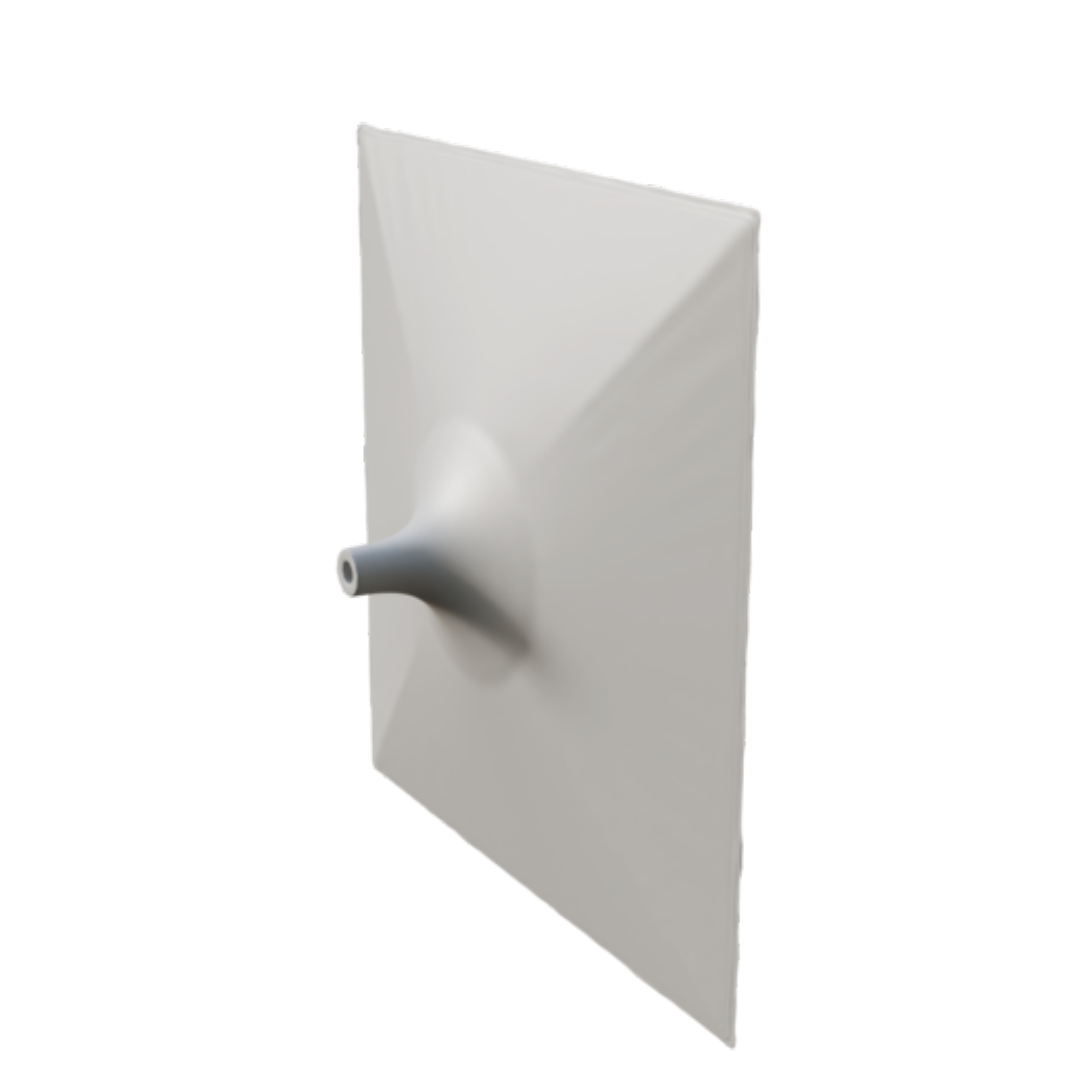} &
    \includegraphics[width=0.22\linewidth]{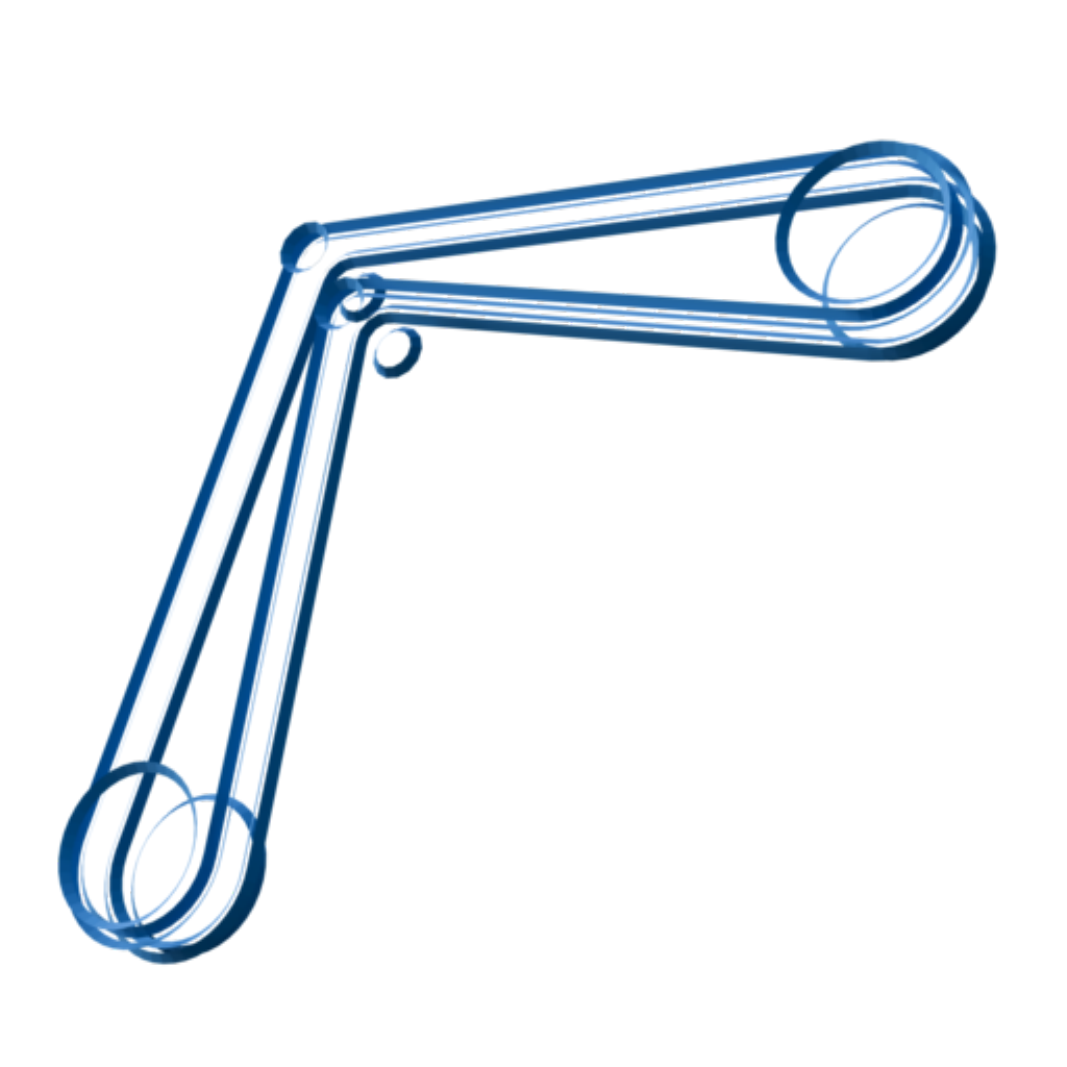} &
    \includegraphics[width=0.22\linewidth]{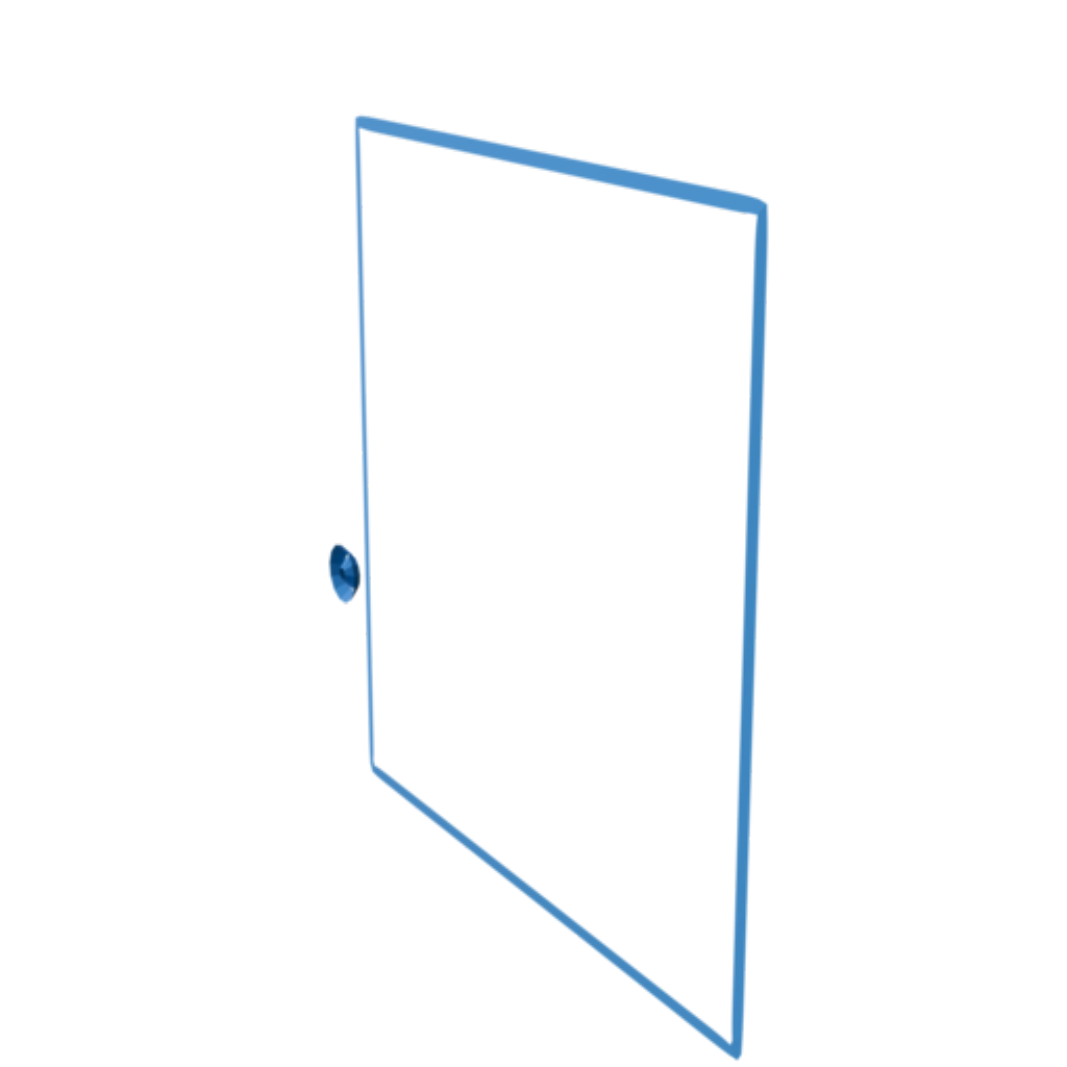}\\
    \raisebox{0.11\linewidth}{\rotatebox[origin=c]{90}{\sffamily\scriptsize{High Genus}}} &
    \includegraphics[width=0.22\linewidth]{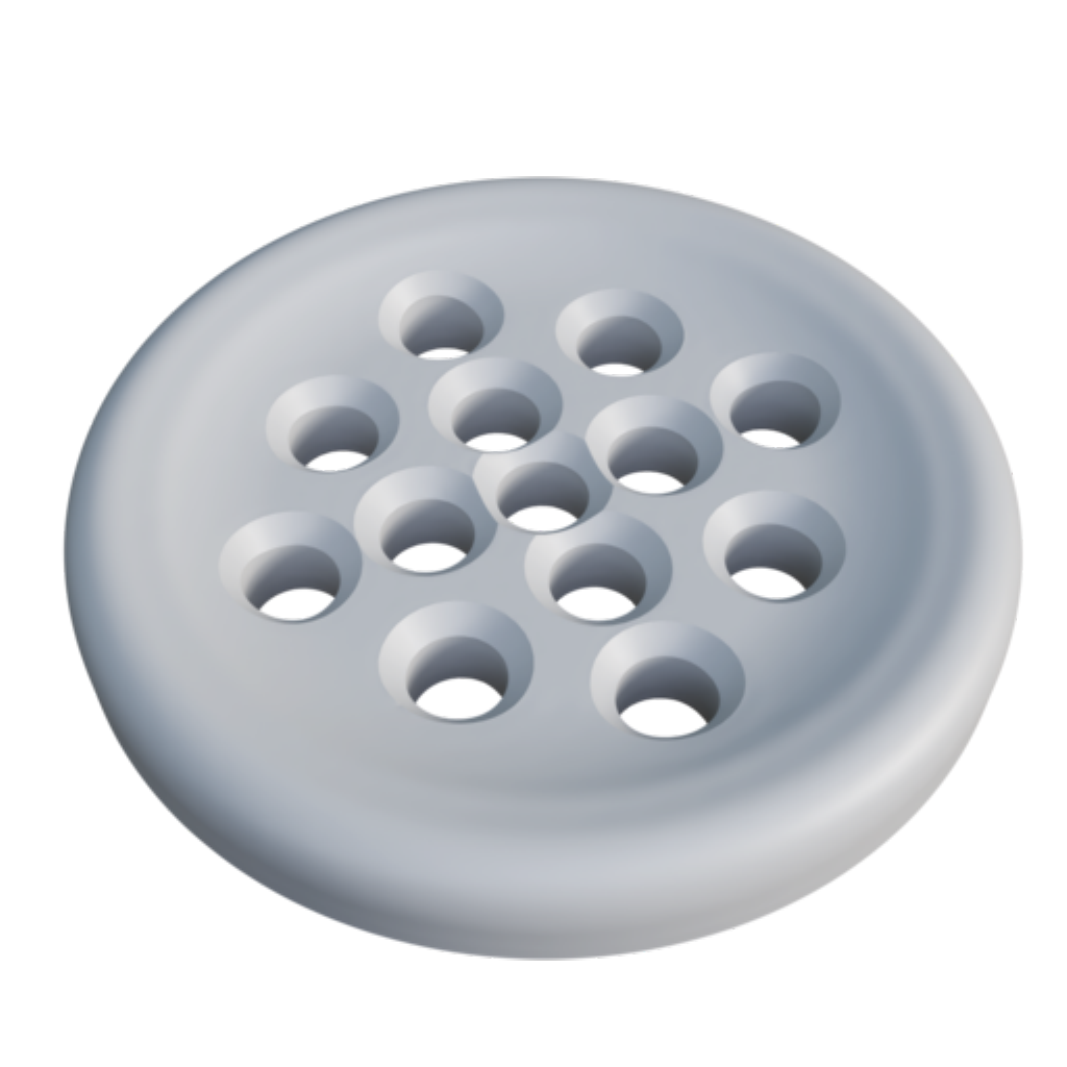} &
    \includegraphics[width=0.22\linewidth]{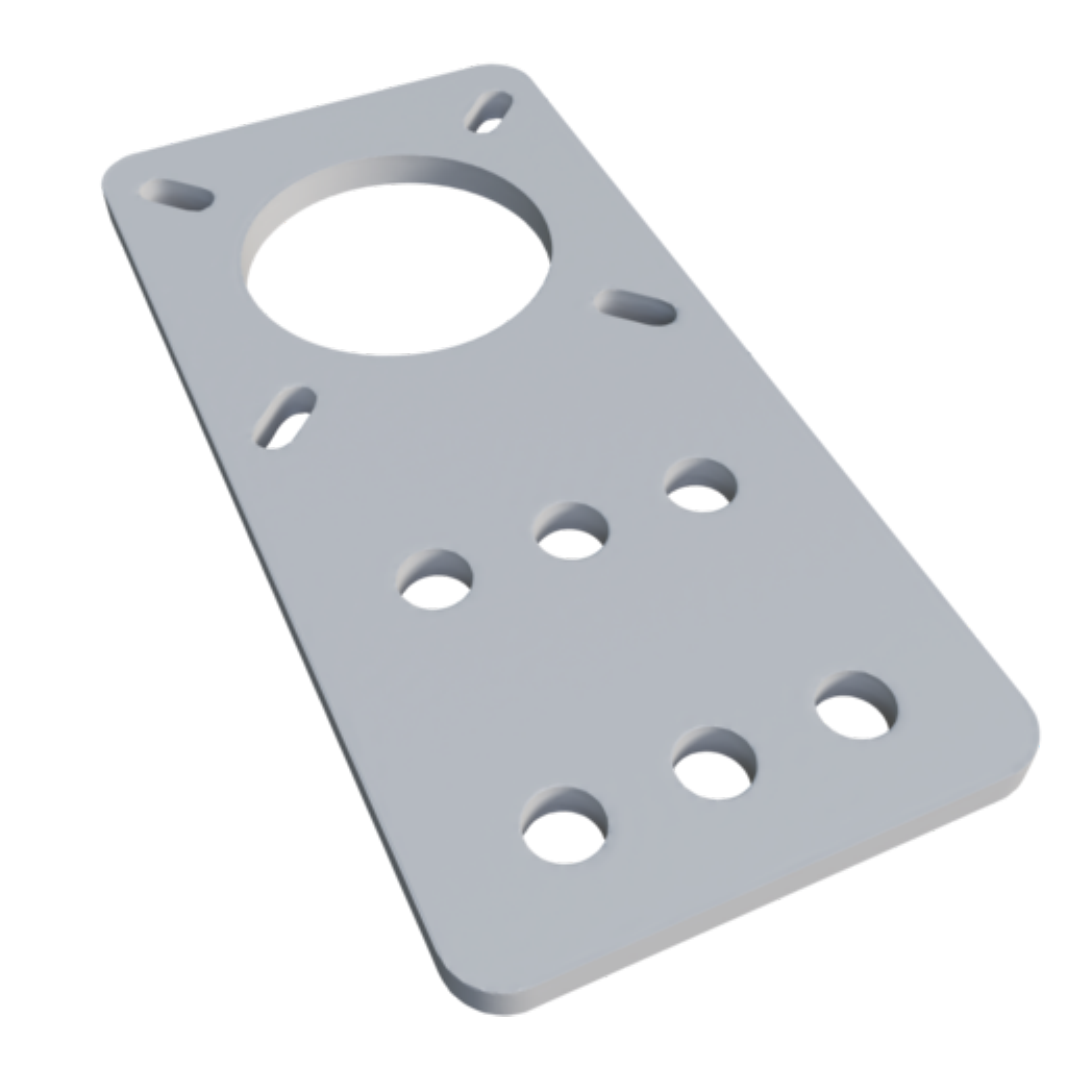} &
    \includegraphics[width=0.22\linewidth]{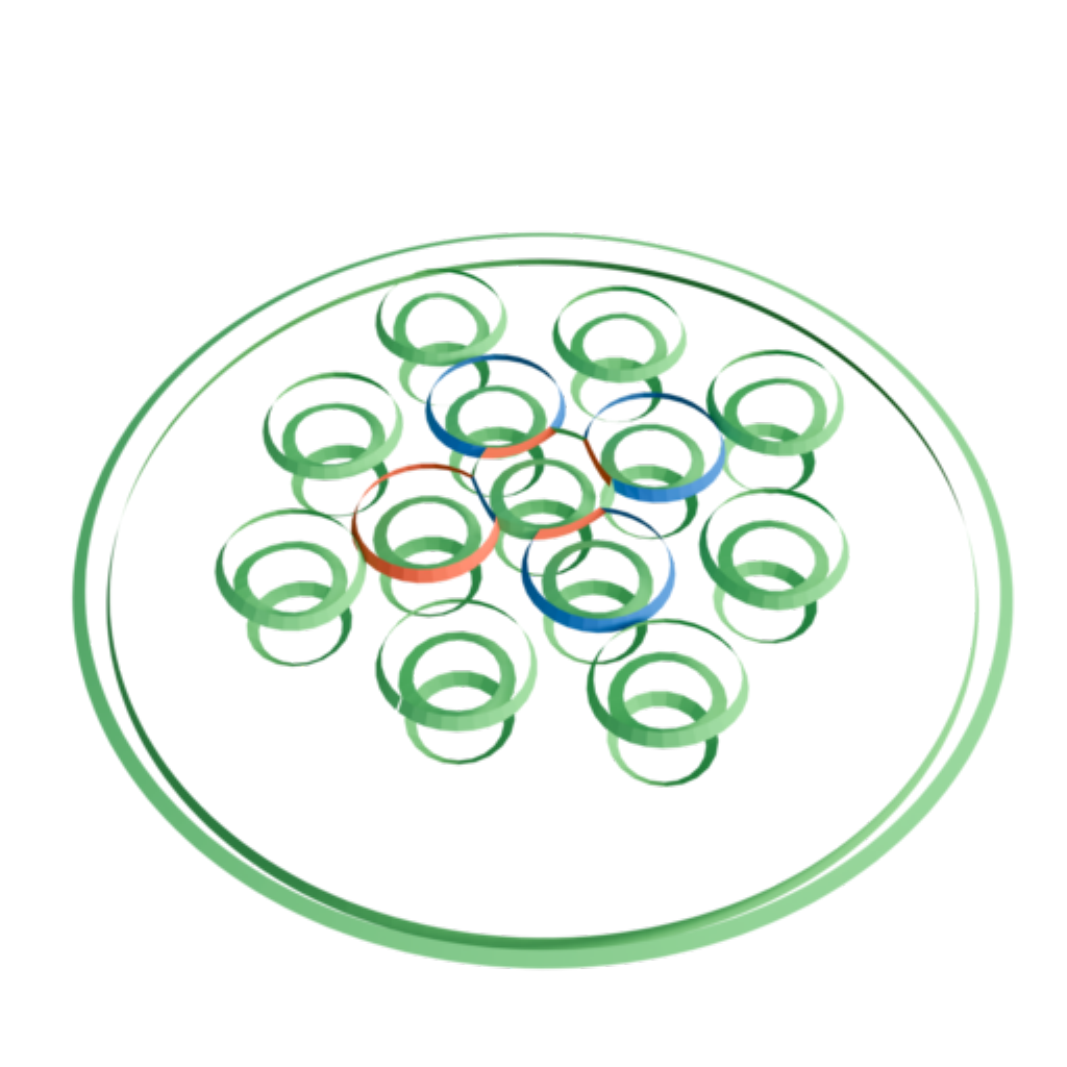} &
    \includegraphics[width=0.22\linewidth]{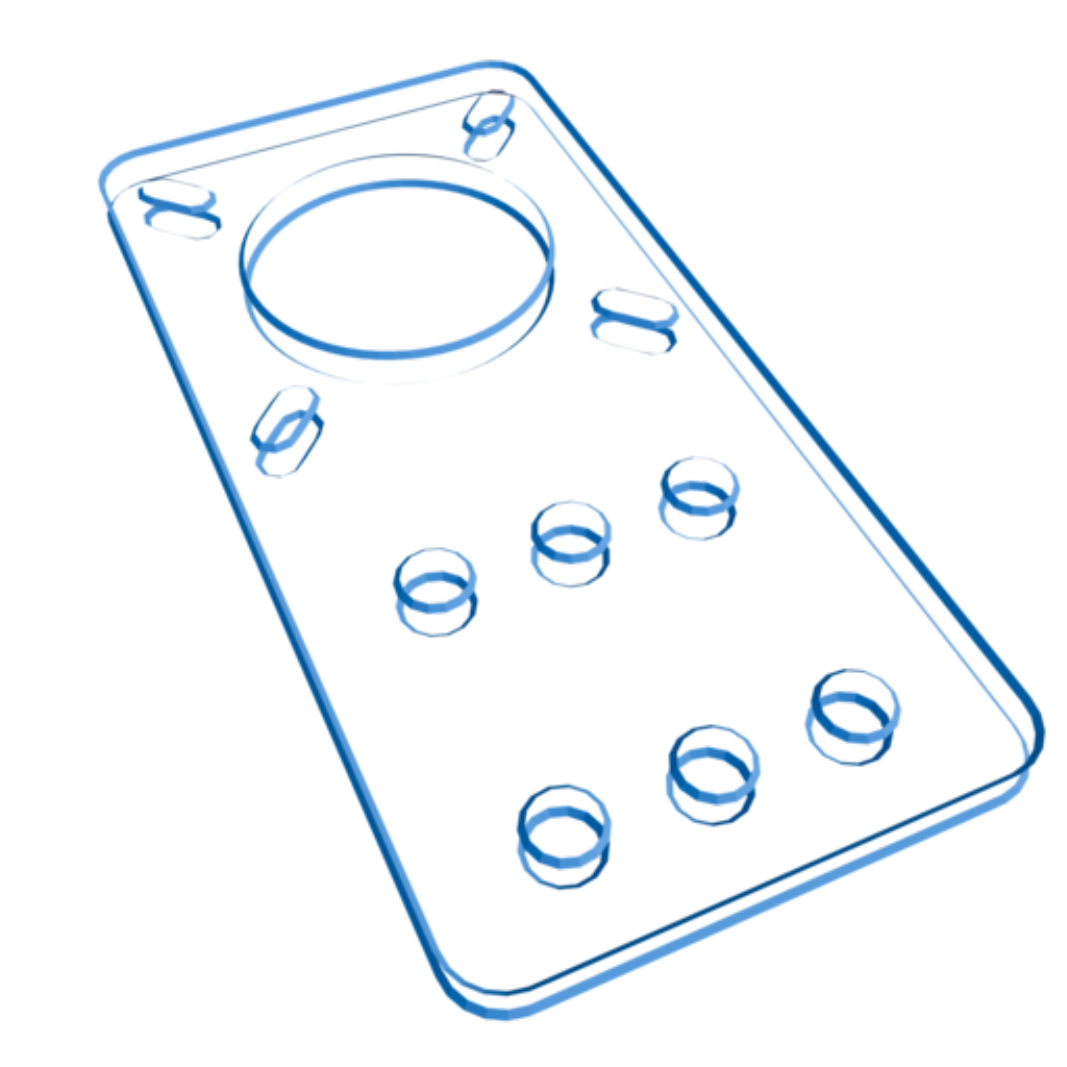}\\
    & \multicolumn{2}{c}{\sffamily Reconstruction} & \multicolumn{2}{c}{\sffamily Feature Set}
\end{tabular}
\caption{{\sharpnet} performance in several challenging cases. The left column shows the reconstruction results, and the right column shows the corresponding sharp feature sets, where each color indicates one dimension of the feature function as defined in Section~\ref{subsec:design}, following the feature-splitting scheme.}
\label{fig:mesh_complex}
\end{figure}

\begin{figure}[!htbp]
\centering
\setlength\tabcolsep{2pt}
\begin{scriptsize}
\begin{tabular}{ccccc}
    \raisebox{0.11\linewidth}{\rotatebox[origin=c]{90}{\sffamily {\nhrep}}} &%
    \includegraphics[width=0.22\linewidth]{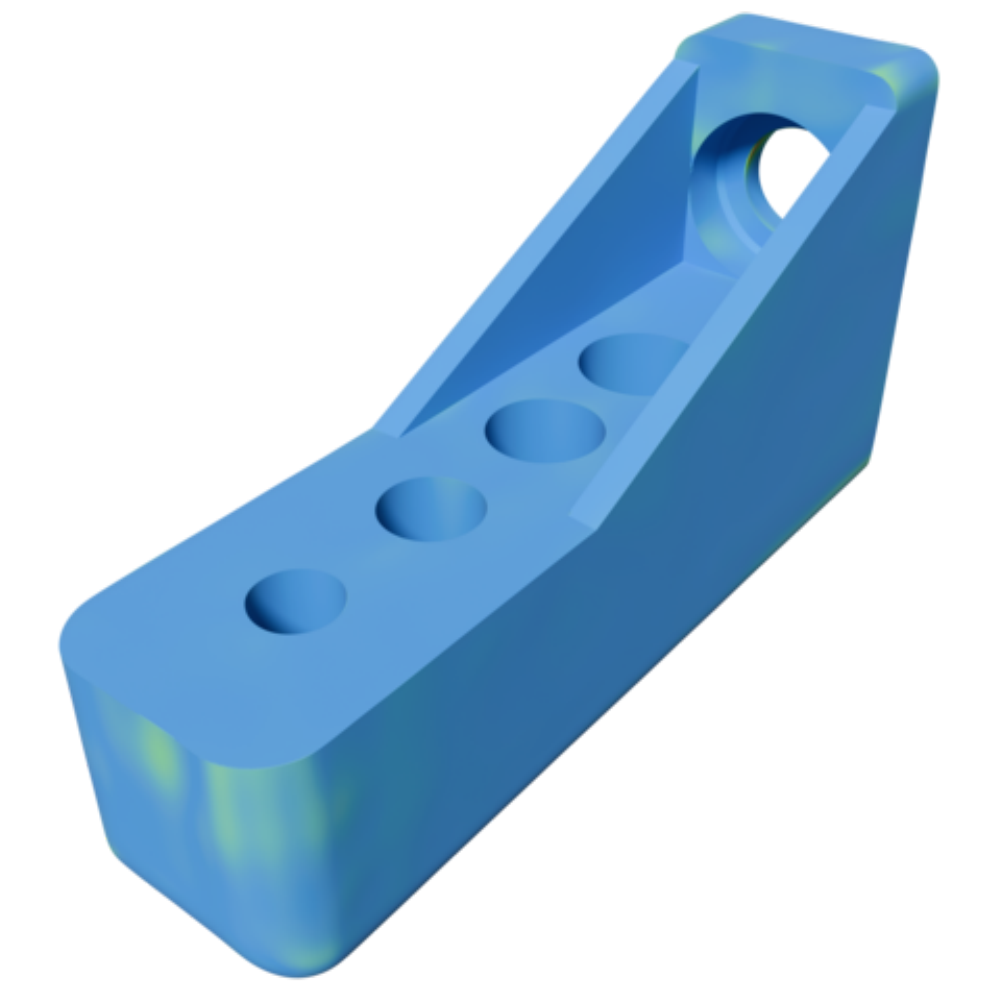} &%
    \includegraphics[width=0.22\linewidth]{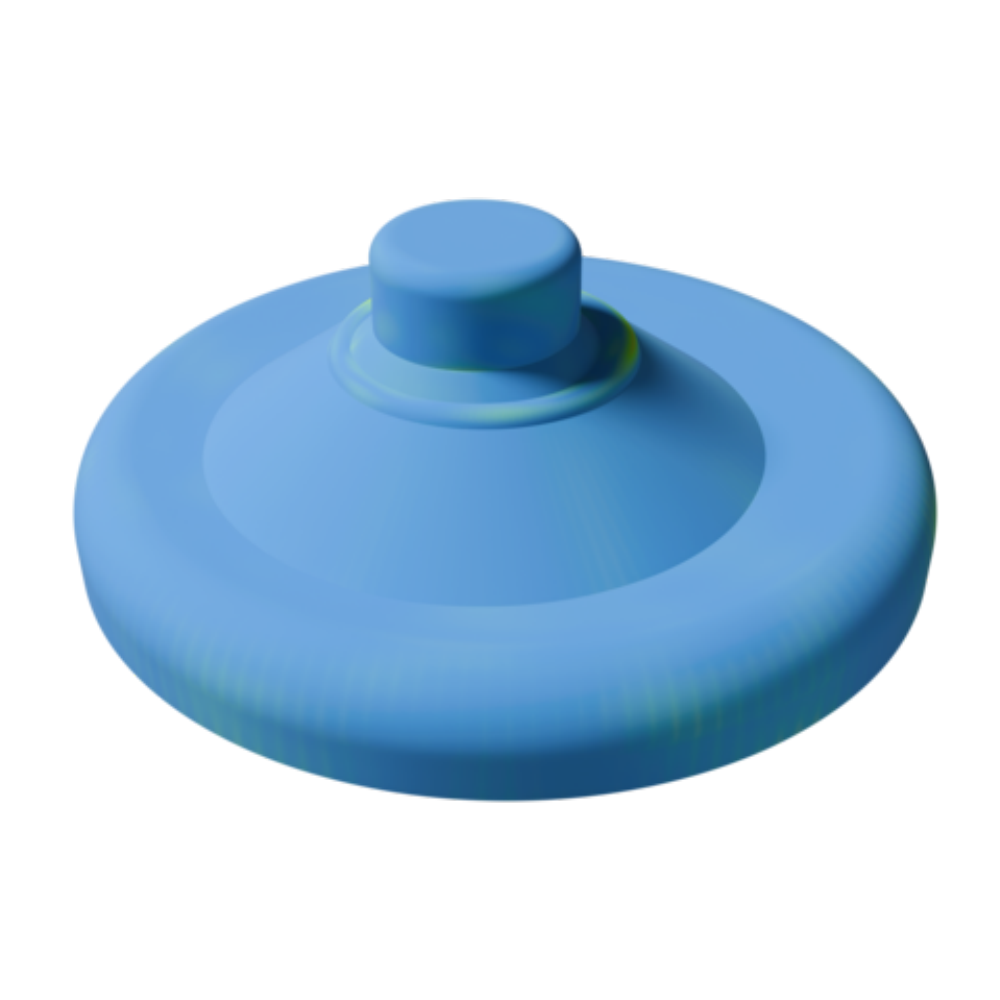} &%
    \includegraphics[width=0.22\linewidth]{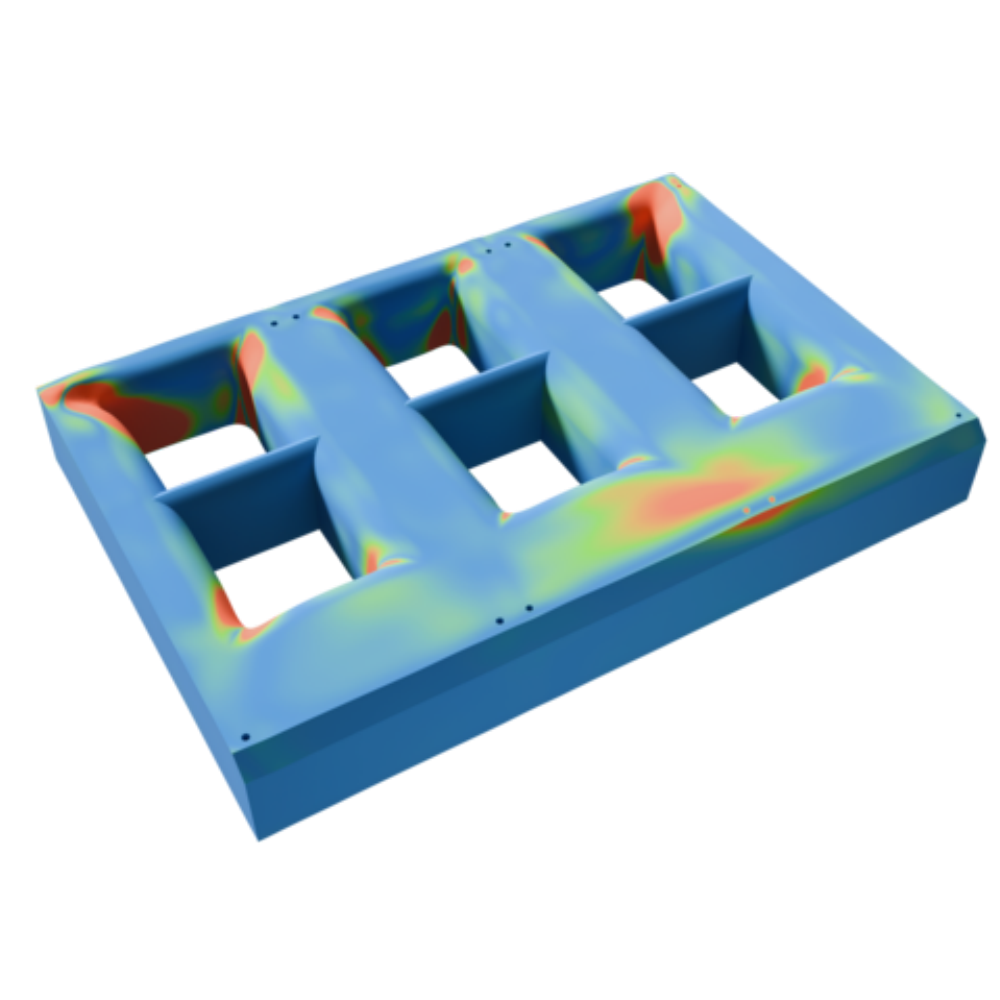} &%
    \includegraphics[width=0.22\linewidth]{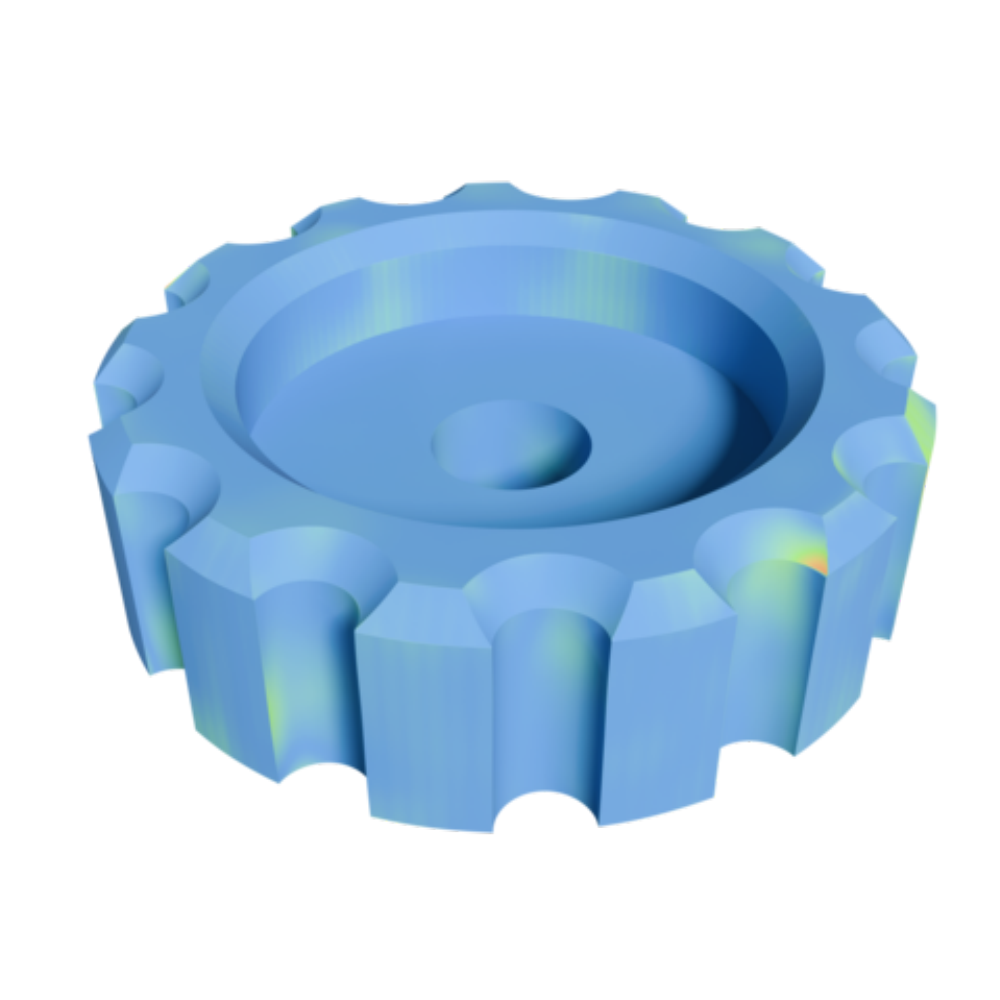} \\
    \makebox[0.01\linewidth]{     } &%
    \makebox[0.22\linewidth]{3.156} &%
    \makebox[0.22\linewidth]{2.914} &%
    \makebox[0.22\linewidth]{3.730} &%
    \makebox[0.22\linewidth]{3.798} \\
    
    \raisebox{0.11\linewidth}{\rotatebox[origin=c]{90}{\sffamily {\sharpnet}}} &%
    \includegraphics[width=0.22\linewidth]{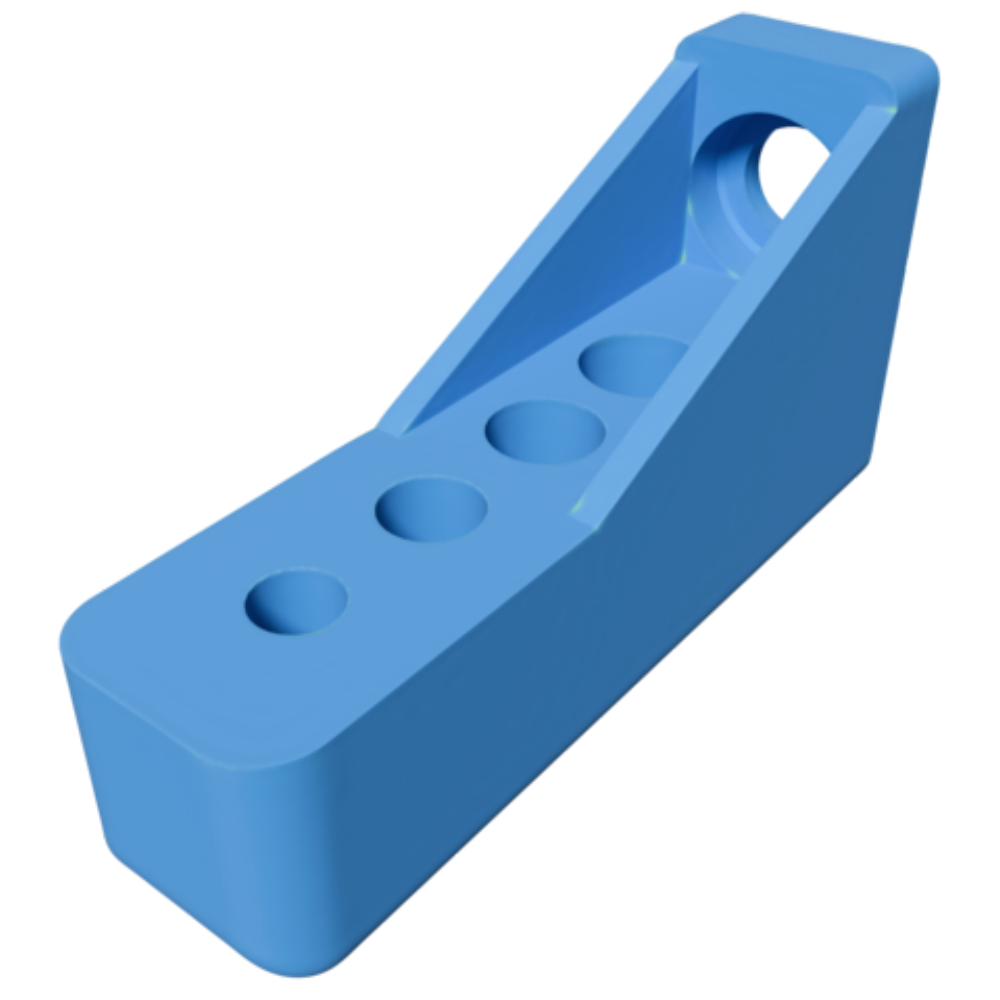} &%
    \includegraphics[width=0.22\linewidth]{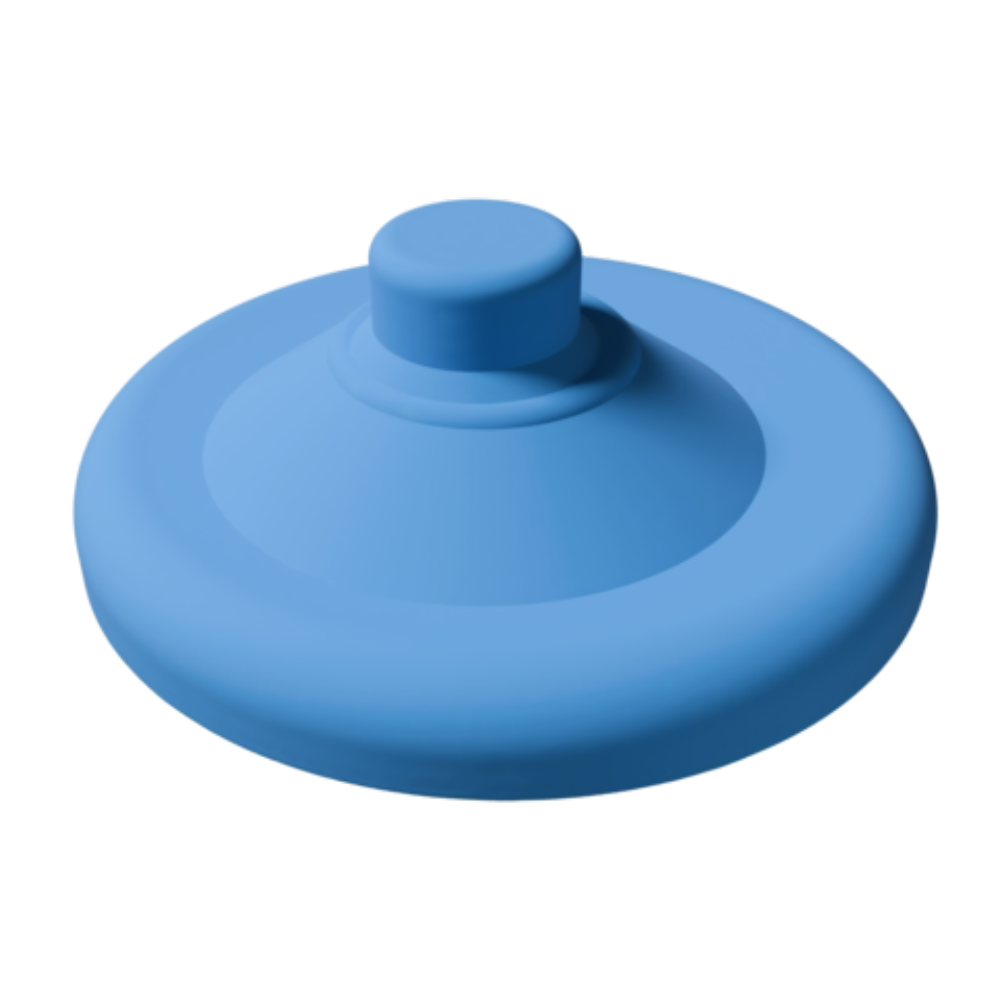} &%
    \includegraphics[width=0.22\linewidth]{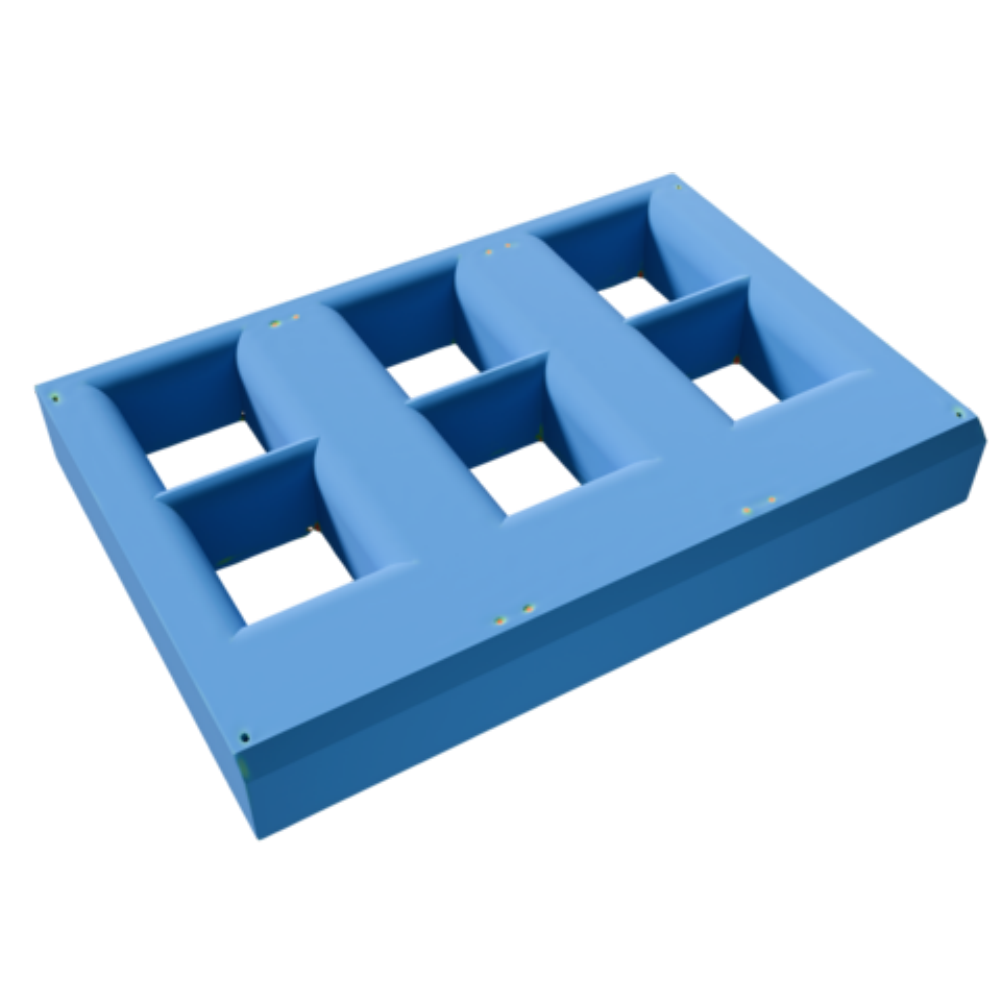} &%
    \includegraphics[width=0.22\linewidth]{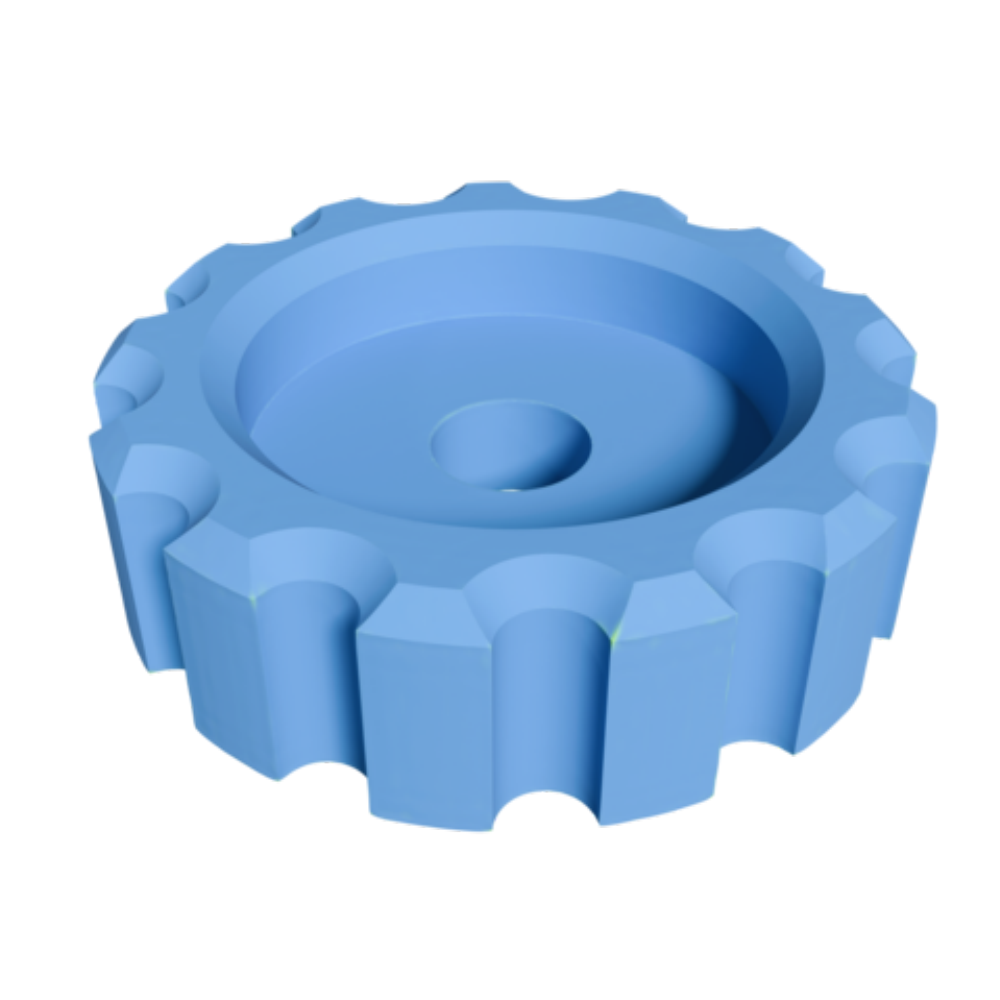} \\
    \makebox[0.01\linewidth]{     } &%
    \makebox[0.22\linewidth]{3.057} &%
    \makebox[0.22\linewidth]{2.590} &%
    \makebox[0.22\linewidth]{3.267} &%
    \makebox[0.22\linewidth]{3.749} \\
    
    \raisebox{0.11\linewidth}{\rotatebox[origin=c]{90}{\sffamily GT}} &%
    \includegraphics[width=0.22\linewidth]{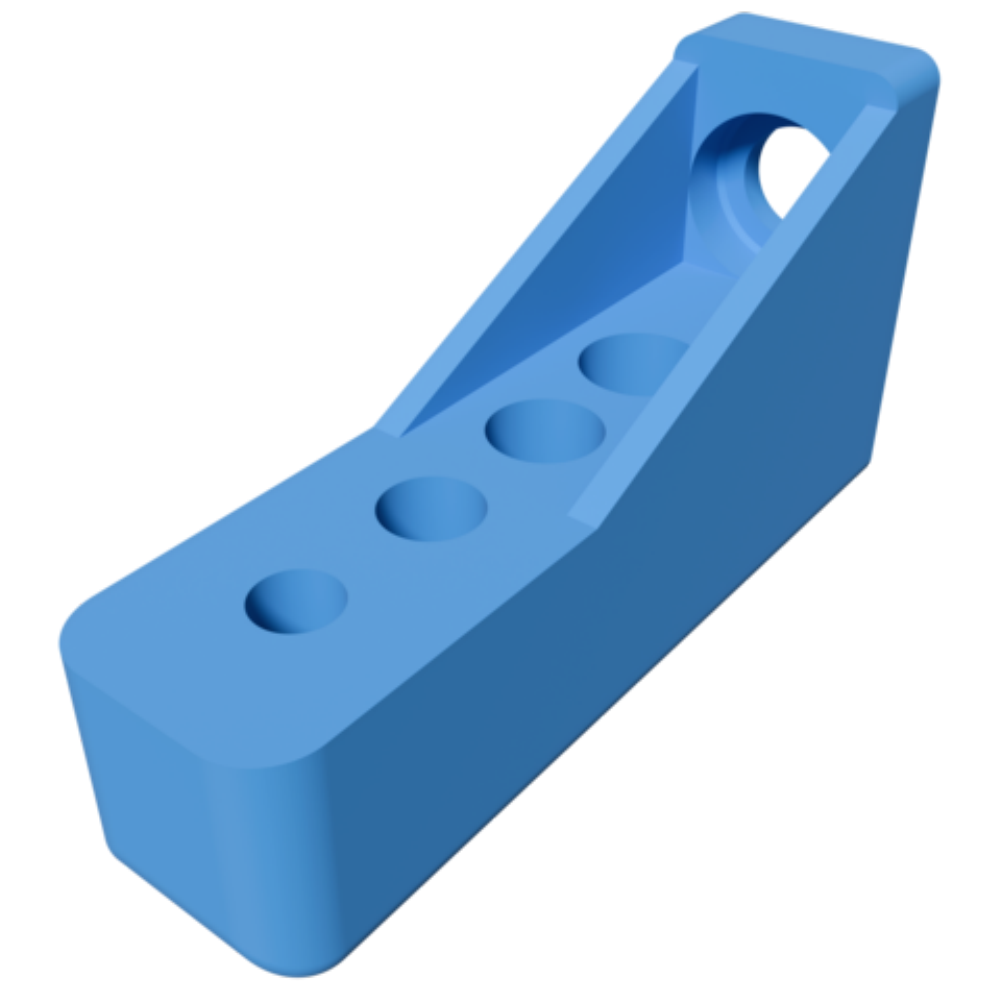} &%
    \includegraphics[width=0.22\linewidth]{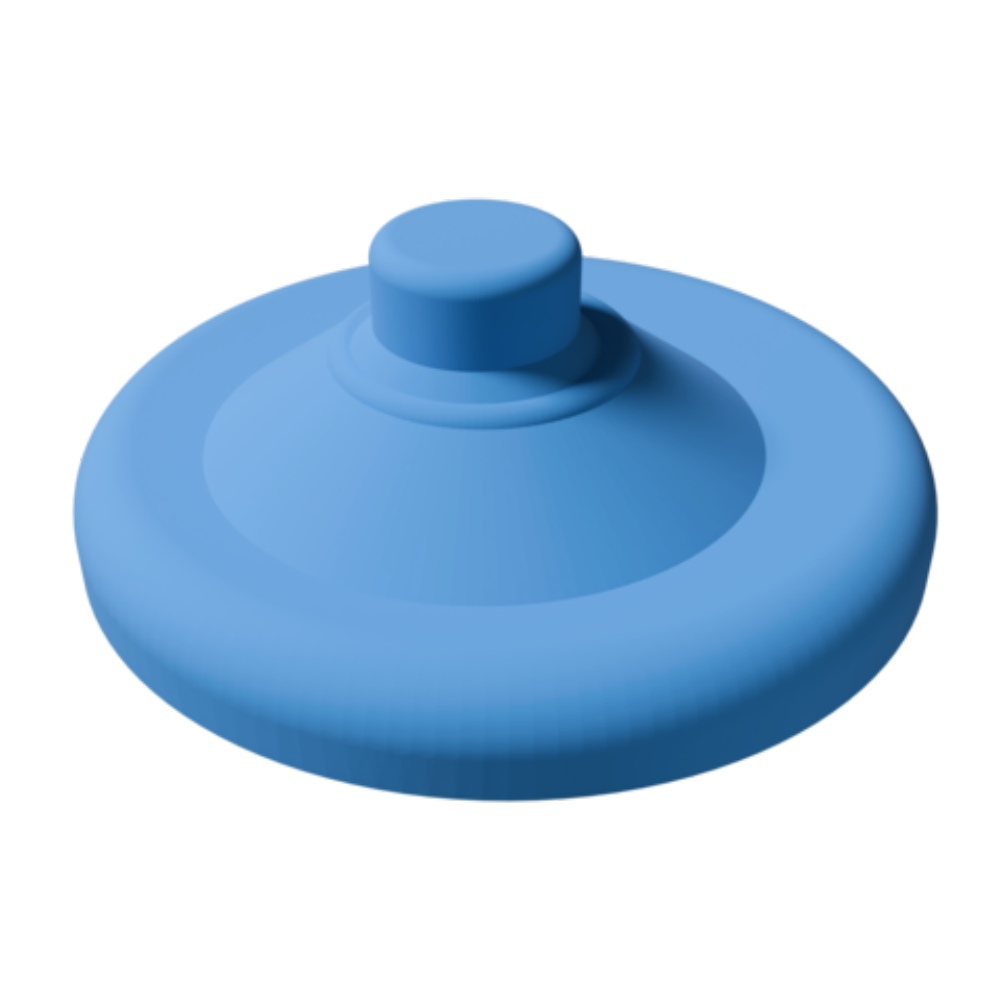} &%
    \includegraphics[width=0.22\linewidth]{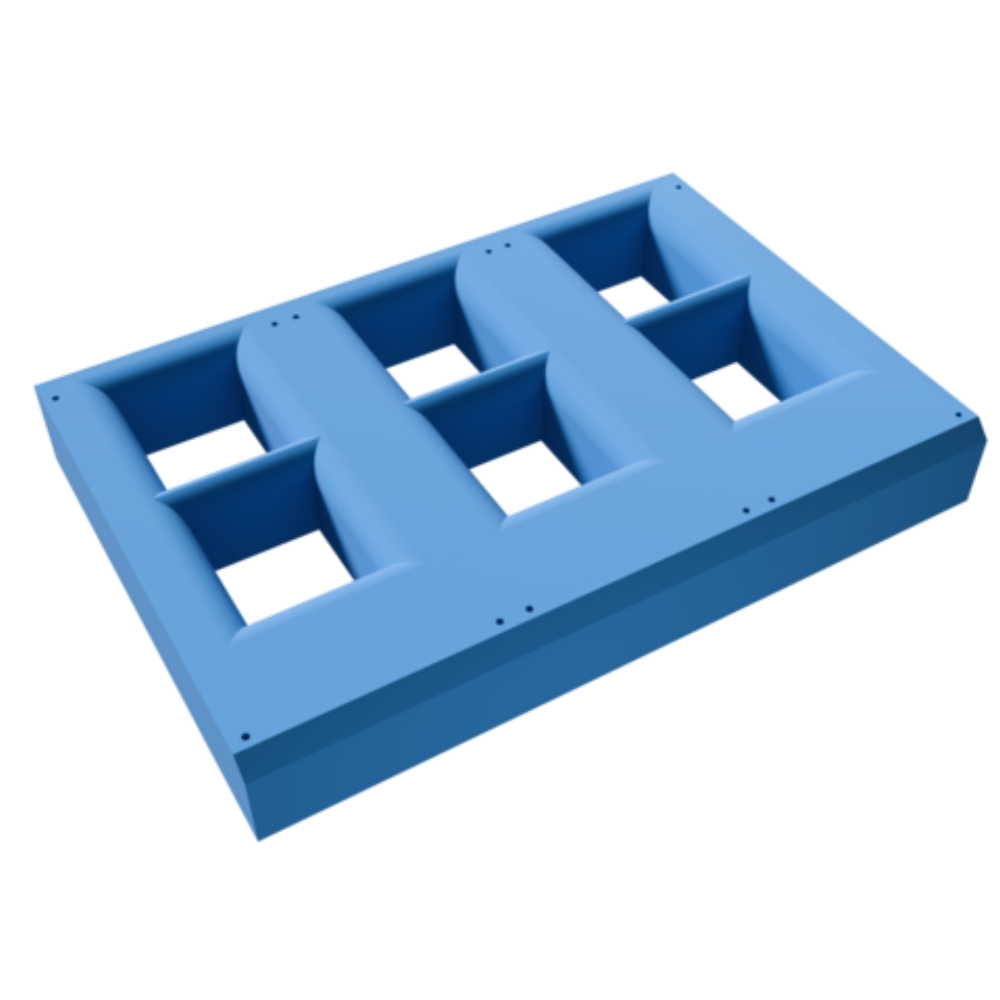} &%
    \includegraphics[width=0.22\linewidth]{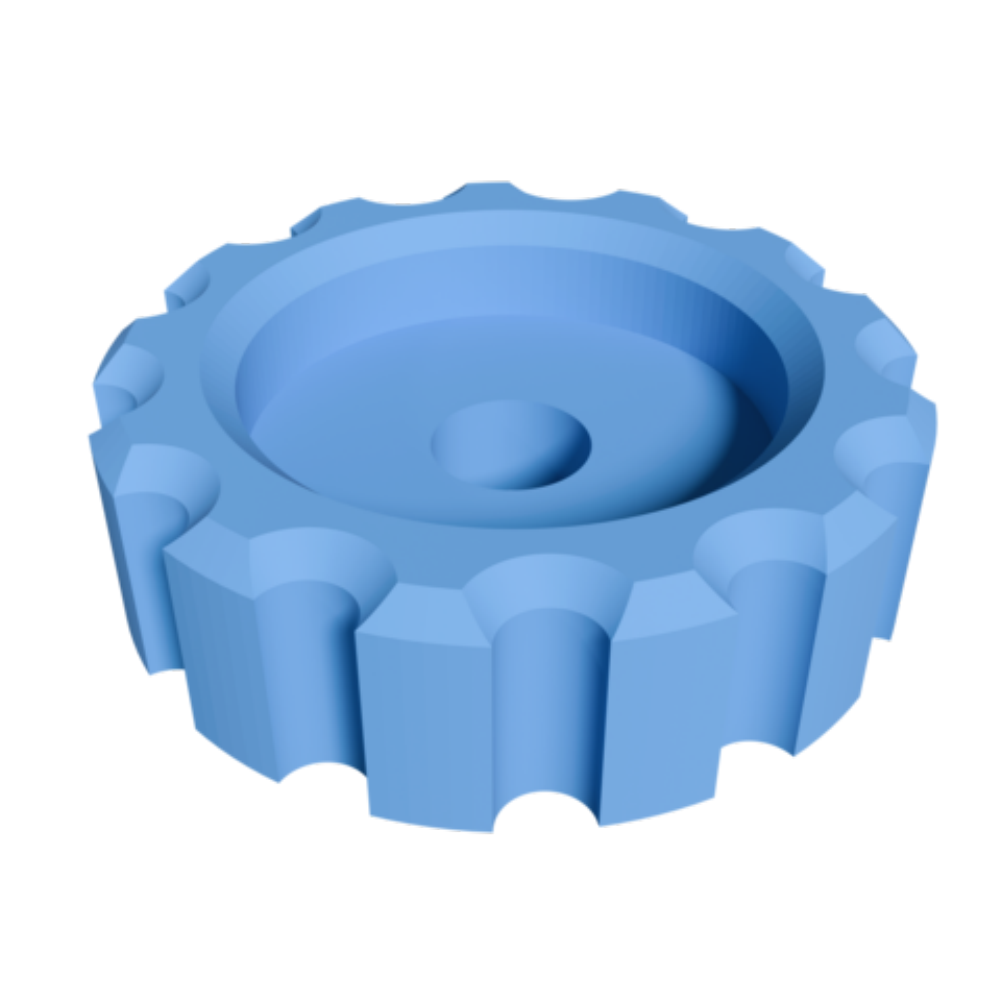} \\
    \makebox[0.01\linewidth]{ } &%
    \makebox[0.22\linewidth]{-} &%
    \makebox[0.22\linewidth]{-} &%
    \makebox[0.22\linewidth]{-} &%
    \makebox[0.22\linewidth]{-} \\
\end{tabular}
\end{scriptsize}
\caption{Illustration of CD errors visualized by a color map. The CD value for each model is shown below its reconstruction. The CD of {\nhrep} is noticeably larger on one of the models due to the presence of open feature curves, exhibiting the same issue shown in Figure~\ref{fig:open_feature}.}
\label{fig:mesh_input_color}
\end{figure}

\paragraph{Complex structures}
Without the need for decomposition or meshing, {\sharpnet} provides substantial flexibility in modeling gradient discontinuities, allowing it to represent a broad spectrum of intricate geometric forms. As shown in Figure~\ref{fig:mesh_complex}, it supports feature curves of arbitrary shapes and connectivity, e.g., complex feature sets, thin-structures, and high-genus geometric shapes. In Figure~\ref{fig:mesh_input_color}, we compare the CD distributions and observe that the CDs of {\nhrep} on high-genus models are noticeably large.

The quantitative results for all 65 models are summarized in Table~\ref{tab:mesh_input}. {\sharpnet} surpasses {\nhrep} across all four evaluation metrics. In addition, both {\nhrep} and Patch-Grid handle sharp features as fixed entities, preventing them from being updated during training. By contrast, the sharp features in our {\sharpnet} are learnable, which makes the framework applicable to a wider variety of tasks, as illustrated in the following subsections.

\begin{table}[!htbp]
\caption{Quantitative statistics of models reconstructed from meshes.}
\label{tab:mesh_input}
\begin{tabular}{ccccc}
\toprule
Method & $\text{CD}^{\times 10^{-3}}$ $\downarrow$  & $\text{HD}^{\times 10^{-2}}$ $\downarrow$ & $\text{NE}^{\circ}$ $\downarrow$ & $\text{FC}^\%$ $\uparrow$ \\
\midrule
{\nhrep}       & 4.273 & 3.081 & 2.327 & 96.59\\
{\sharpnet}     & \textbf{3.788} & \textbf{2.249} & \textbf{2.286} & \textbf{98.43} \\
\bottomrule
\end{tabular}
\end{table}

\begin{figure}
\centering
\begin{subcaptionblock}{.3\linewidth}
    \includegraphics[width=\linewidth]{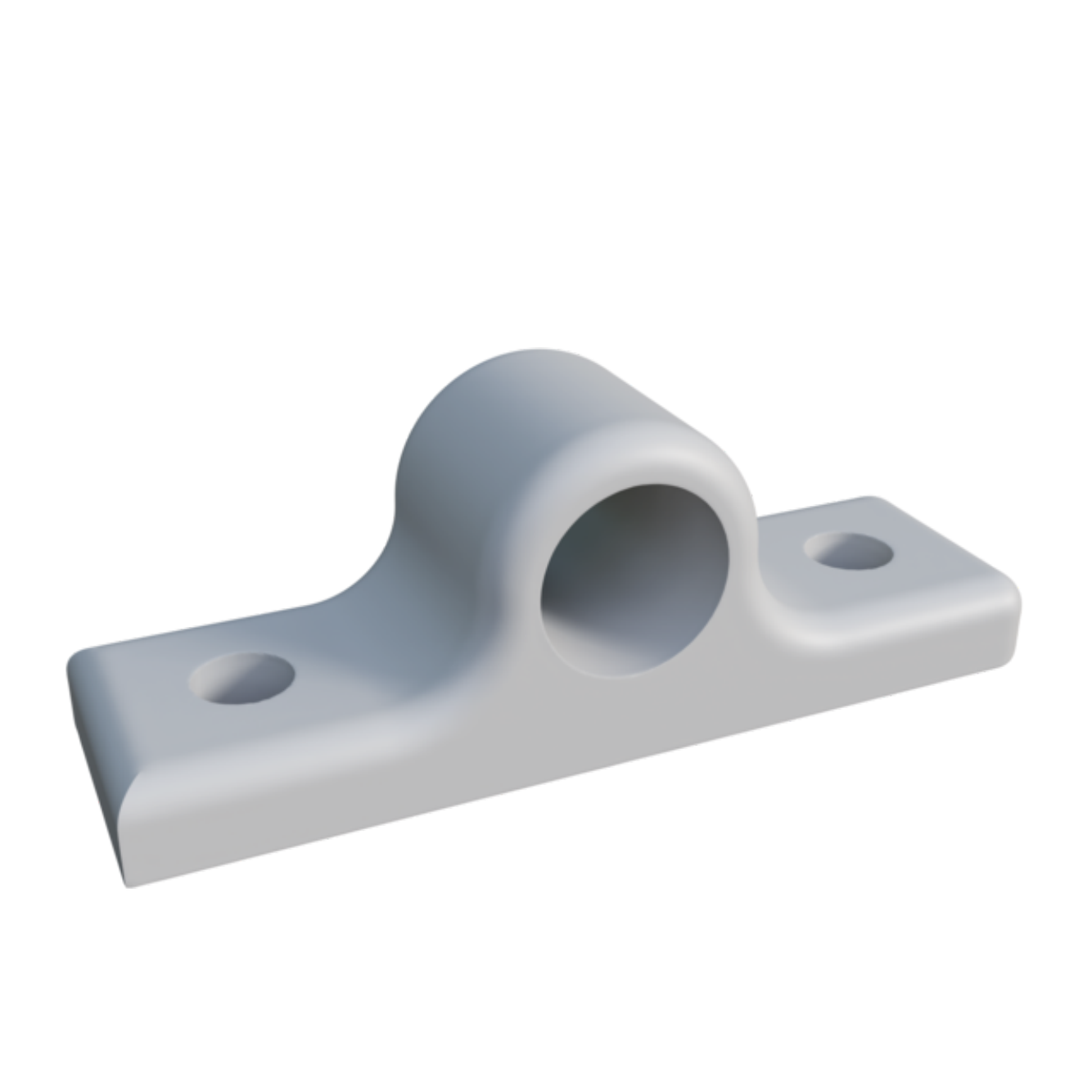}
    \caption*{$A$}
\end{subcaptionblock}
\begin{subcaptionblock}{.3\linewidth}
    \includegraphics[width=\linewidth]{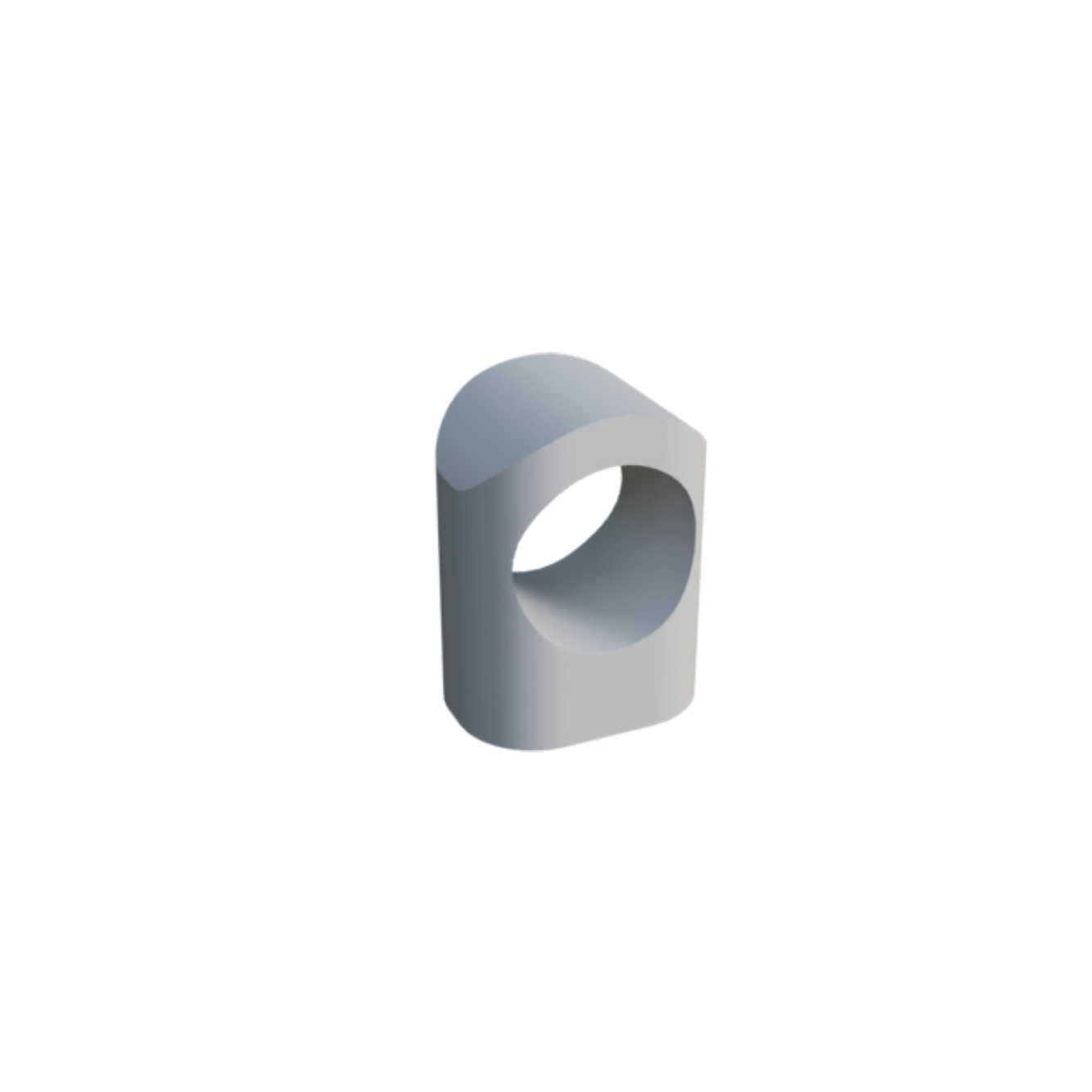}
    \caption*{$A \cap B$}
\end{subcaptionblock}
\begin{subcaptionblock}{.3\linewidth}
    \includegraphics[width=\linewidth]{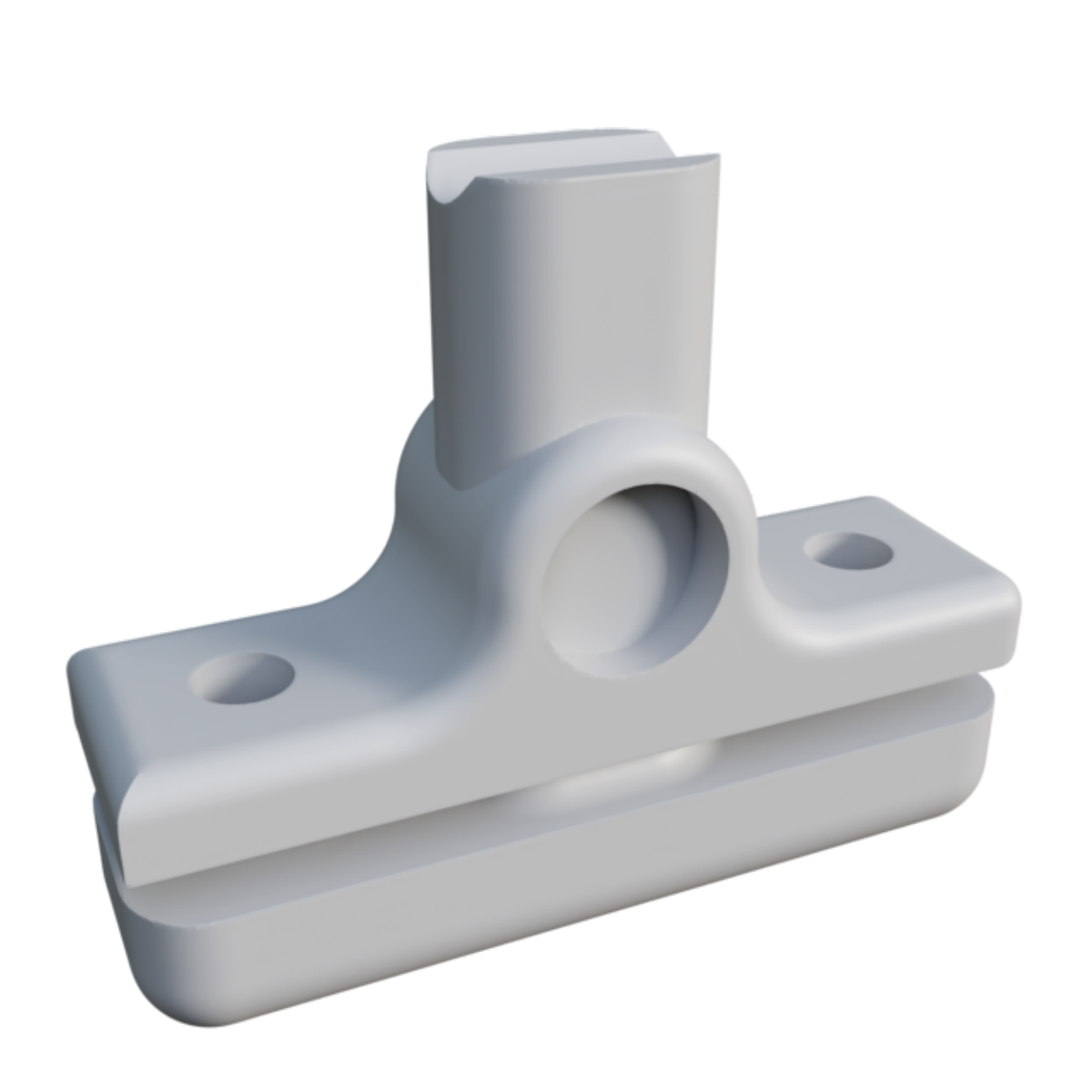}
    \caption*{$A \cup B$}
\end{subcaptionblock}
\begin{subcaptionblock}{.3\linewidth}
    \includegraphics[width=\linewidth]{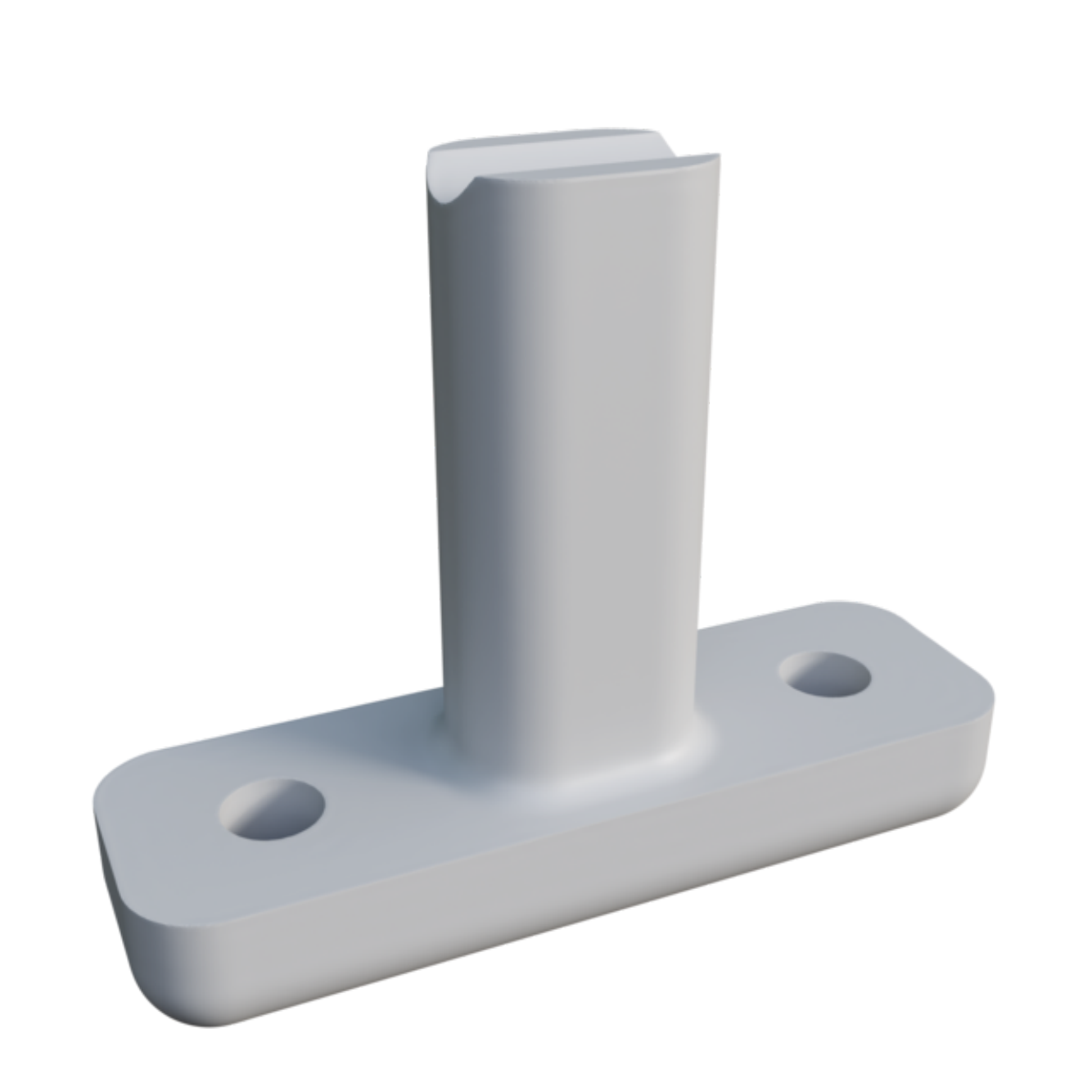}
    \caption*{$B$}
\end{subcaptionblock}
\begin{subcaptionblock}{.3\linewidth}
    \includegraphics[width=\linewidth]{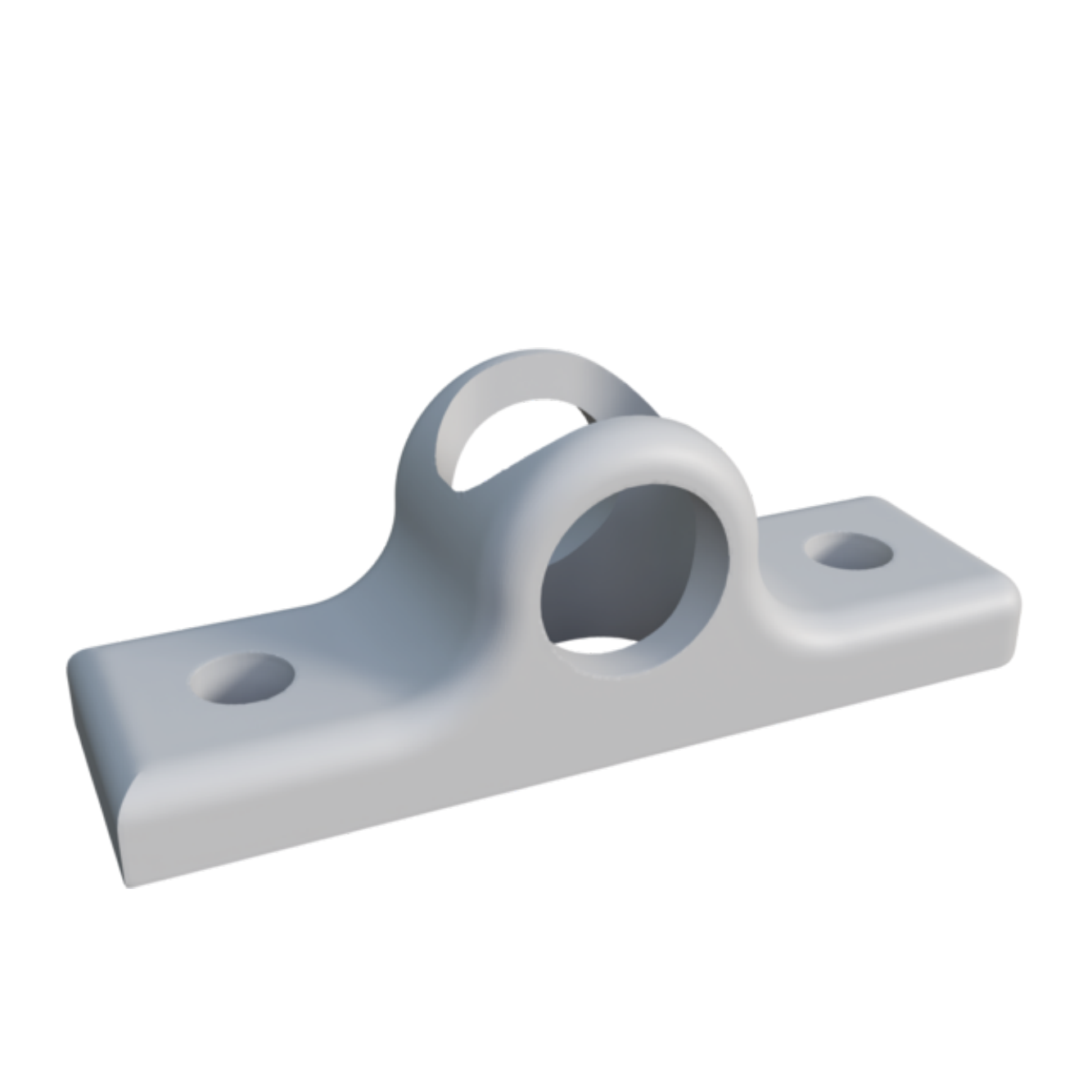}
    \caption*{$A \setminus B$}
\end{subcaptionblock}
\begin{subcaptionblock}{.3\linewidth}
    \includegraphics[width=\linewidth]{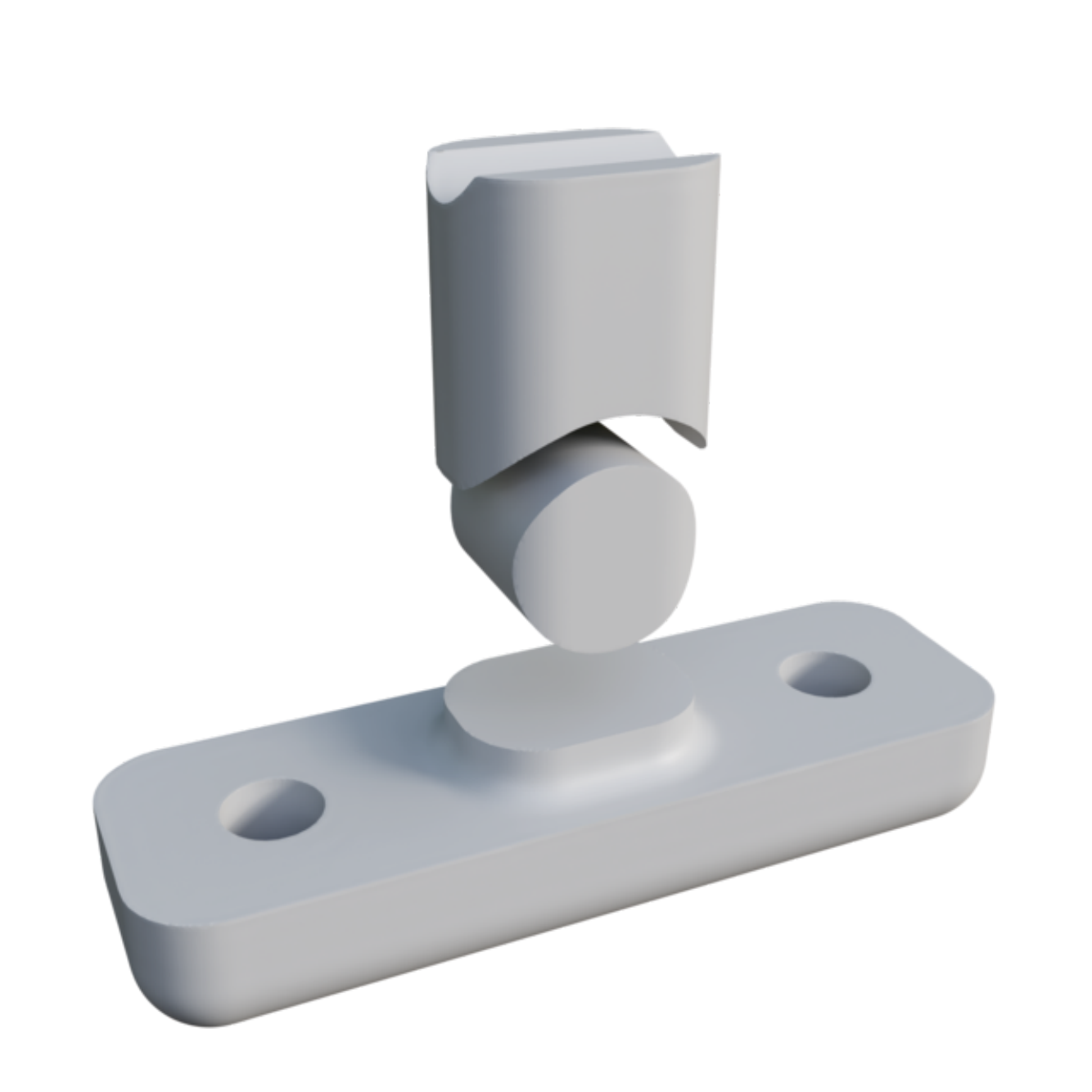}
    \caption*{$B \setminus A$}
\end{subcaptionblock}
\caption{Boolean operations in {\method}. Consider two CAD models, \(A\) and \(B\), represented in {\method} by their signed distance fields \(F_A\) and \(F_B\), respectively. 
Because points inside \(A\) and \(B\) correspond to negative values of \(F_A\) and \(F_B\), the Boolean operations intersection, union, and set difference can be expressed as
\(F_{A \cap B}=\max(F_A,F_B)\), 
\(F_{A \cup B}=\min(F_A,F_B)\), 
\(F_{A \setminus B}=\max(F_A,-F_B)\), 
and \(F_{B \setminus A}=\max(-F_A,F_B)\).}
\label{fig:boolean}
\end{figure}

\paragraph{Boolean operations}
Implicit CAD representations are naturally compatible with boolean operations, which are challenging to process using explicit representations such as splines and meshes. In Figure~\ref{fig:boolean}, we illustrate the application of {\sharpnet} to constructive solid modeling.

\subsection{Reconstruction from Oriented Points}
\label{sec:CAD_points_normals}
We additionally assess our approach using only points together with their corresponding oriented normals as input, and compare it against {\ingp}~\cite{Muller2022INGP}. In contrast, {\nhrep}~\cite{Guo2022NH-Rep} depends on pre-segmented patches and oriented normals and is therefore omitted from this comparison.

We train {\sharpnet} using Equation~\eqref{eqn:3Dloss_points_normals}, where MLP parameters \(\theta\) and feature surface points \(M\) are parameters to be learned. The loss components \(\mathcal{L}_{\text{sur}}\), \(\mathcal{L}_{\text{ext}}\), and \(\mathcal{L}_{\text{ekl}}\) are identical to those defined in Equation~\eqref{eqn:3Dloss_mesh}. The regularization term \(\mathcal{L}_\mathcal{R}(M)\) is the same as in Equation~\eqref{eqn:2Dloss}, but it is now imposed on the boundary curves \(\partial M\), rather than on \(M\) itself as in the 3D case, where \(M\) denotes surfaces. \(\mathcal{L}_\mathcal{R}\) is used to prevent \(M\) from self-intersection and folding.
\begin{equation}
\label{eqn:3Dloss_points_normals}
    \mathcal{L}(\theta,M) =
        \alpha_{\text{sur}} \cdot \mathcal{L}_{\text{sur}} +
        \alpha_{\text{ext}} \cdot \mathcal{L}_{\text{ext}} +
        \alpha_{\text{ekl}} \cdot \mathcal{L}_{\text{ekl}} +
        \alpha_{\text{nor}} \cdot \mathcal{L}_{\text{nor}} +
        \alpha_\mathcal{R} \cdot \mathcal{L}_\mathcal{R}.
\end{equation}

The parameters are set to \(\alpha_{\text{sur}}=7000\), \(\alpha_{\text{ext}}=600\), \(\alpha_{\text{ekl}}=35\), \(\alpha_{\text{nor}}=15\), \(\alpha_\mathcal{R}=10\) in our experiments.

The sharp features \(M\) in Equation~\eqref{eqn:3Dloss_points_normals} are learnable parameters that need to be initialized before training. We employ the pre-trained NerVE network~\cite{Zhu2023NerVE} to infer sharp features from the input point cloud. The procedure for initializing \(M\) from a point cloud is explained in detail in Appendix~\ref{sec:feature_surface_points}.

\begin{figure*}
    \centering
    \includegraphics[width=0.12\linewidth]{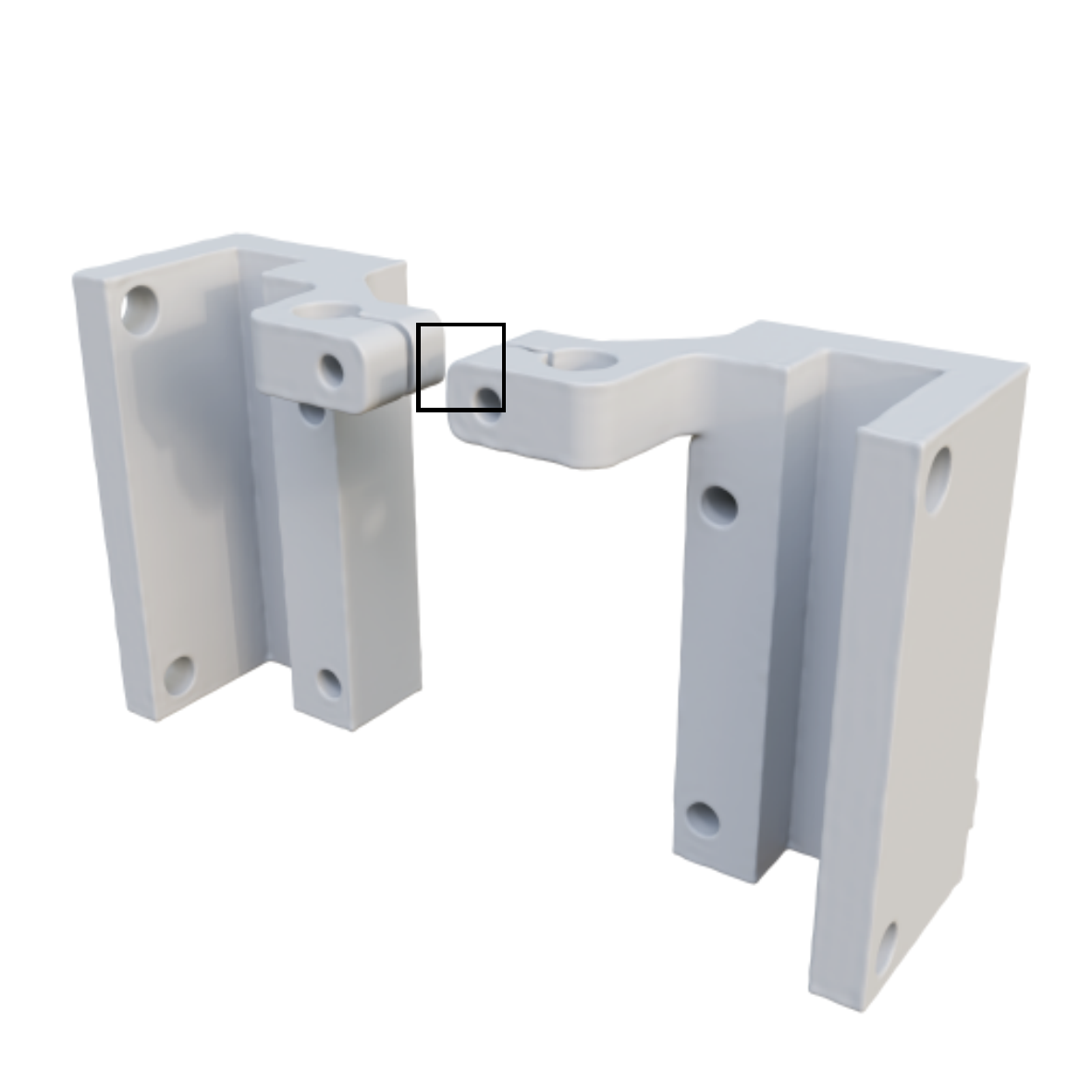}
    \includegraphics[width=0.12\linewidth]{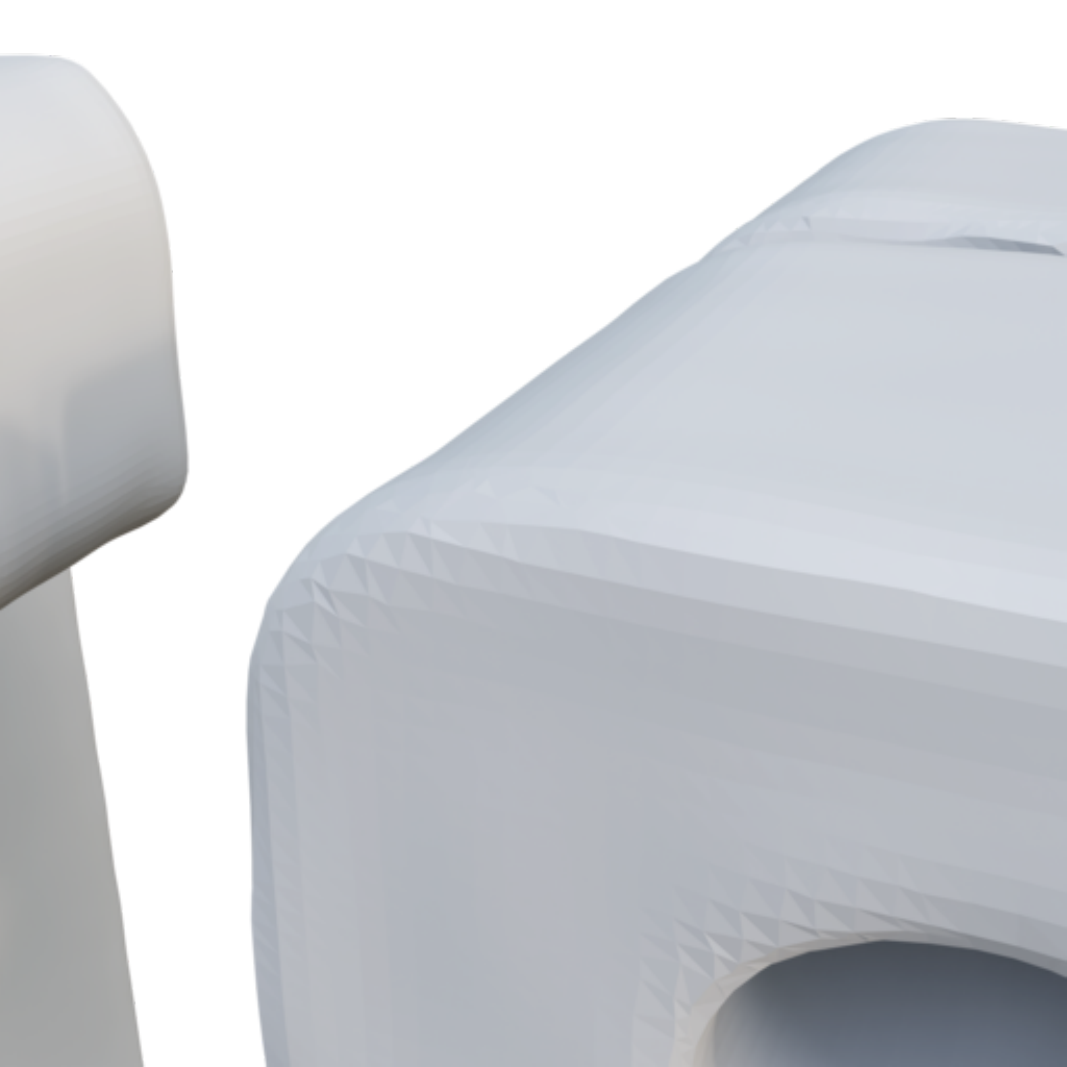}
    \includegraphics[width=0.12\linewidth]{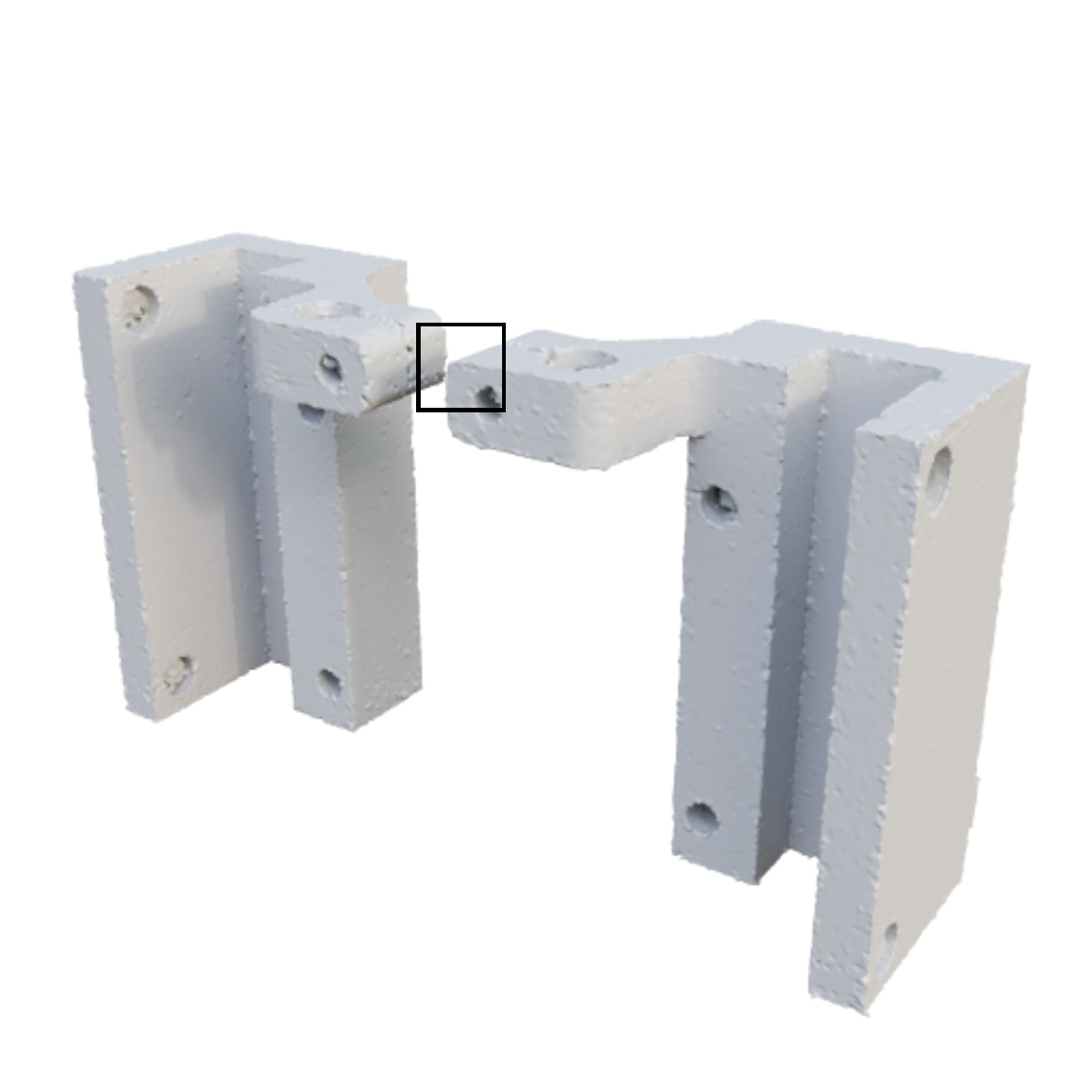}
    \includegraphics[width=0.12\linewidth]{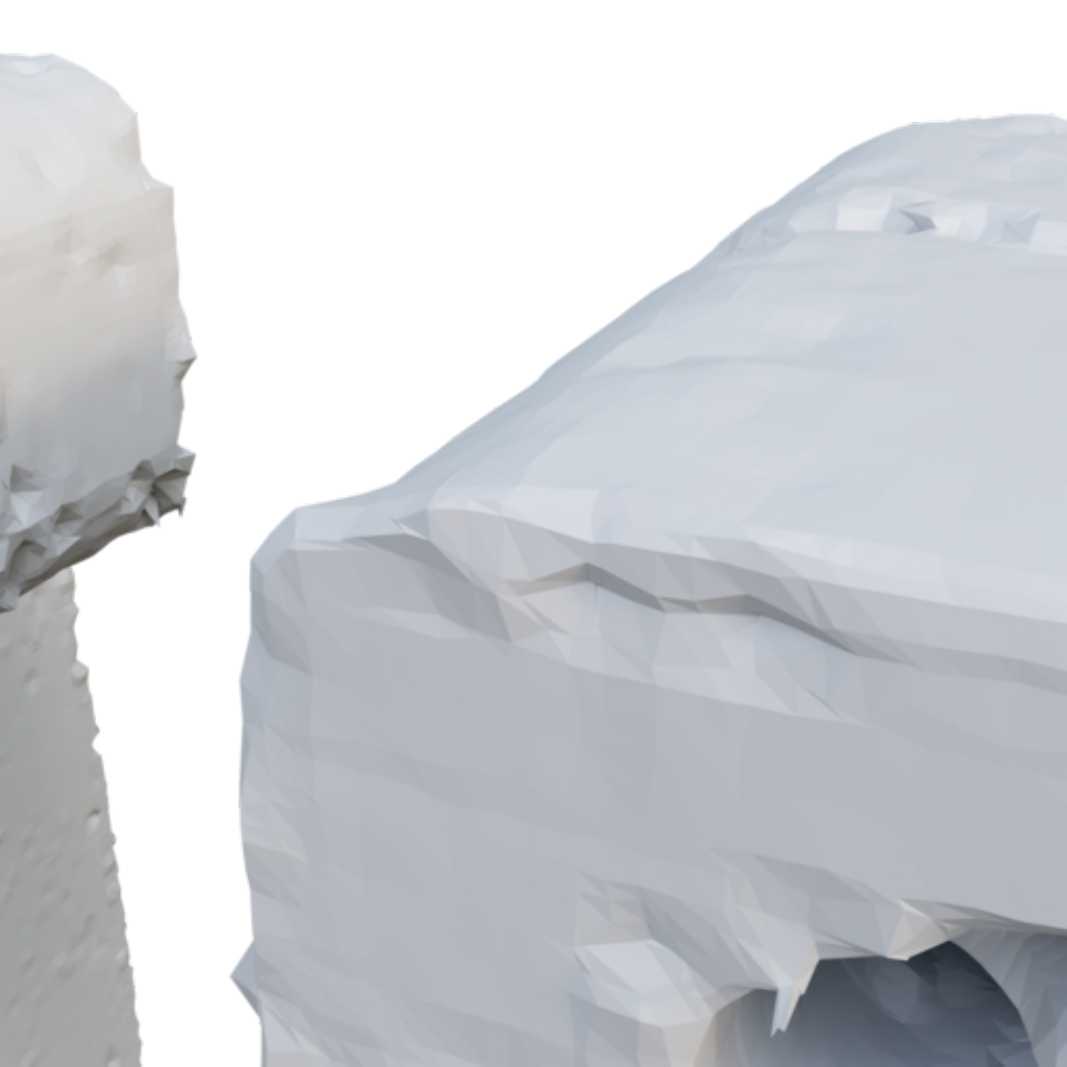}
    \includegraphics[width=0.12\linewidth]{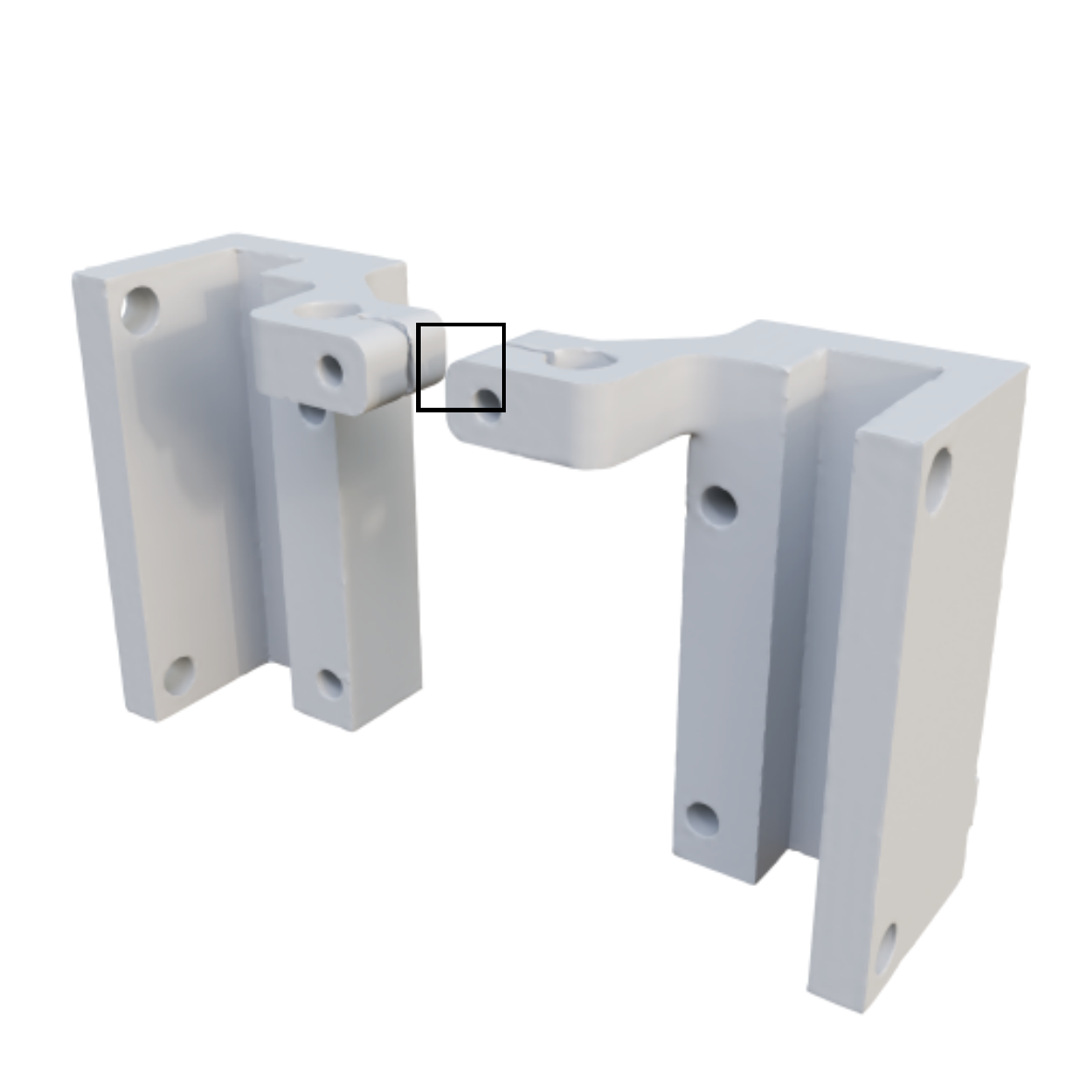}
    \includegraphics[width=0.12\linewidth]{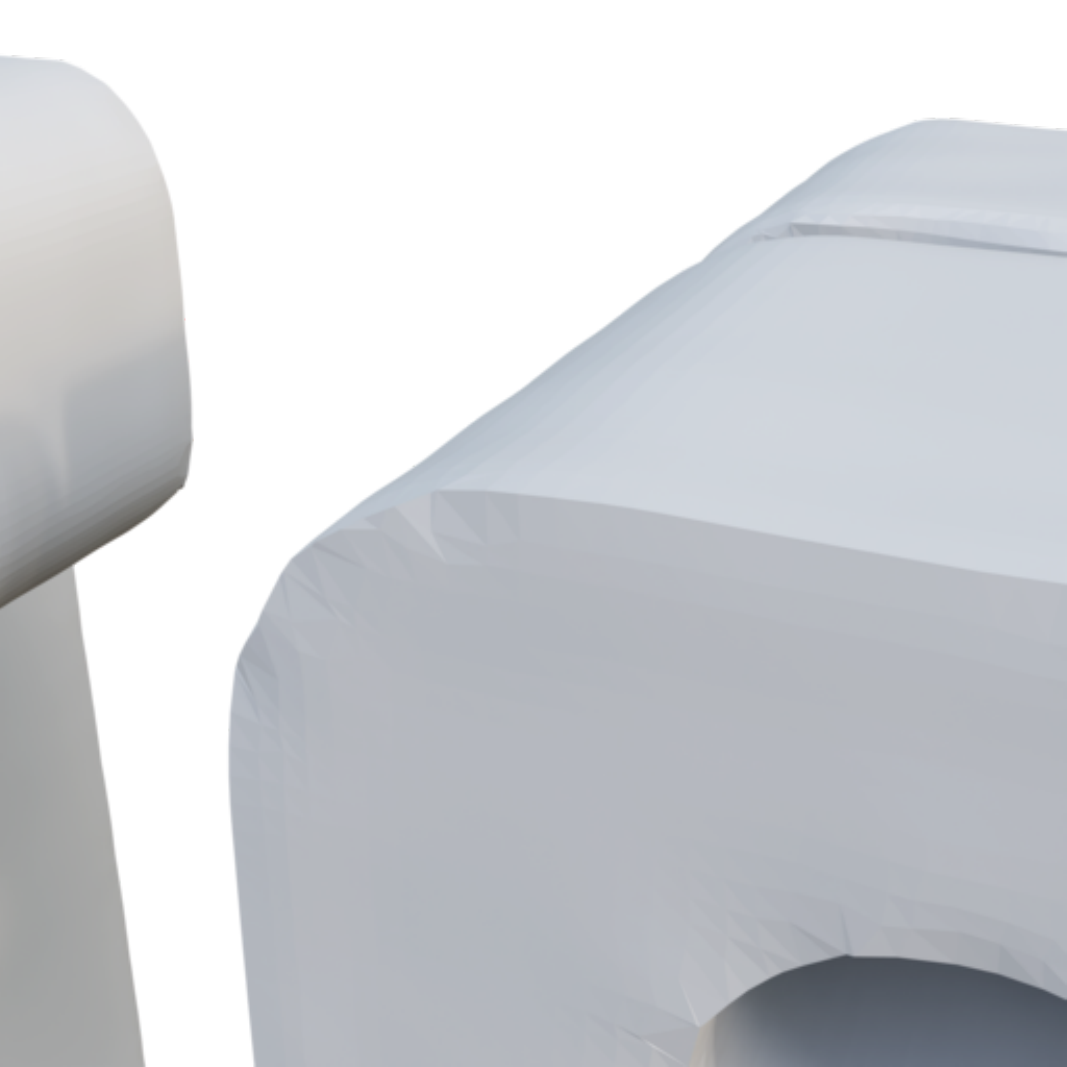}
    \includegraphics[width=0.12\linewidth]{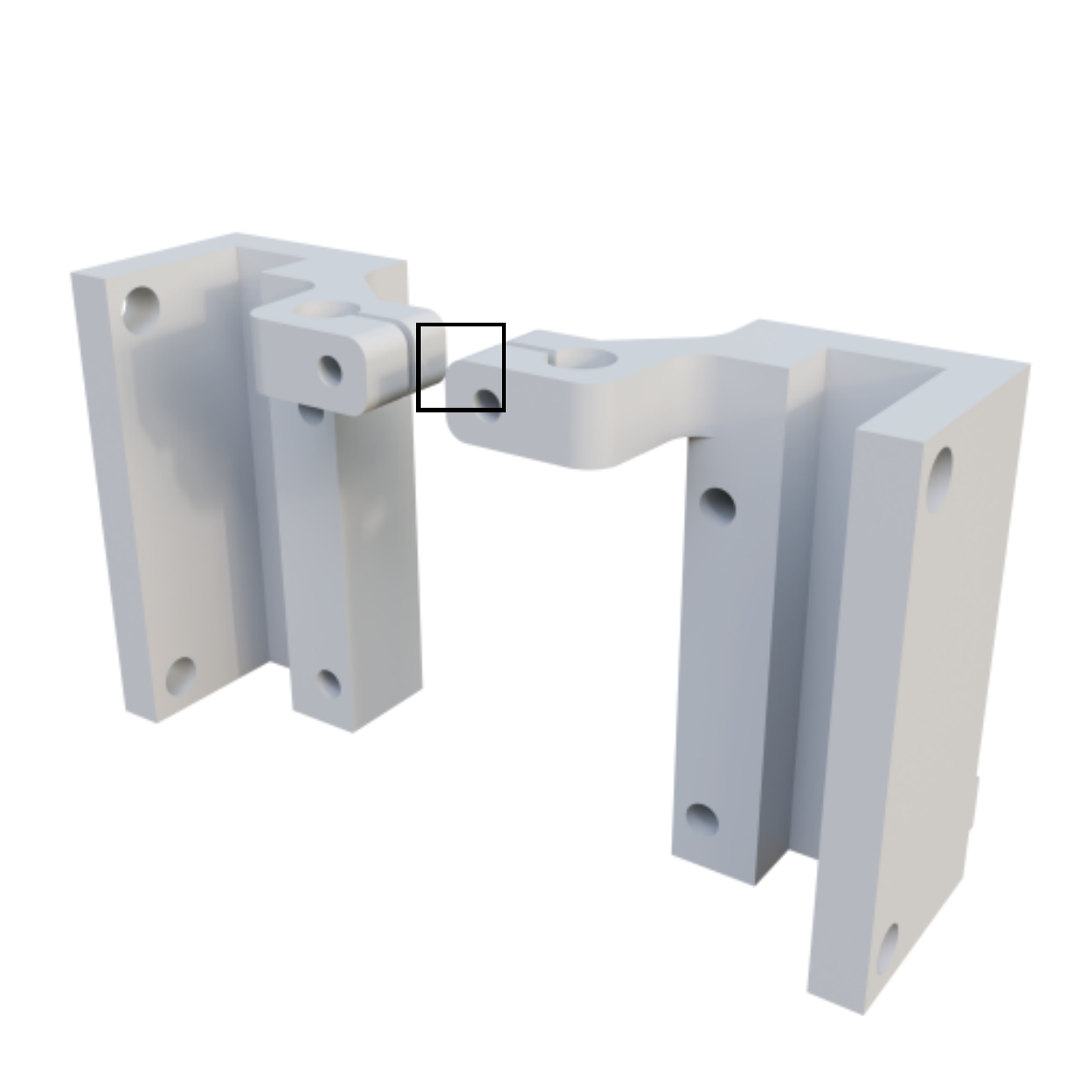}
    \includegraphics[width=0.12\linewidth]{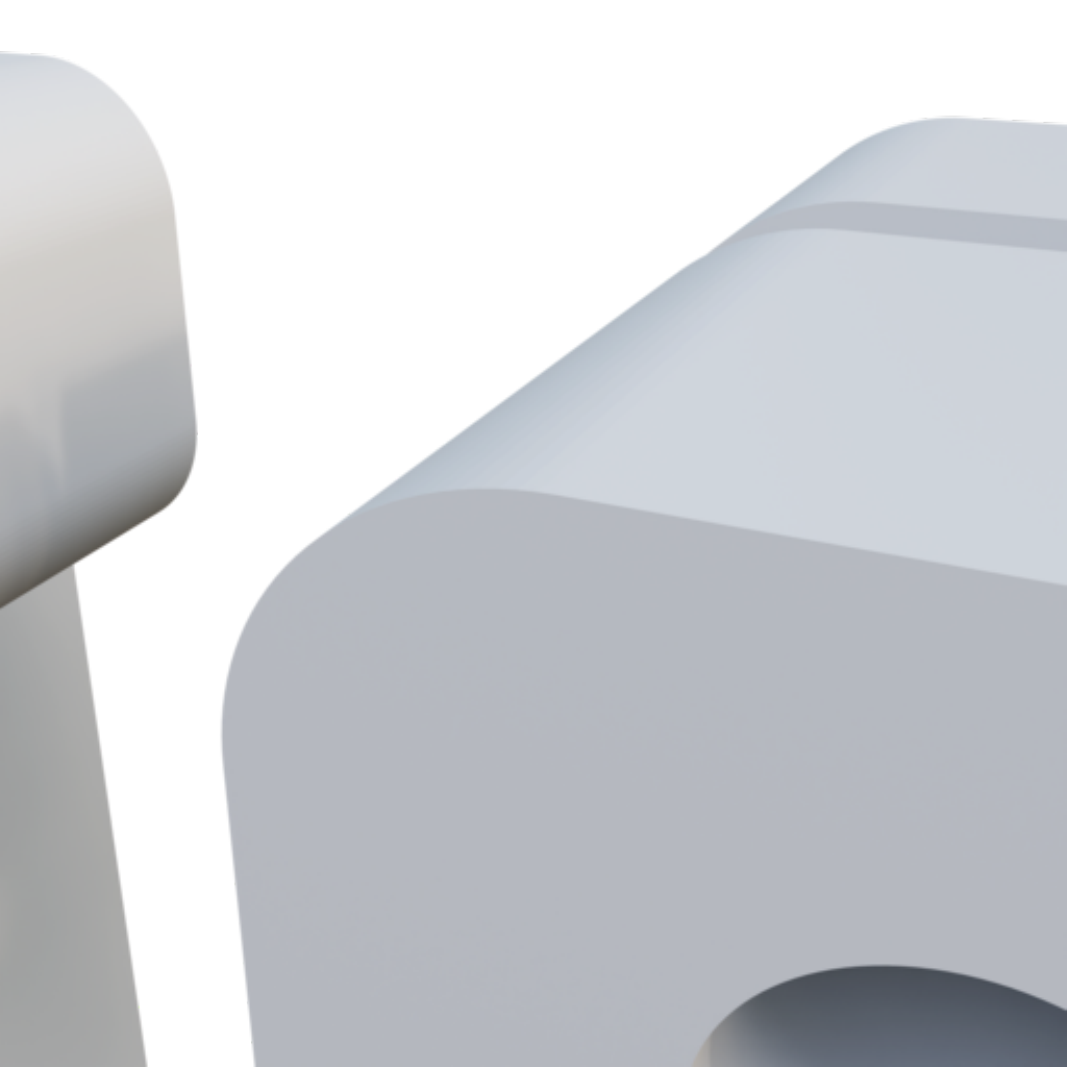}\\
    \makebox[0.24\linewidth]{\text{3.422}}
    \makebox[0.24\linewidth]{\text{3.340}}
    \makebox[0.24\linewidth]{\text{3.050}}
    \makebox[0.24\linewidth]{\text{-}}\\
    \vspace{6pt}
    \includegraphics[width=0.12\linewidth]{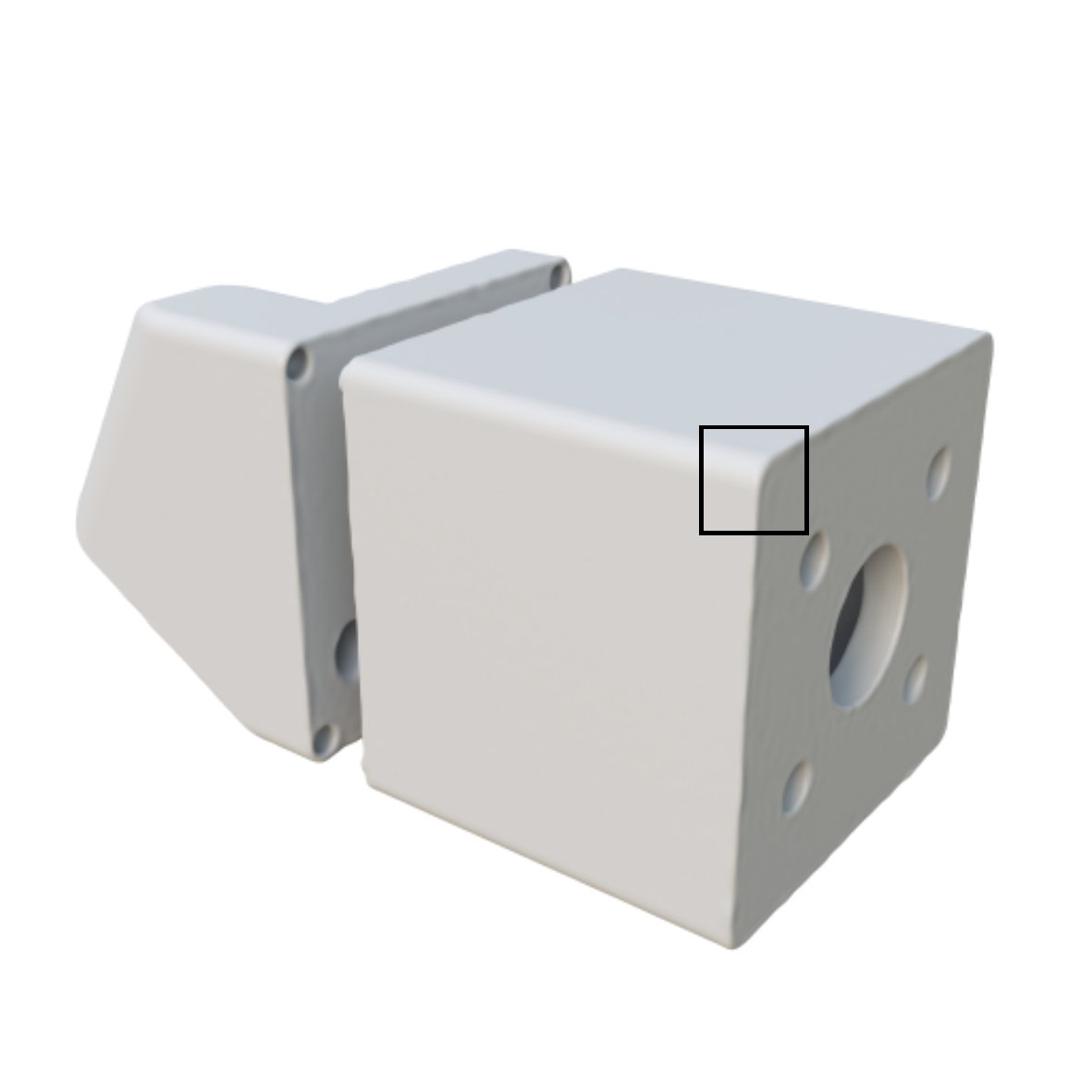}    \includegraphics[width=0.12\linewidth]{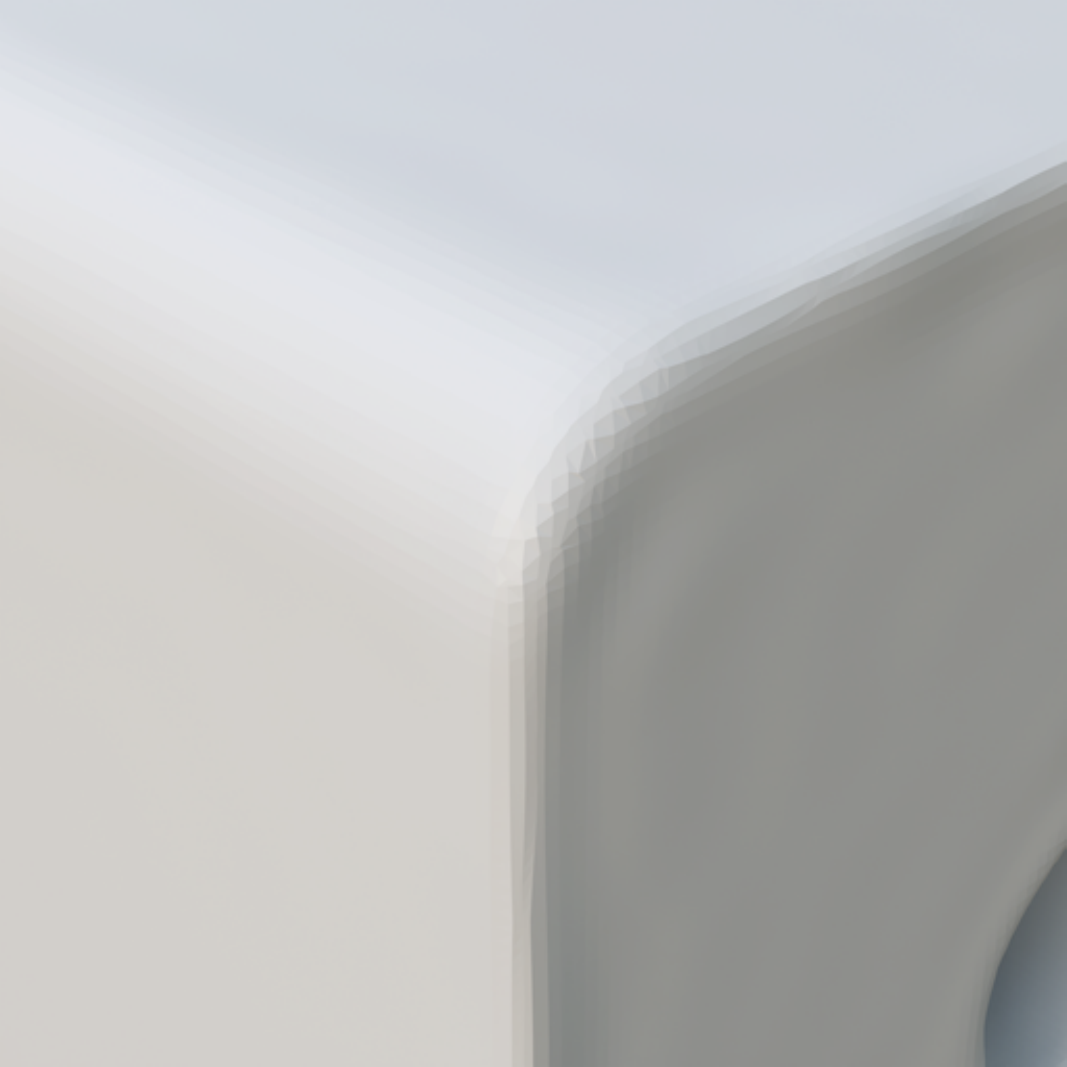}
    \includegraphics[width=0.12\linewidth]{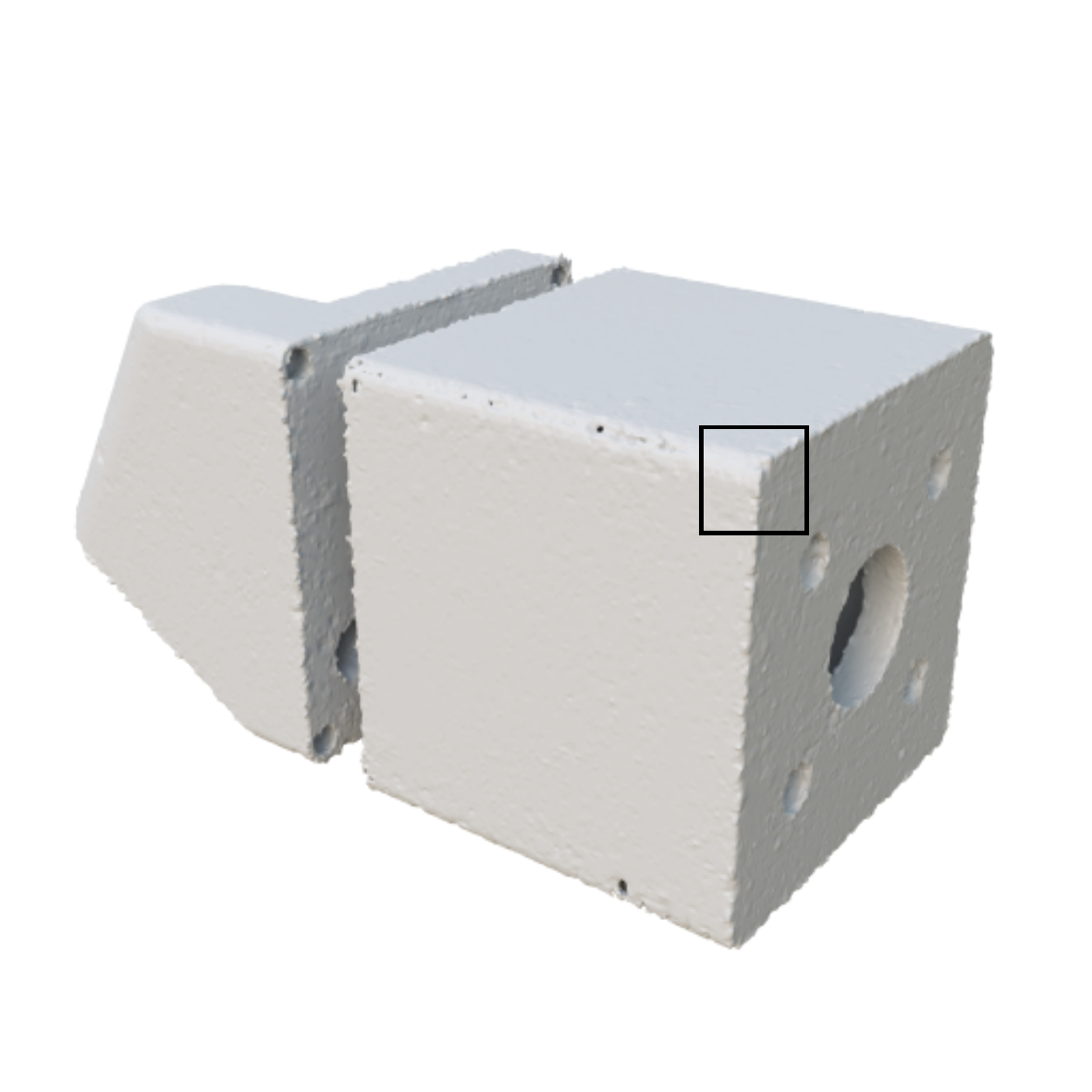}    \includegraphics[width=0.12\linewidth]{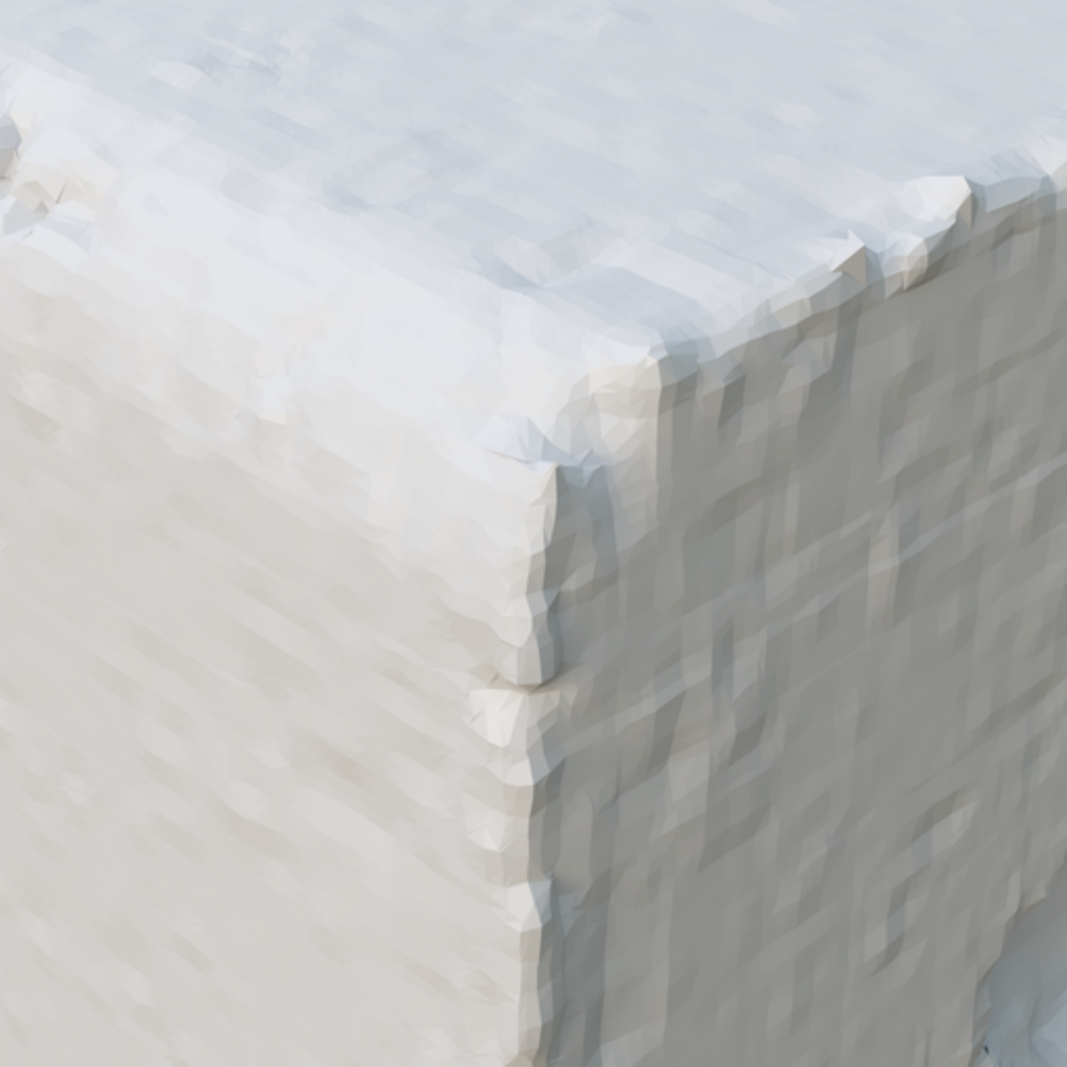}
    \includegraphics[width=0.12\linewidth]{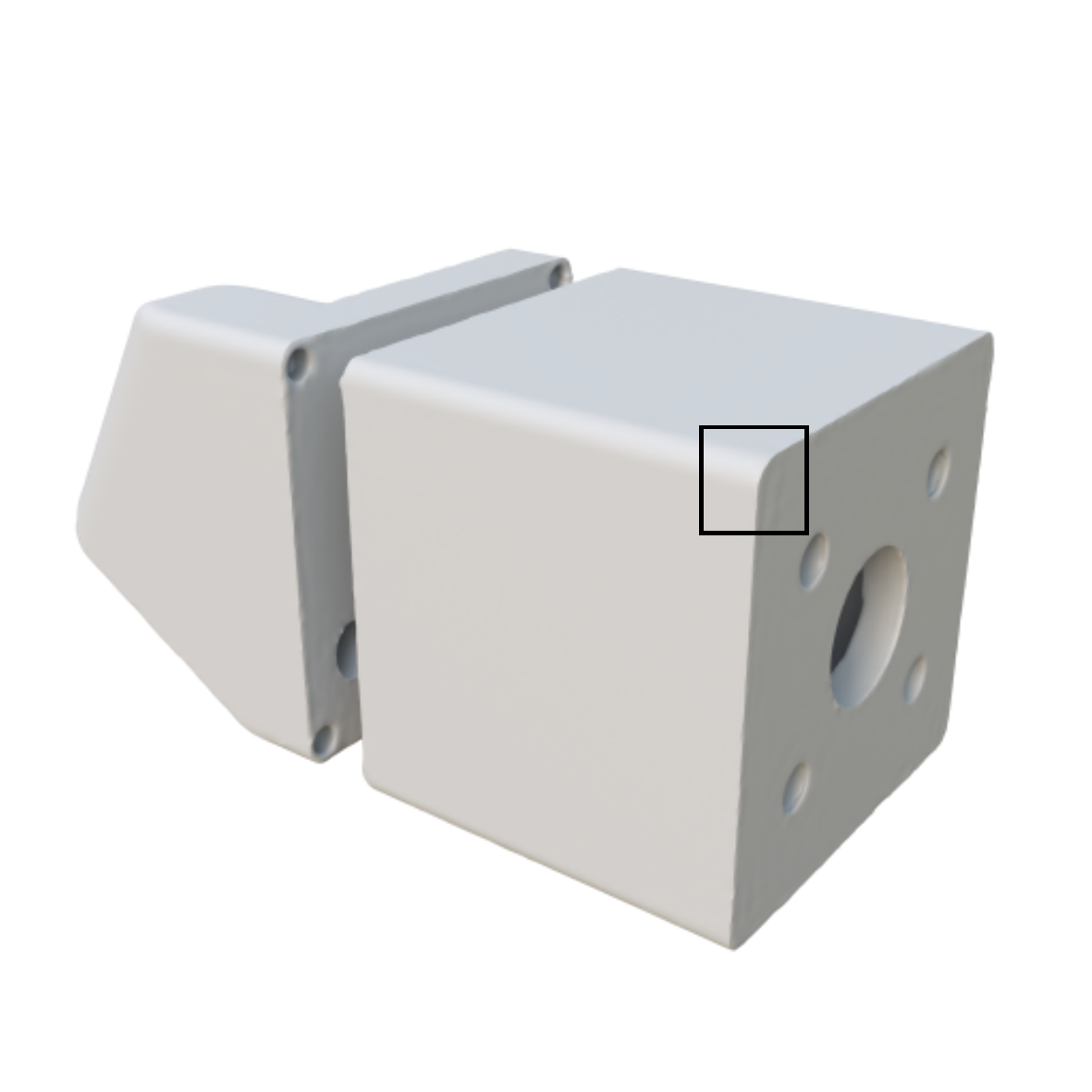}    \includegraphics[width=0.12\linewidth]{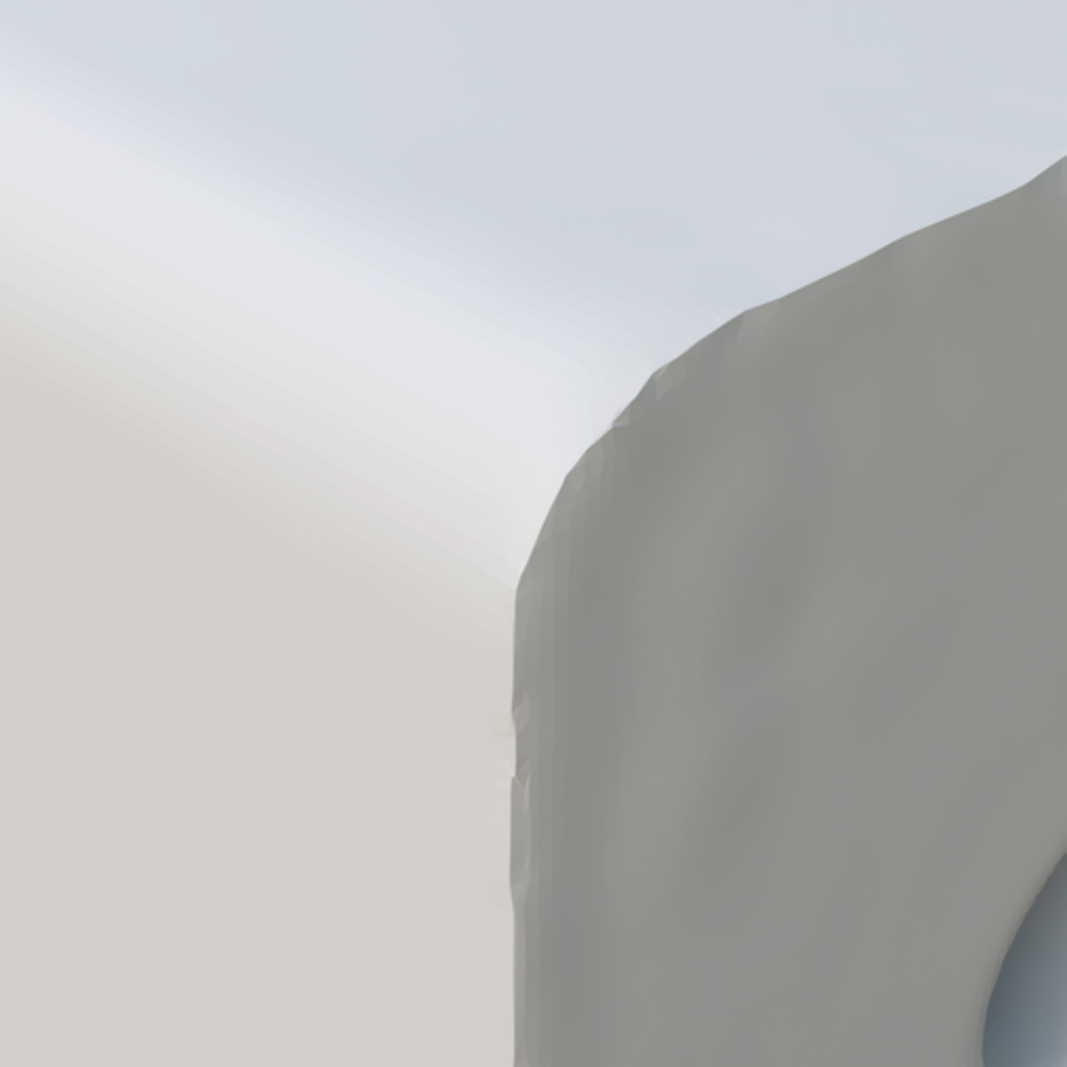}
    \includegraphics[width=0.12\linewidth]{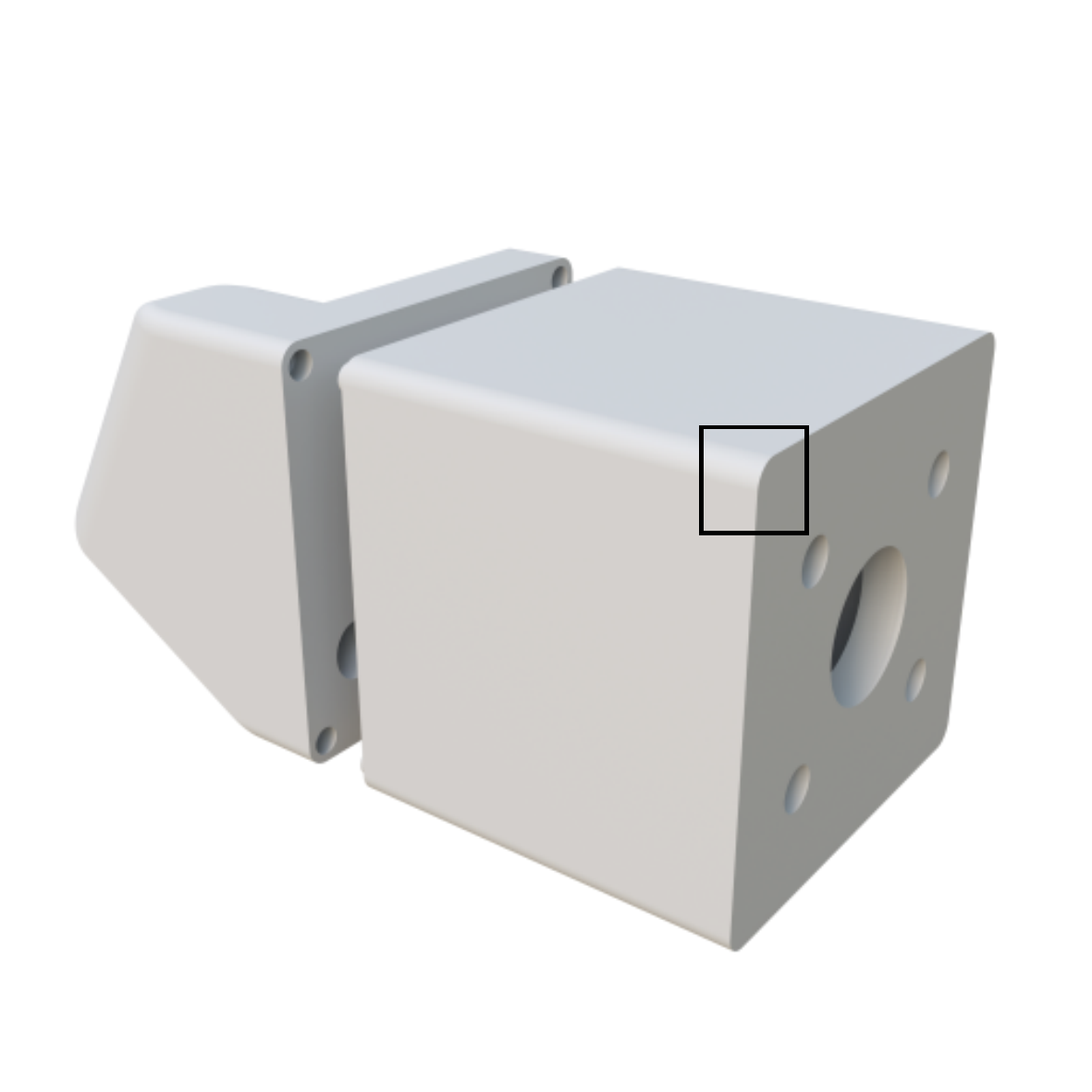}    \includegraphics[width=0.12\linewidth]{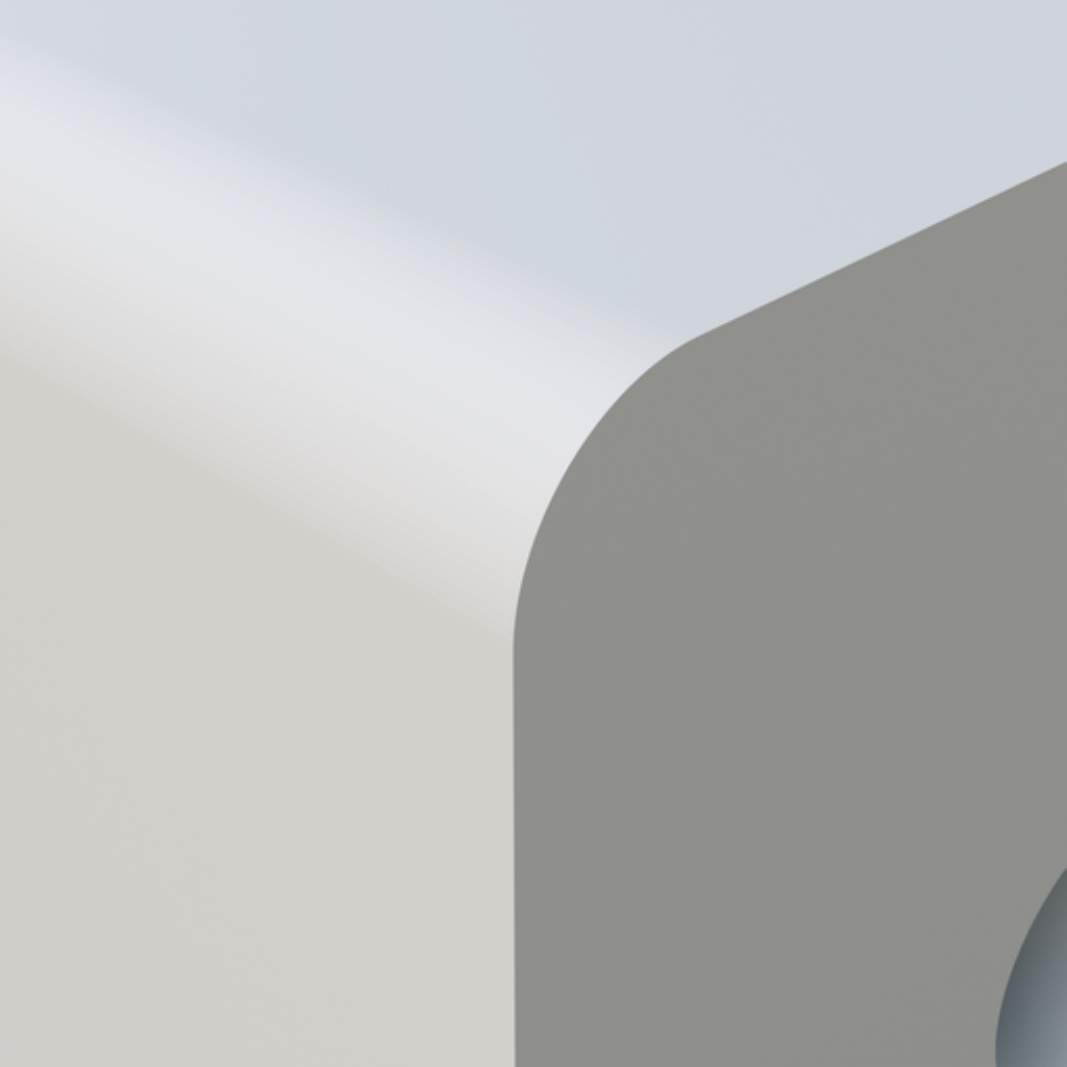}\\
    \makebox[0.24\linewidth]{\text{4.752}}
    \makebox[0.24\linewidth]{\text{4.496}}
    \makebox[0.24\linewidth]{\text{4.240}}
    \makebox[0.24\linewidth]{\text{-}}\\
    \vspace{6pt}
    \includegraphics[width=0.12\linewidth]{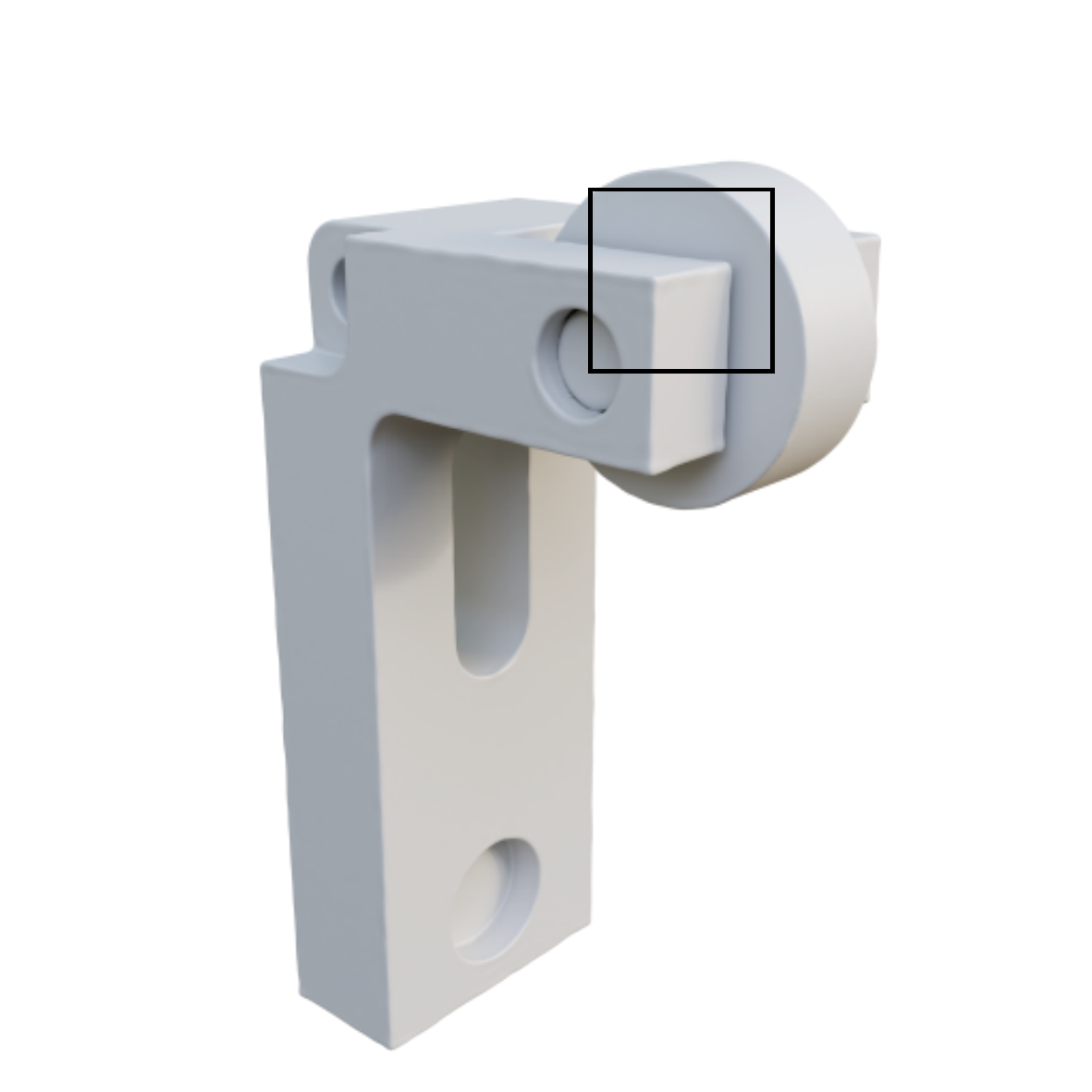}    \includegraphics[width=0.12\linewidth]{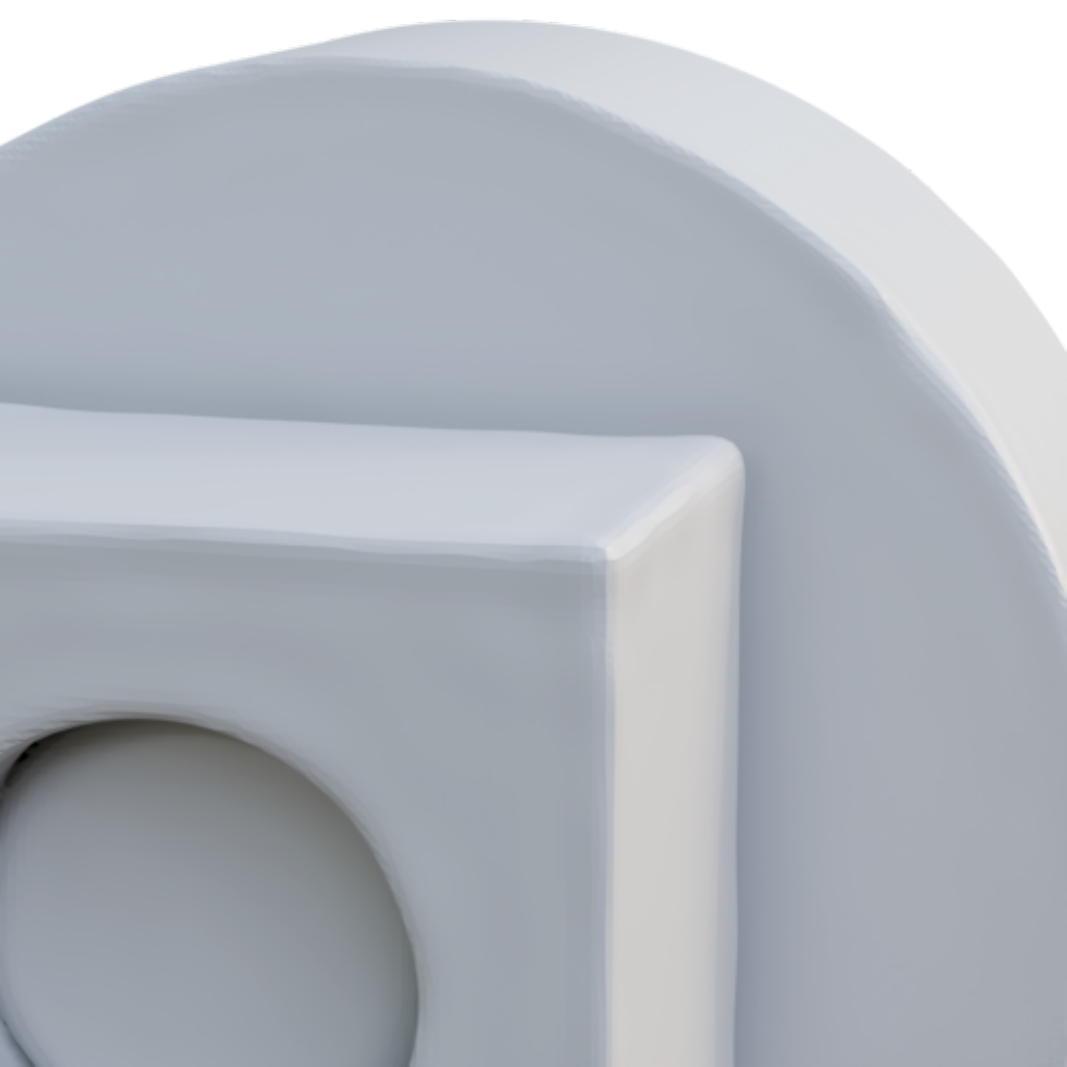}
    \includegraphics[width=0.12\linewidth]{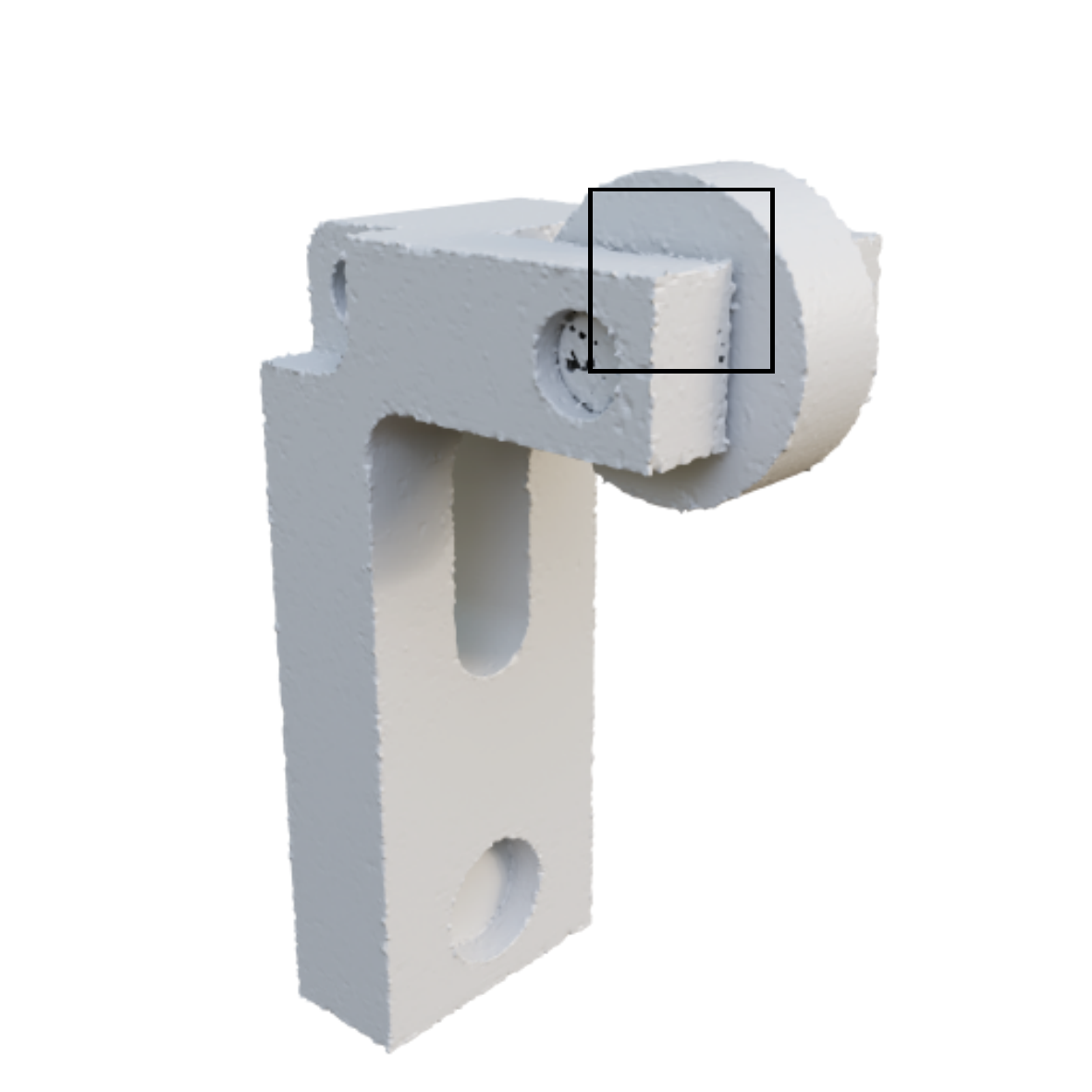}    \includegraphics[width=0.12\linewidth]{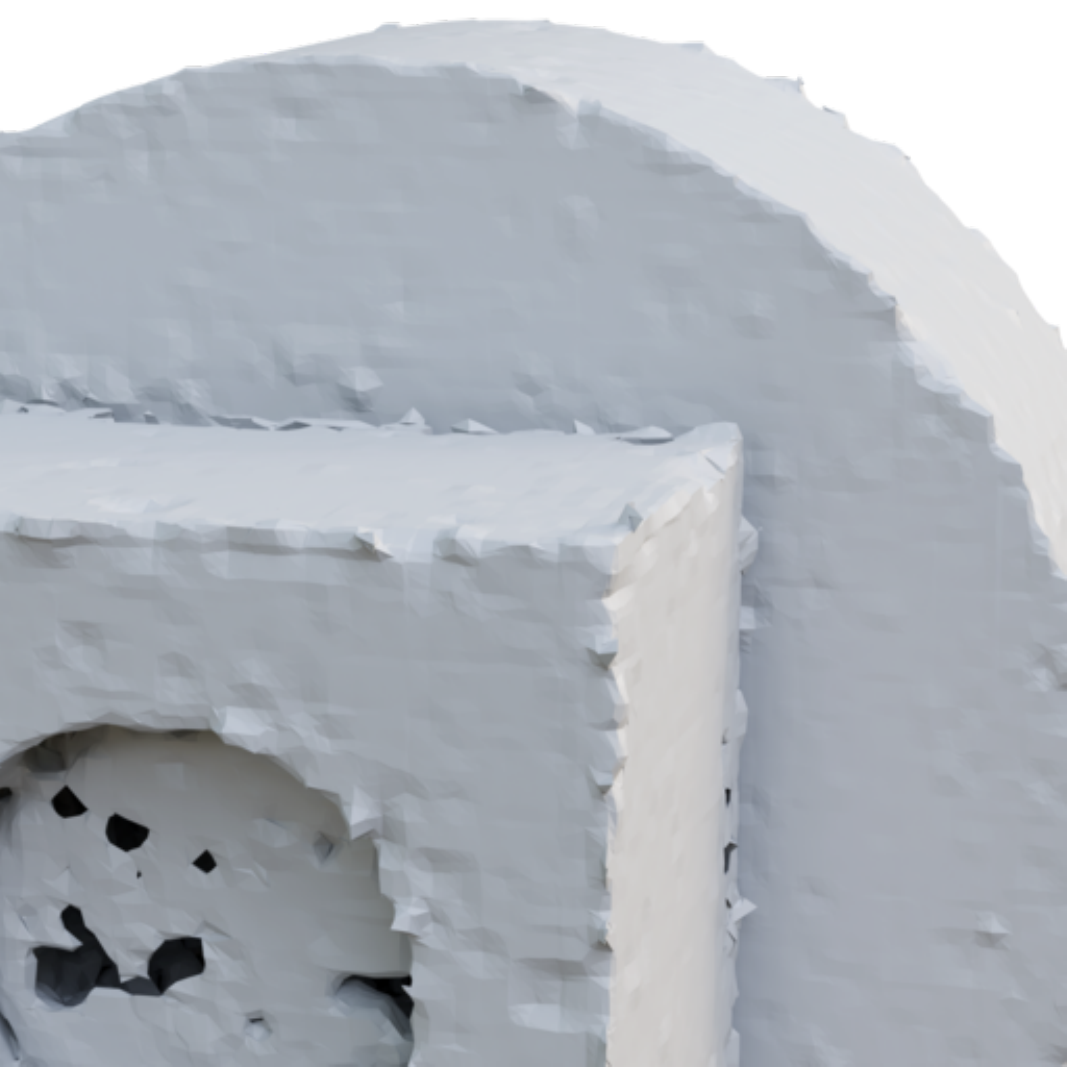}
    \includegraphics[width=0.12\linewidth]{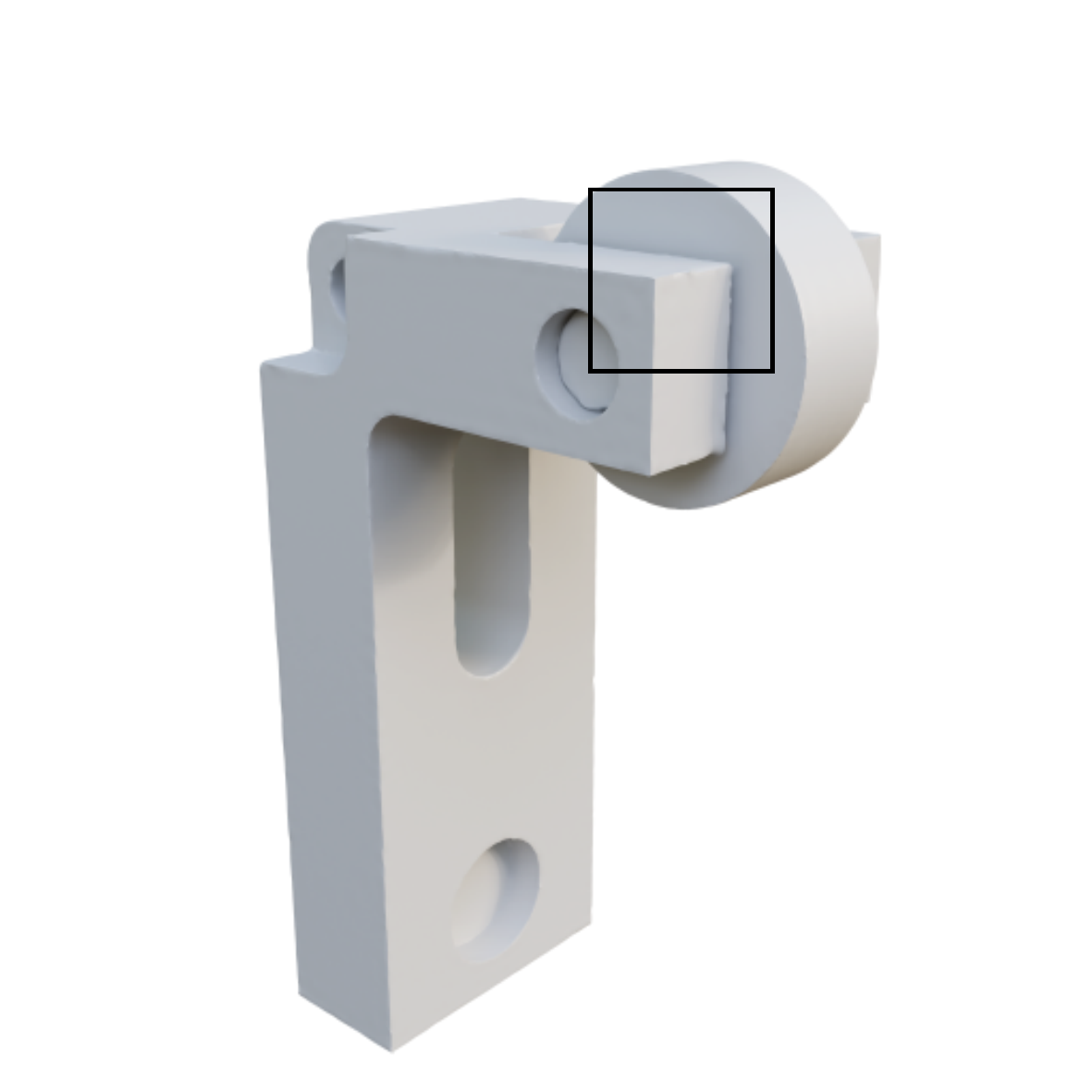}
    \includegraphics[width=0.12\linewidth]{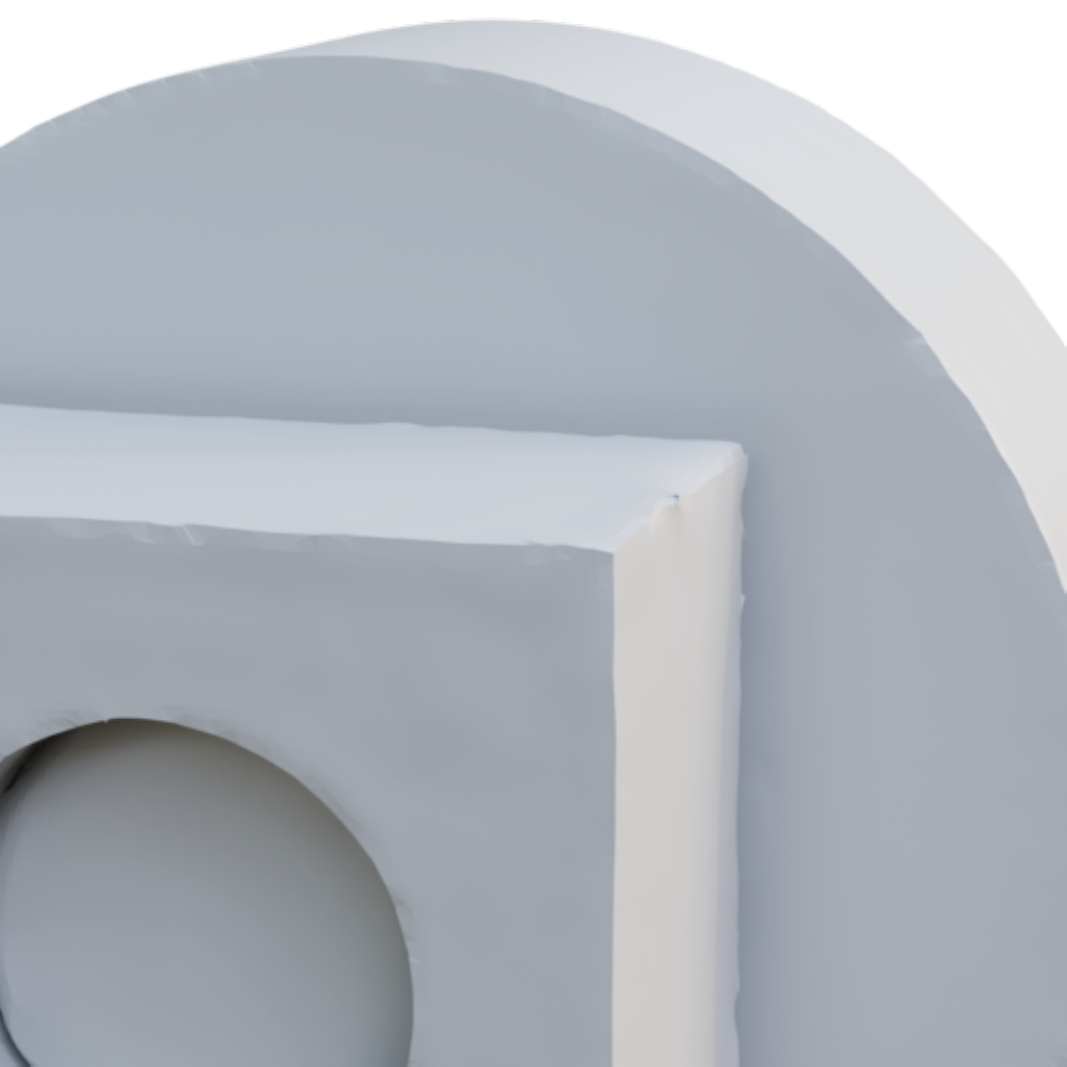}
    \includegraphics[width=0.12\linewidth]{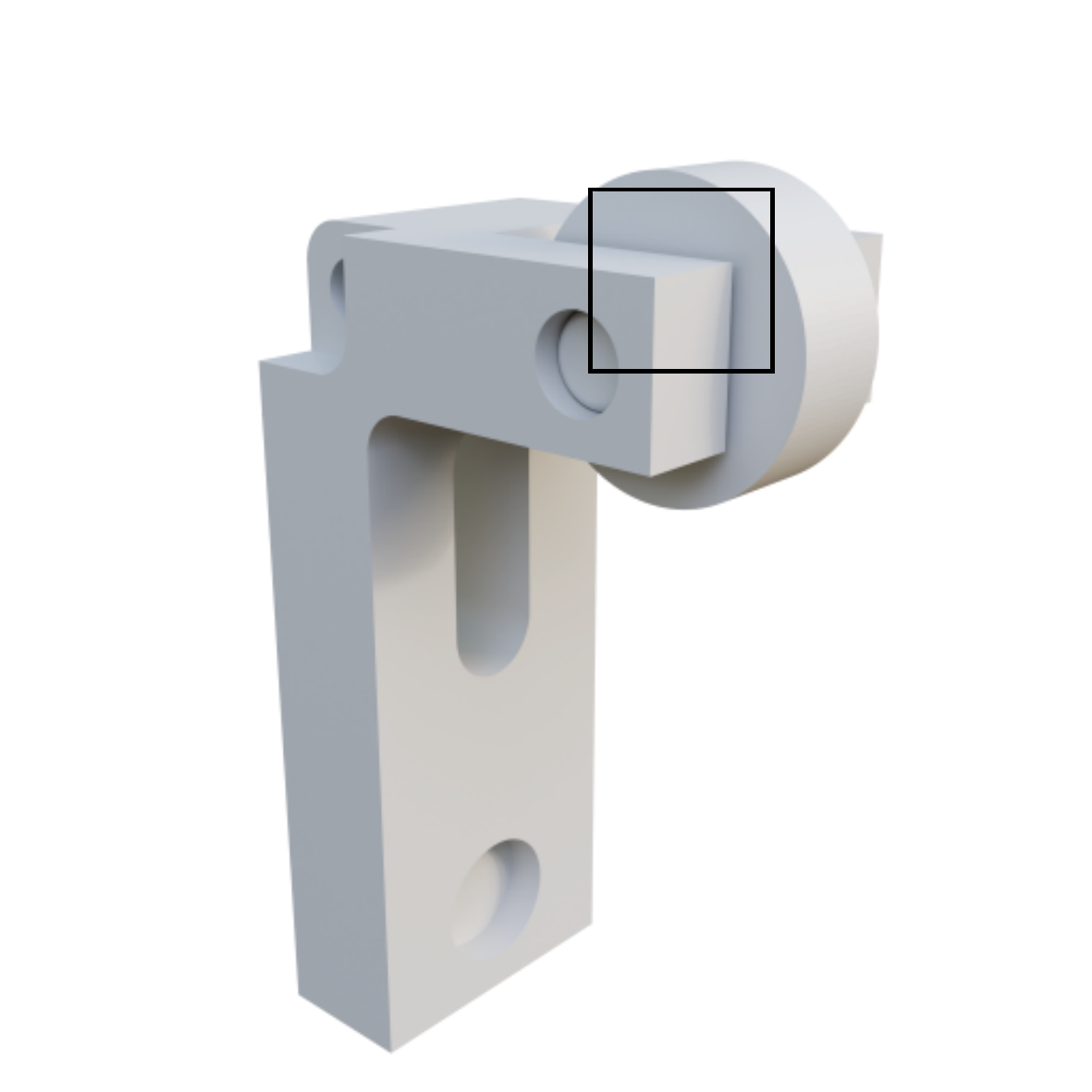}
    \includegraphics[width=0.12\linewidth]{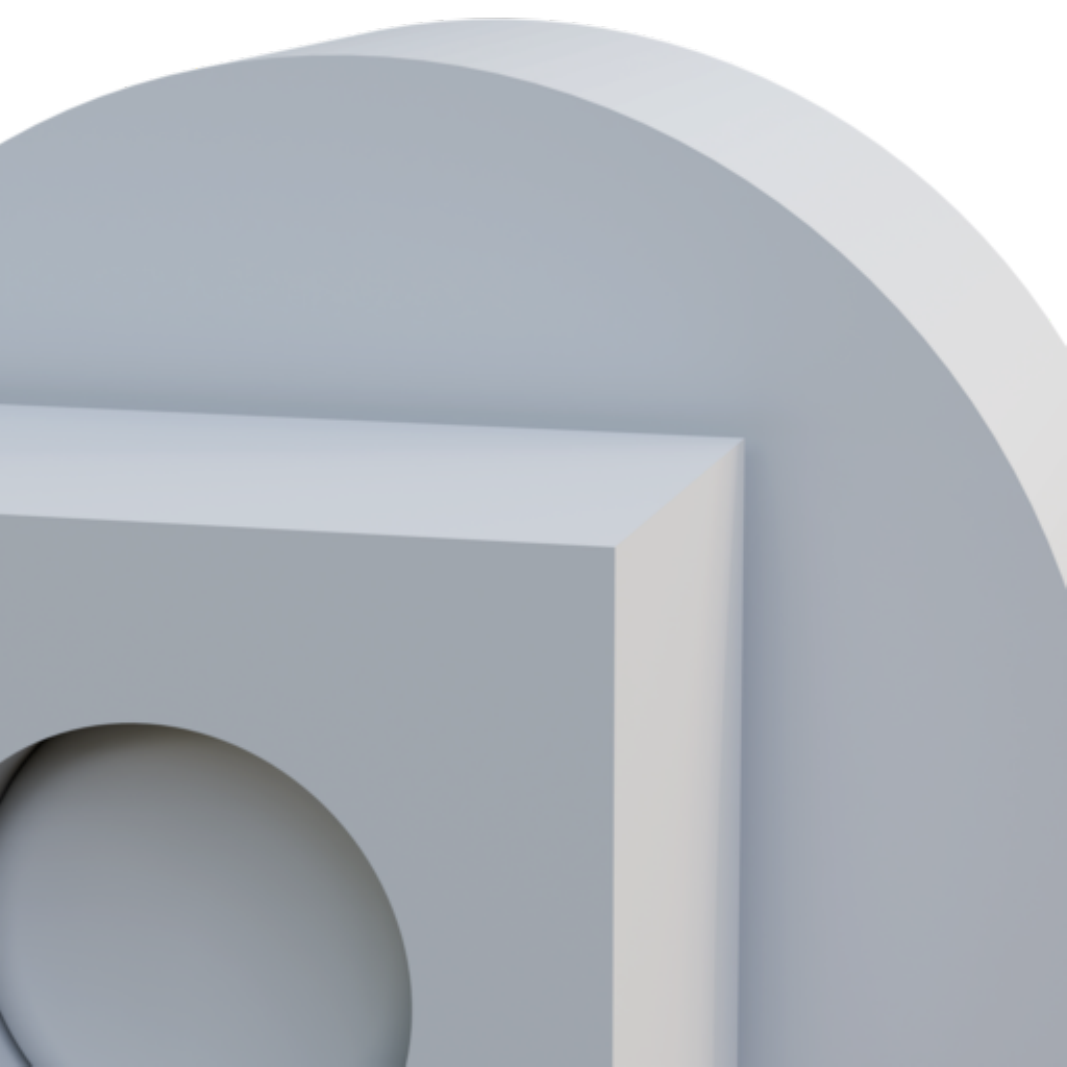}\\
    \makebox[0.24\linewidth]{\text{3.608}}
    \makebox[0.24\linewidth]{\text{3.511}}
    \makebox[0.24\linewidth]{\text{3.183}}
    \makebox[0.24\linewidth]{\text{-}}\\
    \vspace{6pt}
    \includegraphics[width=0.12\linewidth]{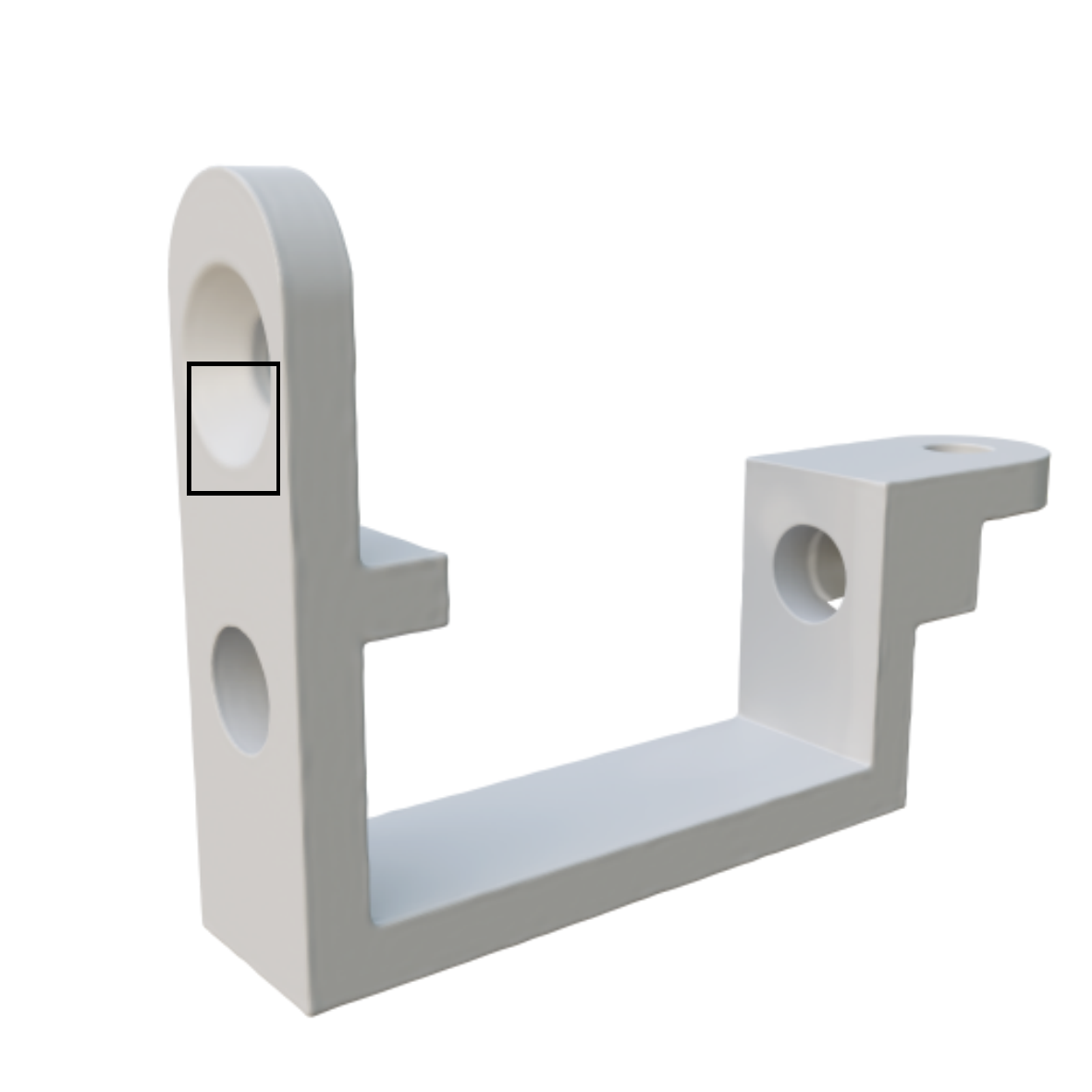}    \includegraphics[width=0.12\linewidth]{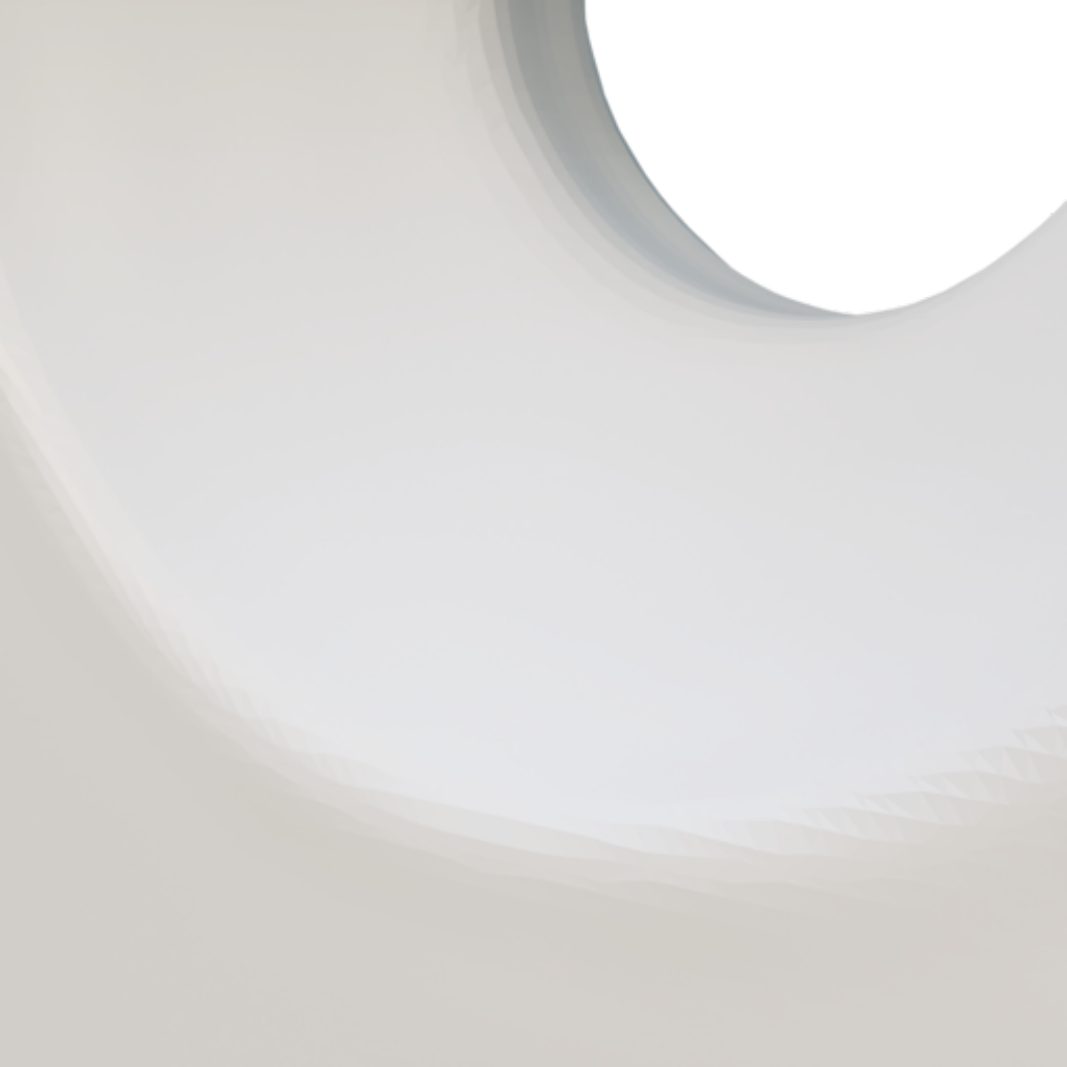}
    \includegraphics[width=0.12\linewidth]{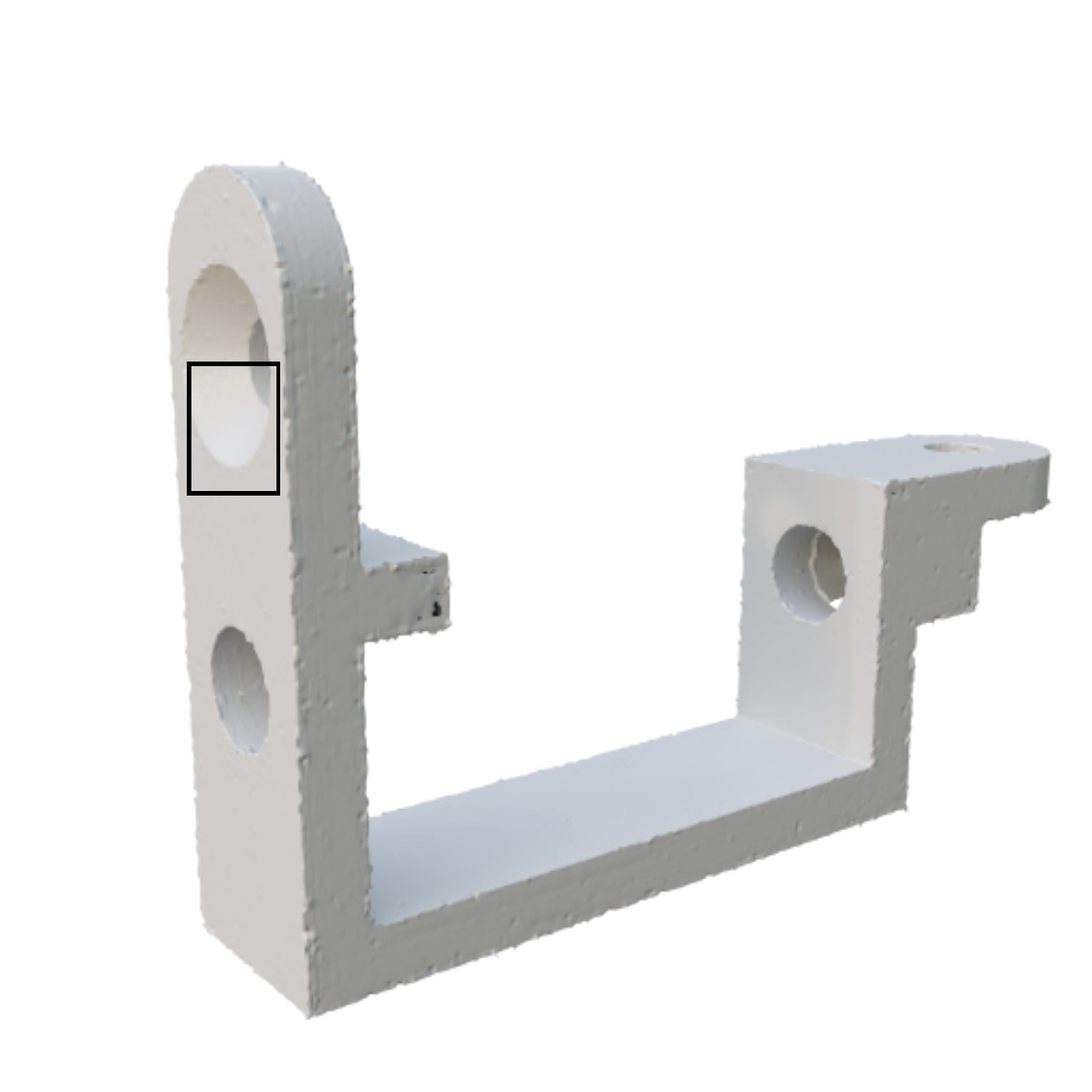}
    \includegraphics[width=0.12\linewidth]{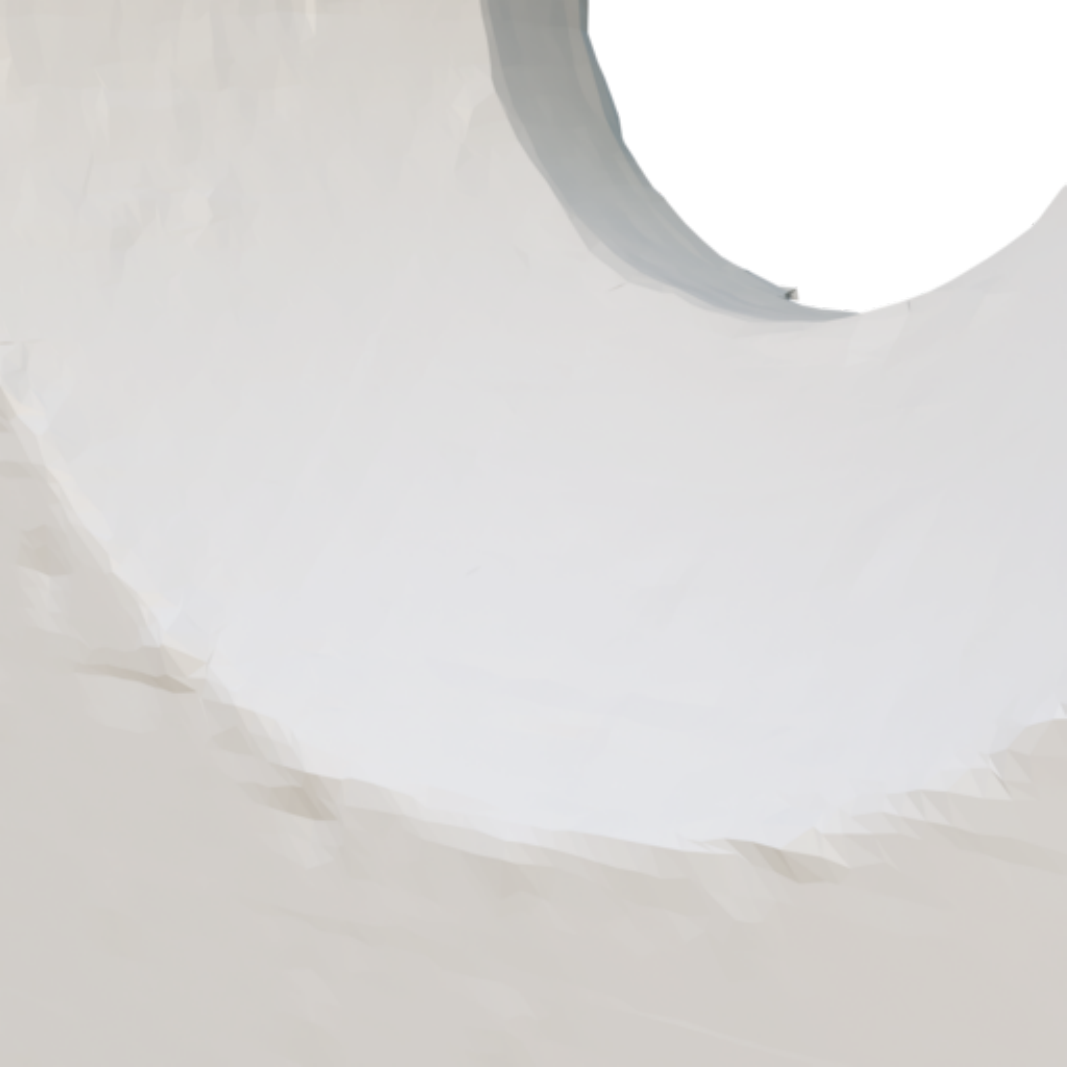}
    \includegraphics[width=0.12\linewidth]{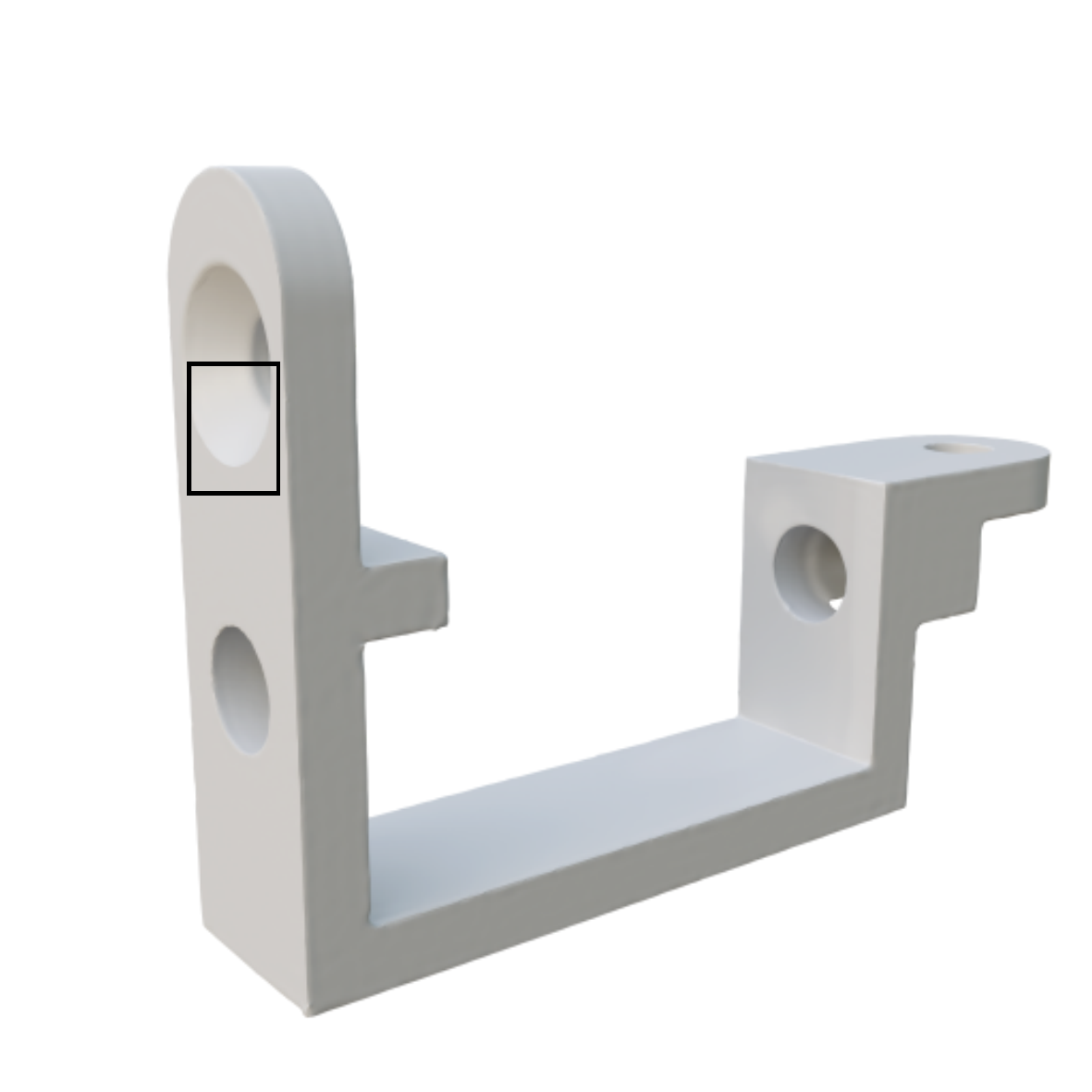}    \includegraphics[width=0.12\linewidth]{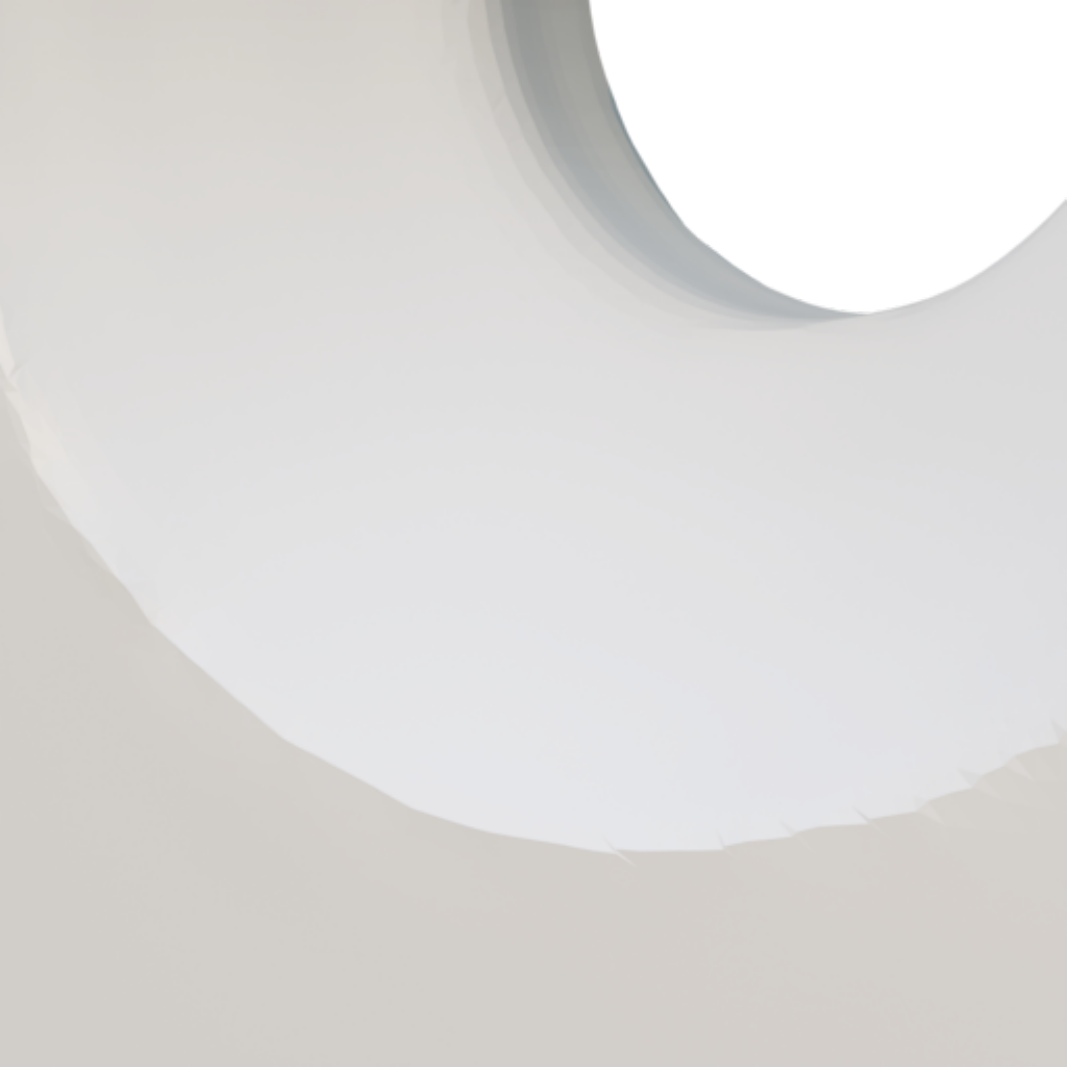}
    \includegraphics[width=0.12\linewidth]{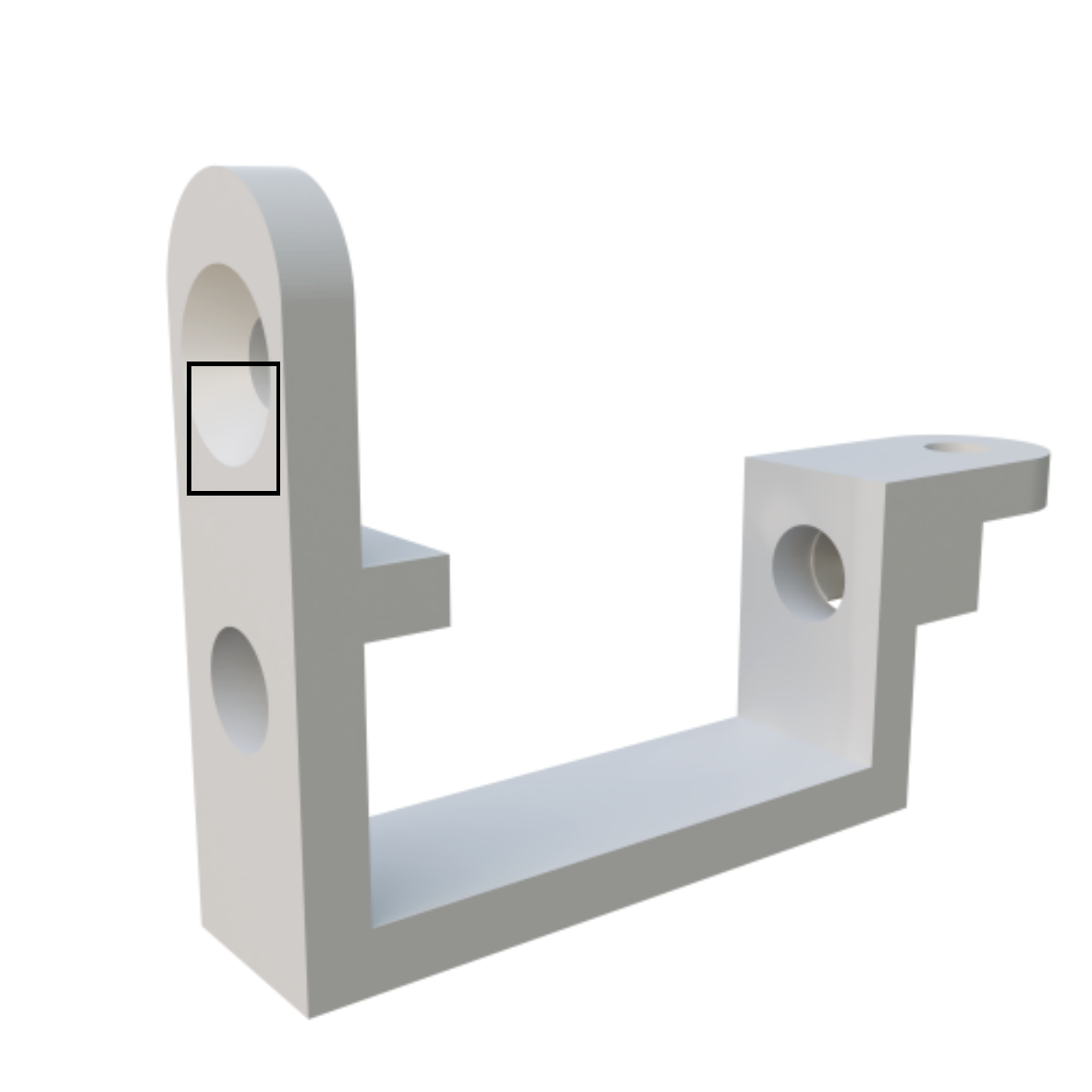}
    \includegraphics[width=0.12\linewidth]{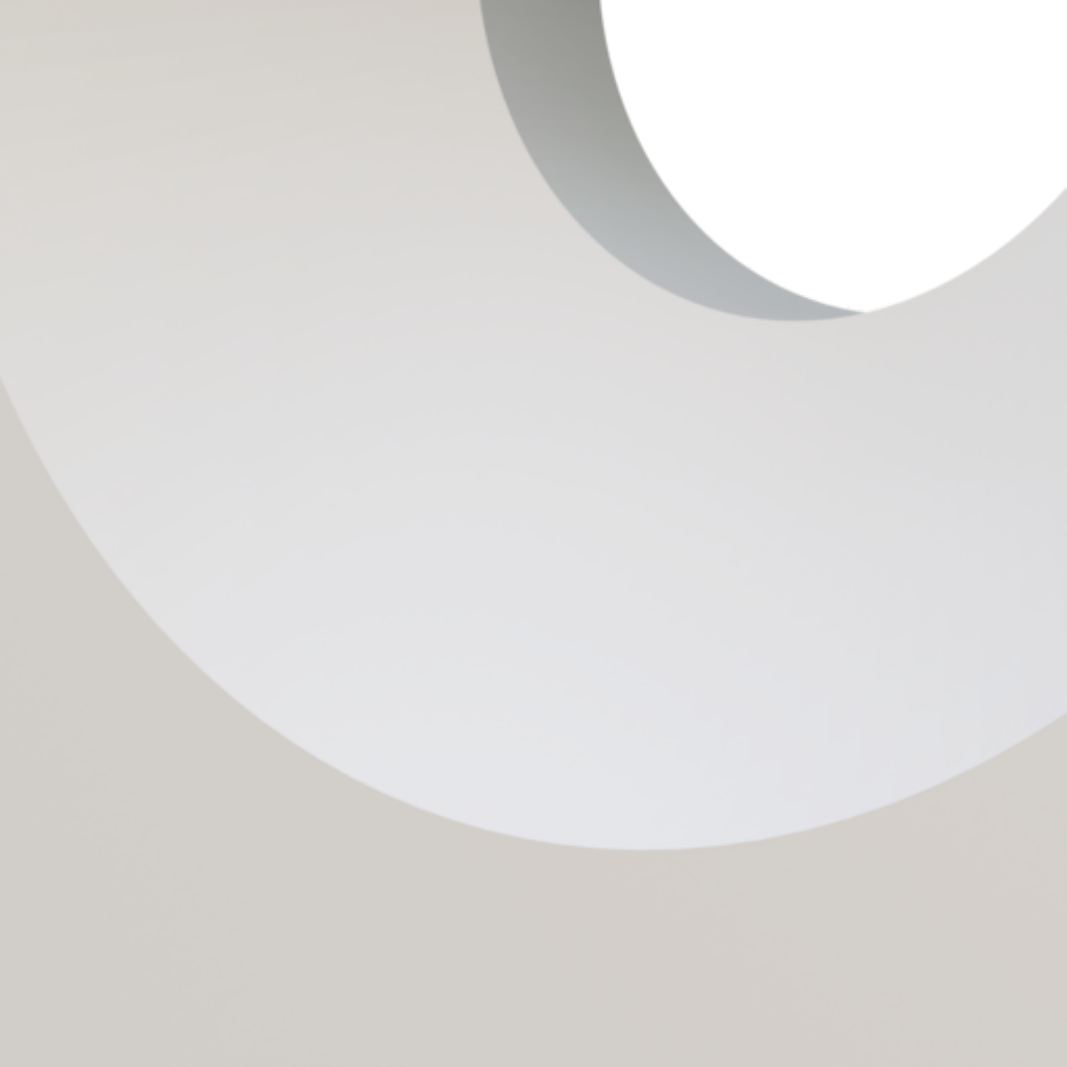}\\
    \makebox[0.24\linewidth]{\text{2.619}}
    \makebox[0.24\linewidth]{\text{2.456}}
    \makebox[0.24\linewidth]{\text{2.336}}
    \makebox[0.24\linewidth]{\text{-}}\\
    \vspace{6pt}
    \makebox[0.24\linewidth]{\sffamily SIREN}
    \makebox[0.24\linewidth]{\sffamily \ingp}
    \makebox[0.24\linewidth]{\sffamily \sharpnet}
    \makebox[0.24\linewidth]{\sffamily GT}
    \caption{CAD models reconstructed from points and their associated oriented normals. The CD for each complete model is shown below each model.}
    \label{fig:cad_pointcloud_with_normals}
\end{figure*}

\begin{figure}
\centering
\setlength\tabcolsep{2pt}
\begin{scriptsize}
\begin{tabular}{ccccc}

    \raisebox{0.095\linewidth}{\rotatebox[origin=c]{90}{\sffamily\siren}} &%
    \includegraphics[width=0.19\linewidth]{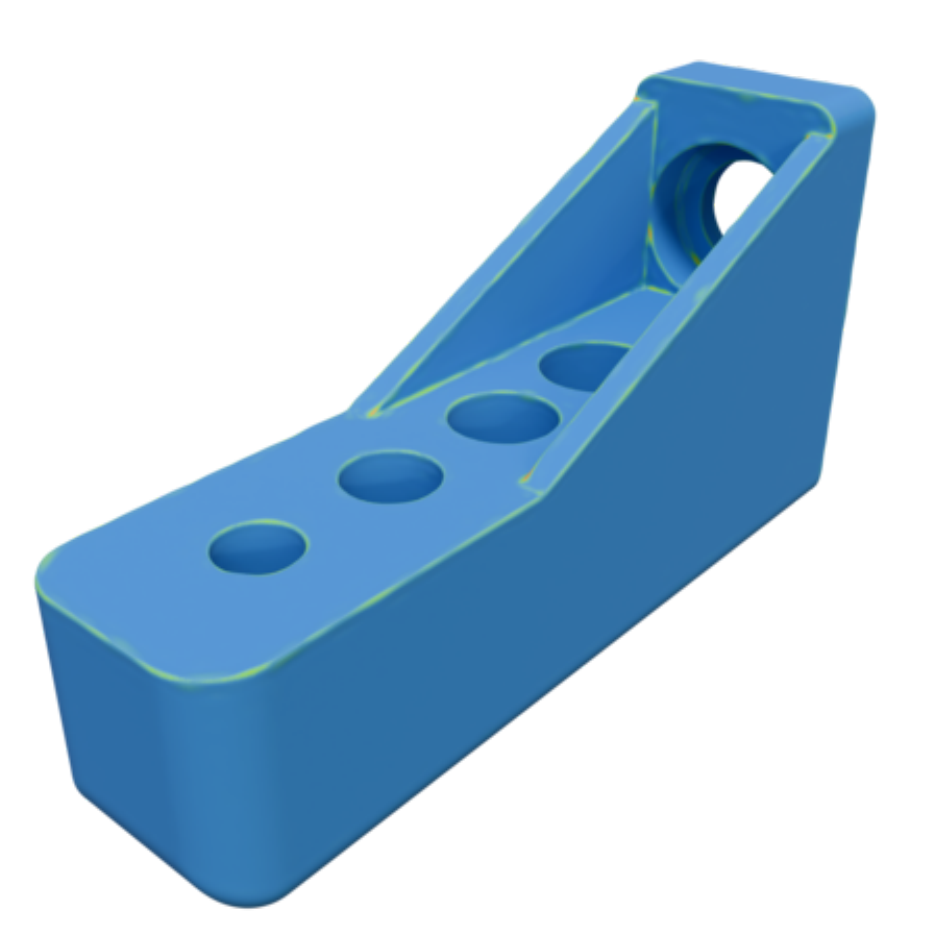} &%
    \includegraphics[width=0.19\linewidth]{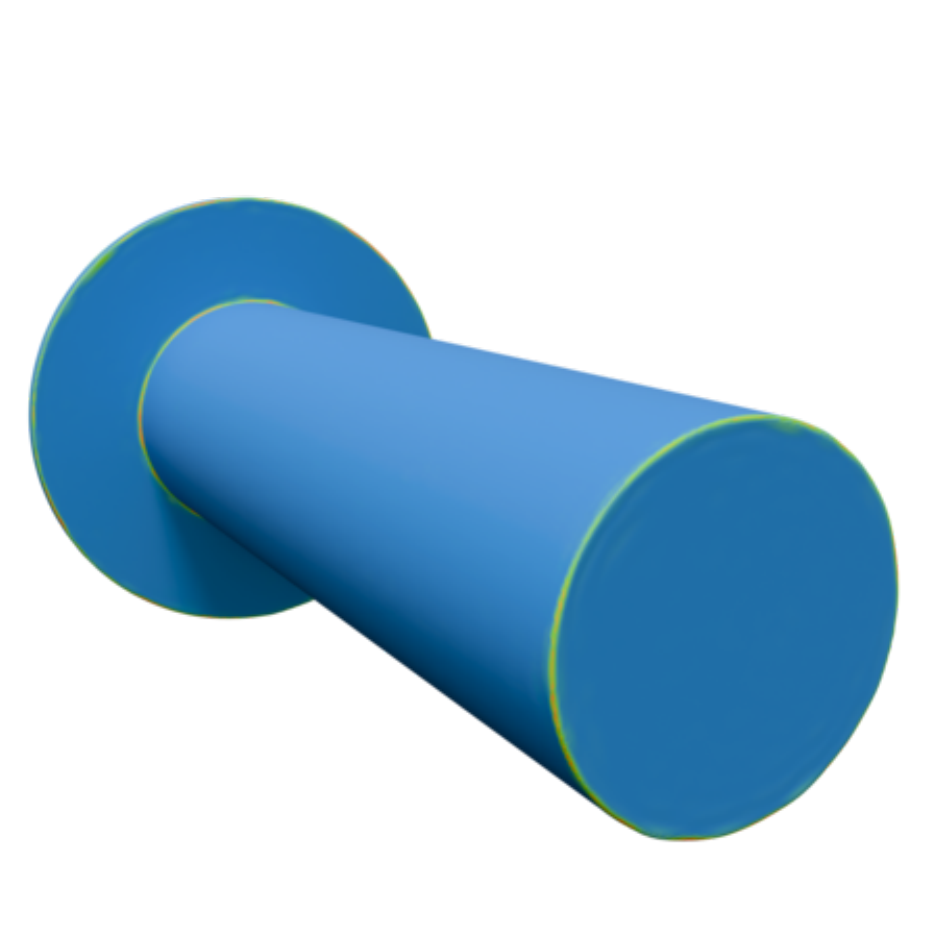} &%
    \includegraphics[width=0.19\linewidth]{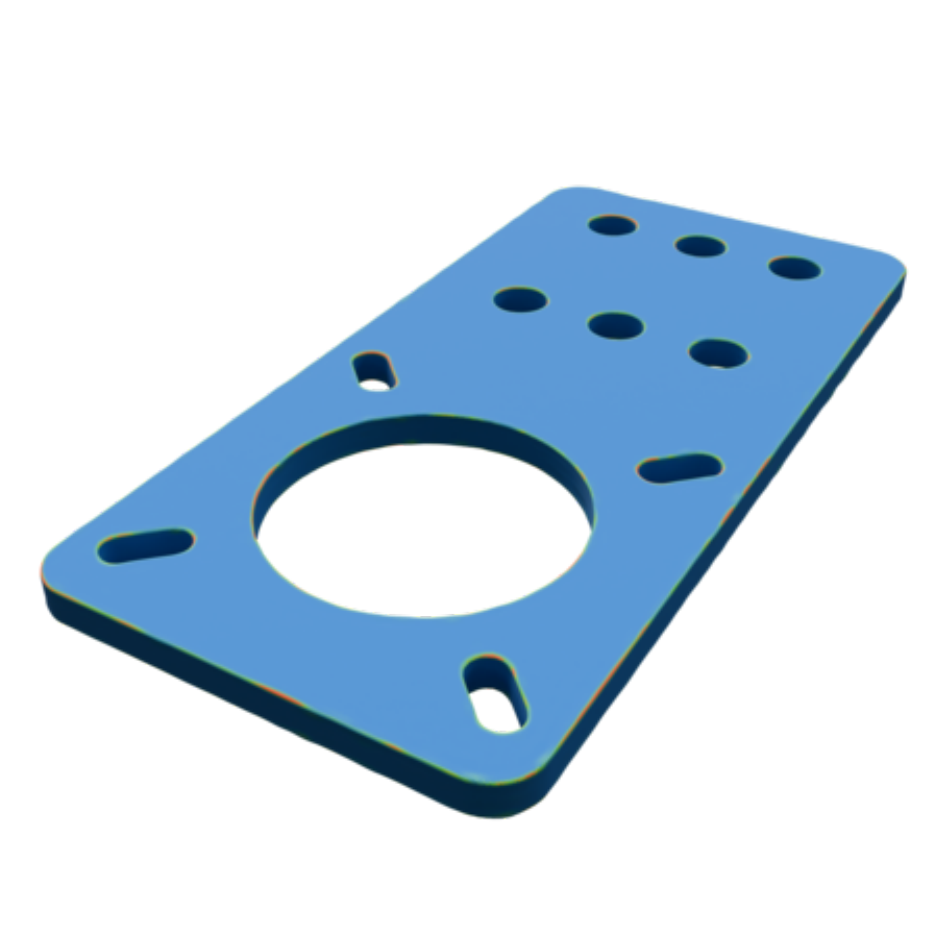} &%
    \includegraphics[width=0.19\linewidth]{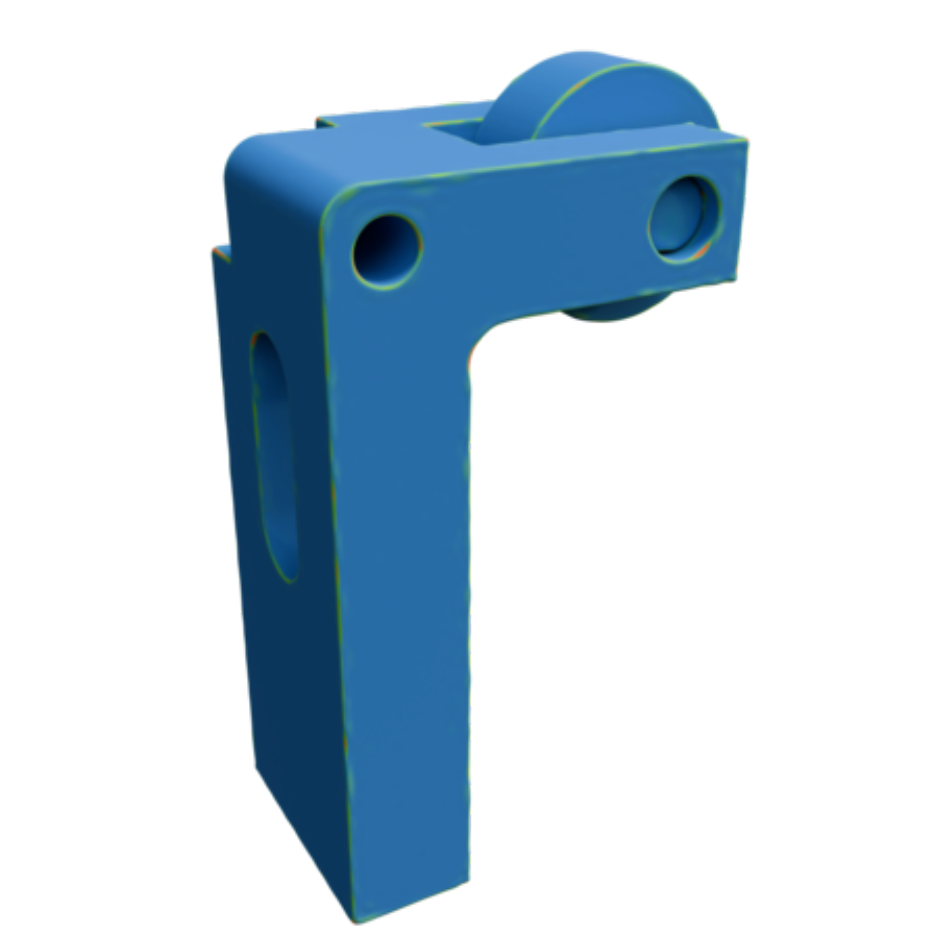} \\
    \makebox[0.01\linewidth]{} &%
    \makebox[0.22\linewidth]{3.424} &%
    \makebox[0.22\linewidth]{2.774} &%
    \makebox[0.22\linewidth]{2.506} &%
    \makebox[0.22\linewidth]{3.608} \\

    \raisebox{0.095\linewidth}{\rotatebox[origin=c]{90}{\sffamily\ingp}} &%
    \includegraphics[width=0.19\linewidth]{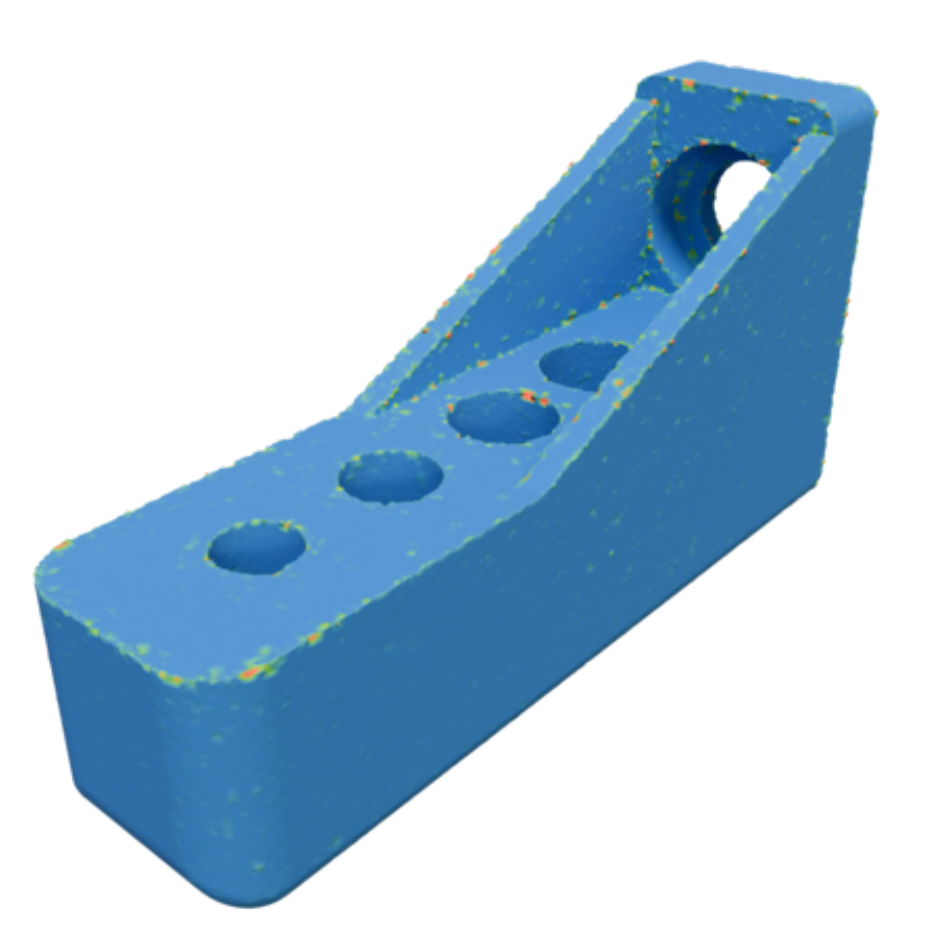} &%
    \includegraphics[width=0.19\linewidth]{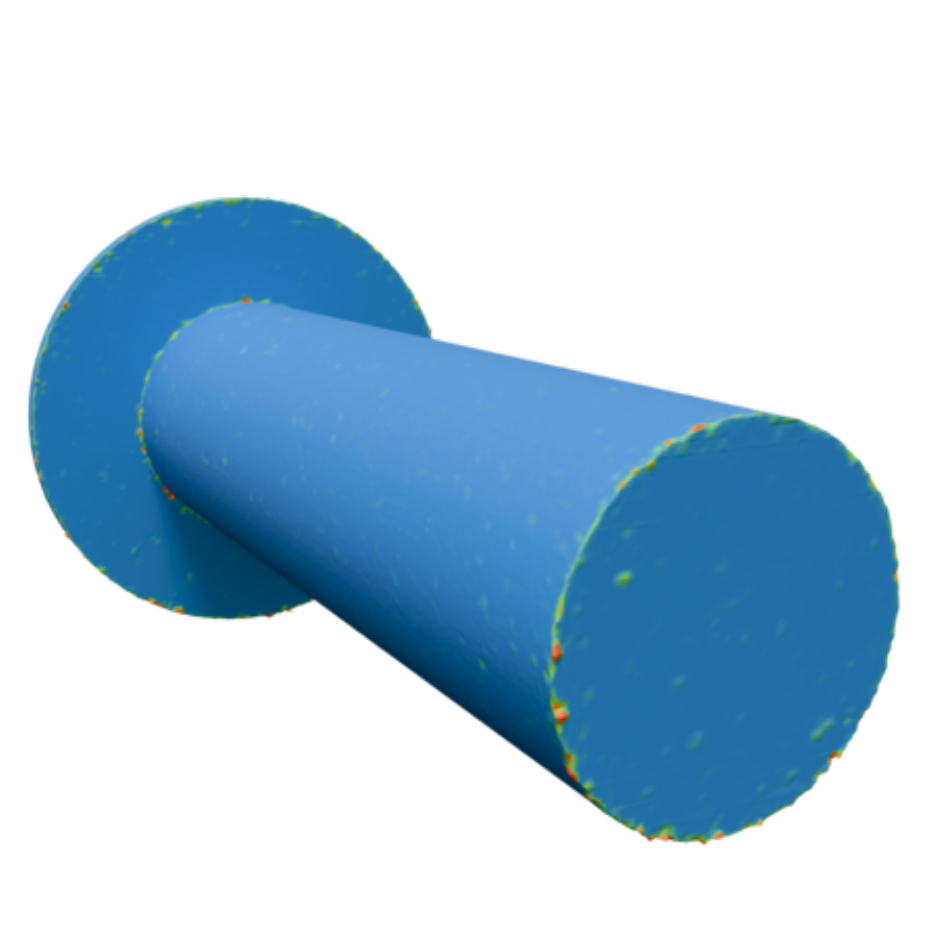} &%
    \includegraphics[width=0.19\linewidth]{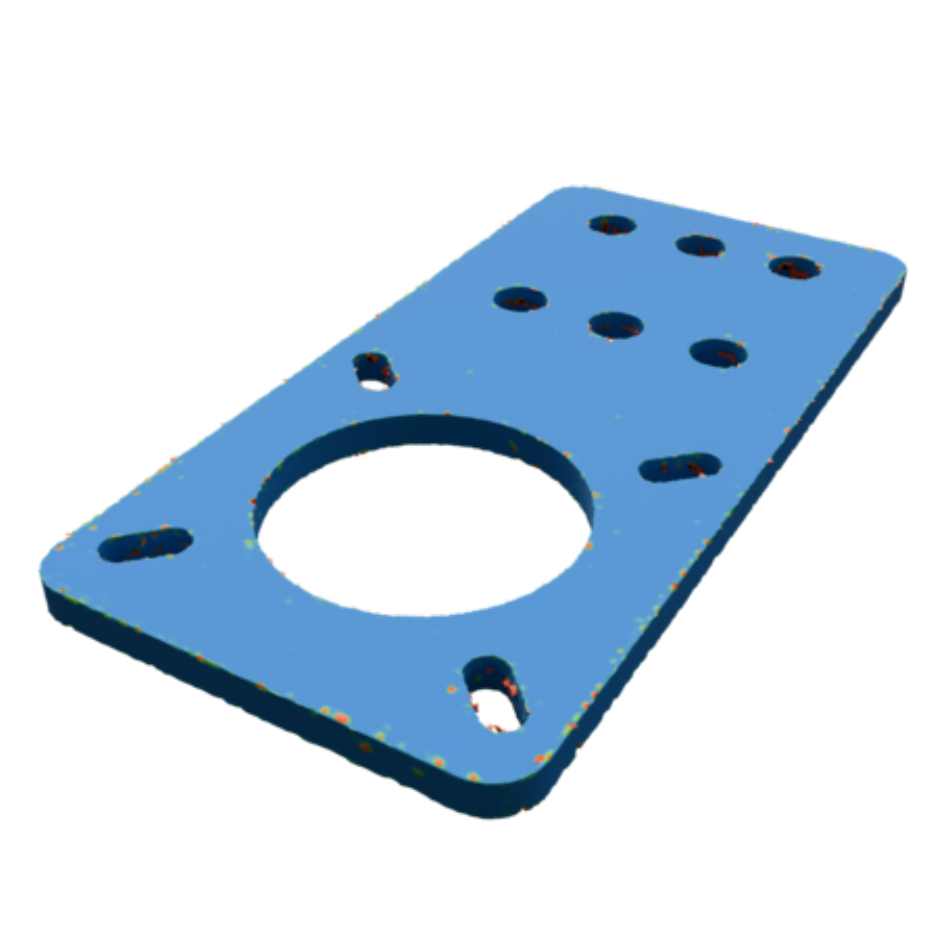} &%
    \includegraphics[width=0.19\linewidth]{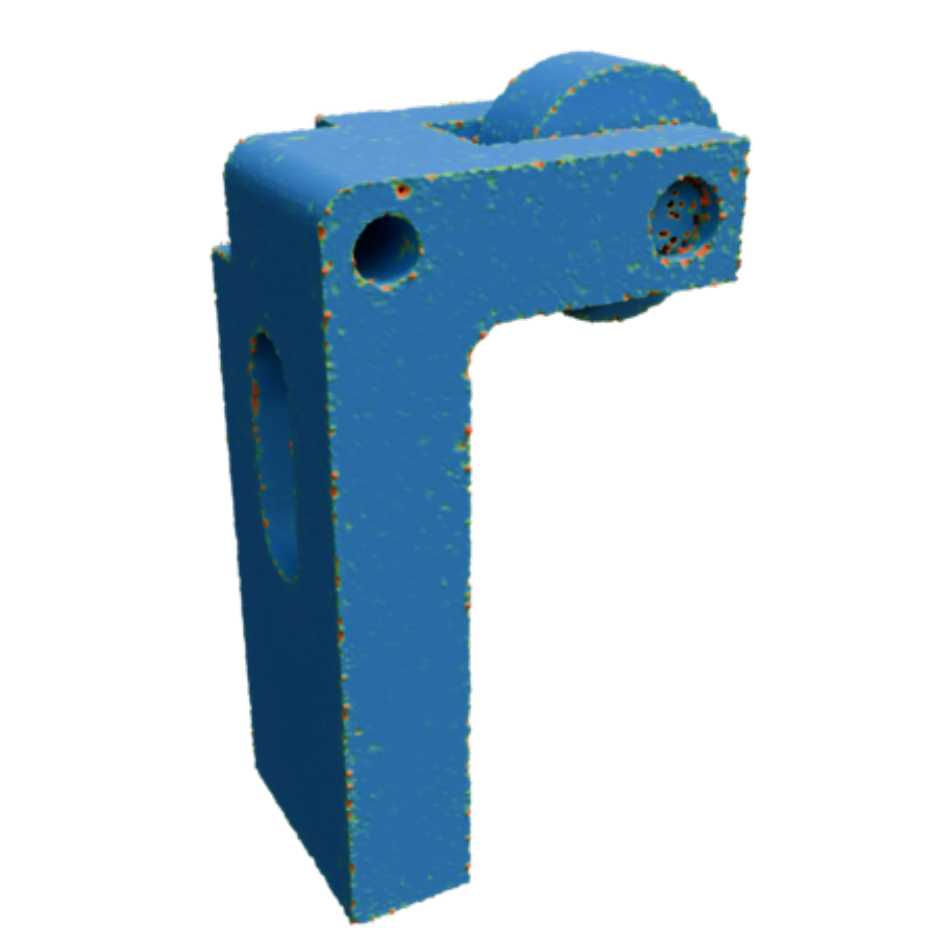} \\
    \makebox[0.01\linewidth]{} &%
    \makebox[0.22\linewidth]{3.076} &%
    \makebox[0.22\linewidth]{2.478} &%
    \makebox[0.22\linewidth]{2.397} &%
    \makebox[0.22\linewidth]{3.511} \\

    \raisebox{0.095\linewidth}{\rotatebox[origin=c]{90}{\sffamily\sharpnet}} &%
    \includegraphics[width=0.19\linewidth]{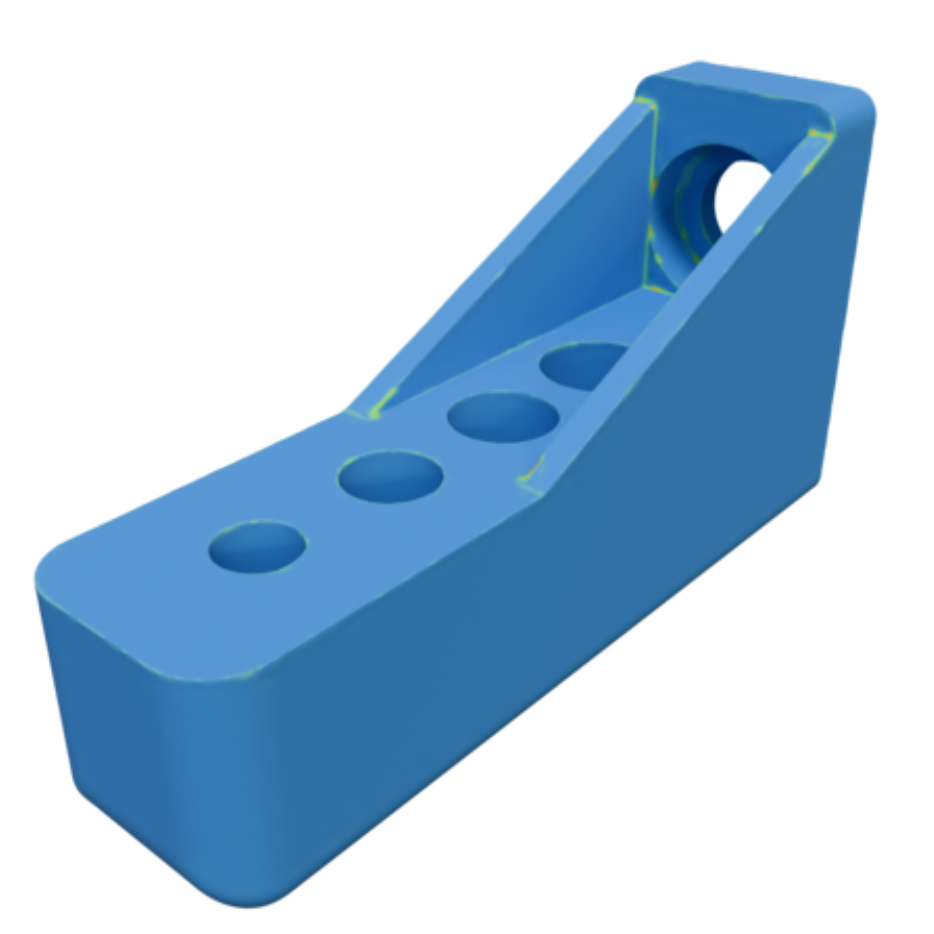} &%
    \includegraphics[width=0.19\linewidth]{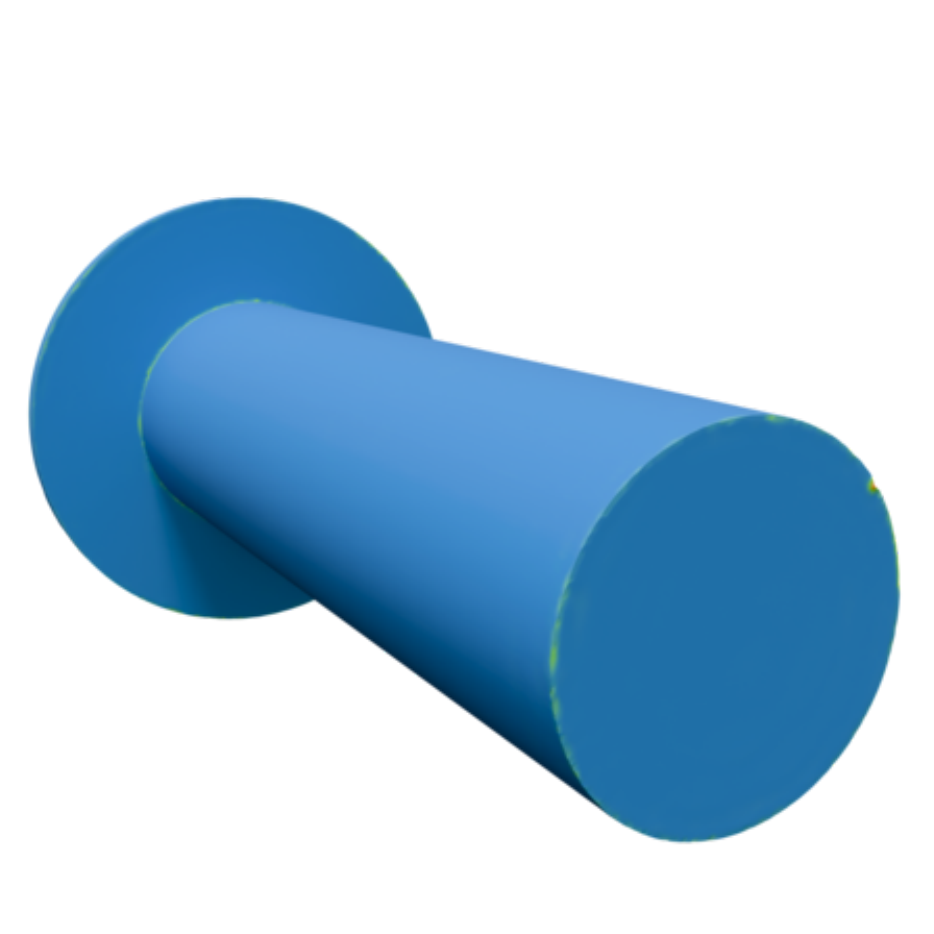} &%
    \includegraphics[width=0.19\linewidth]{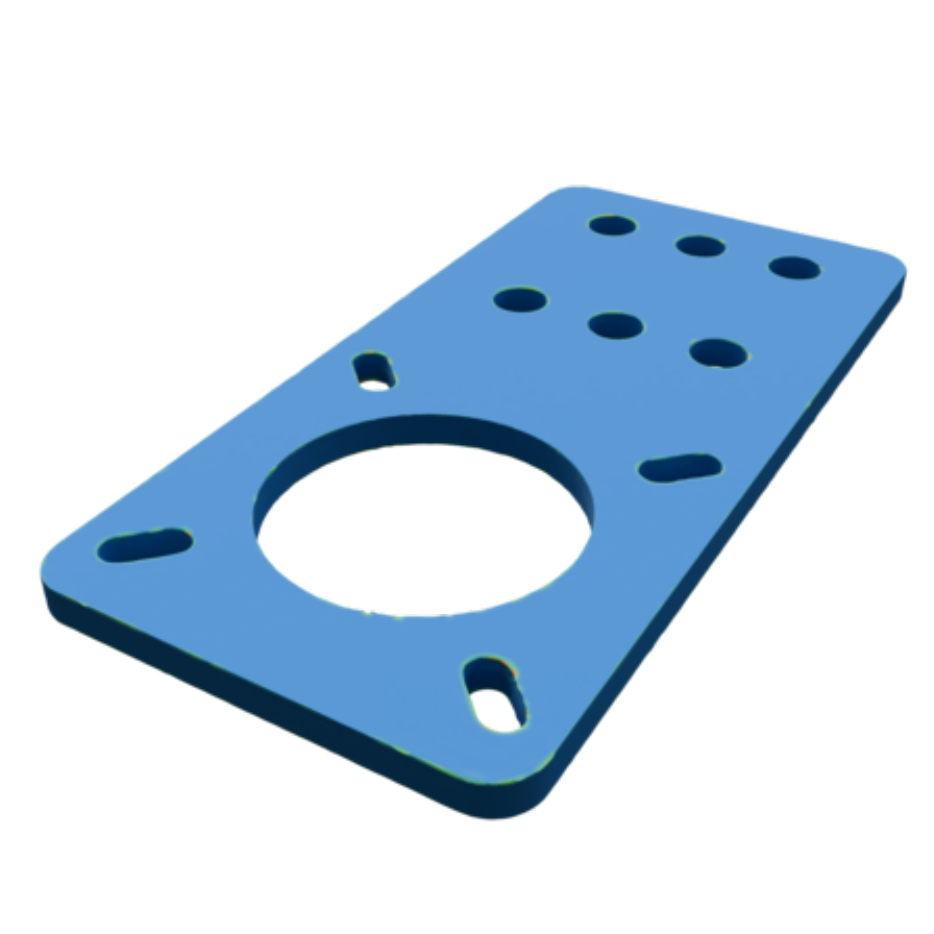} &%
    \includegraphics[width=0.19\linewidth]{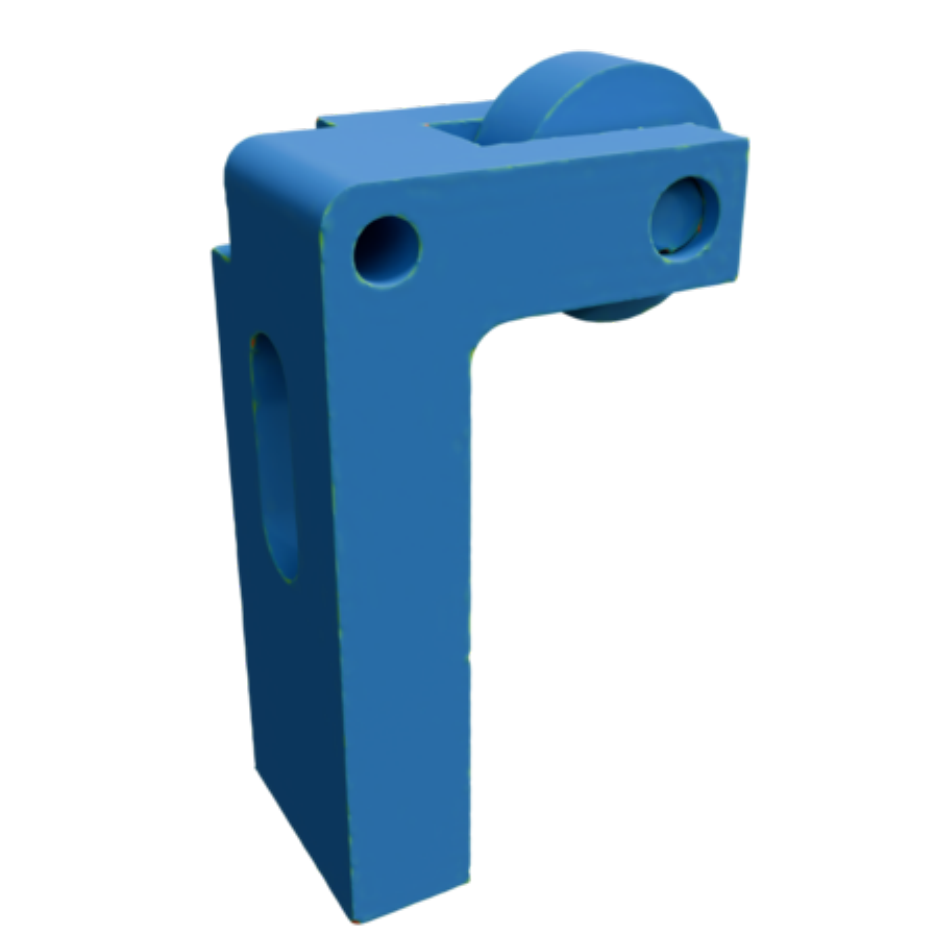} \\
    \makebox[0.01\linewidth]{} &%
    \makebox[0.22\linewidth]{3.053} &%
    \makebox[0.22\linewidth]{2.471} &%
    \makebox[0.22\linewidth]{2.209} &%
    \makebox[0.22\linewidth]{3.183} \\

\end{tabular}
\end{scriptsize}
\caption{Illustration of CD errors by color map. The CD is shown below each model. {\ingp} produces many noisy points, particularly in regions around sharp edges.}
\label{fig:normal_input_color}
\end{figure}

As illustrated in Figure~\ref{fig:cad_pointcloud_with_normals}, when points and oriented normals instead of ground truth distances are used as constraints, the surfaces generated by {\ingp}~\cite{Muller2022INGP} clearly exhibit bumpiness, because {\ingp} adopts multiresolution hash grid encoding with linear interpolation resulting in only $C^0$-continuity at grid boundaries. Figure~\ref{fig:normal_input_color} visualizes CD errors using a color map, revealing that the {\ingp} errors are especially large near sharp edges.

\begin{figure}
\begin{subcaptionblock}{.32\linewidth}
    \centering
    \includegraphics[width=\linewidth]{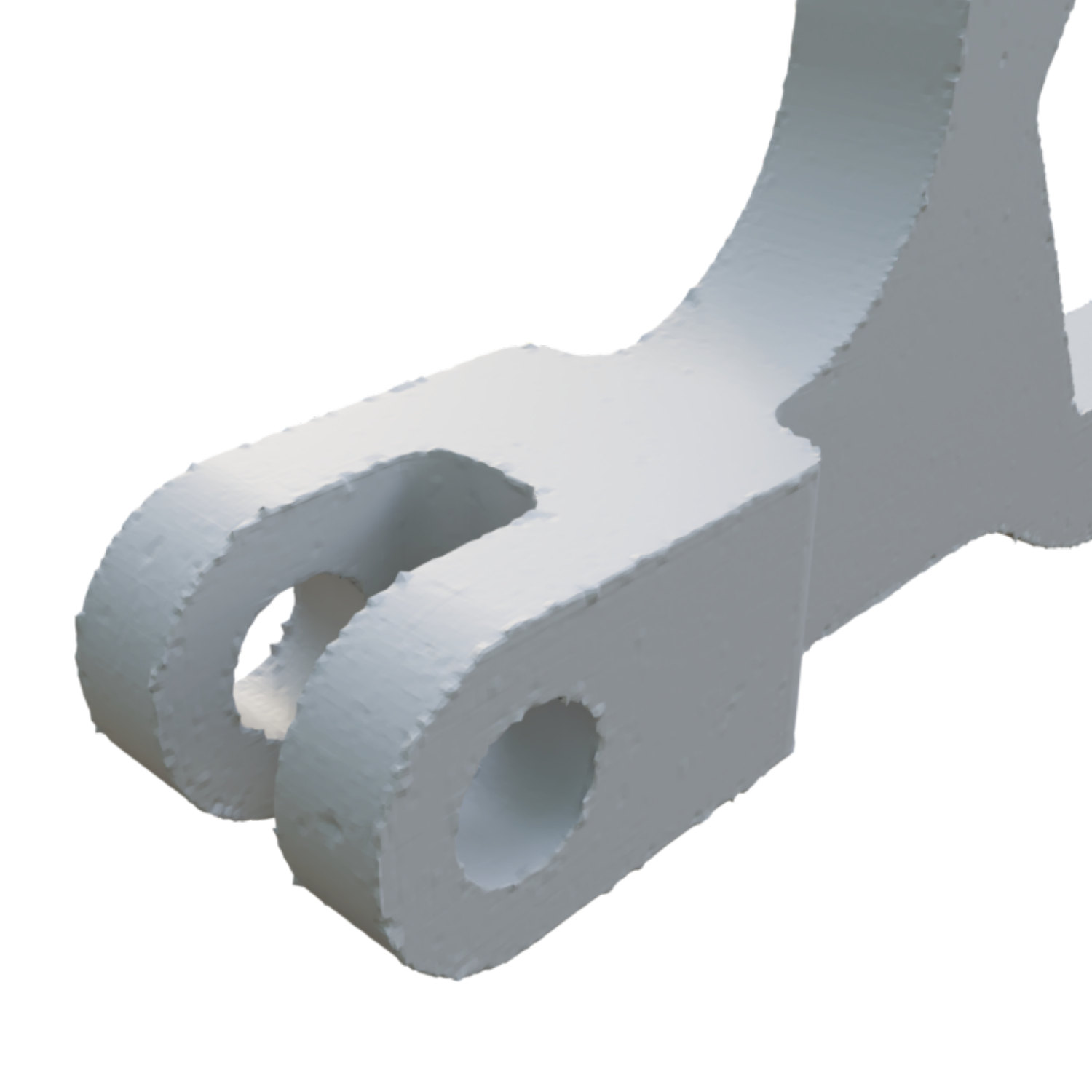}\\[.0625\linewidth]
    \includegraphics[width=\linewidth]{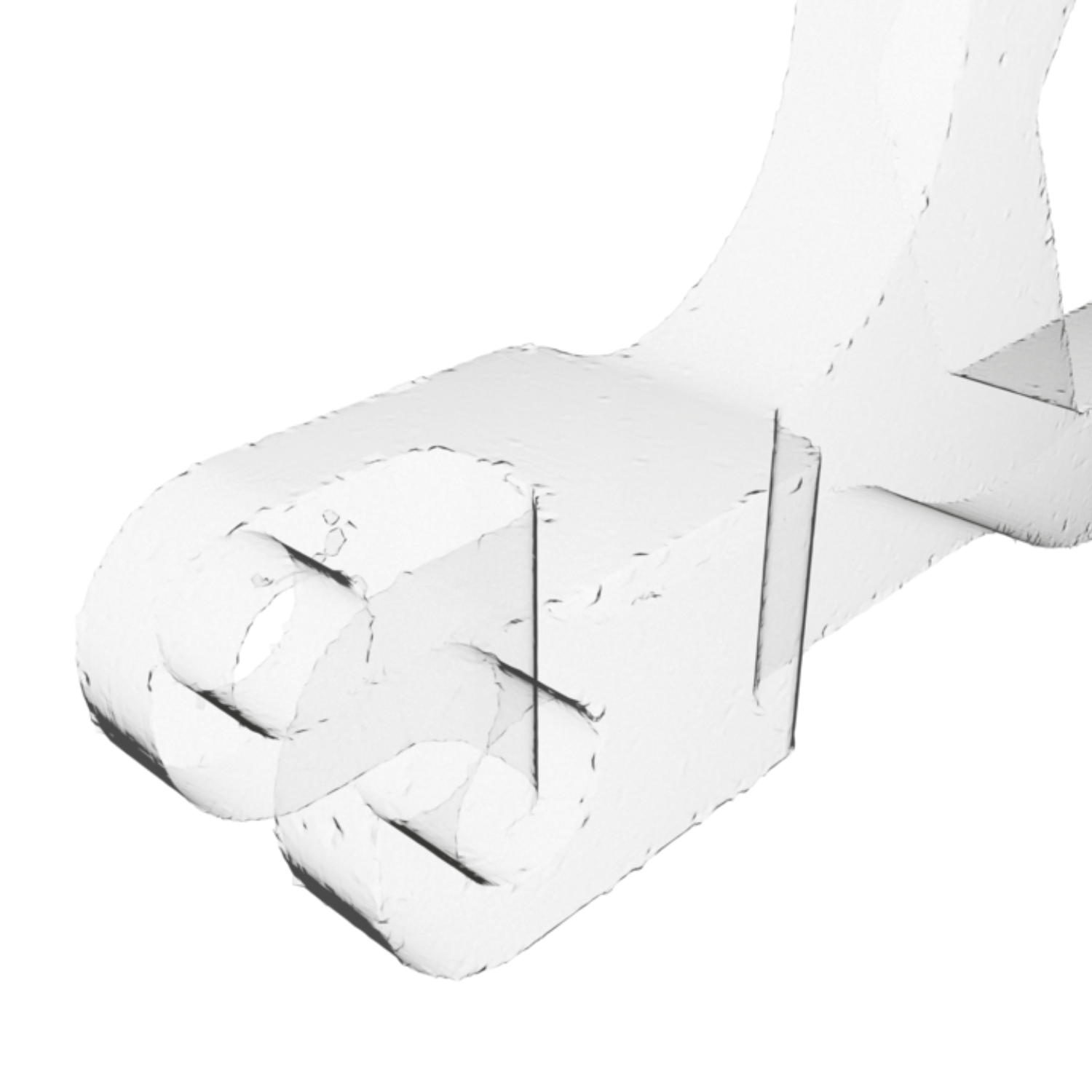}
    \caption{}
    \label{fig:instant_ngp_wo_normal (normal)}
\end{subcaptionblock}\hfill%
\begin{subcaptionblock}{.32\linewidth}
    \centering
    \includegraphics[width=\linewidth]{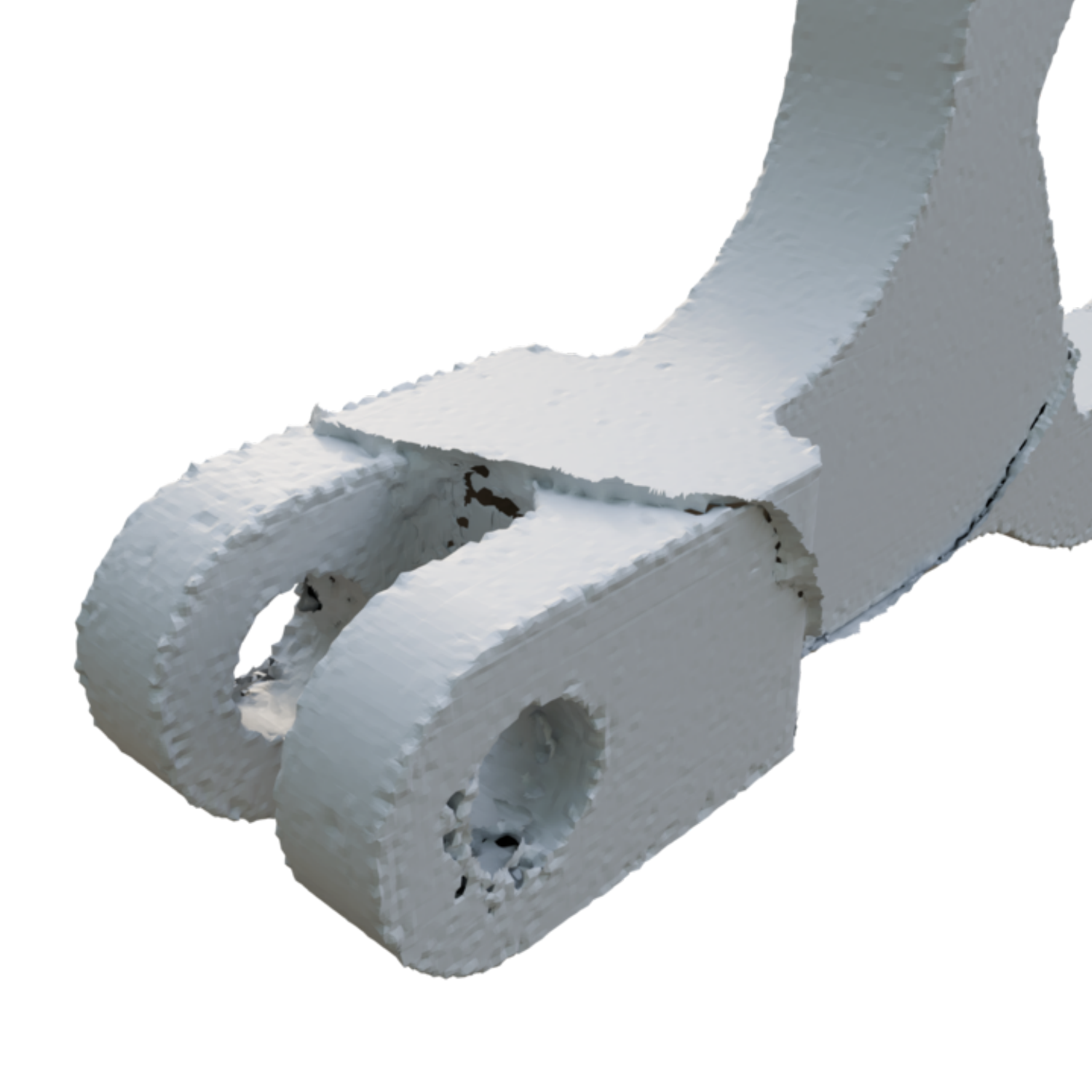}\\[.0625\linewidth]
    \includegraphics[width=\linewidth]{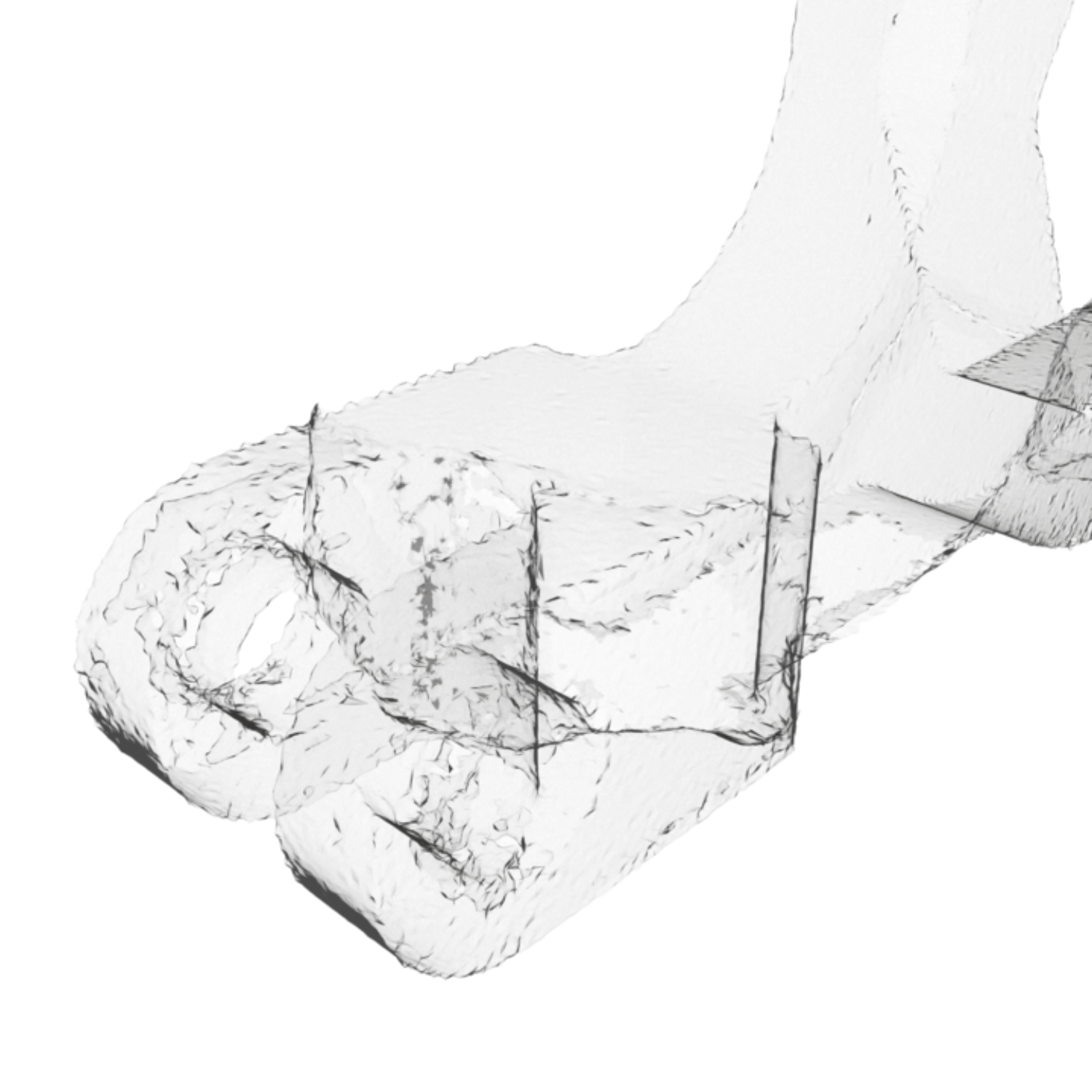}
    \caption{}
    \label{fig:instant_ngp_wo_normal (pointcloud)}
\end{subcaptionblock}\hfill%
\begin{subcaptionblock}{.32\linewidth}
    \centering
    \includegraphics[width=\linewidth]{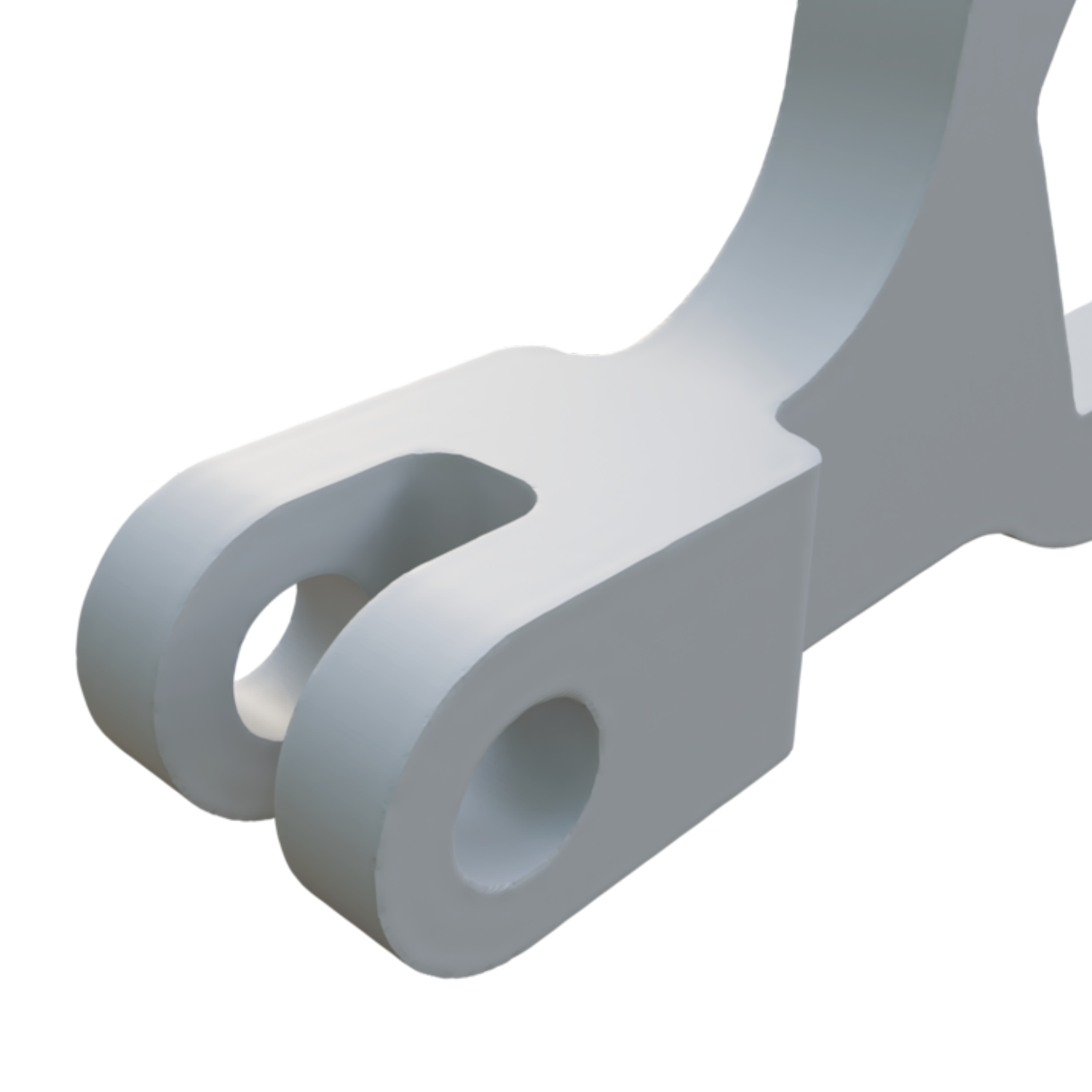}\\[.0625\linewidth]
    \includegraphics[width=\linewidth]{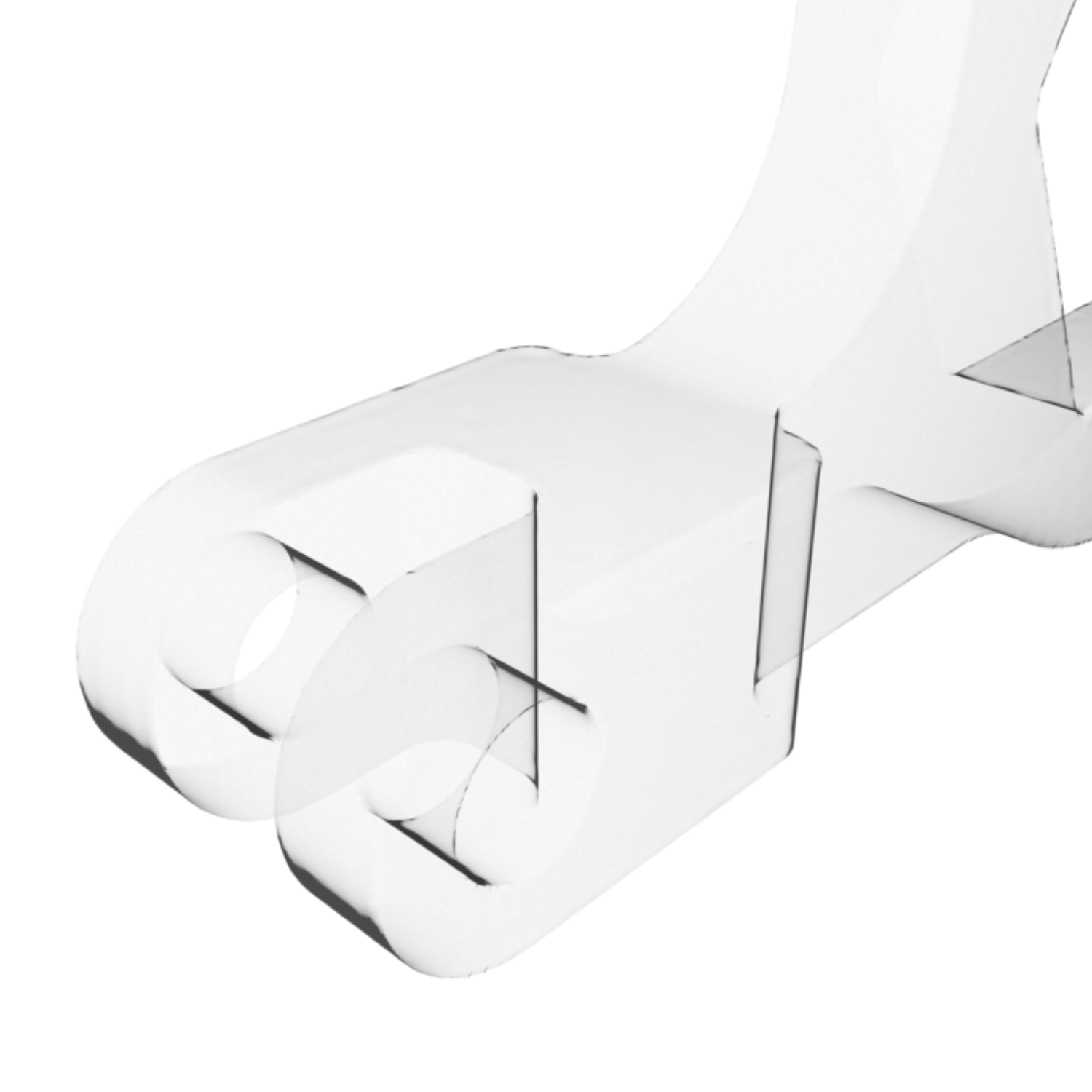}
    \caption{}
    \label{fig:instant_ngp_wo_normal (sharp)}
\end{subcaptionblock}
\caption{%
(\subref{fig:instant_ngp_wo_normal (normal)})~{\ingp}~\cite{Muller2022INGP} achieves good results under ground-truth SDF supervision, but in point-cloud reconstruction tasks, even with normal supervision, its lack of direct local feature correlation often leads to small artifacts away from the sampled points. 
(\subref{fig:instant_ngp_wo_normal (pointcloud)})~Without oriented-normal supervision, {\ingp} has a high probability of producing local normal flips, causing surface breakage at the boundary of normal flipped region, or degenerating into learning a UDF, resulting in missing or double-layered surfaces. 
(\subref{fig:instant_ngp_wo_normal (sharp)})~In contrast, {\sharpnet}, equipped with learnable features, performs robustly even without normal supervision.}
\label{fig:instant_ngp_without_normal}
\end{figure}

As illustrated in Figure~\ref{fig:instant_ngp_without_normal}, when oriented normal supervision is removed, the performance of {\ingp} degrades significantly, resulting in a mixture of UDF and SDF, because its grid features are spatially localized and lack awareness of the overall geometry. A detailed analysis of {\sharpnet} trained without normal supervision is provided in Section~\ref{sec:CAD_points}. Owing to its MLP-based architecture, {\sharpnet} is well-suited for learning SDFs from point-only input.

The statistical metrics for the 100 testing models are summarized in Table~\ref{tab:points_normals_input}. {\sharpnet} clearly surpasses both SIREN and {\ingp} in performance.

\begin{table}[!htbp]
\caption{Quantitative statistics of models reconstructed from point clouds with oriented normals.}
\label{tab:points_normals_input}
\begin{tabular}{ccccc}
\toprule
Method & $\cramped{\text{CD}^{\times 10^{-3}}}$ $\downarrow$  & $\cramped{\text{HD}^{\times 10^{-2}}}$ $\downarrow$ & $\text{NE}^{\circ}$ $\downarrow$ & $\text{FC}^\%$ $\uparrow$ \\
\midrule
{\siren}        & 4.532 & 4.119 & 4.524 & 96.28\\
{\ingp}  & 4.214 & 3.325 & 8.557 & 95.88\\
{\sharpnet}     & \textbf{3.807} & \textbf{2.234} & \textbf{2.918} & \textbf{98.09} \\
\bottomrule
\end{tabular} 
\end{table}

\subsection{Reconstruction from Raw Points}
\label{sec:CAD_points}

To further evaluate our method, we perform CAD reconstruction using only point clouds as input. NeurCADRecon~\cite{Dong2024NeurCADRecon} generates CAD models from point clouds by enforcing zero Gaussian curvature based solely on point cloud information. We compare our approach with the NeurCADRecon and SIREN baselines under the setting of point-cloud-only input.

We optimize {\sharpnet} using the loss in Equation~\eqref{eqn:3Dloss_points}, which is analogous to the loss in Equation~\eqref{eqn:3Dloss_points_normals}, except that the normal component $\mathcal{L}_{\text{nor}}$ is omitted.
\begin{equation}
\label{eqn:3Dloss_points}
    \mathcal{L}(\theta, M) =
        \alpha_{\text{sur}} \cdot \mathcal{L}_{\text{sur}} +
        \alpha_{\text{ext}} \cdot \mathcal{L}_{\text{ext}} +
        \alpha_{\text{ekl}} \cdot \mathcal{L}_{\text{ekl}} +
        \alpha_\mathcal{R} \cdot \mathcal{L}_\mathcal{R}.
\end{equation}

The parameters are set to \(\alpha_{\text{sur}}=7000\), \(\alpha_{\text{ext}}=600\), \(\alpha_{\text{ekl}}=50\), and \(\alpha_\mathcal{R}=10\) in the experiment. The initialization of \(M\) is the same as that in Section~\ref{sec:CAD_points_normals}.

\begin{figure*}
    \centering
    \includegraphics[width=0.12\linewidth]{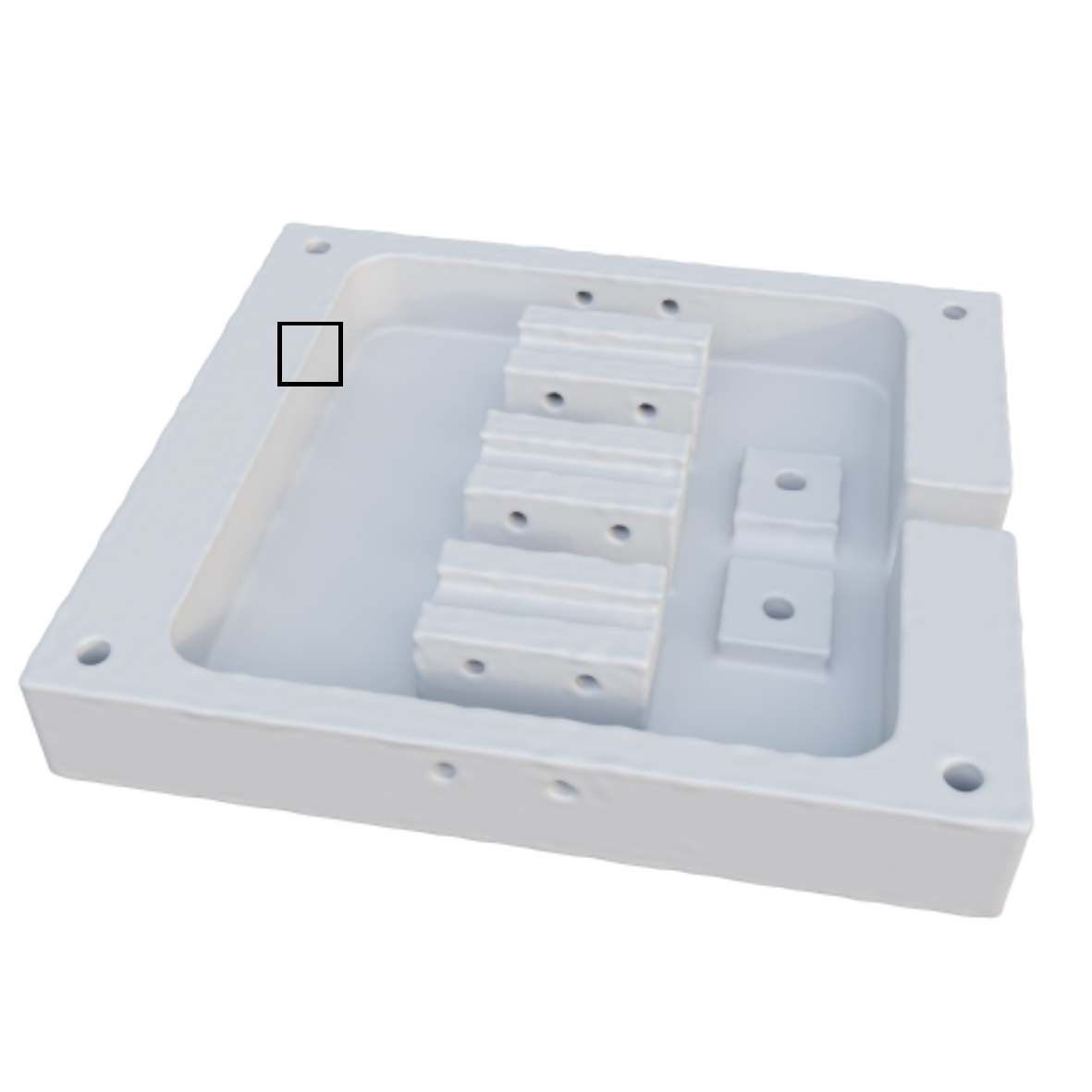}    \includegraphics[width=0.12\linewidth]{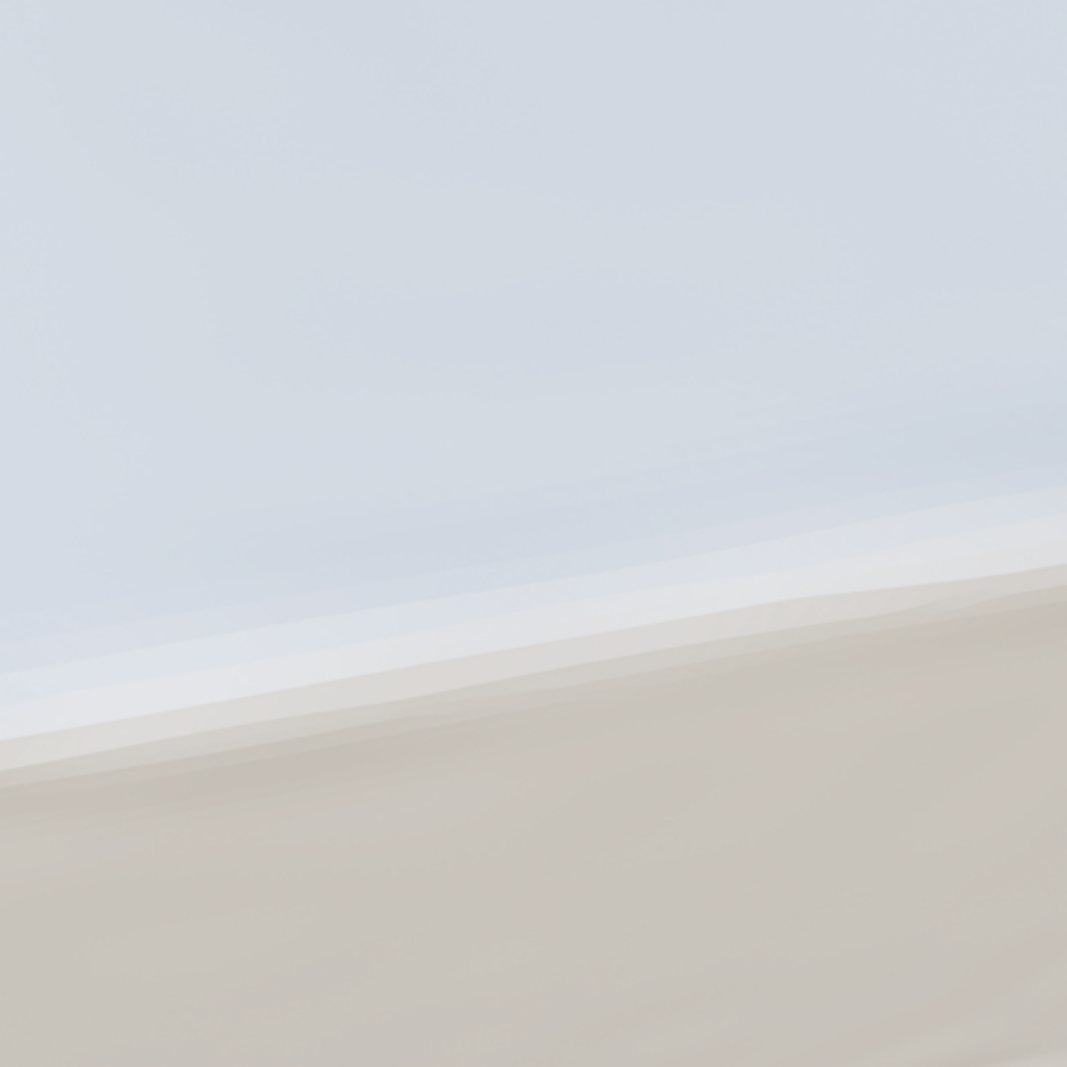}
    \includegraphics[width=0.12\linewidth]{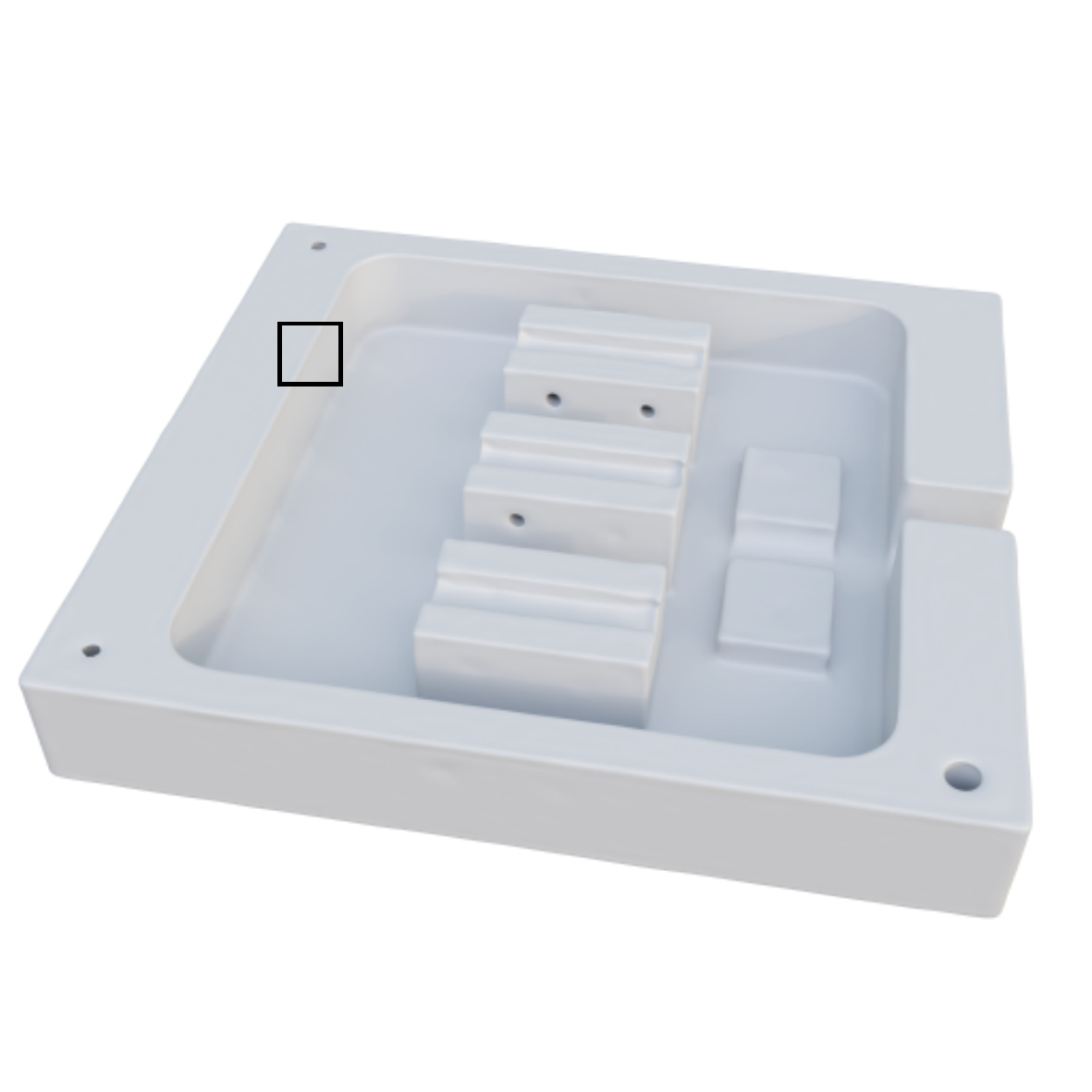}    \includegraphics[width=0.12\linewidth]{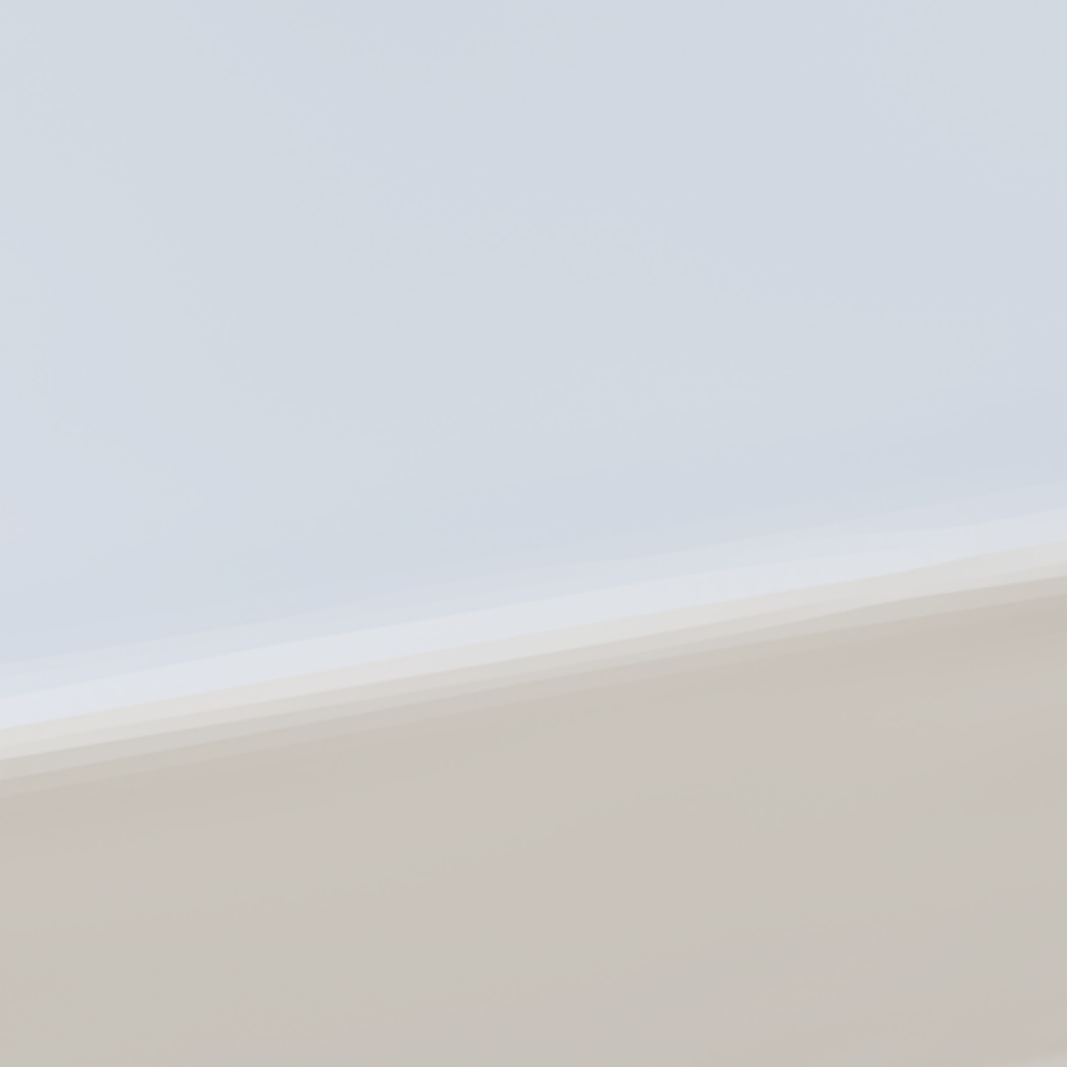}
    \includegraphics[width=0.12\linewidth]{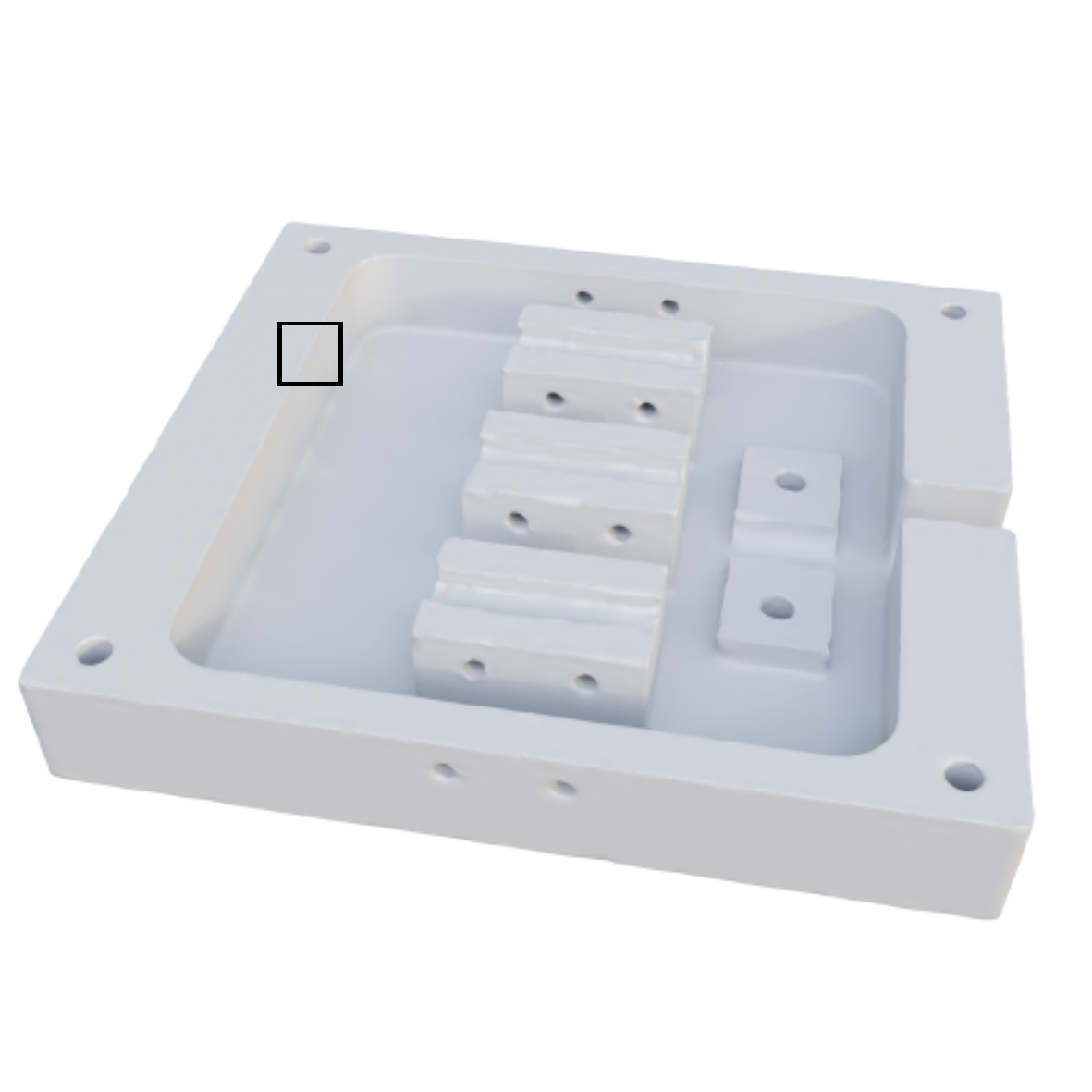}    \includegraphics[width=0.12\linewidth]{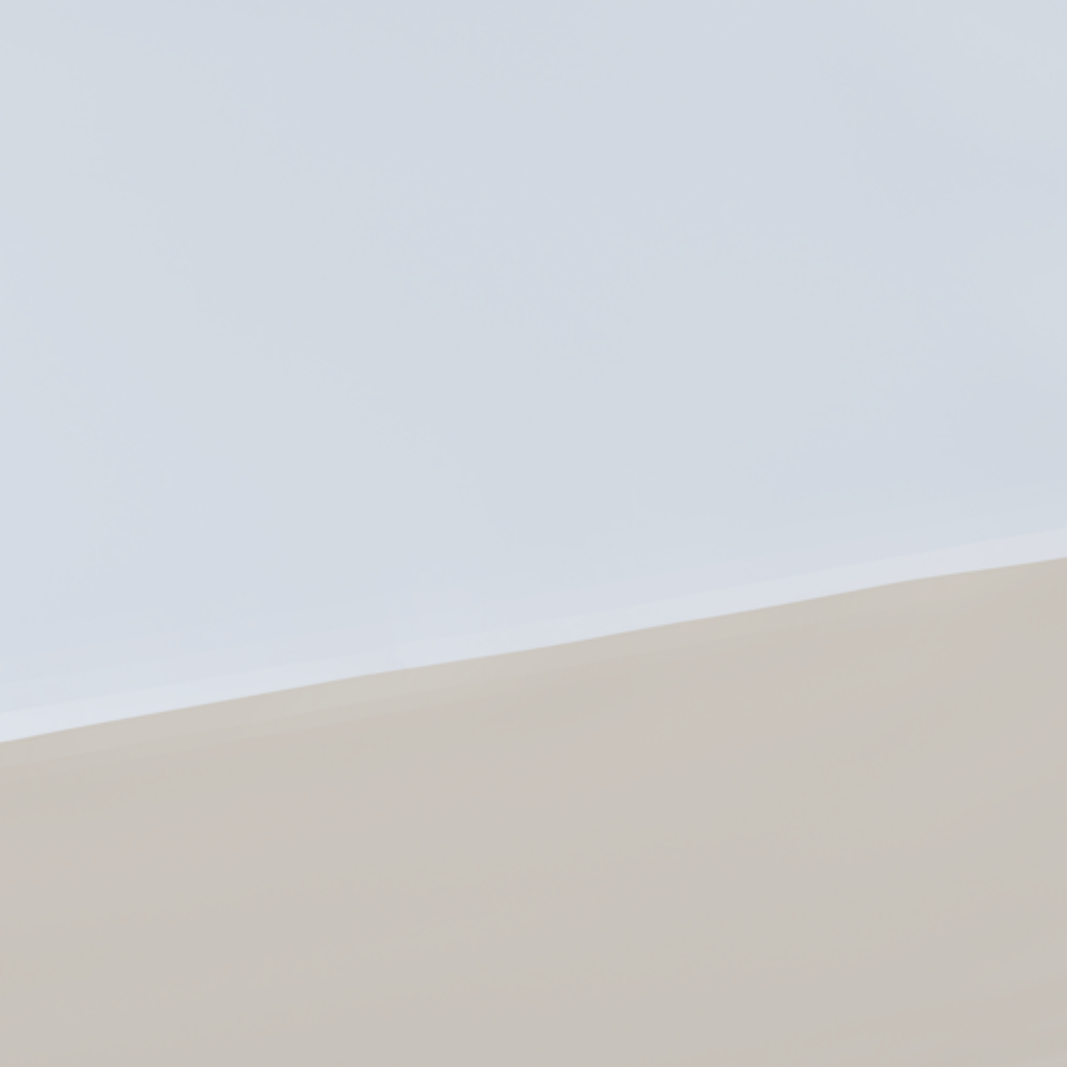}
    \includegraphics[width=0.12\linewidth]{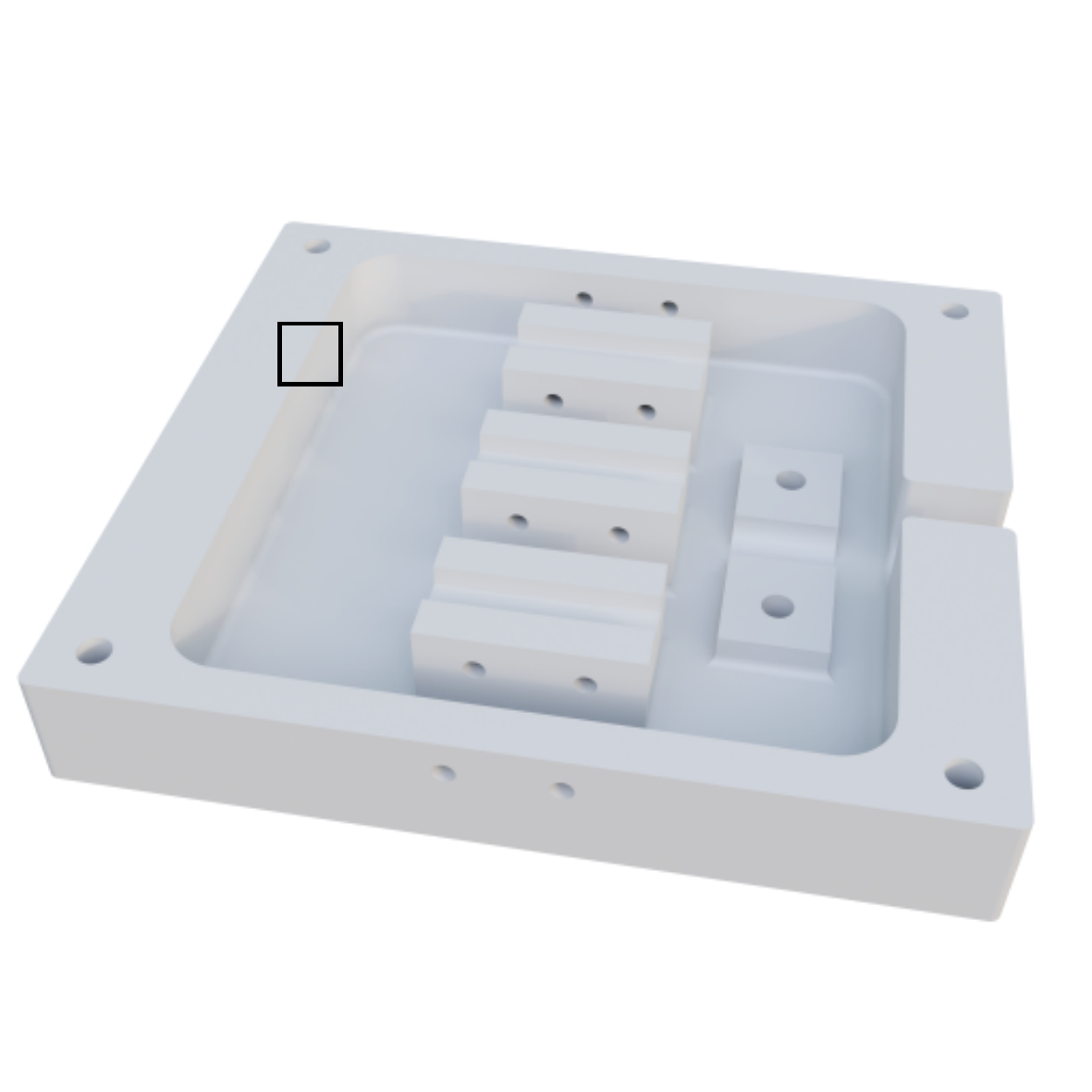}
    \includegraphics[width=0.12\linewidth]{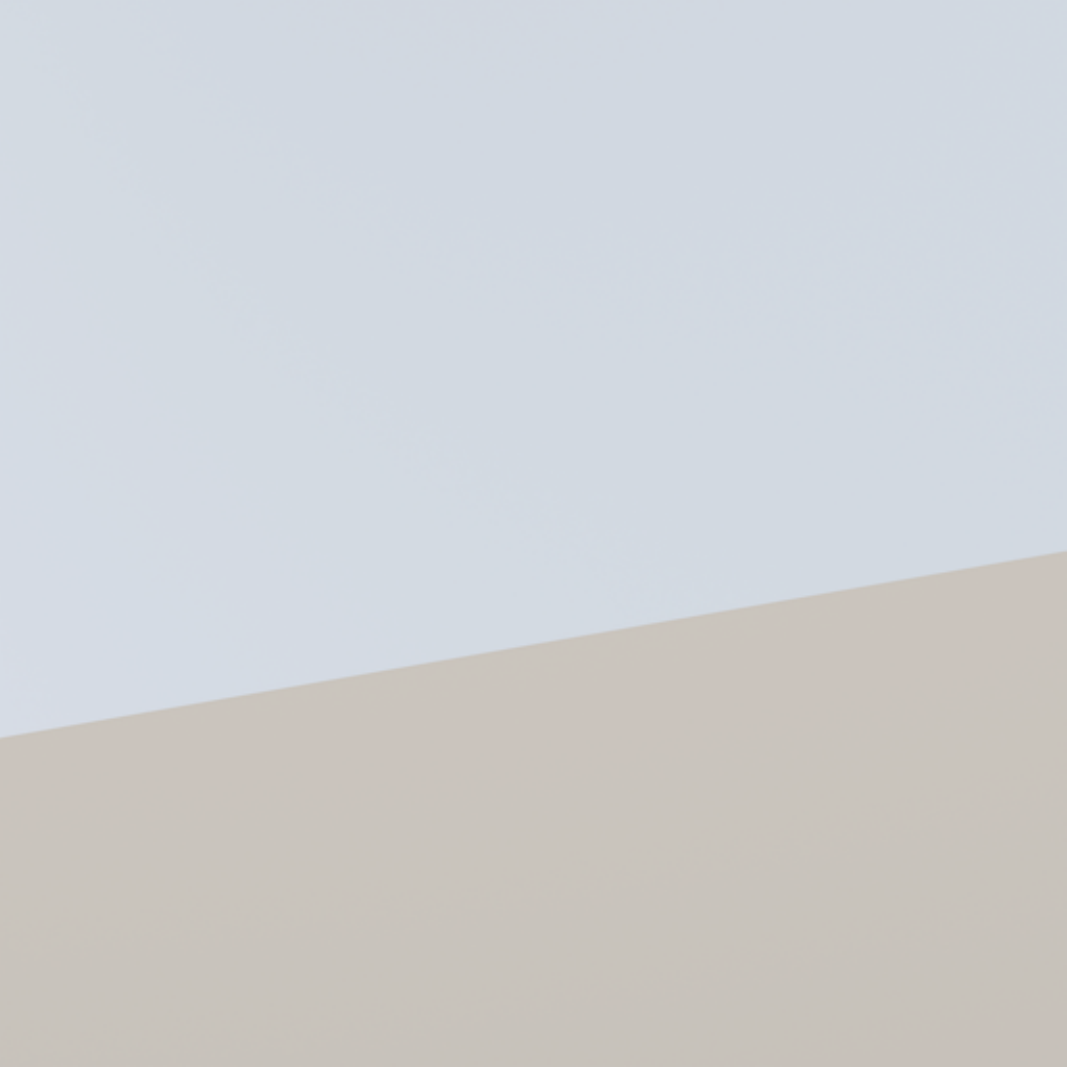}\\
    \makebox[0.24\linewidth]{\text{4.536}}
    \makebox[0.24\linewidth]{\text{4.096}}
    \makebox[0.24\linewidth]{\text{4.033}}
    \makebox[0.24\linewidth]{\text{-}}\\
    \vspace{6pt}
    \includegraphics[width=0.12\linewidth]{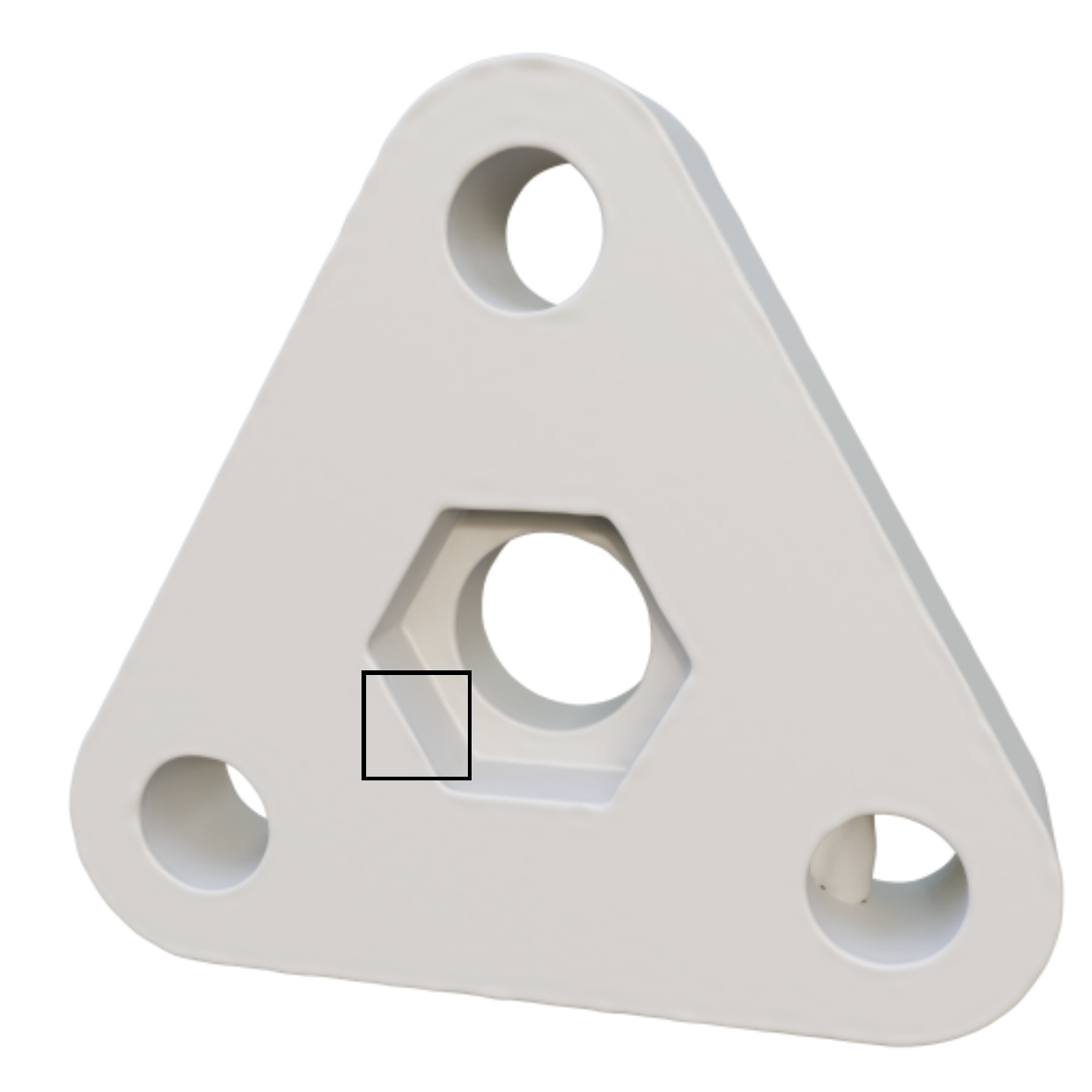}
    \includegraphics[width=0.12\linewidth]{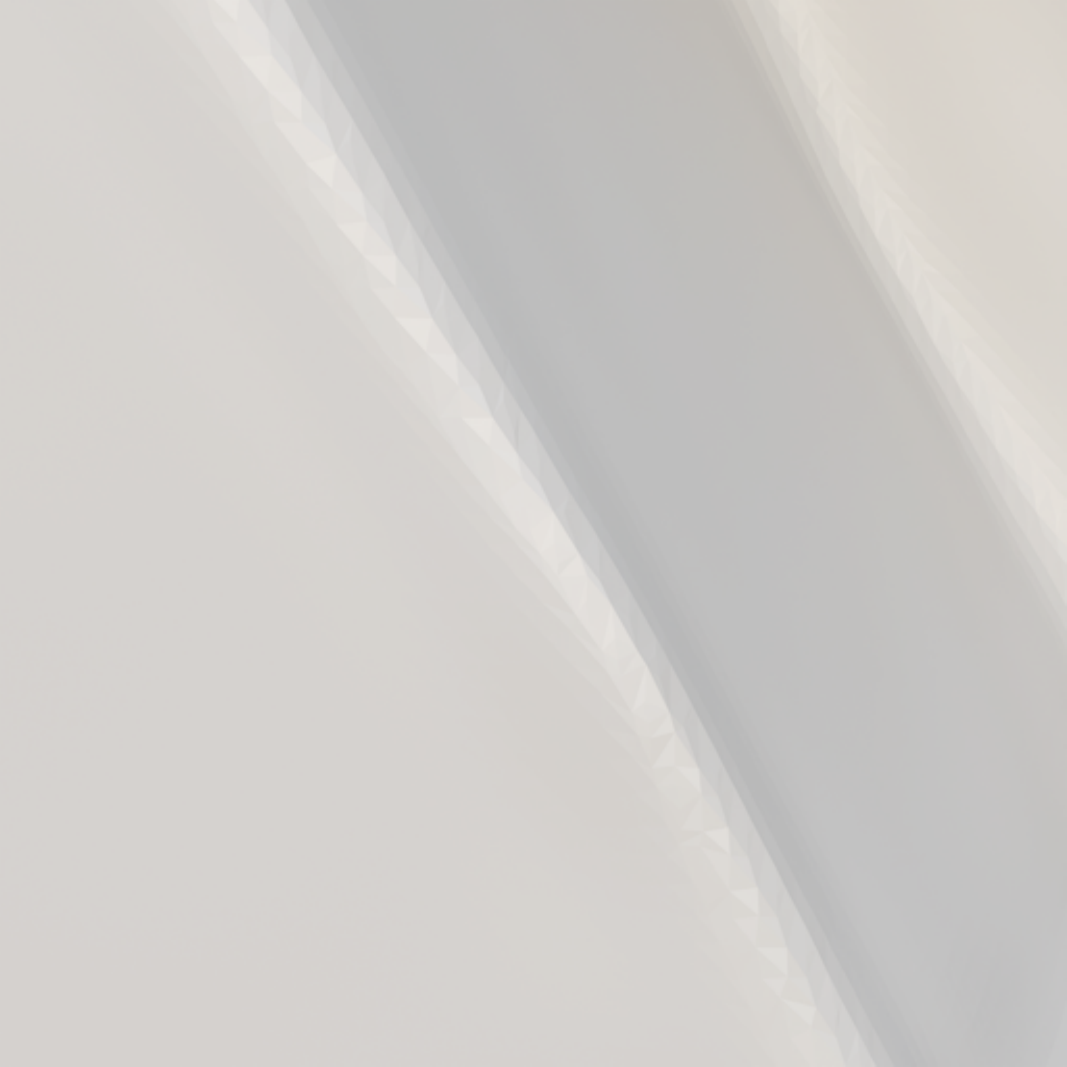}
    \includegraphics[width=0.12\linewidth]{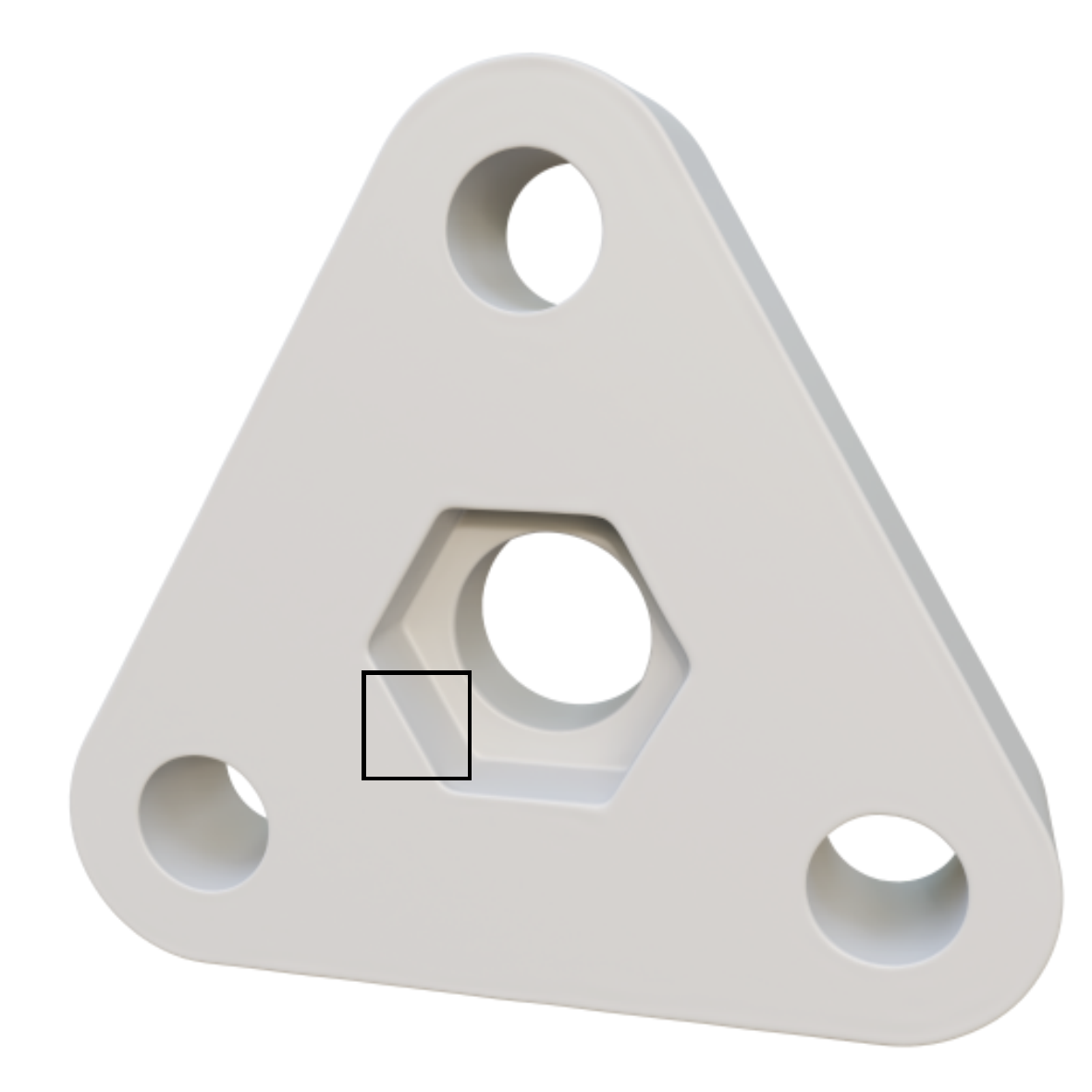}    \includegraphics[width=0.12\linewidth]{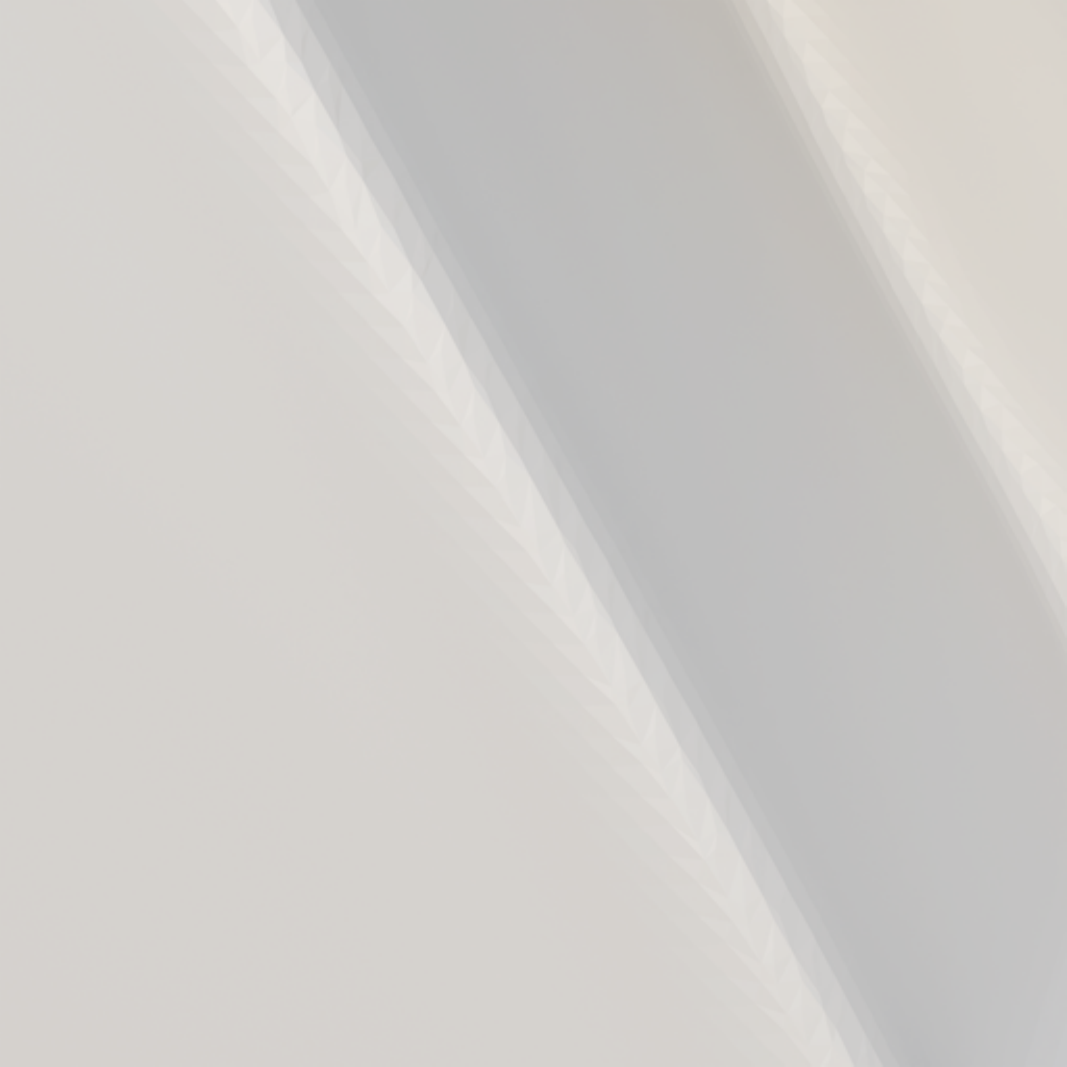}
    \includegraphics[width=0.12\linewidth]{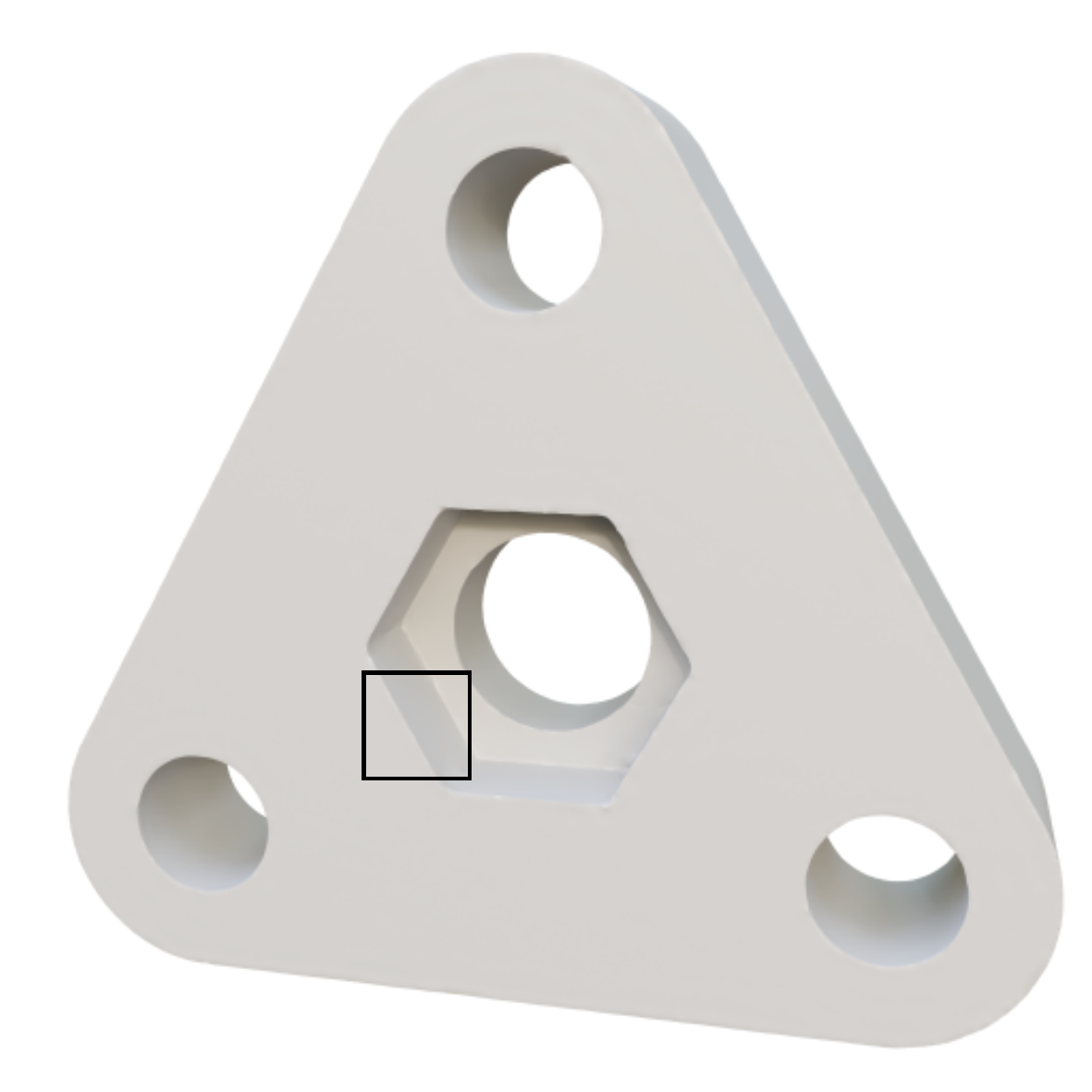}    \includegraphics[width=0.12\linewidth]{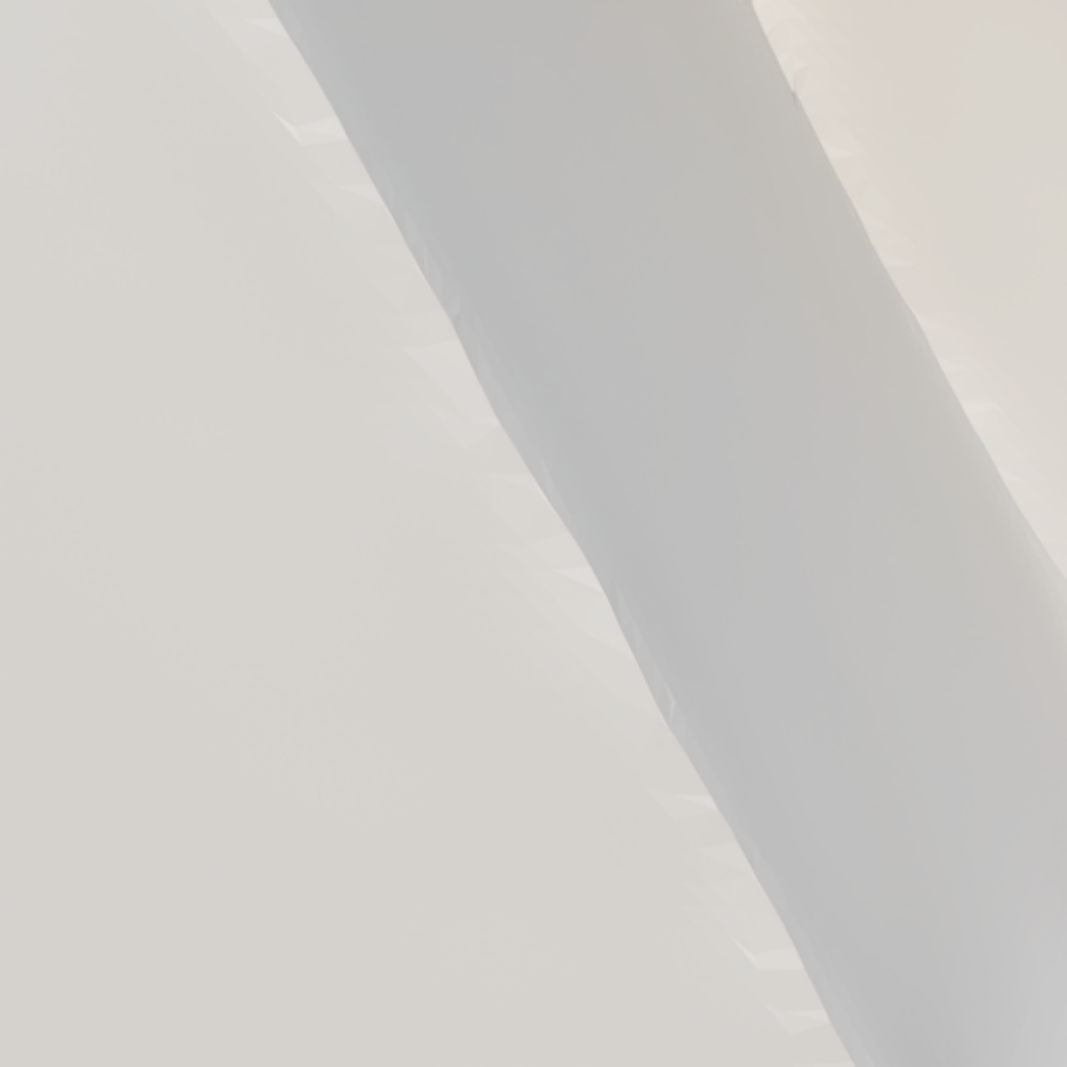}
    \includegraphics[width=0.12\linewidth]{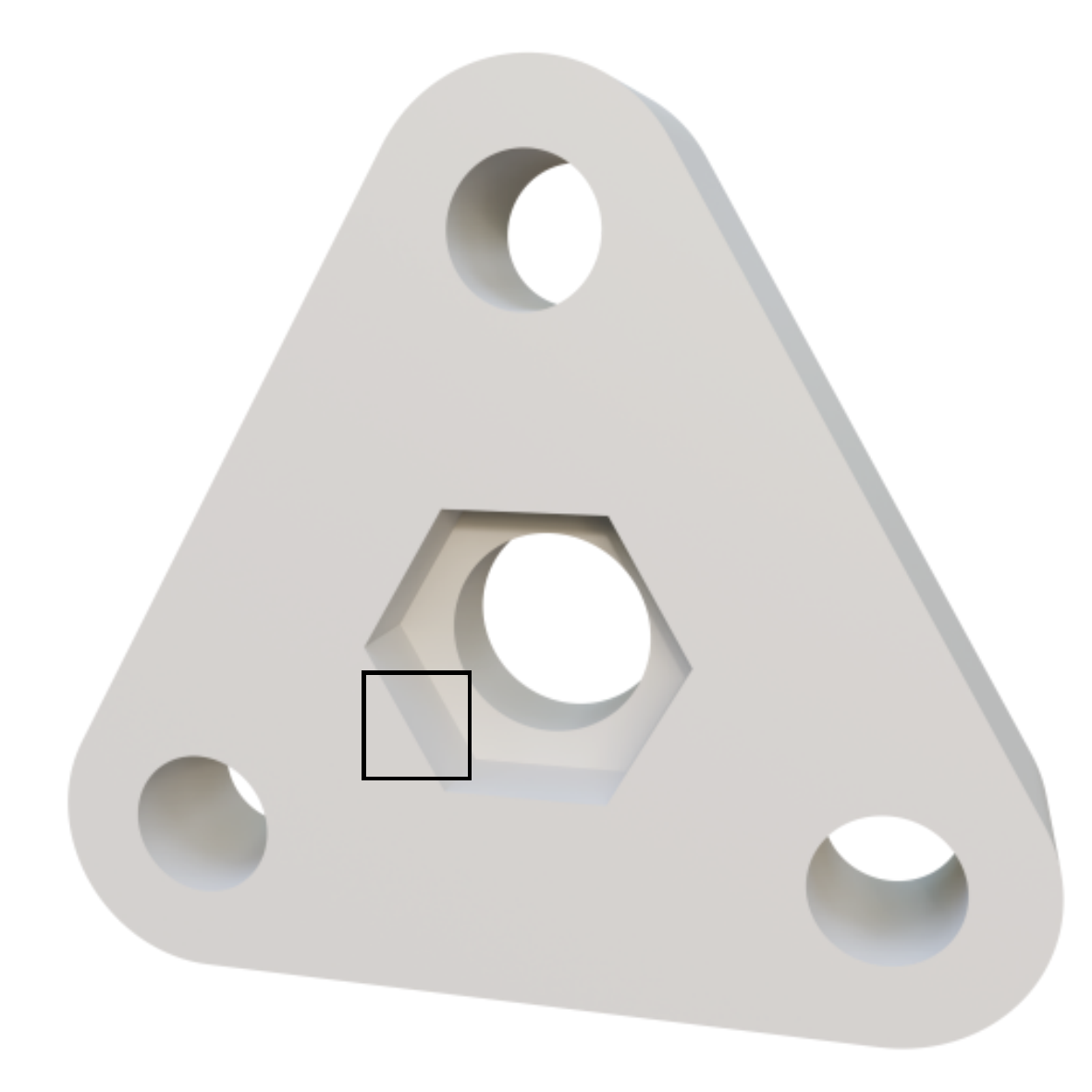}
    \includegraphics[width=0.12\linewidth]{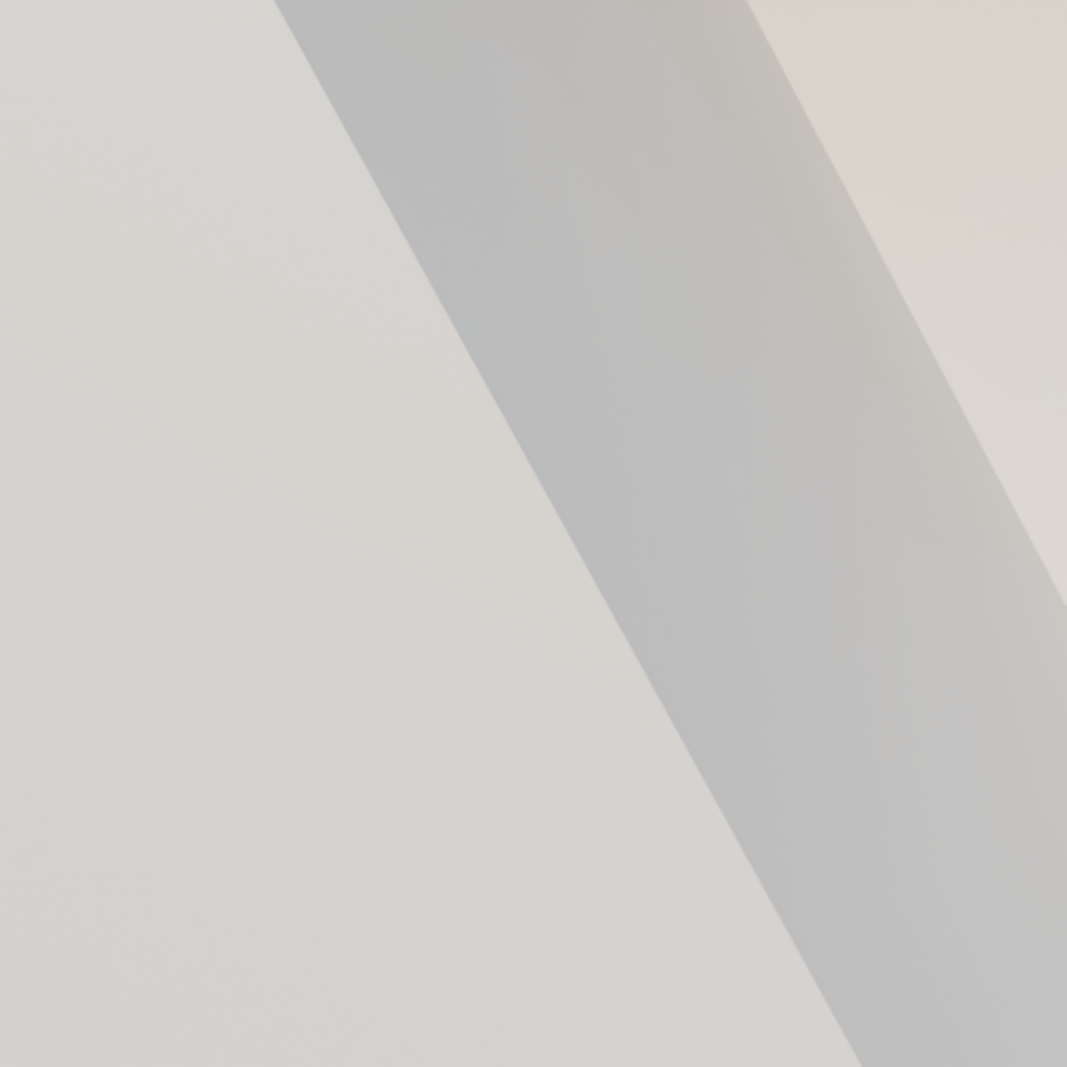}\\
    \makebox[0.24\linewidth]{\text{3.713}}
    \makebox[0.24\linewidth]{\text{3.261}}
    \makebox[0.24\linewidth]{\text{3.255}}
    \makebox[0.24\linewidth]{\text{-}}\\
    \vspace{6pt}
    \includegraphics[width=0.12\linewidth]{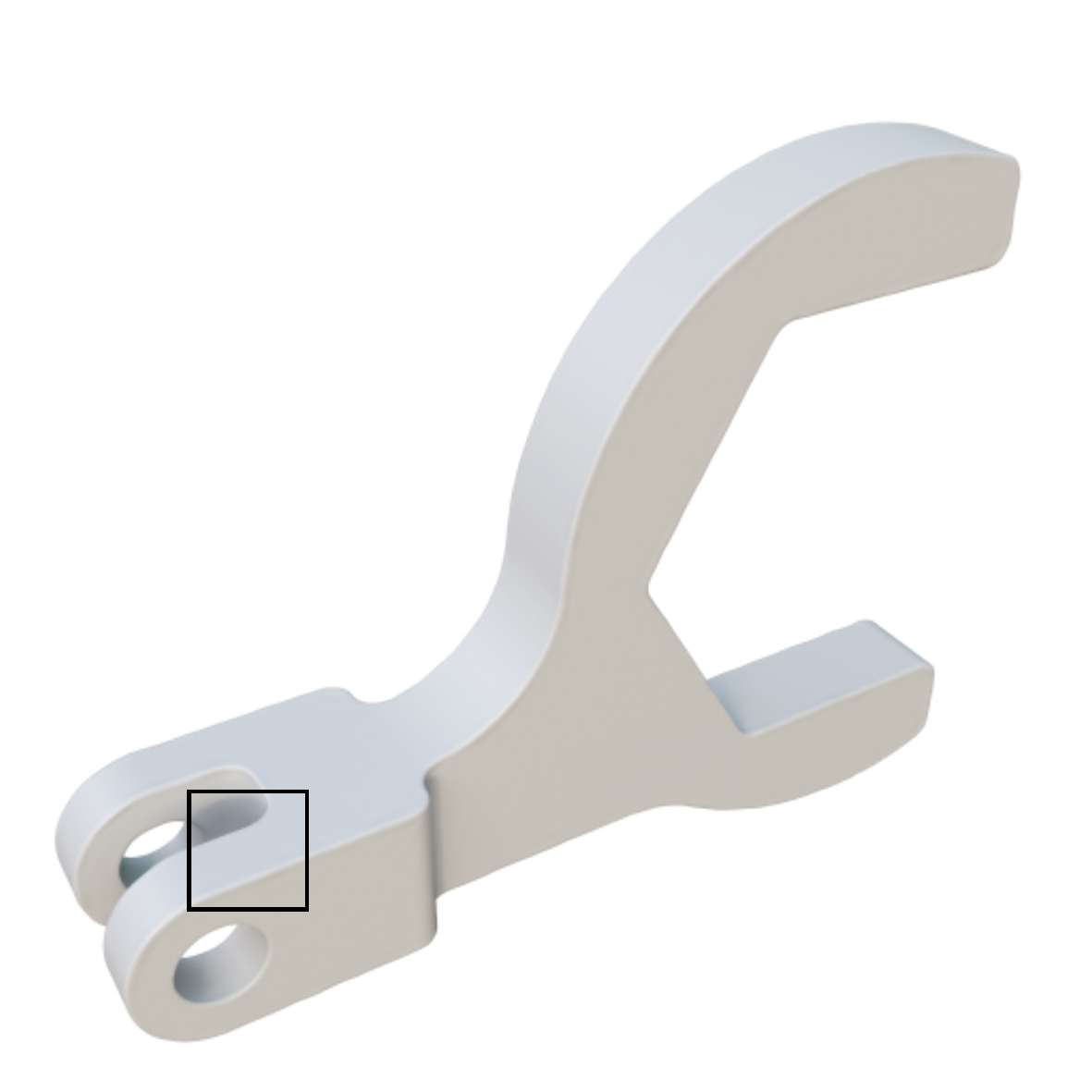}    \includegraphics[width=0.12\linewidth]{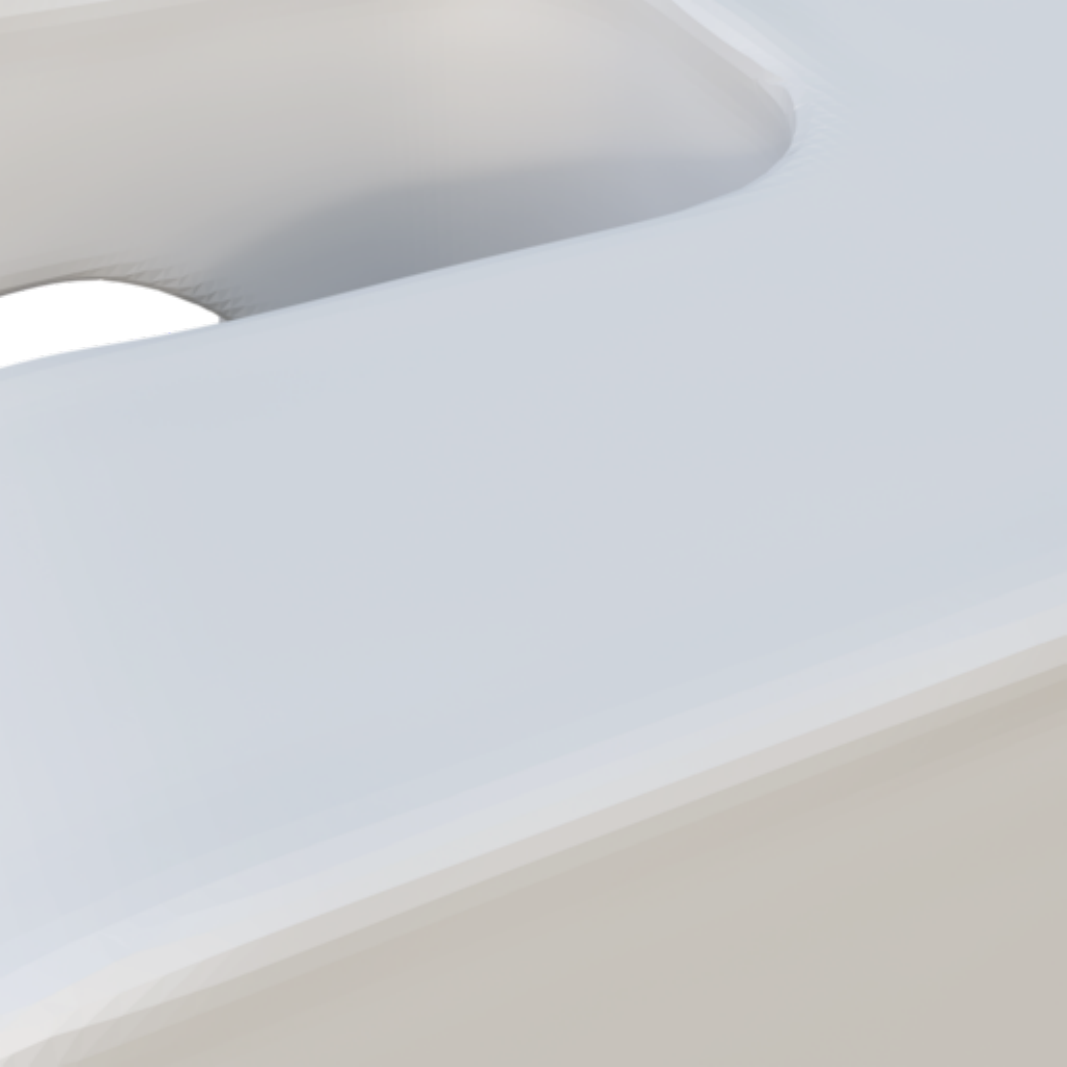}
    \includegraphics[width=0.12\linewidth]{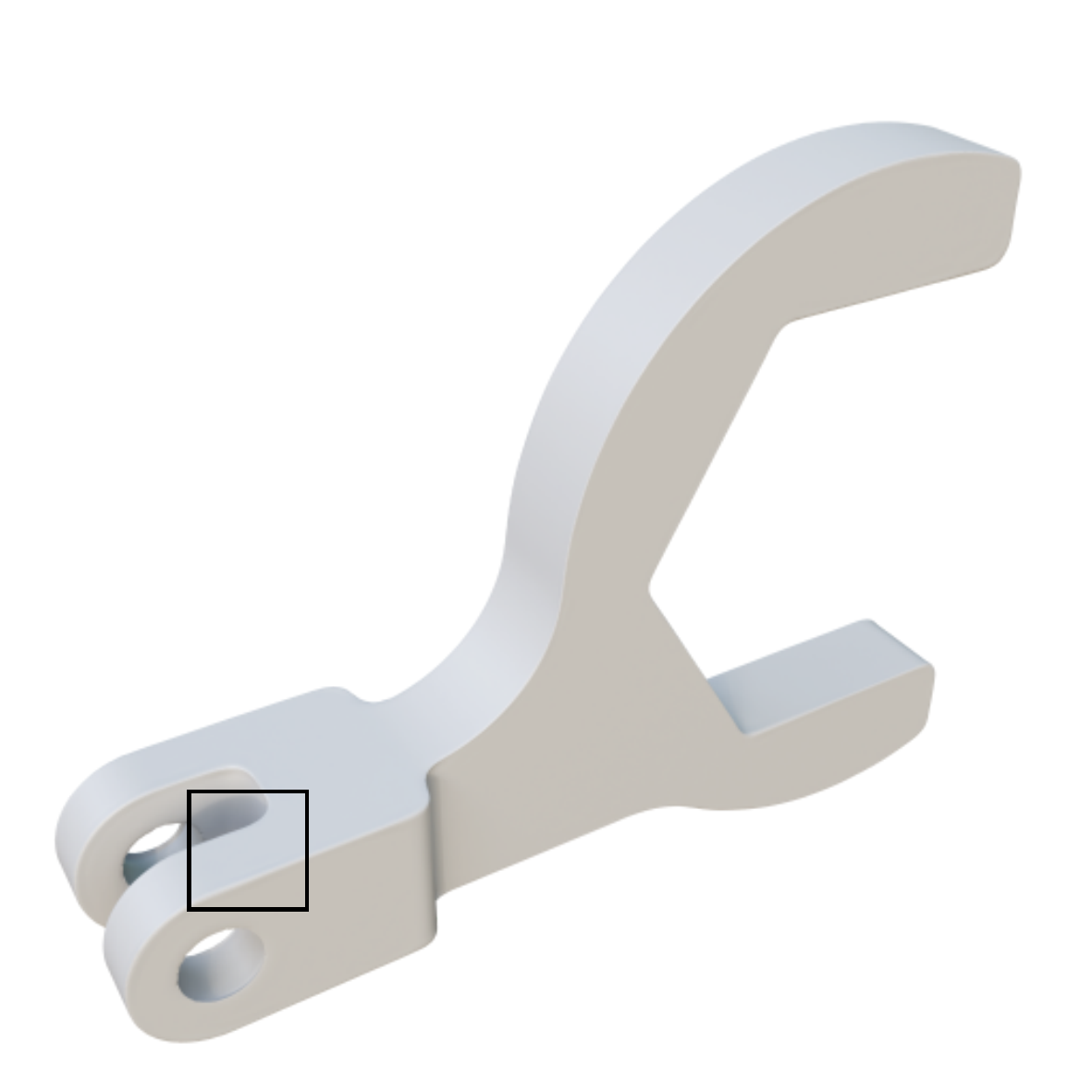}    \includegraphics[width=0.12\linewidth]{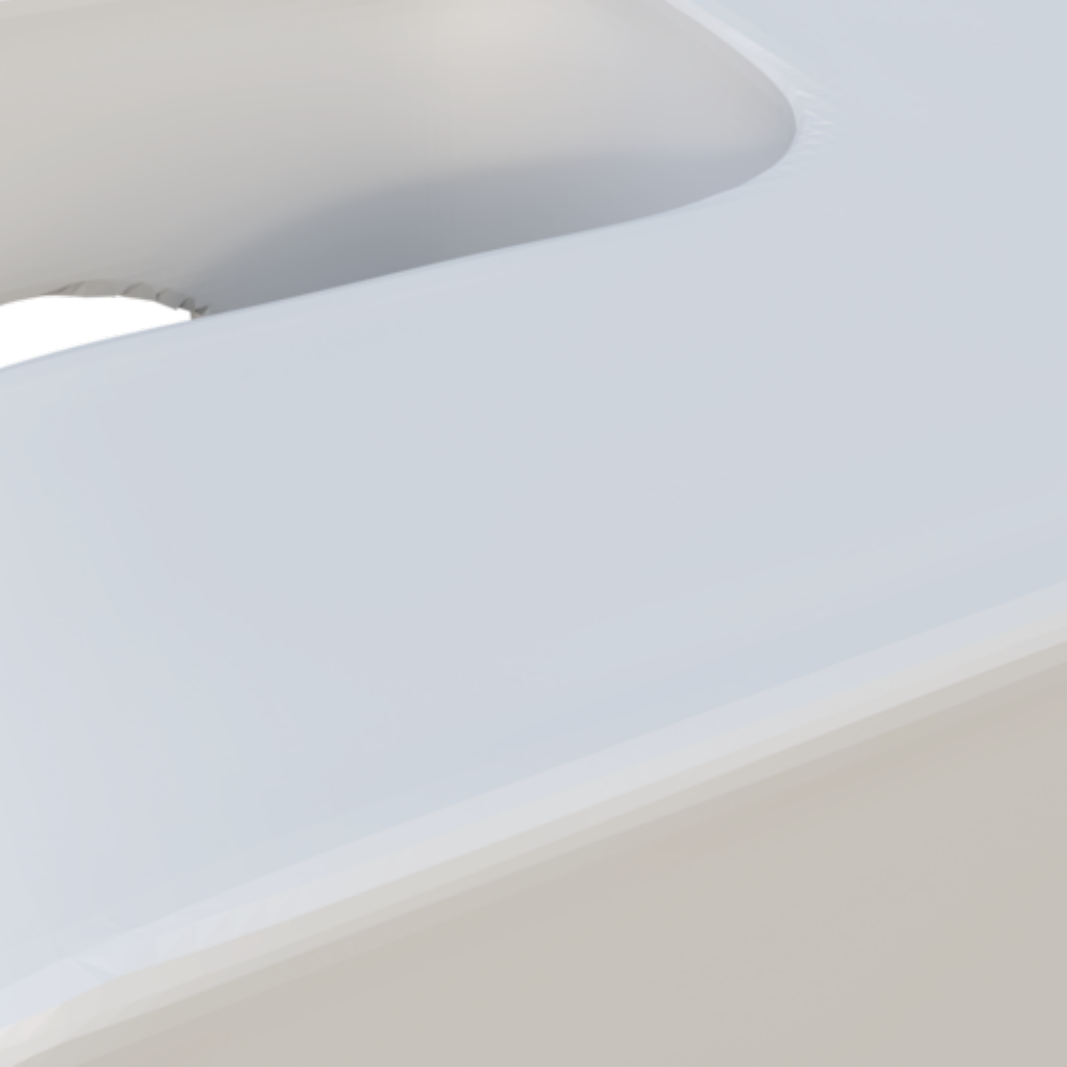}
    \includegraphics[width=0.12\linewidth]{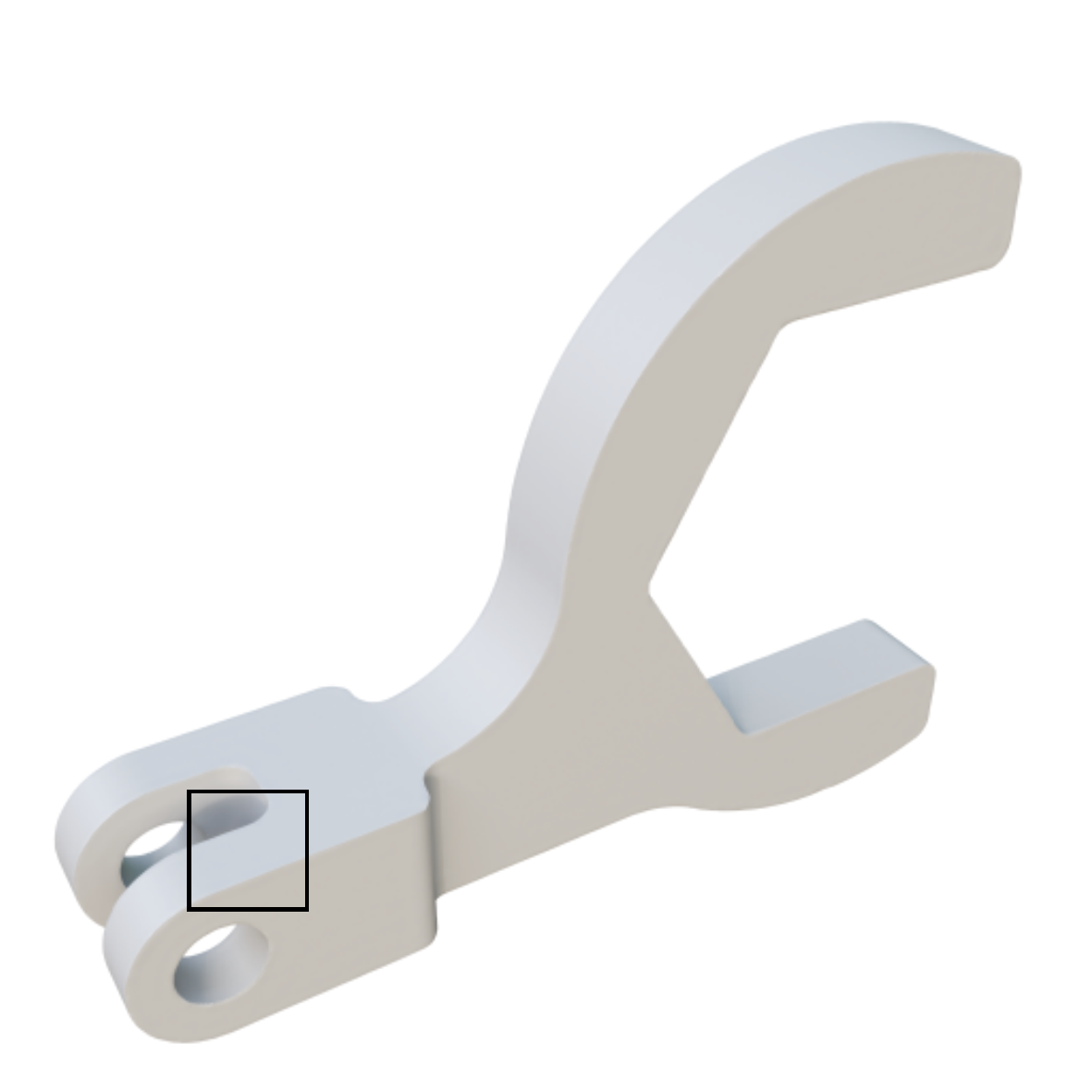}    \includegraphics[width=0.12\linewidth]{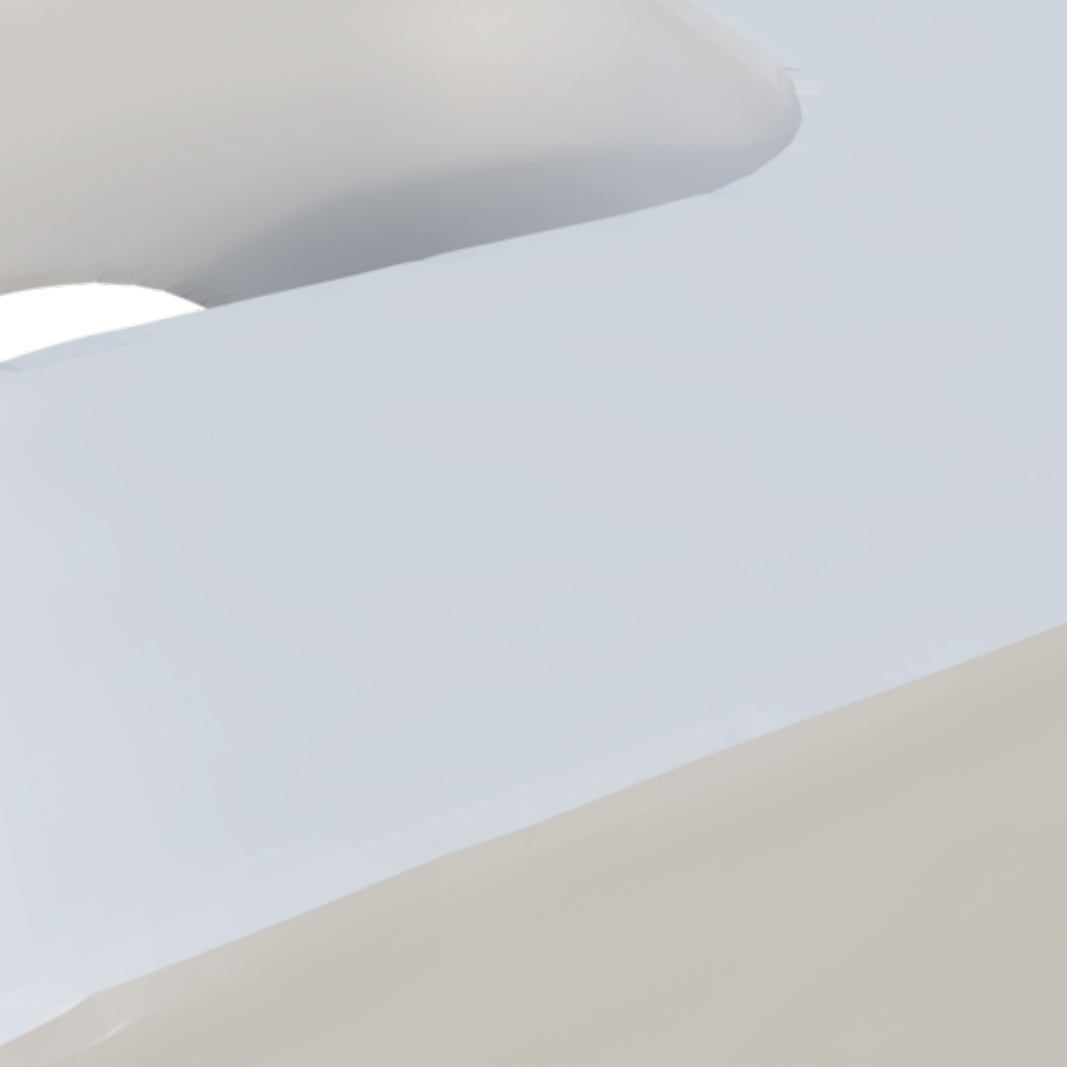}
    \includegraphics[width=0.12\linewidth]{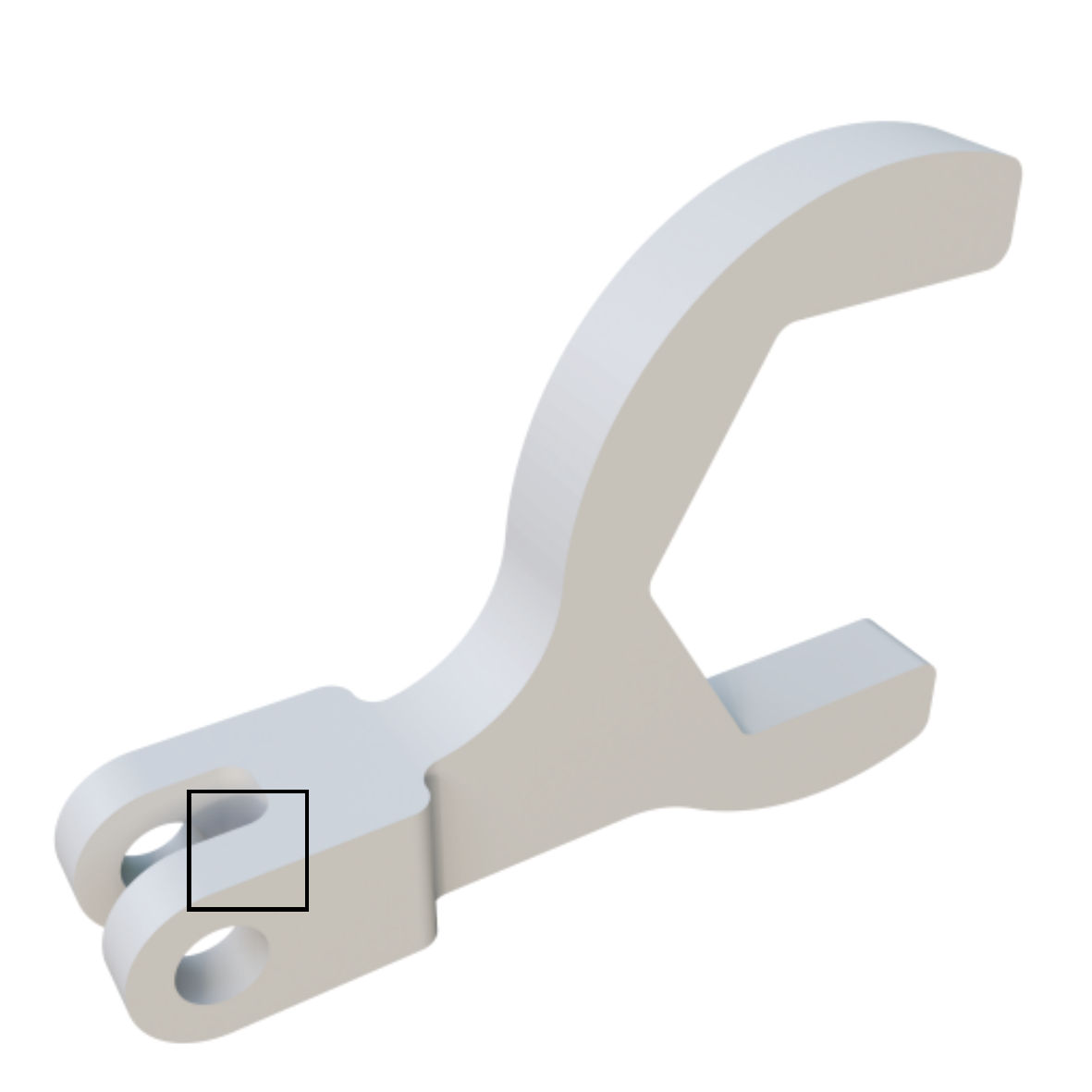}
    \includegraphics[width=0.12\linewidth]{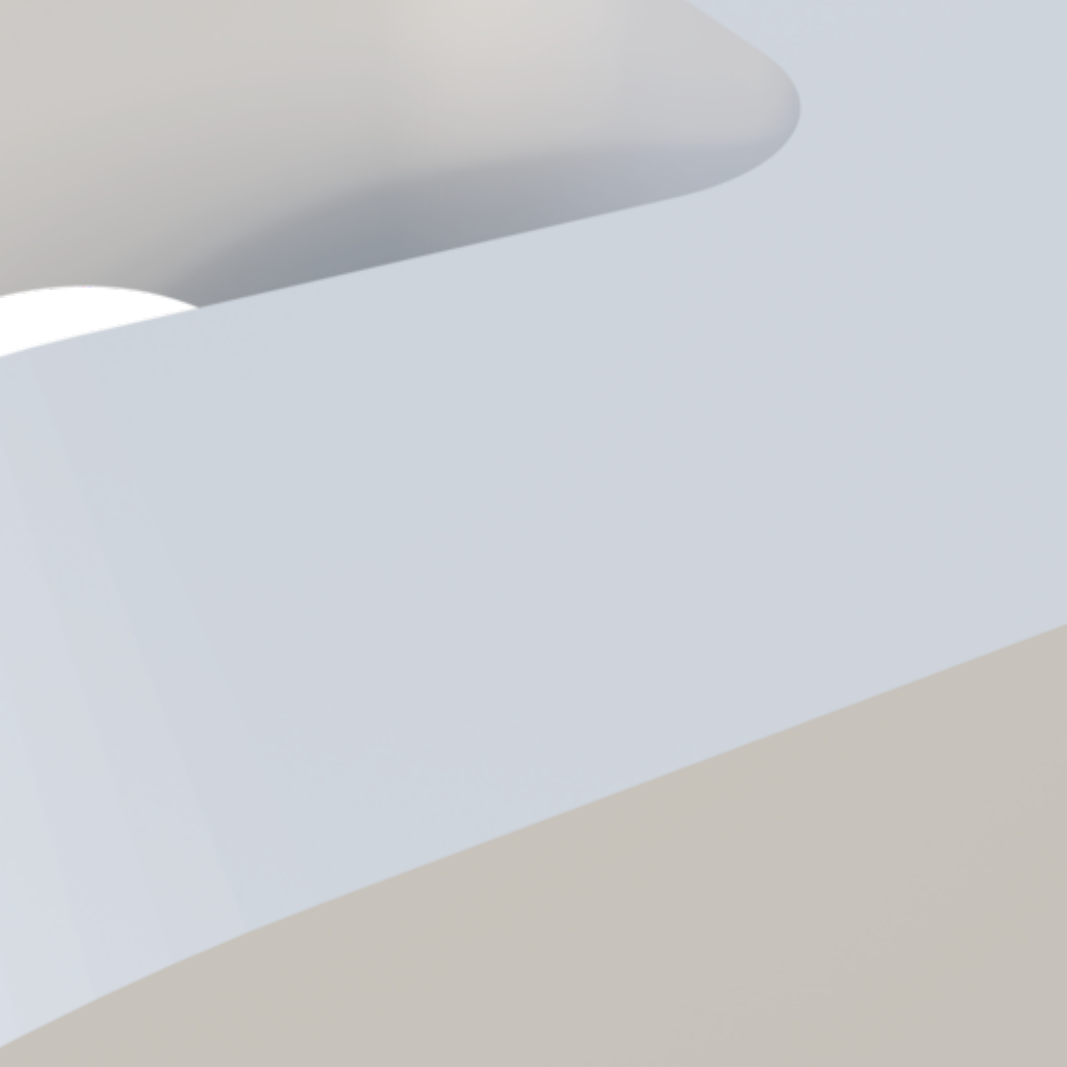}\\
    \makebox[0.24\linewidth]{\text{2.591}}
    \makebox[0.24\linewidth]{\text{2.316}}
    \makebox[0.24\linewidth]{\text{2.293}}
    \makebox[0.24\linewidth]{\text{-}}\\
    \vspace{6pt}
    \includegraphics[width=0.12\linewidth]{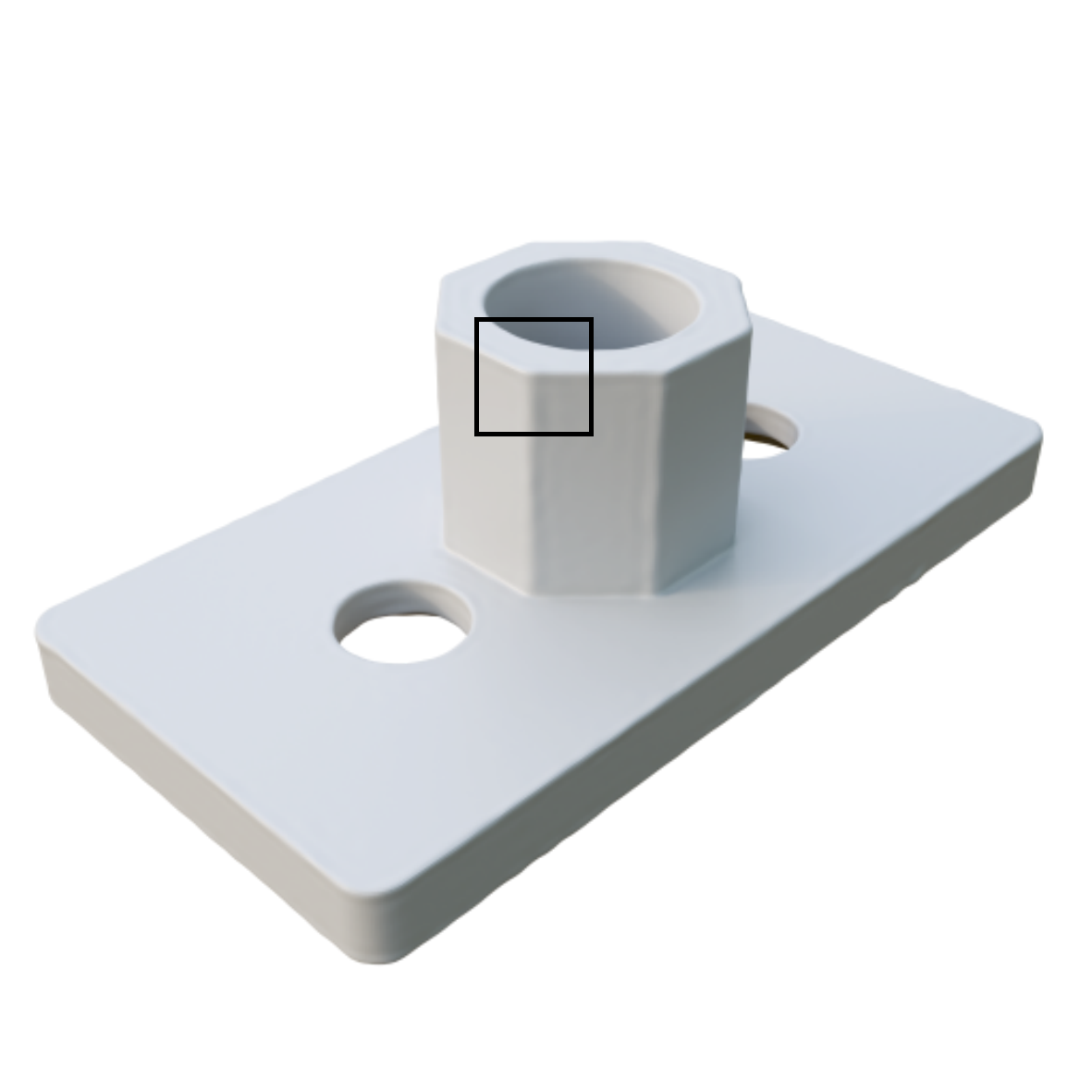}    \includegraphics[width=0.12\linewidth]{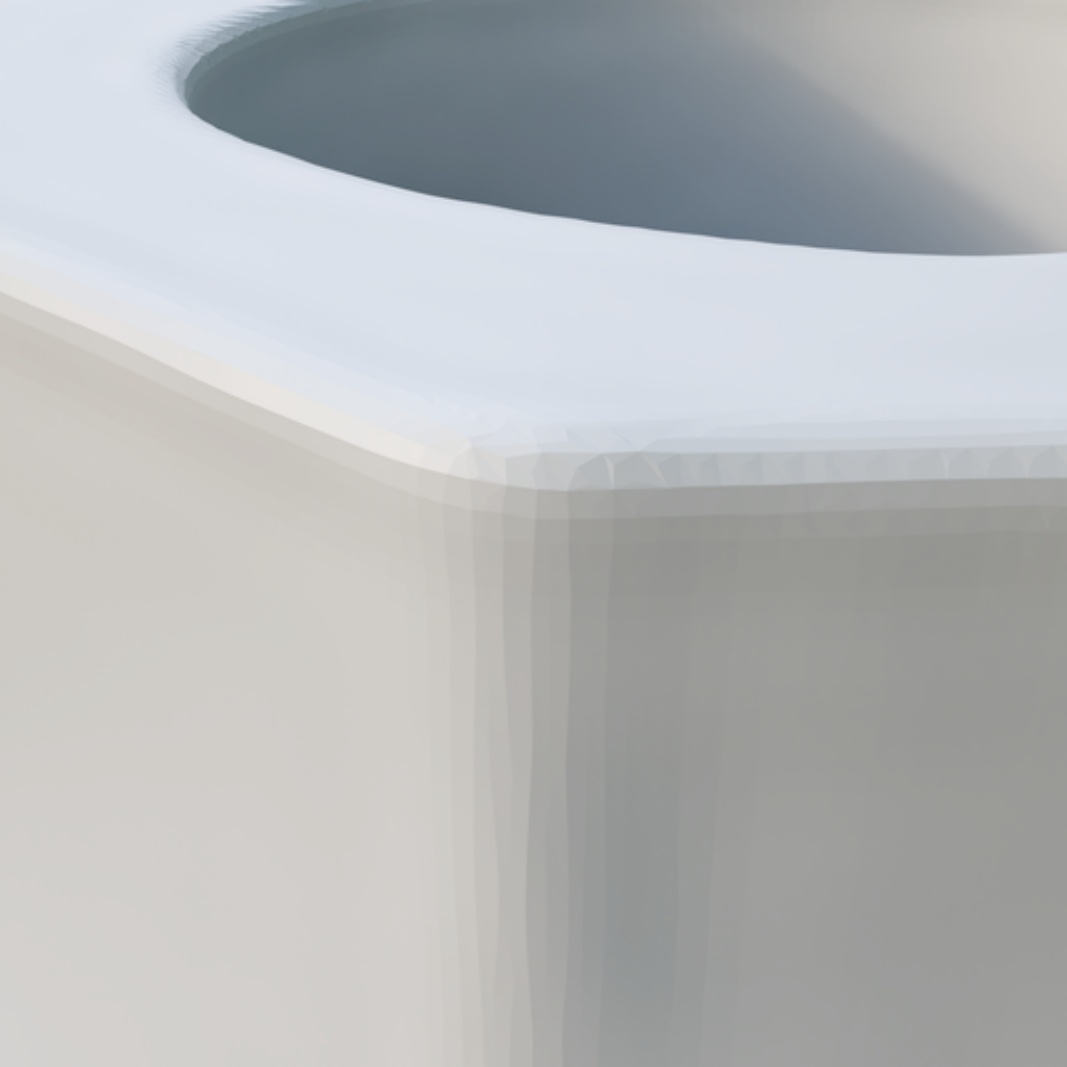}
    \includegraphics[width=0.12\linewidth]{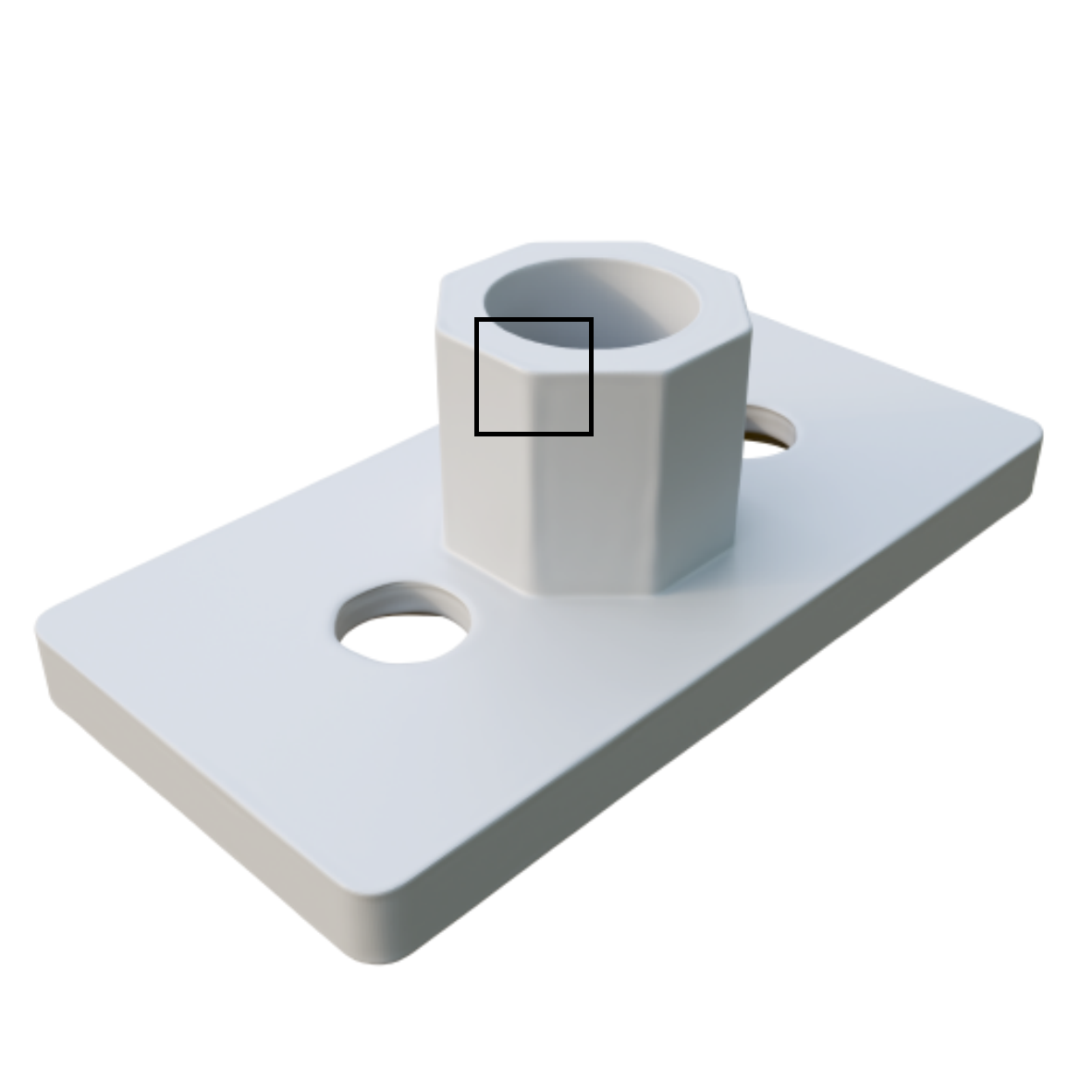}    \includegraphics[width=0.12\linewidth]{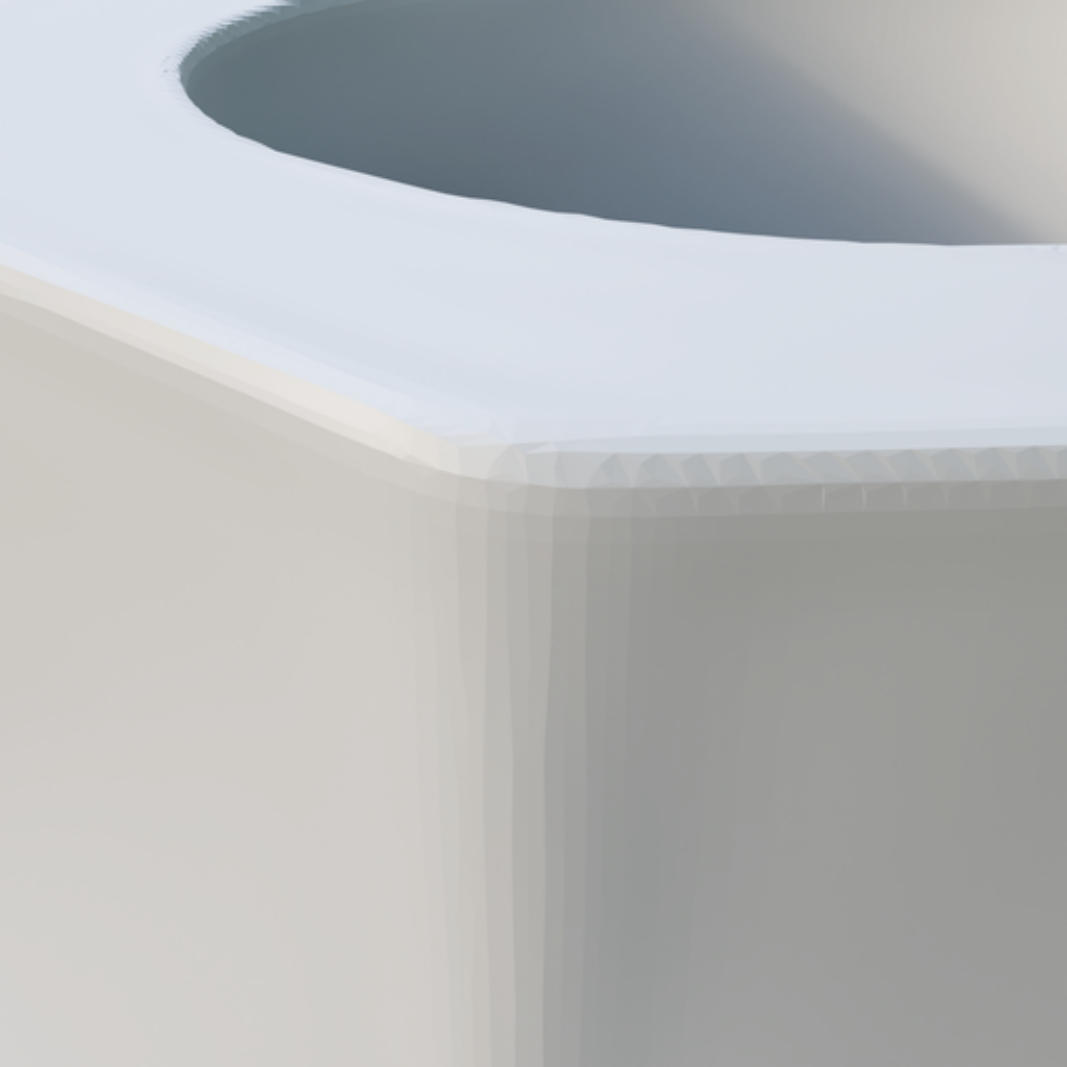}
    \includegraphics[width=0.12\linewidth]{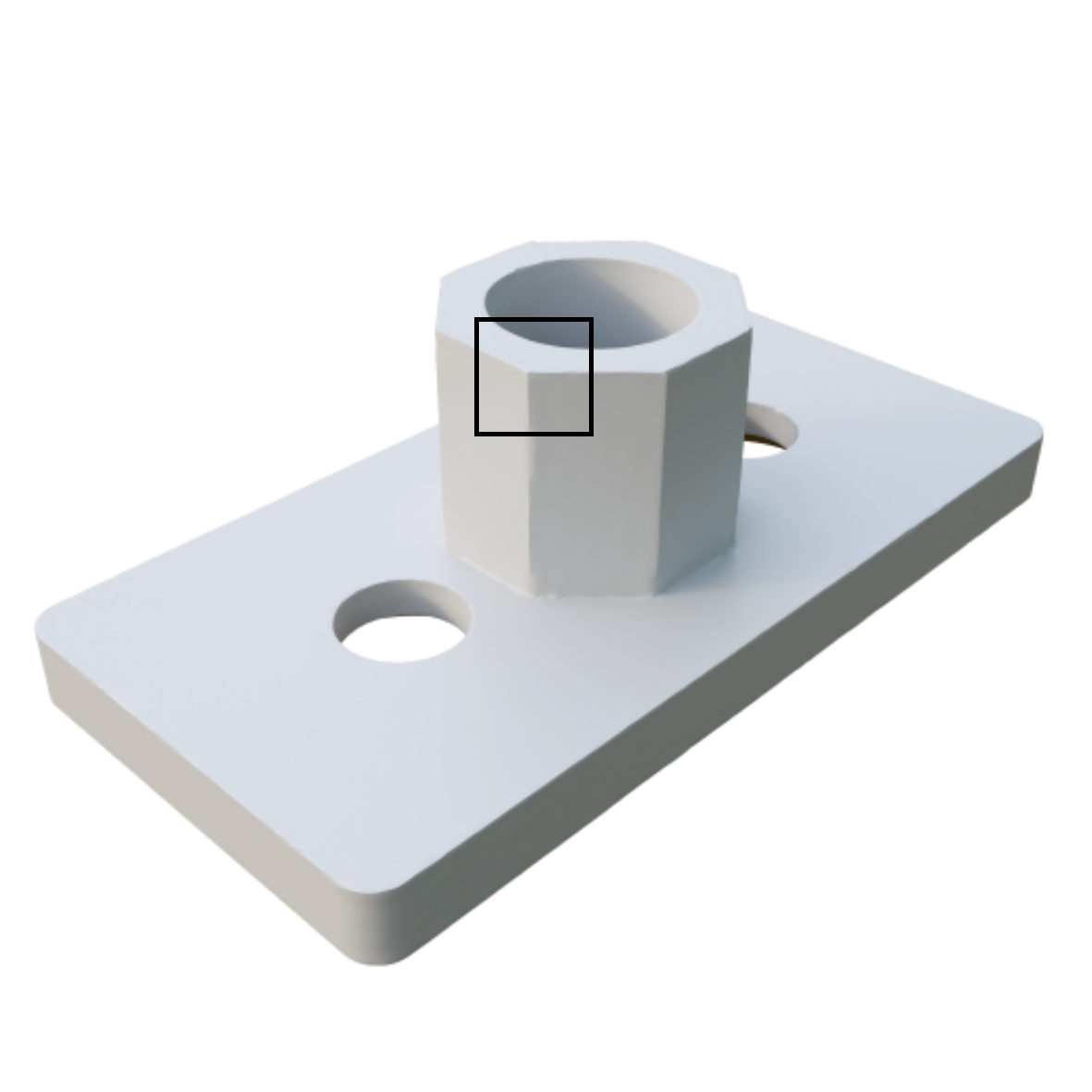}    \includegraphics[width=0.12\linewidth]{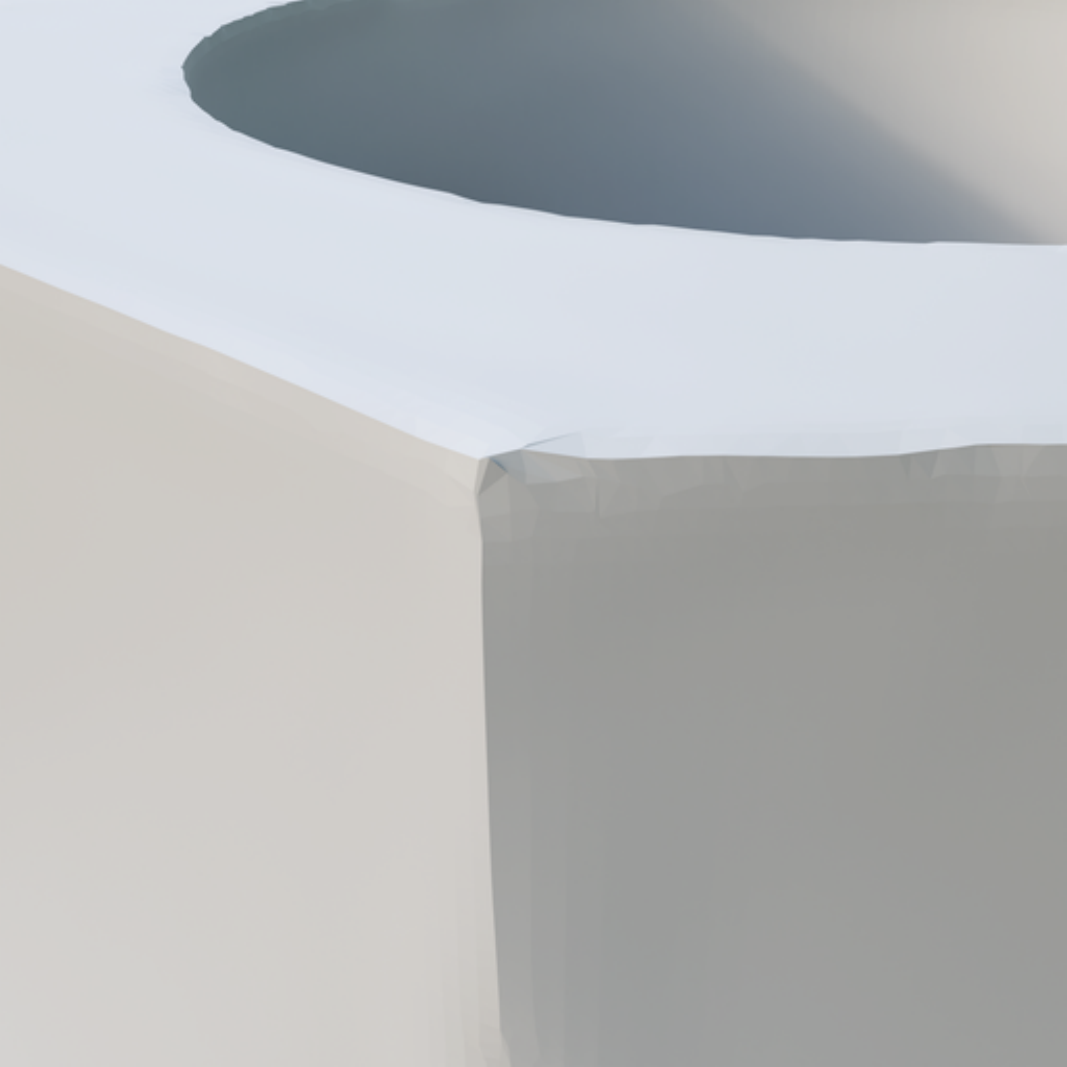}
    \includegraphics[width=0.12\linewidth]{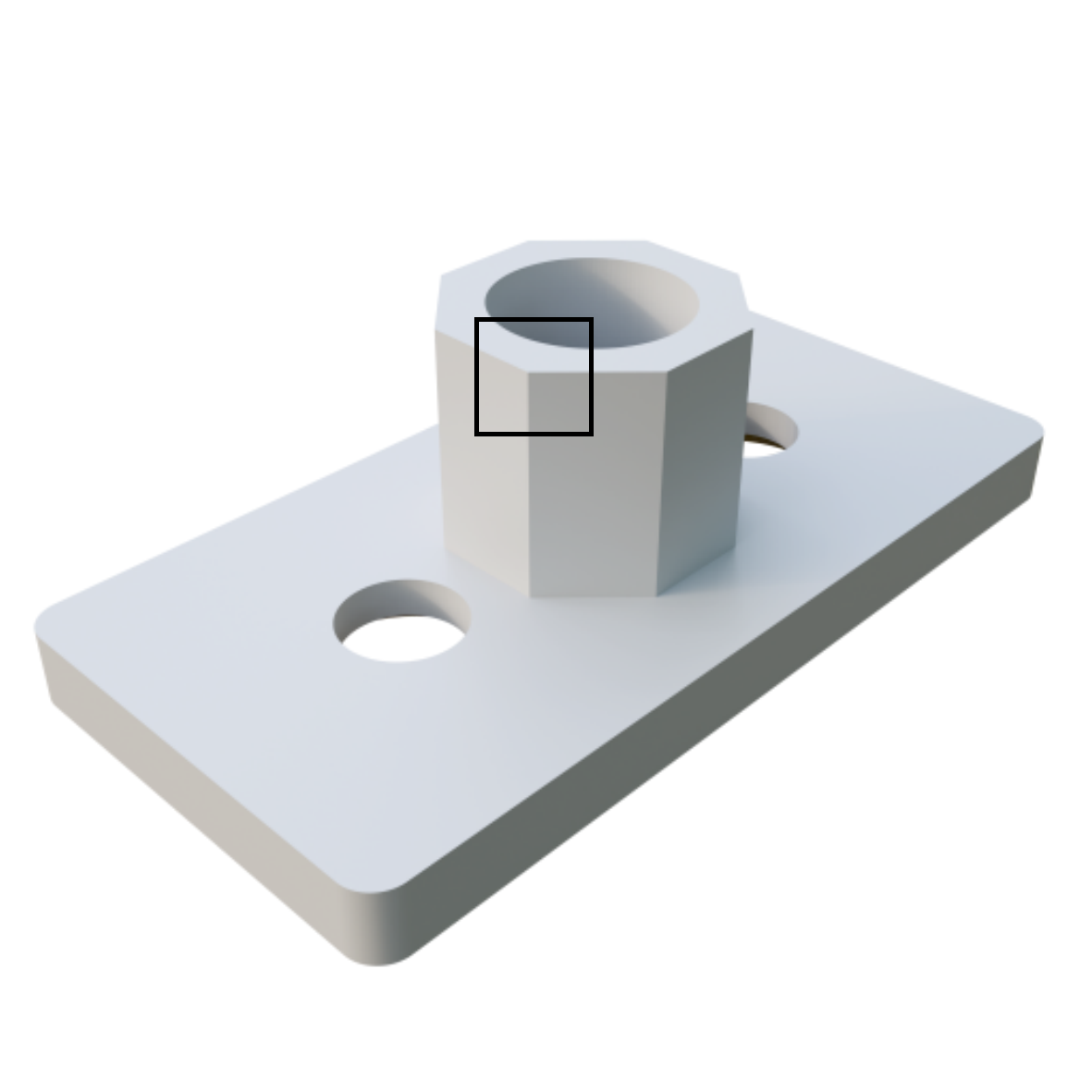}
    \includegraphics[width=0.12\linewidth]{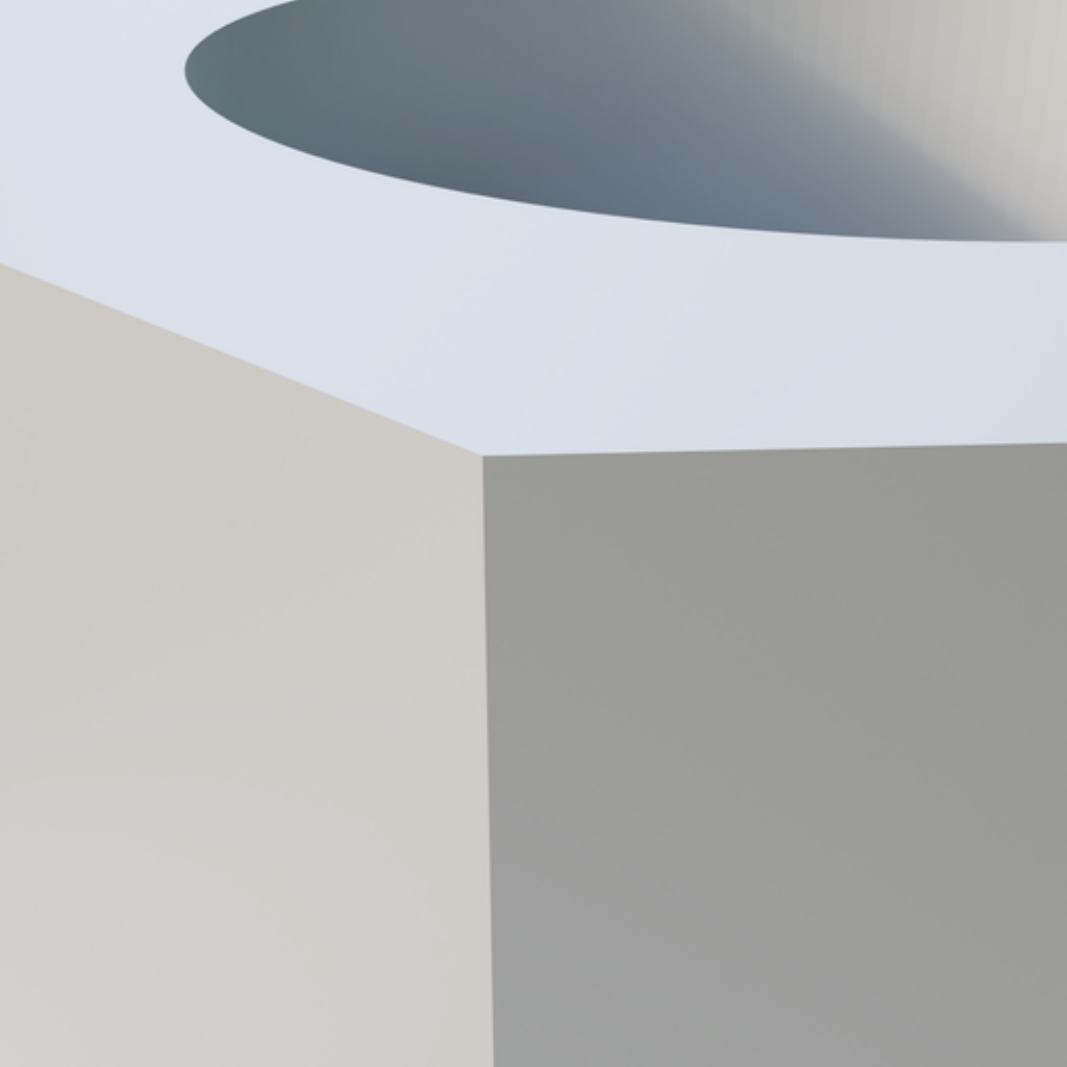}\\
    \makebox[0.24\linewidth]{\text{3.225}}
    \makebox[0.24\linewidth]{\text{2.880}}
    \makebox[0.24\linewidth]{\text{2.850}}
    \makebox[0.24\linewidth]{\text{-}}\\
    \vspace{6pt}
    \makebox[0.24\linewidth]{\sffamily \text{SIREN}}
    \makebox[0.24\linewidth]{\sffamily \text{NeurCADRecon}}
    \makebox[0.24\linewidth]{\sffamily \text{\sharpnet}}
    \makebox[0.24\linewidth]{\sffamily \text{GT}}
    \caption{CAD models reconstructed from points. The CD for each complete model is shown below each model.}
    \label{fig:cad_pointcloud}
\end{figure*}

\begin{figure}
\centering
\setlength\tabcolsep{2pt}
\begin{scriptsize}
\begin{tabular}{ccccc}
    \raisebox{0.095\linewidth}{\rotatebox[origin=c]{90}{\sffamily\siren}} &%
    \includegraphics[width=0.19\linewidth]{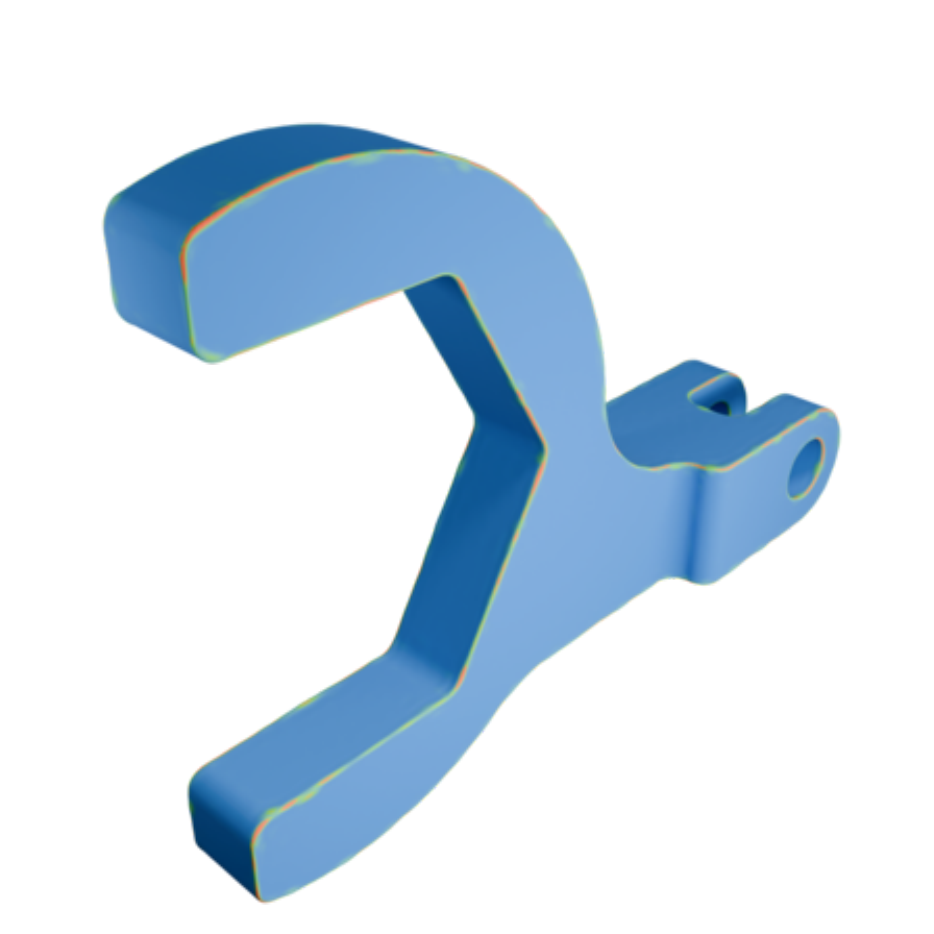} &%
    \includegraphics[width=0.19\linewidth]{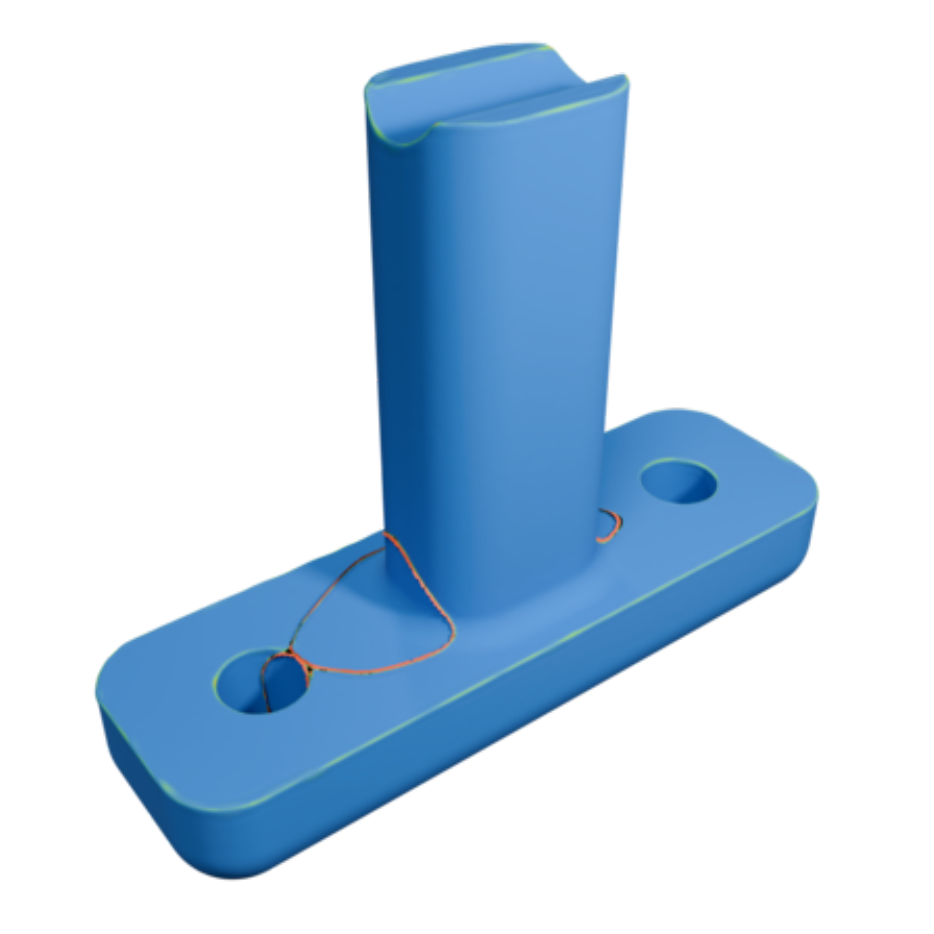} &%
    \includegraphics[width=0.19\linewidth]{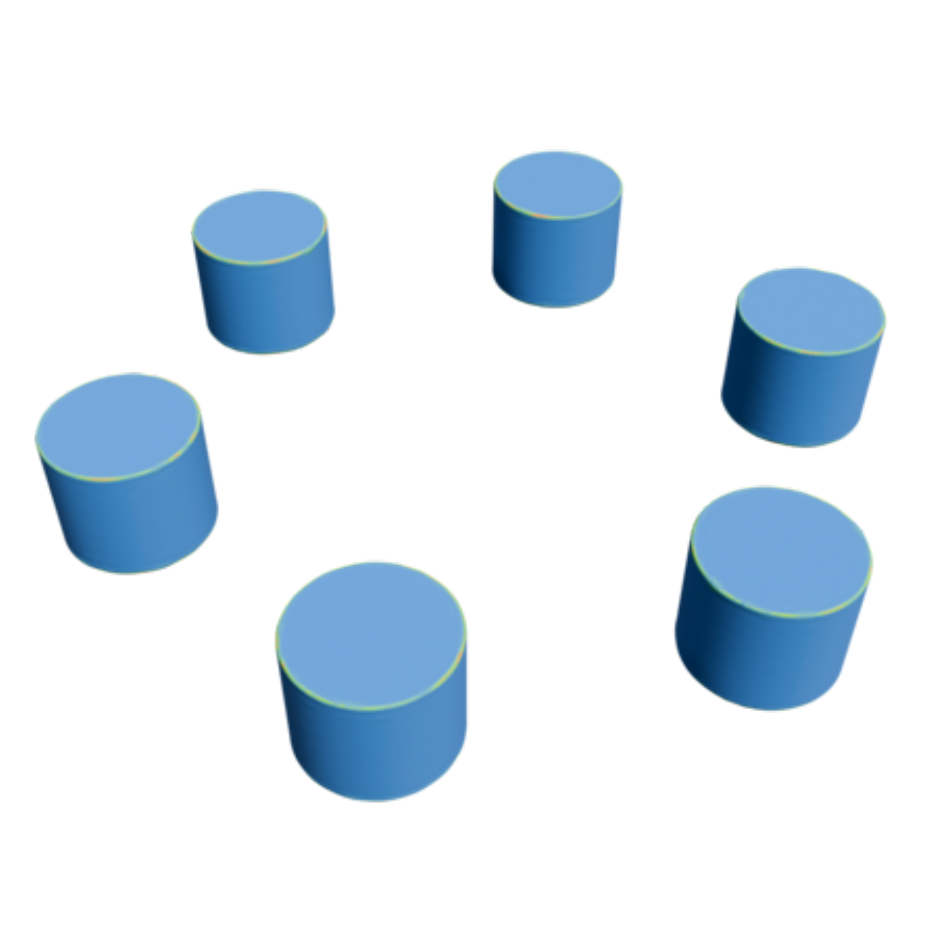} &%
    \includegraphics[width=0.19\linewidth]{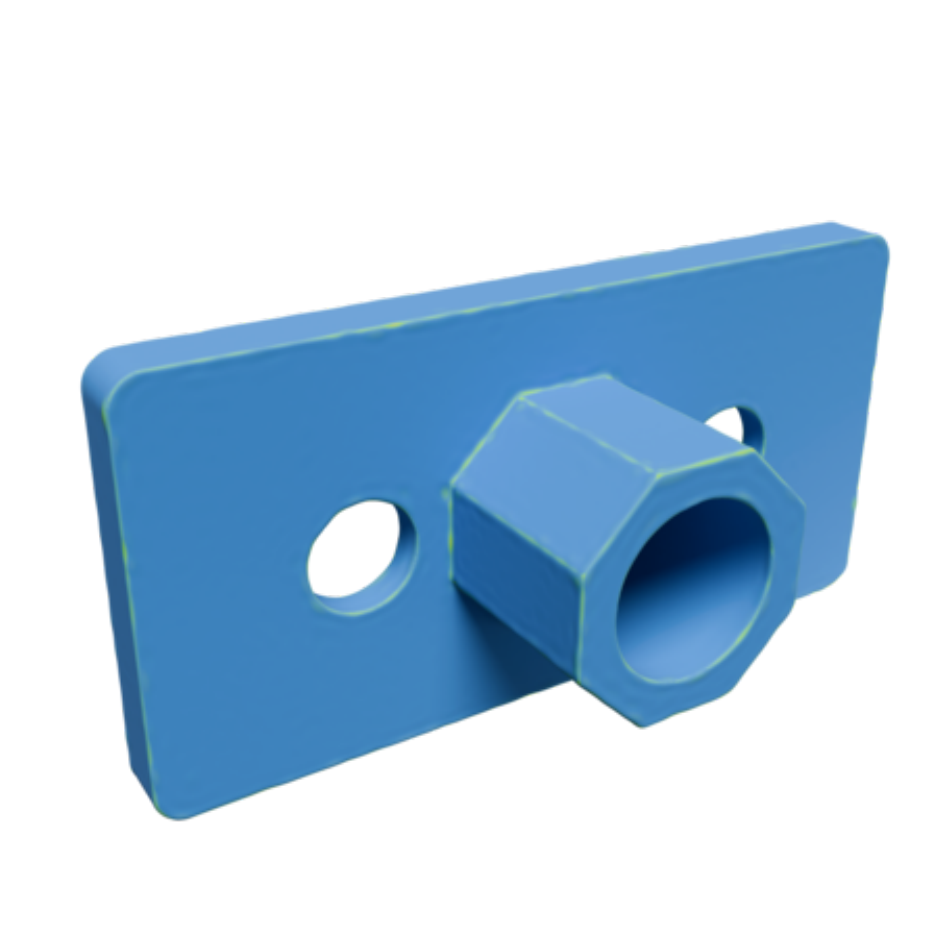} \\
    \makebox[0.01\linewidth]{} &%
    \makebox[0.22\linewidth]{2.451} &%
    \makebox[0.22\linewidth]{2.978} &%
    \makebox[0.22\linewidth]{2.126} &%
    \makebox[0.22\linewidth]{3.025} \\

    \raisebox{0.095\linewidth}{\rotatebox[origin=c]{90}{\sffamily NeurCADRecon}} &%
    \includegraphics[width=0.19\linewidth]{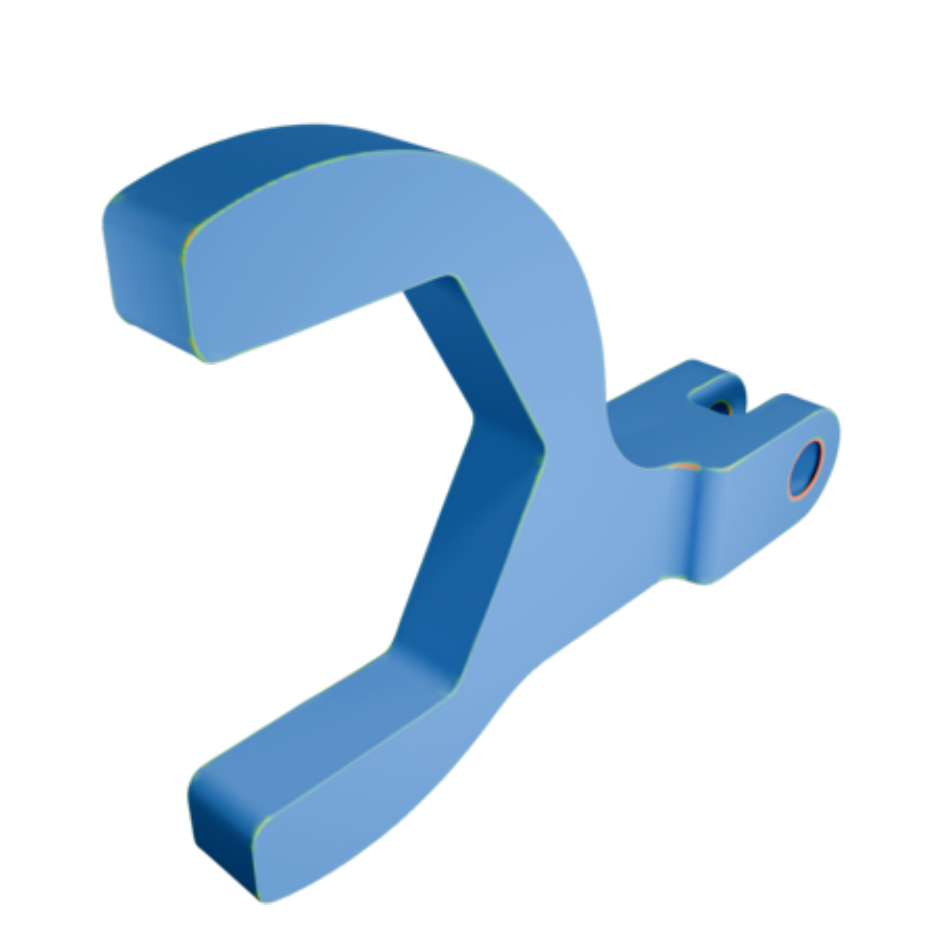} &%
    \includegraphics[width=0.19\linewidth]{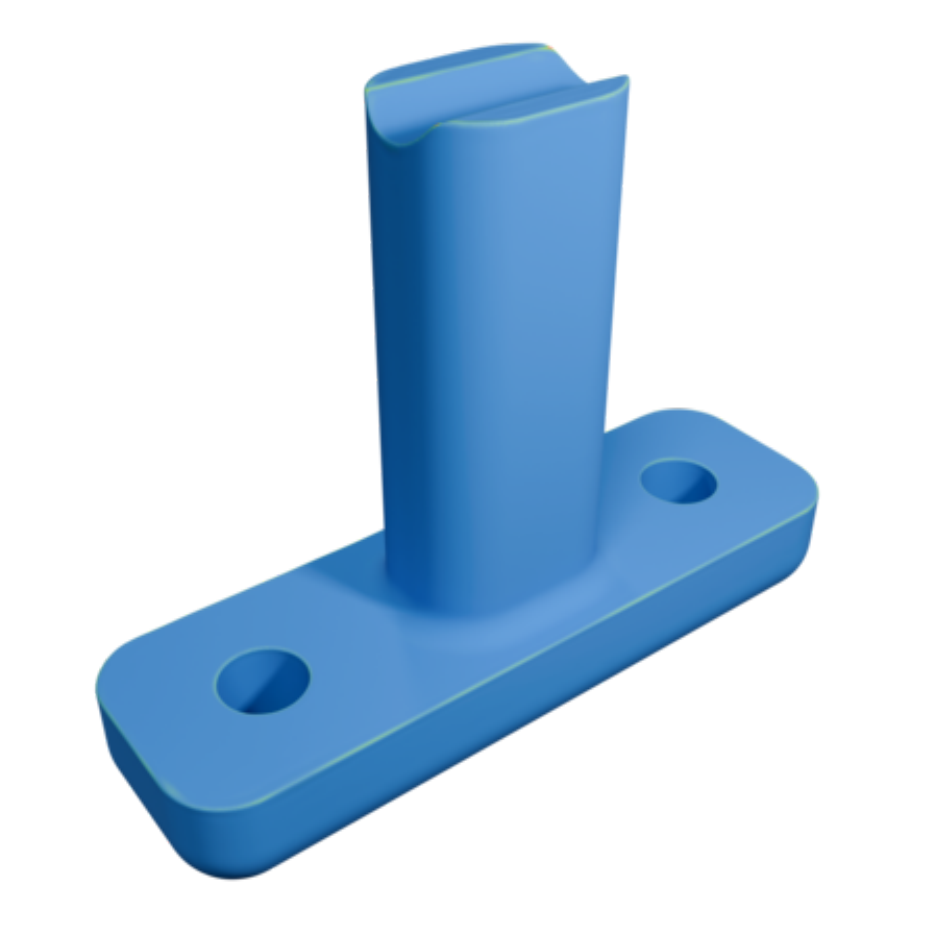} &%
    \includegraphics[width=0.19\linewidth]{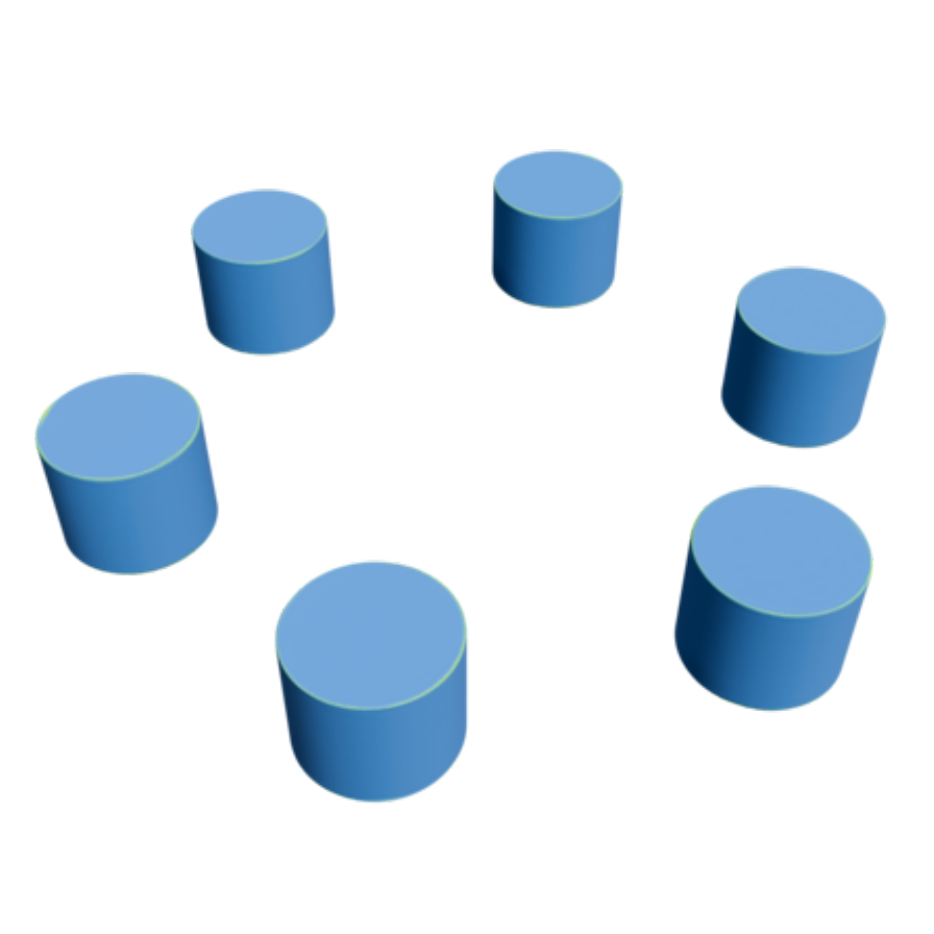} &%
    \includegraphics[width=0.19\linewidth]{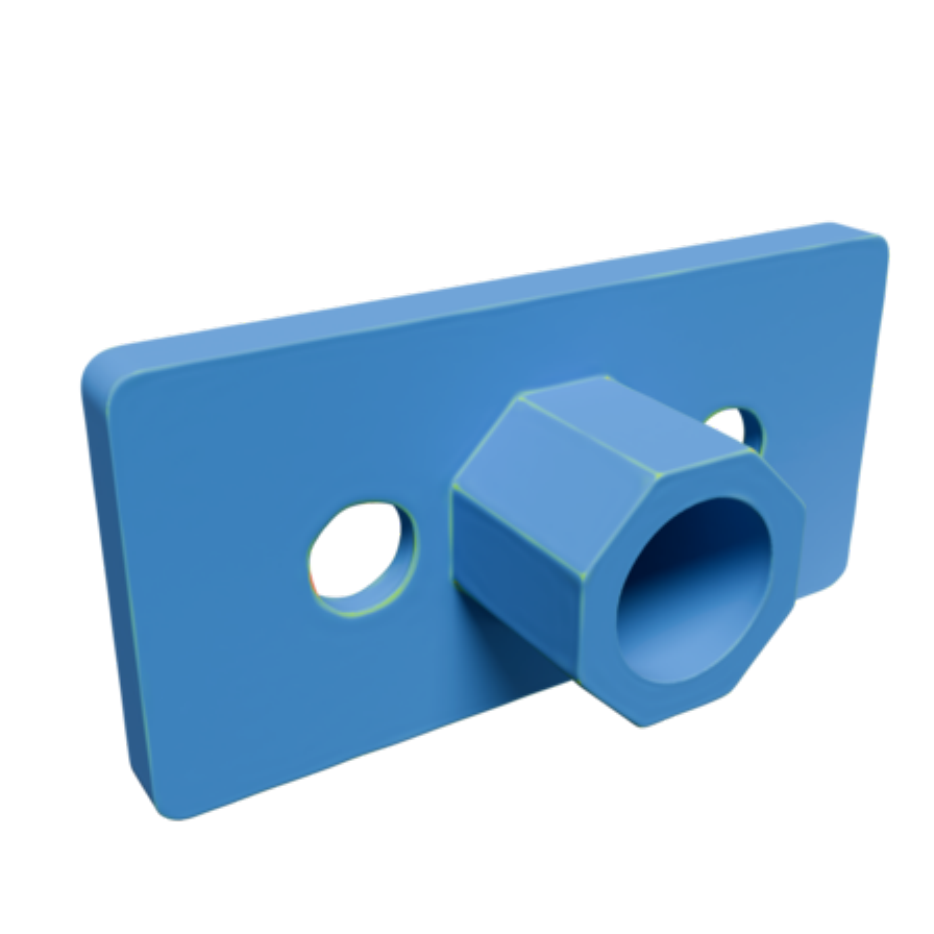} \\
    \makebox[0.01\linewidth]{} &%
    \makebox[0.22\linewidth]{2.316} &%
    \makebox[0.22\linewidth]{2.742} &%
    \makebox[0.22\linewidth]{2.008} &%
    \makebox[0.22\linewidth]{2.880} \\

    \raisebox{0.095\linewidth}{\rotatebox[origin=c]{90}{\sffamily\sharpnet}} &%
    \includegraphics[width=0.19\linewidth]{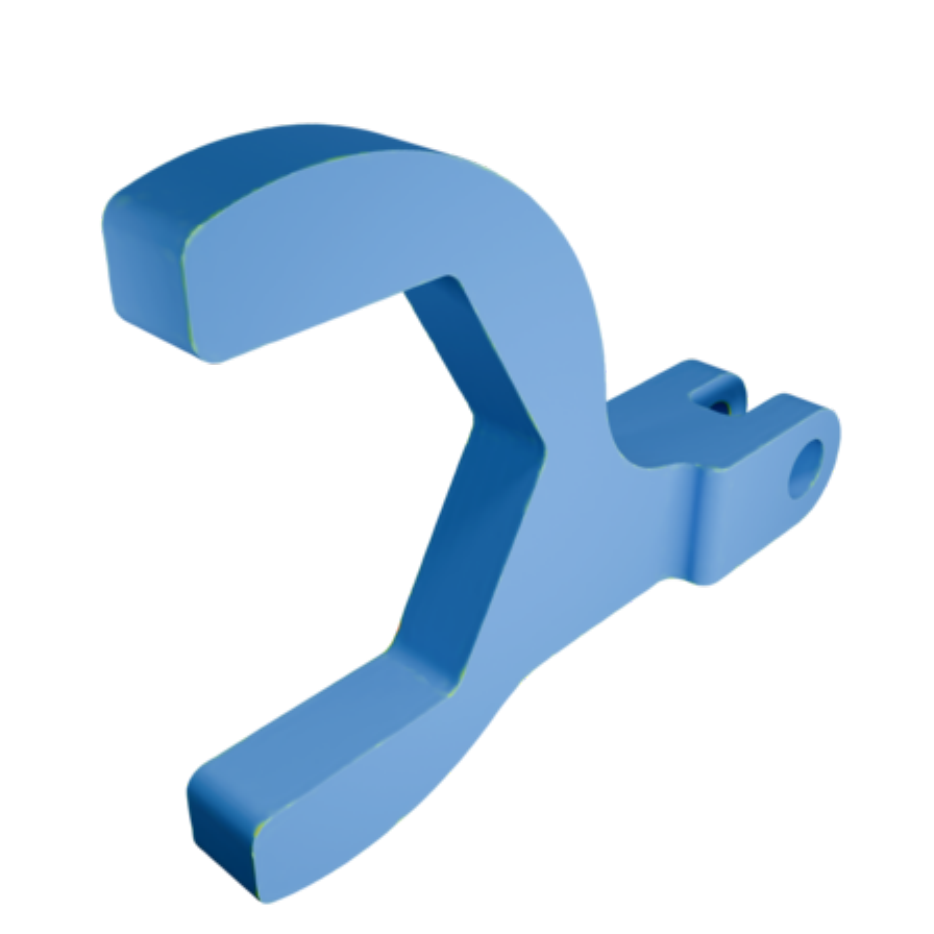} &%
    \includegraphics[width=0.19\linewidth]{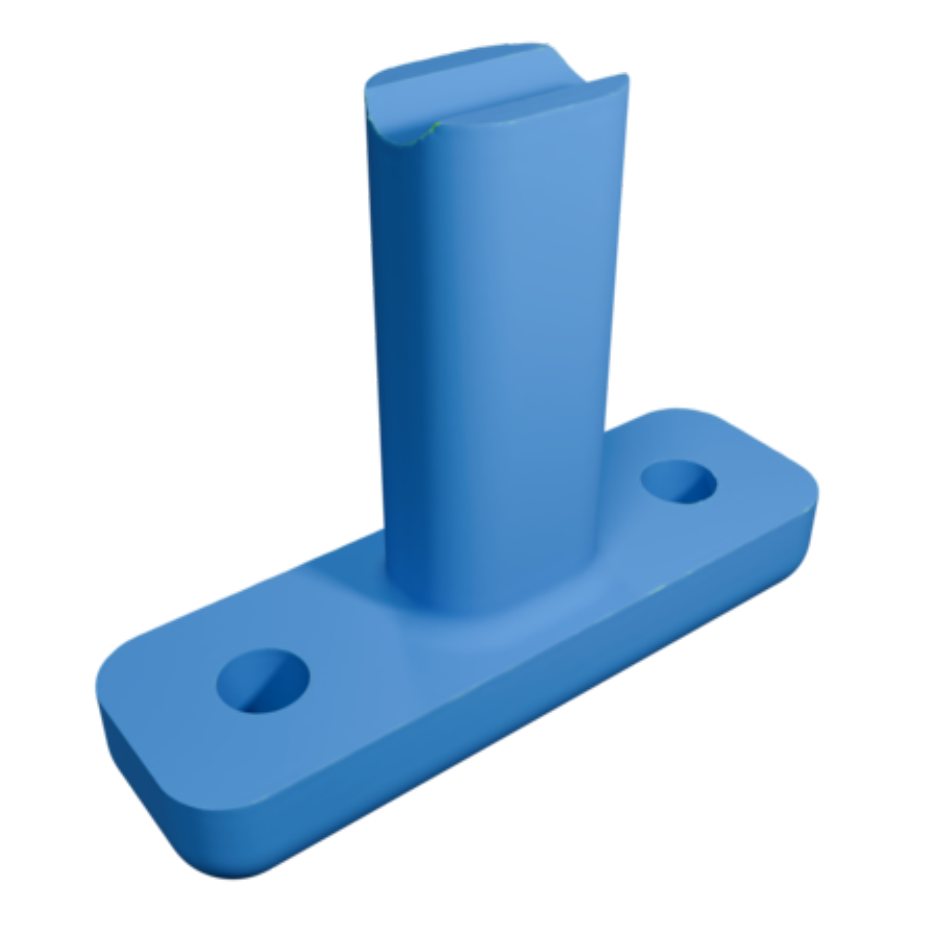} &%
    \includegraphics[width=0.19\linewidth]{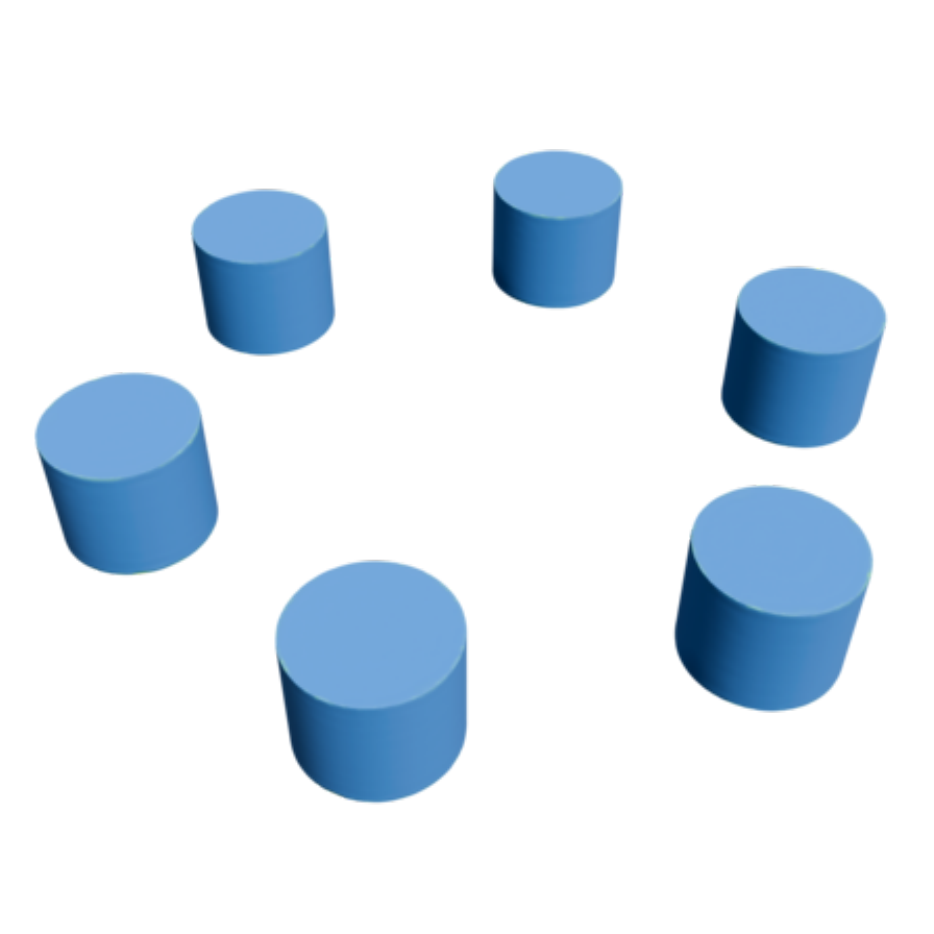} &%
    \includegraphics[width=0.19\linewidth]{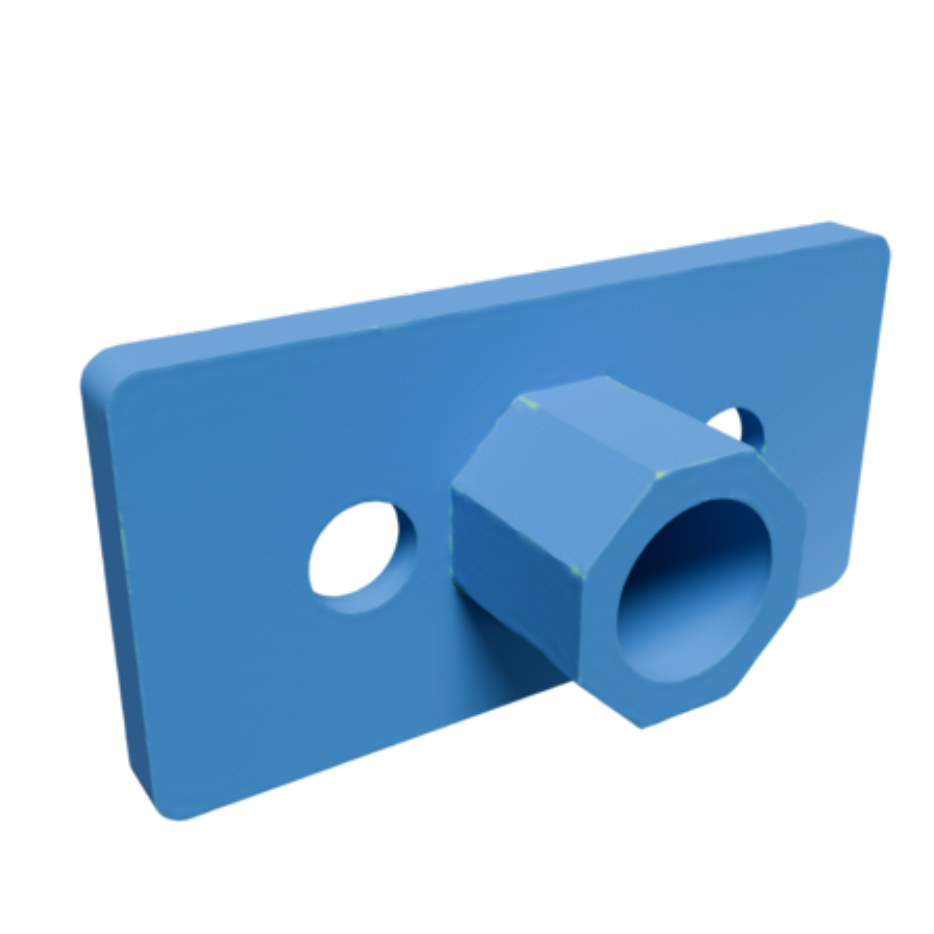} \\
    \makebox[0.01\linewidth]{} &%
    \makebox[0.22\linewidth]{2.293} &%
    \makebox[0.22\linewidth]{2.740} &%
    \makebox[0.22\linewidth]{2.005} &%
    \makebox[0.22\linewidth]{2.850} \\

\end{tabular}
\end{scriptsize}
\caption{Illustration of CD errors by color map. The CD is shown below each model. NeurCADRecon exhibits noticeably large errors along sharp edges, because its plain MLP architecture is unable to accurately model perfectly sharp edges.
}
\label{fig:pointcloud_input_color}
\end{figure}

Figure~\ref{fig:cad_pointcloud} presents the comparison results. While zero Gaussian curvature is advantageous for recovering sharp features, NeurCADRecon~\cite{Dong2024NeurCADRecon} still fails to reproduce clear, well-defined sharp structures because of the intrinsic smoothness of its MLP-based representation. The distributions of CD errors are shown in Figure~\ref{fig:pointcloud_input_color}. The quantitative statistics of the 100 testing models are listed in Table~\ref{tab:points_input}. Due to its controllable $C^0$-continuous representation, {\sharpnet} surpasses NeurCADRecon in both visual quality and quantitative metrics. In fact, {\sharpnet} has the potential to be a superior backbone for NeurCADRecon.

\begin{table}[!htbp]
\caption{Quantitative statistics of models reconstructed solely from points.}
\label{tab:points_input}
\begin{tabular}{ccccc}
\toprule
Method & $\cramped{\text{CD}^{\times 10^{-3}}}$ $\downarrow$  & $\cramped{\text{HD}^{\times 10^{-2}}}$ $\downarrow$ & $\text{NE}^{\circ}$ $\downarrow$ & $\text{FC}^\%$ $\uparrow$ \\
\midrule
SIREN        & 4.593 & 4.798 & 5.877 & 95.84\\
NeurCADRecon & 4.354 & 5.515 & 5.388 & 96.03\\
{\sharpnet}     & \textbf{4.129} & \textbf{4.004} & \textbf{3.839} & \textbf{97.21} \\
\bottomrule
\end{tabular}
\end{table}

\subsection{Ablation Studies}
\label{subsec:ablation}

\begin{figure}[!htbp]
    \centering
    \raisebox{0.11\linewidth}{
    \rotatebox[origin=c]{90}{\sffamily \small{Fixed Feature}}
    }
    \includegraphics[width=0.22\linewidth]{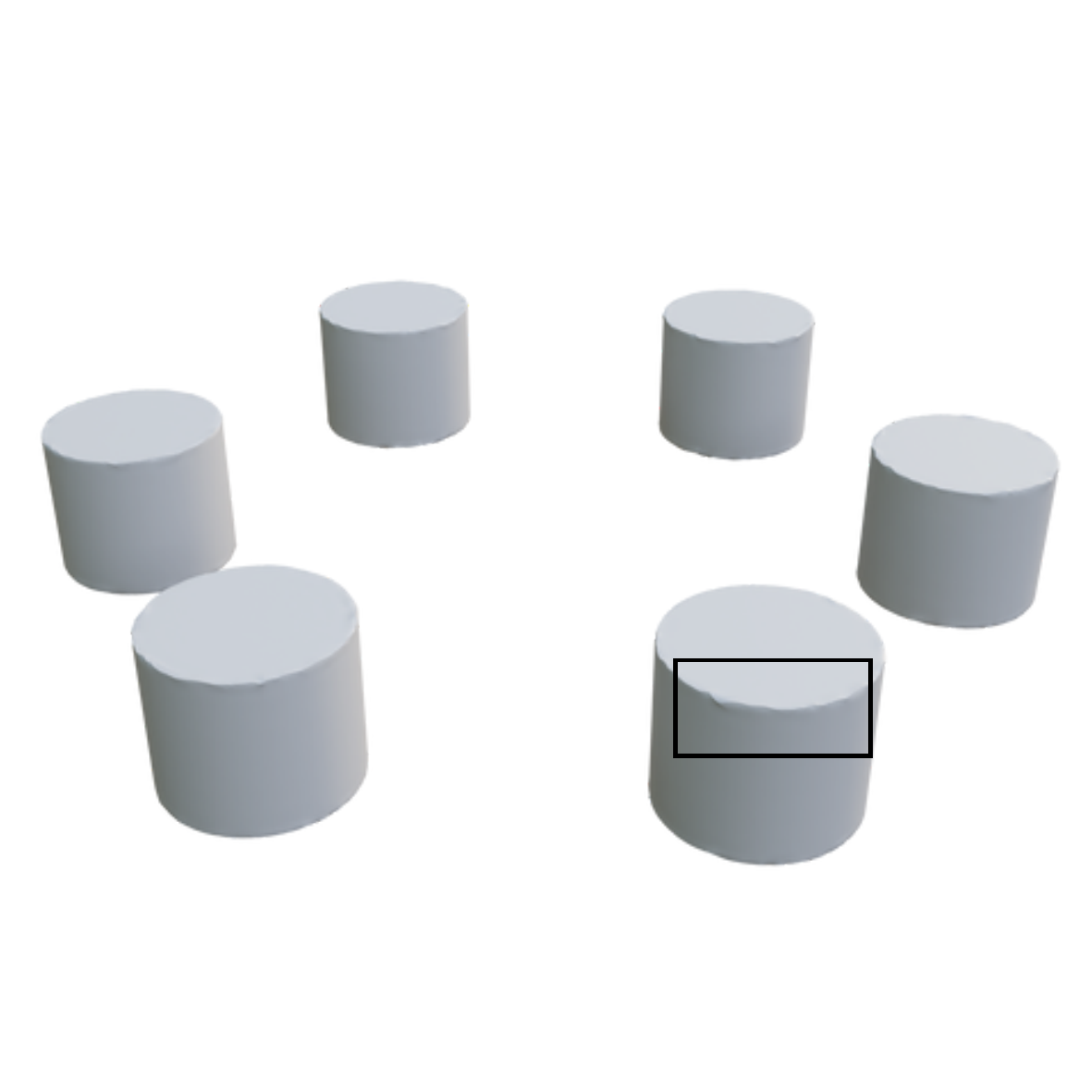}
    \includegraphics[width=0.22\linewidth]{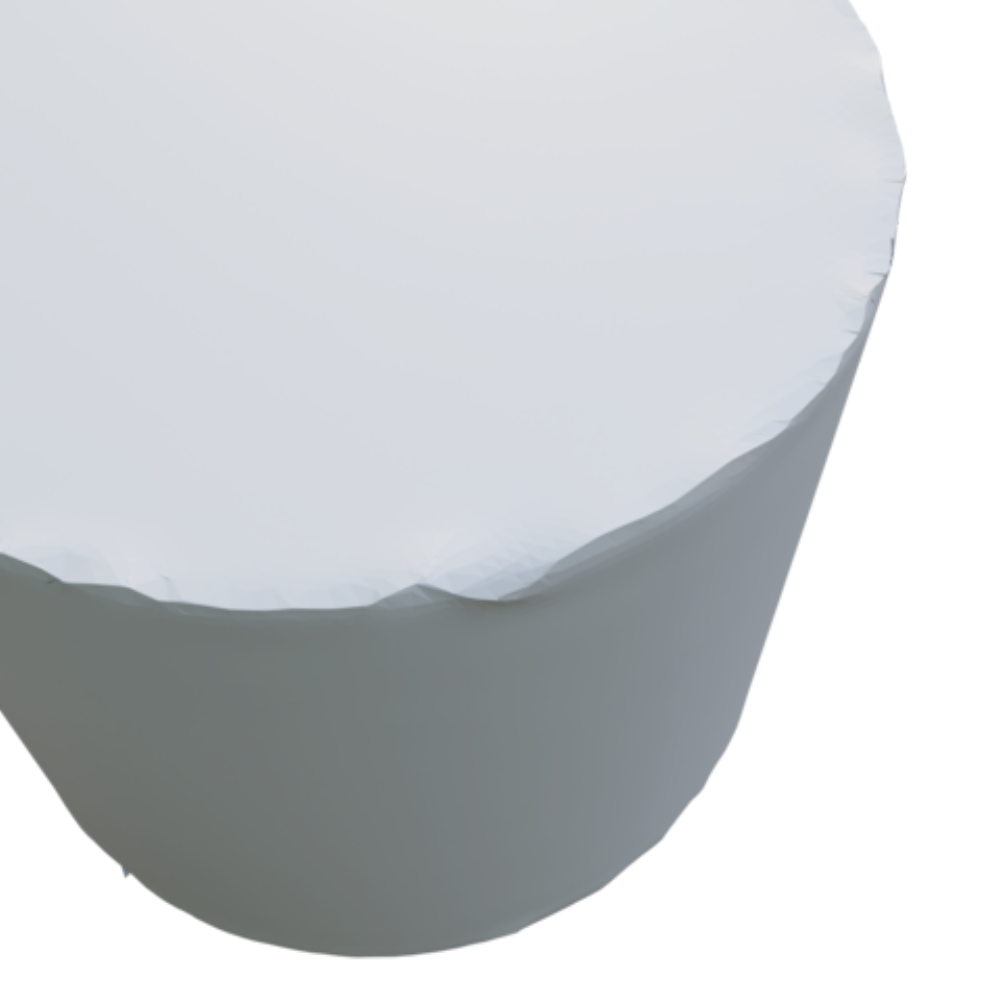}
    \includegraphics[width=0.22\linewidth]{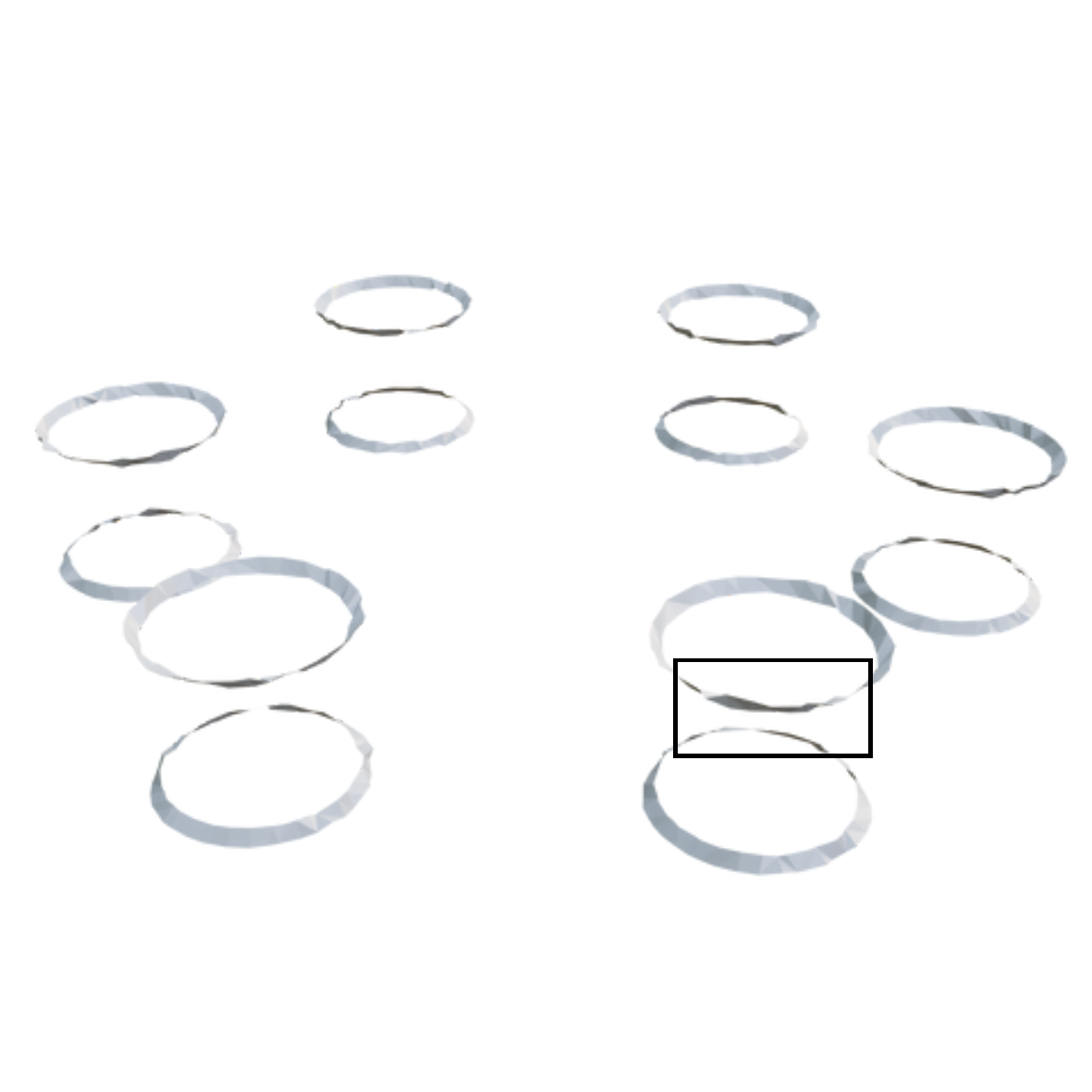}
    \includegraphics[width=0.22\linewidth]{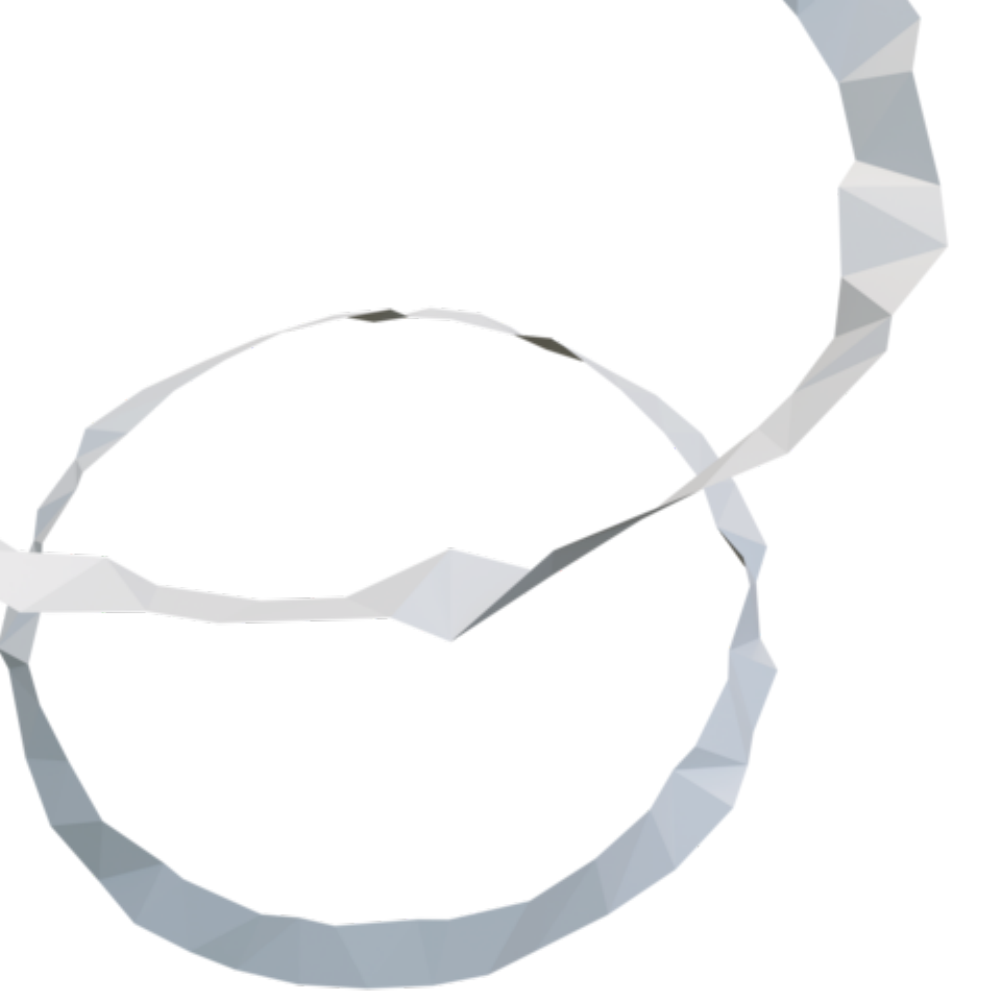}\\
    \raisebox{0.11\linewidth}{
    \rotatebox[origin=c]{90}{\sffamily \small{Learned Feature}}
    }
    \includegraphics[width=0.22\linewidth]{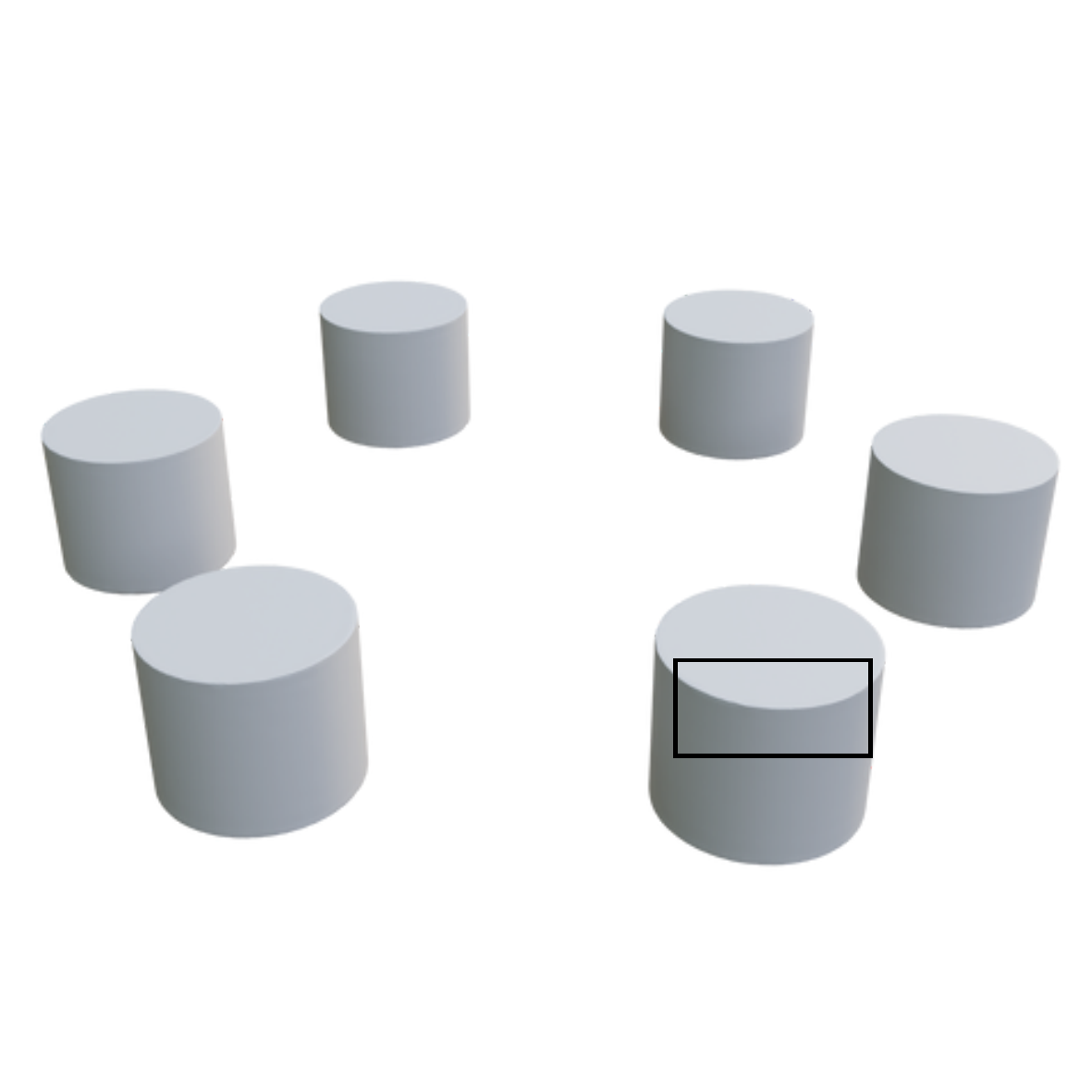}
    \includegraphics[width=0.22\linewidth]{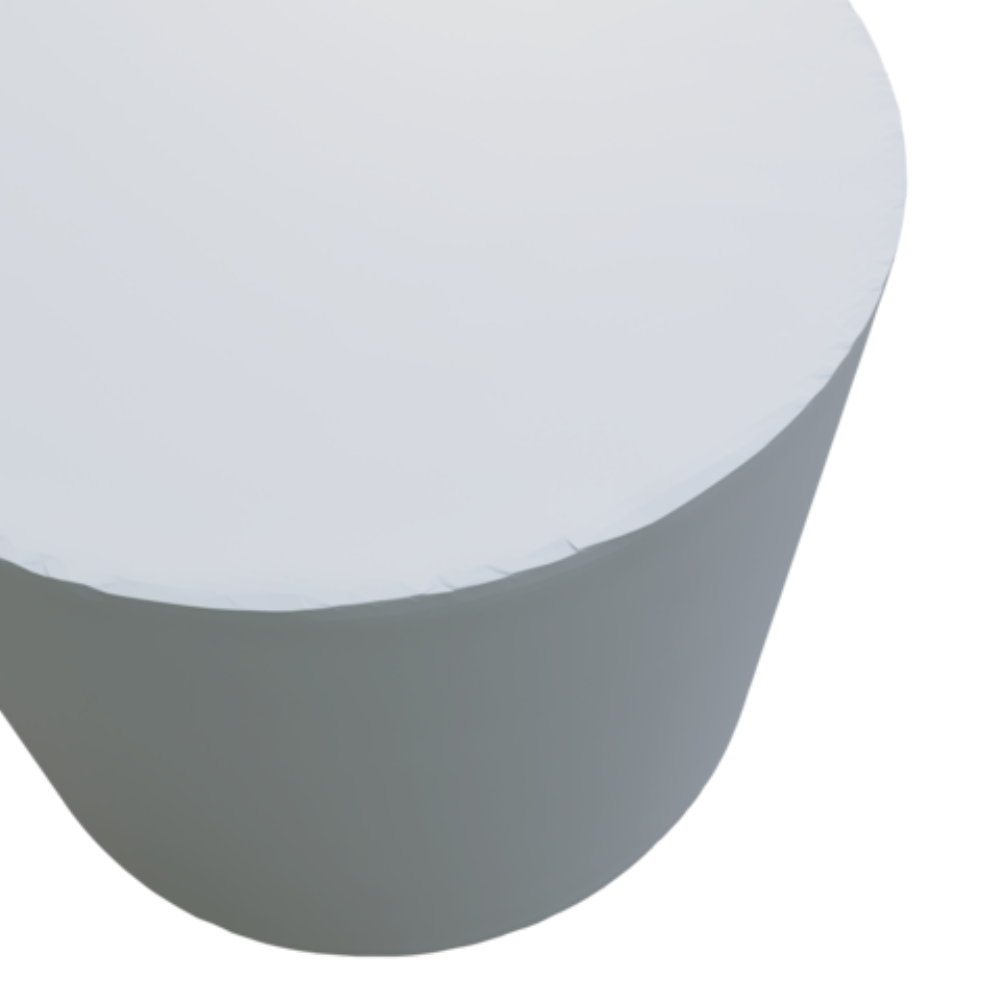}
    \includegraphics[width=0.22\linewidth]{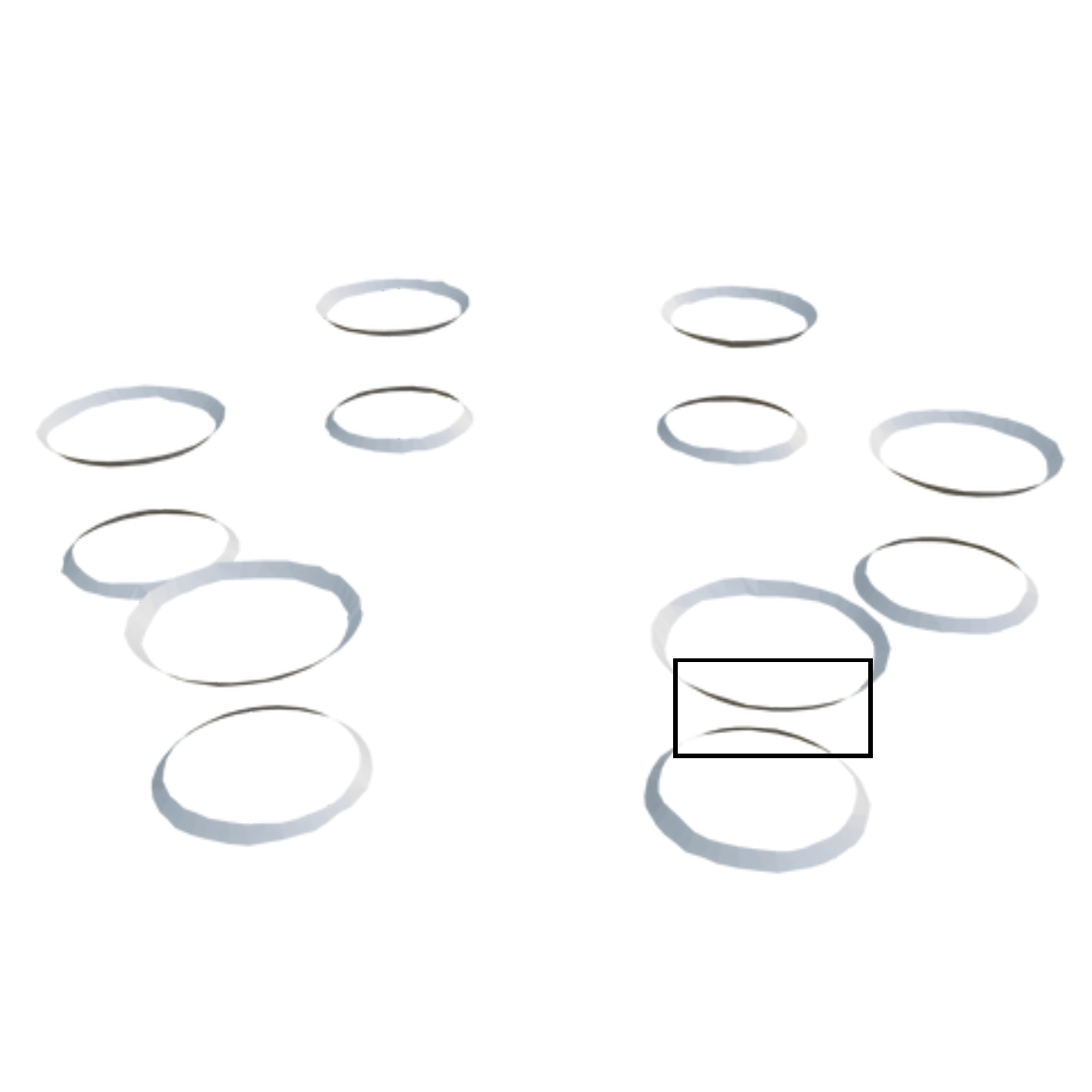}
    \includegraphics[width=0.22\linewidth]{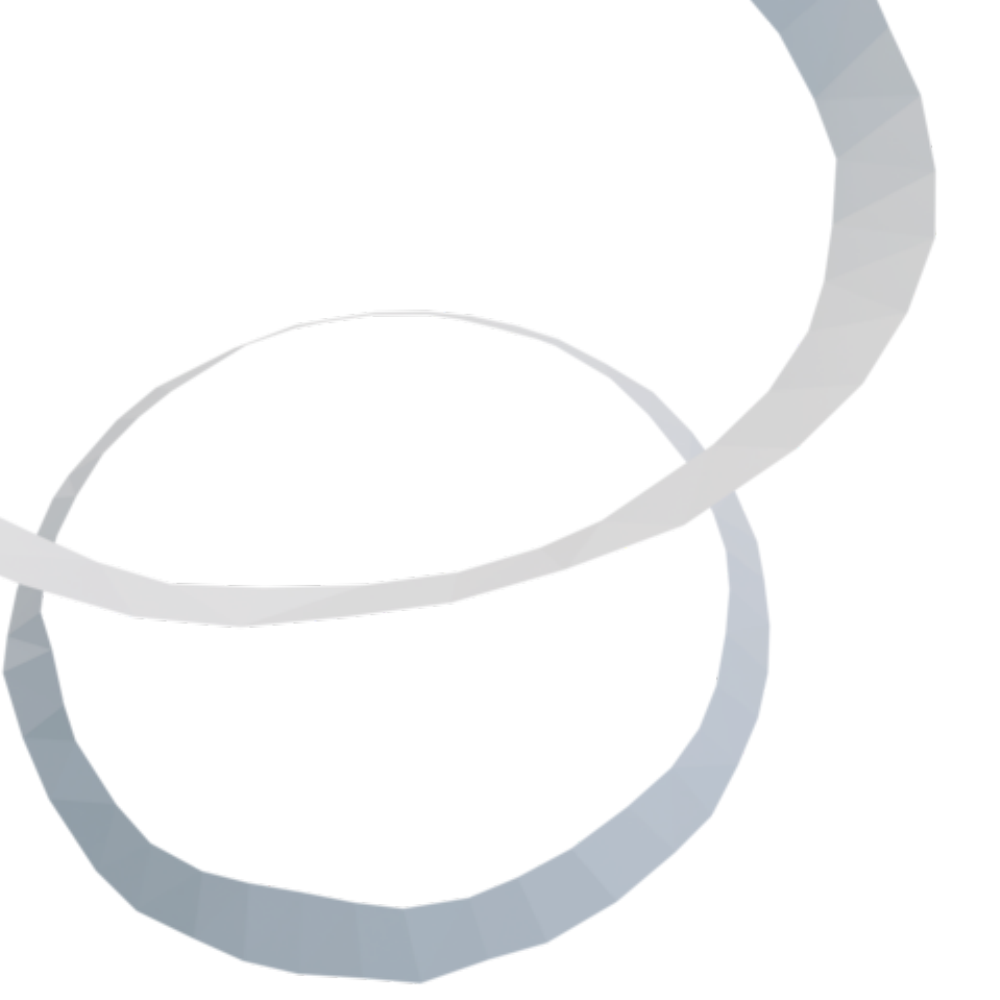}\\
    \makebox[0.08\linewidth]{}
    \makebox[0.44\linewidth]{\sffamily Reconstruction}
    \makebox[0.44\linewidth]{\sffamily Feature Set}
    \caption{Feature set initialization obtained from point clouds is often inaccurate, and using it without optimization leads to noticeable artifacts around sharp edges in the CAD models. In contrast, optimizing the feature set significantly improves reconstruction quality around sharp features.}
    \label{fig:feature_learning}
\end{figure}

\begin{figure}
    \centering
    \includegraphics[width=\linewidth]{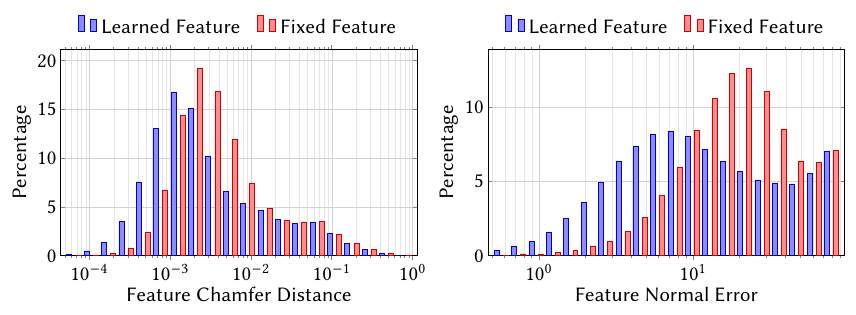}
    \caption{Histograms of the FCD and FNE distributions on a logarithmic scale. Both distance and angular errors exhibit a prominently shift toward zero after optimizing \(M\). This suggests that optimizing \(M\) allows {\sharpnet} to localize \(C^0\) features more accurately.}
    \label{fig:feature_learning_histogram}
\end{figure}

\paragraph{Learning of feature set $M$}
To evaluate the impact of feature set learning, we reconstruct CAD models from input points and oriented normals using the fixed initialized feature surface. During optimization of the loss function in Equation~\eqref{eqn:3Dloss_points}, the location of the points of $M$ is kept fixed. The comparison in Figure~\ref{fig:feature_learning} illustrates that, without a learnable $M$, noticeable artifacts appear along sharp edges, while allowing the learning of $M$ largely eliminates these artifacts.
To quantitatively assess the impact of learning, let \(S\) be the zero-level set of {\sharpnet}. We obtain the zero-level curve \(L\) of {\sharpnet} restricted to the feature set \(M\), given by \(L = S \cap M\), which represents the collection of non-smooth points on the reconstructed surface. We then compute the Chamfer distance between \(L\) and the ground-truth sharp edges, which we denote as the feature Chamfer distance (FCD). In addition, we evaluate the normals of \(M\) at sample points on \(L\) and compare them against the ground-truth normals; we refer to this quantity as the feature normal error (FNE) in degrees.
Figure~\ref{fig:feature_learning_histogram} shows the FCD and FNE distributions in the entire data set obtained with and without feature learning. It is clear that feature learning leads to substantial reductions in FCD and FNE compared to the use of fixed features.

\paragraph{Mollifier}
The performance of {\sharpnet} depends mainly on the complexity of the feature set \(M\). Let \(m\) denote the number of triangles in \(M\). When training on \(n\) sampled points, the number of the global integral terms in Equation~\eqref{eqn:feature} is \(O(mn)\), which becomes a bottleneck when \(M\) is relatively complex. The mollifier transforms the global integral into the local integral in Equation~\eqref{eqn:feature_local}. For each sample point, the algorithm scans all triangles on $M$ to identify those lying within the specified mollifier radius, and then evaluates the integrals only over this subset. Although this selection step still incurs an overall complexity of $O(mn)$, it is inexpensive in practice, and it does not track gradients or contribute to backpropagation. Since the global integral is reduced to local integrals and the number of triangles within the radius for any sample point is roughly constant, the number of the local integral terms is effectively reduced to approximately $O(n)$, substantially decreasing both computation time and memory requirements. We train {\sharpnet} on a single NVIDIA RTX 4090 (24\,GB) GPU and measure both training speed and memory consumption with and without the mollifier.

With the mollifier, under the setting of Section~\ref{sec:CAD_mesh}, {\sharpnet} trains in about 20--34 minutes (23 minutes on average) and consumes 1.2--3.2\,GB of GPU memory (1.9\,GB on average).
Without the mollifier, the Eikonal loss \(\mathcal{L}_{\mathrm{ekl}}\) triggers out-of-memory errors during backpropagation due to its reliance on second-order derivatives. Even when reducing the number of sampled points per epoch to one tenth of the original, around $40\%$ of experiments still encounter memory overflow. In this setting, the median training time rises to 110 minutes and the median memory usage reaches 15\,GB. Considering that the non-mollified version uses only one tenth of the original samples, the mollifier yields nearly two orders of magnitude improvement in both training speed and memory efficiency.

\begin{figure}
\centering
\begin{subcaptionblock}{.49\linewidth}
    \centering
    \begin{minipage}{0.69\linewidth}
        \includegraphics[width=\linewidth]{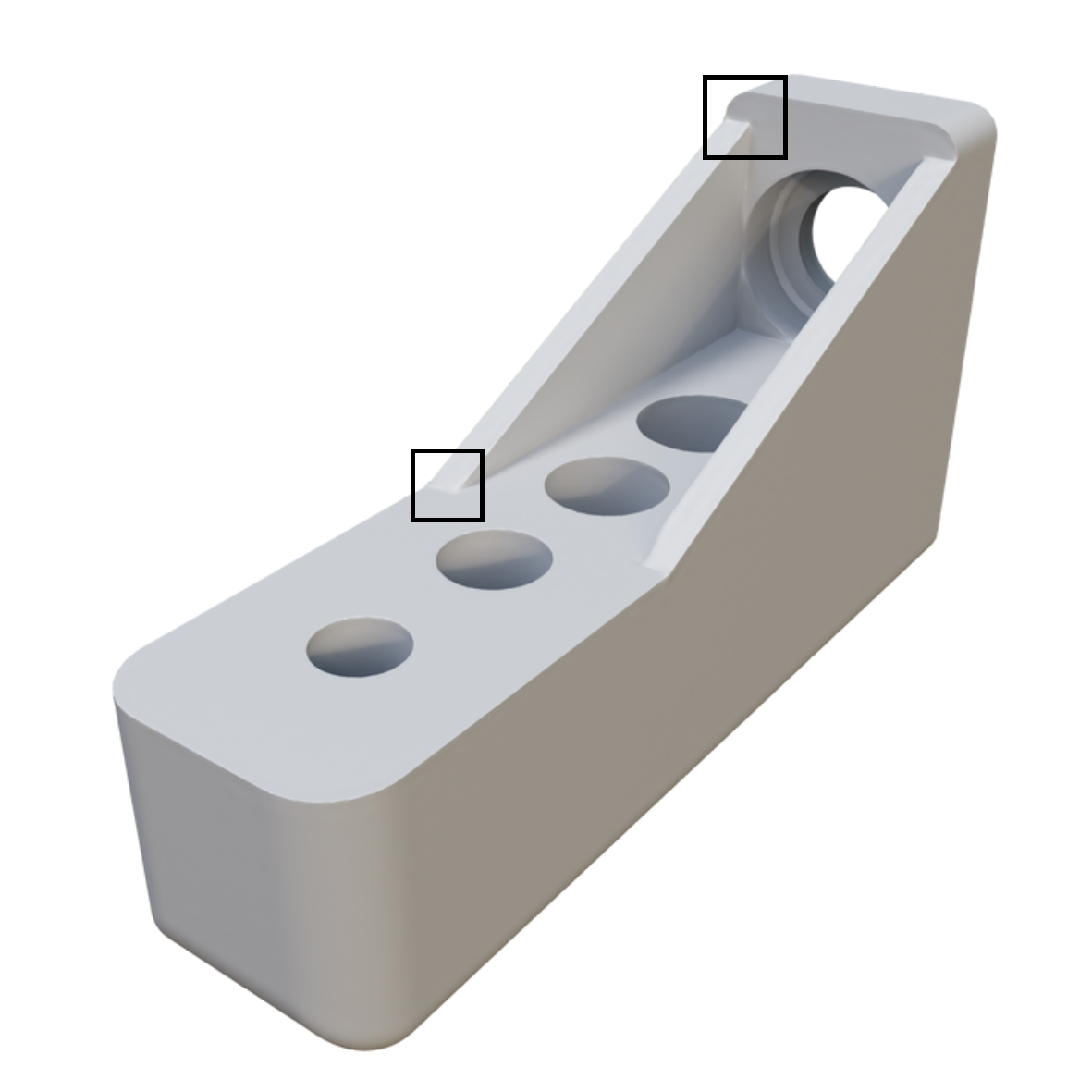}
    \end{minipage}
    \begin{minipage}{0.29\linewidth}
        \includegraphics[width=\linewidth]{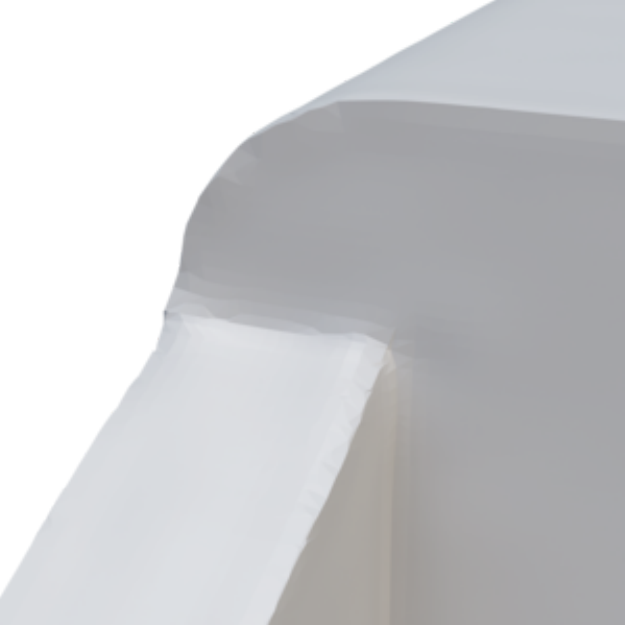}
        \vspace{5pt}
        \includegraphics[width=\linewidth]{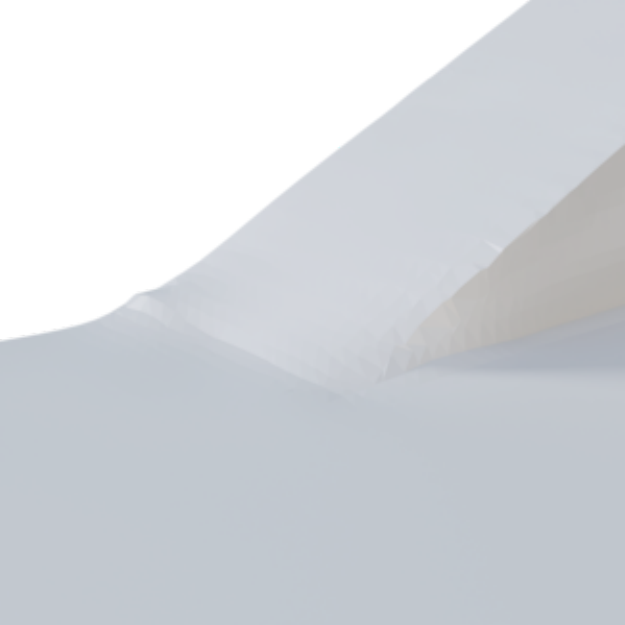}
    \end{minipage}
    \caption*{{\method} w/o Splitting}
\end{subcaptionblock}
\begin{subcaptionblock}{.49\linewidth}
    \centering
    \begin{minipage}{0.69\linewidth}
        \includegraphics[width=\linewidth]{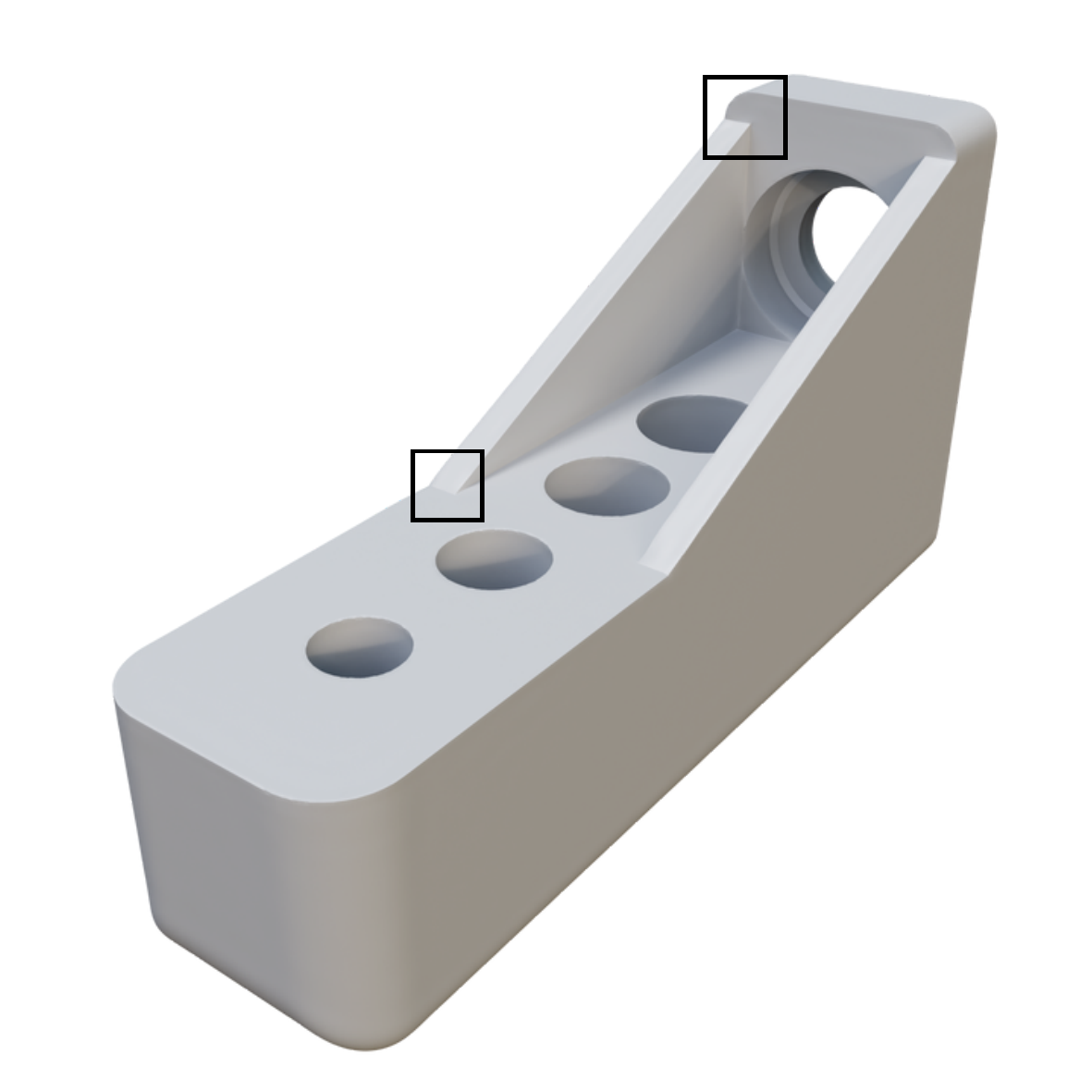}
    \end{minipage}
    \begin{minipage}{0.29\linewidth}
        \includegraphics[width=\linewidth]{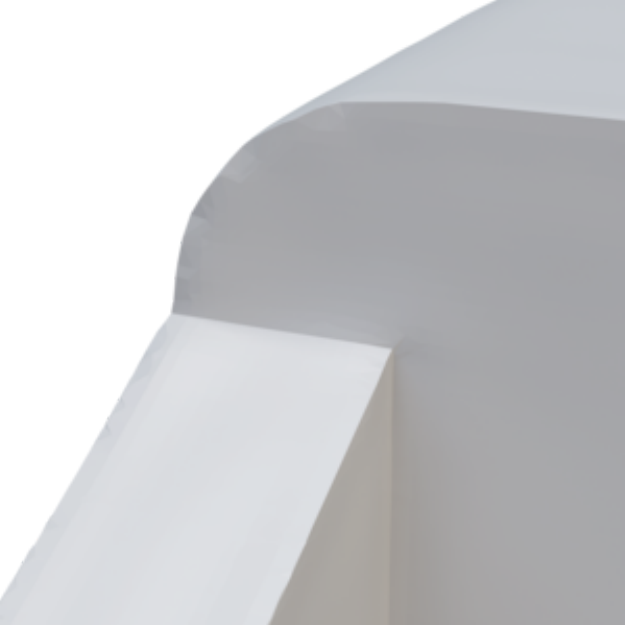}
        \vspace{5pt}
        \includegraphics[width=\linewidth]{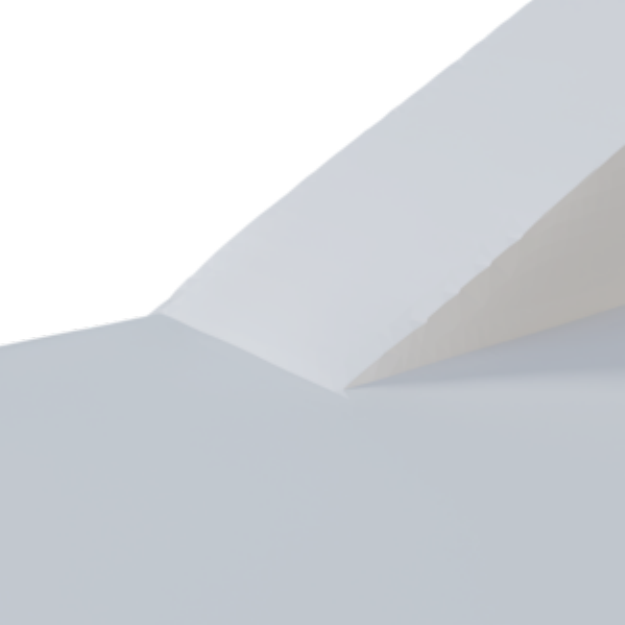}
    \end{minipage}
    \caption*{{\method} w/ Splitting}
\end{subcaptionblock}
\caption{Without applying sharp curve splitting at joints, the corners of models produce artifacts, because of the discontinuity of gradient differences at these corners.}
\label{fig:split_feature}
\end{figure}

\paragraph{Splitting of feature set $M$}
The splitting of $M$ enables the representation of a jump in the derivative difference. The jump may appear at sharp corners where concave and convex sharp curves meet. As illustrated in Figure~\ref{fig:split_feature}, visible artifacts arise at these corners if no splitting is applied. Once splitting is performed, the corners are accurately reconstructed, demonstrating that {\sharpnet} can capture not only a normal derivative jump but also the discontinuity of the normal derivative jump.

\subsection{Limitations}

\begin{figure}
\centering
\begin{subcaptionblock}{.48\linewidth}
    \centering
    \includegraphics[width=0.49\linewidth]{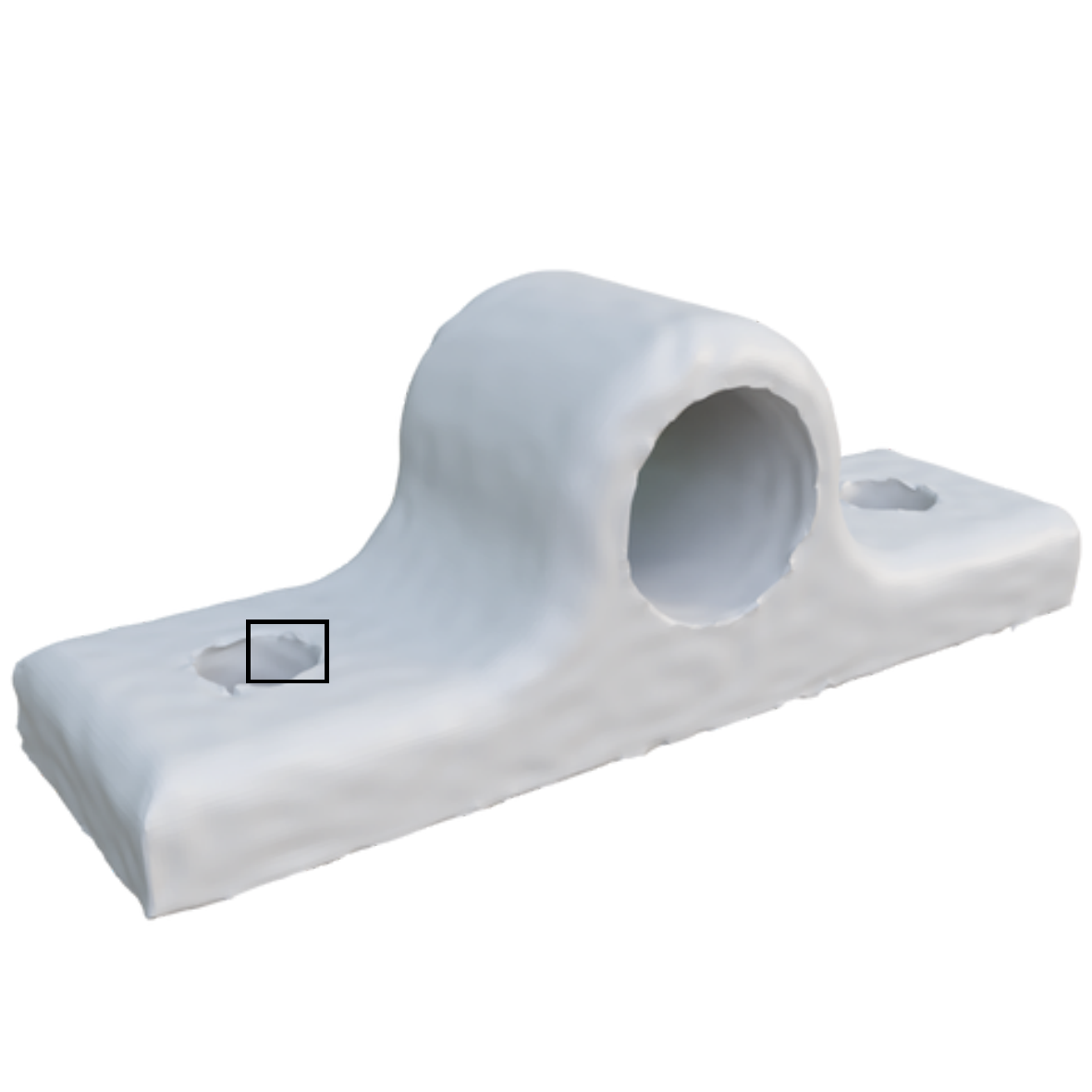}
    \includegraphics[width=0.49\linewidth]{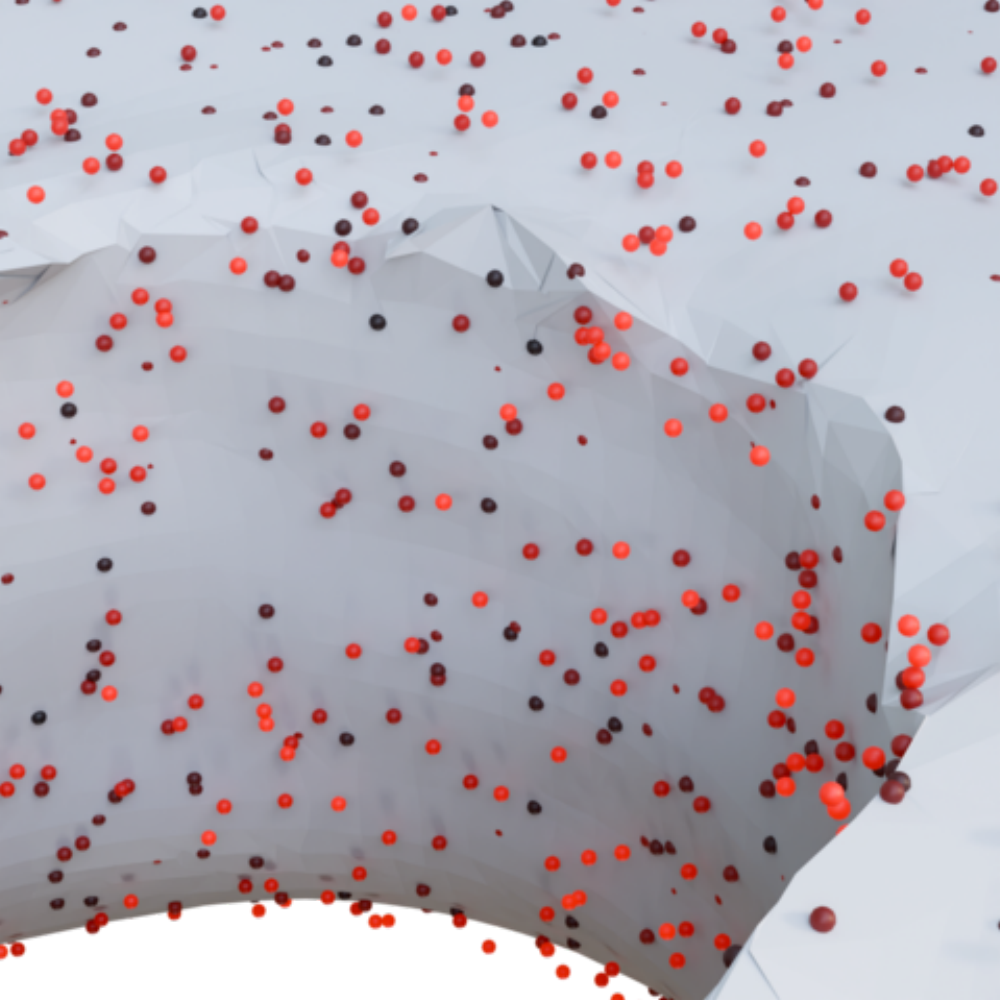}
    \caption{Learned feature}
    \label{fig:noised_pointcloud_learned}
\end{subcaptionblock}
\begin{subcaptionblock}{.48\linewidth}
    \centering
    \includegraphics[width=0.49\linewidth]{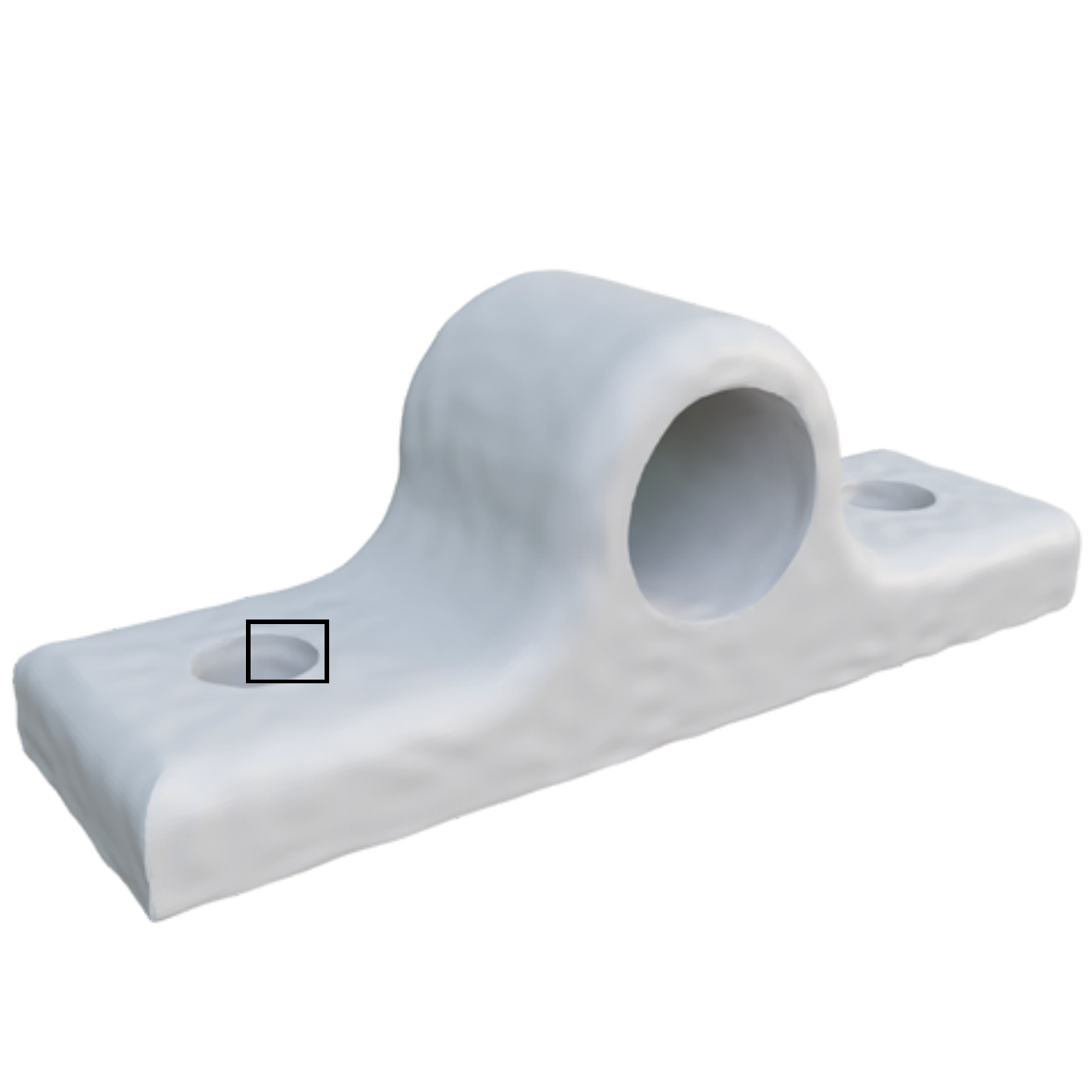}
    \includegraphics[width=0.49\linewidth]{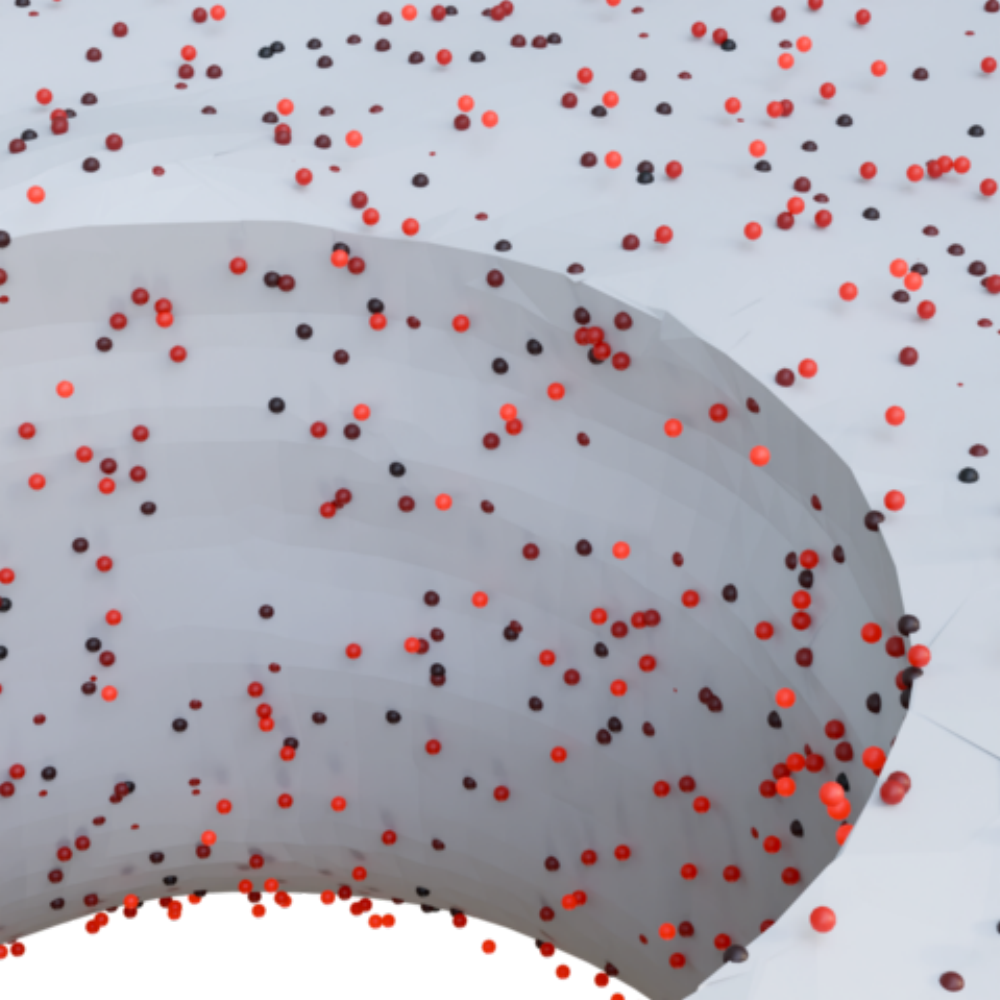}
    \caption{Fixed ground-truth feature}
    \label{fig:noised_pointcloud_fixed}
\end{subcaptionblock}
\caption{Limitations: feature set learning is sensitive to noise in the input point cloud. We visualize the reconstructed surfaces using the learned feature surface $M$ in~(\subref{fig:noised_pointcloud_learned}) and the fixed ground-truth surface $M$ in~(\subref{fig:noised_pointcloud_fixed}), respectively, in the presence of noisy input. The input points are colored according to their point-to-surface distances. When the ground-truth feature set $M$ is given, {\method} can recover sharp edges reasonably well despite the noise. However, learning the feature set $M$ directly from noisy points is challenging, which leads to degraded  reconstruction quality.}
\label{fig:noised_pointcloud}
\end{figure}

In principle, the feature surface $M$ can be any curve or surface. In practice, however, we discretize it into polylines or triangular meshes to enable efficient evaluation of the integrals, leveraging the fact that the integrals of Green's function admit closed-form expressions on line segments and triangles.

The feature surface $M$ must be known in advance or provided through an approximate initialization; it cannot be learned directly from scratch. In addition, we assume that the initial feature surface $M$ has the correct topology, since topological defects (e.g., cracks or holes) cannot be corrected during the optimization process.

In practice, we observe that the reconstruction of sharp features is highly sensitive to noise in the input point cloud. When the feature surface $M$ is initialized directly from noisy data, high noise levels affect multiple stages of the pipeline. First, noise degrades the extraction of feature curves using existing point‑cloud methods, such as NerVE~\cite{Zhu2023NerVE}. Second, estimating guiding directions from nearest‑neighbor neighborhoods becomes unstable. Finally, during joint optimization, noisy samples provide unreliable supervision, preventing the feature surface from being accurately learned, as illustrated in Figure~\ref{fig:noised_pointcloud}(\subref{fig:noised_pointcloud_learned}). By contrast, when the ground‑truth feature surface $M$ is provided and fixed during training, noise primarily affects the smoothness of the reconstructed surface, while the sharp features themselves remain largely accurate, as shown in Figure~\ref{fig:noised_pointcloud}(\subref{fig:noised_pointcloud_fixed}).

\section{Conclusions \& Future Directions}
\label{sec:discussions}
We introduce {\sharpnet}, a fully controllable $C^0$-continuous MLP-based representation. {\sharpnet} maintains $C^0$ continuity on the designated feature surface while remaining smooth elsewhere. Our meshless representation is more flexible and more precise than other representations, such as mesh-based approaches, which often introduce unnecessary $C^0$ artifacts in regions that should be smooth. Moreover, our feature function has a differentiable closed-form expression, enabling the feature surface itself to be learned jointly with the MLP parameters. We assess {\sharpnet} on various 2D tasks and CAD reconstruction problems. Our method can learn implicit neural CAD representations that preserve sharp edges, both when sharp edges are provided and when no sharp edge priors are available. Comparisons with several state-of-the-art baselines show that {\sharpnet} achieves superior performance.

{\sharpnet} can be extended to model non-differentiable features in time-varying signals or spatiotemporal fields, such as sharp transitions in motion or deformation data. Another promising direction is to represent higher-order discontinuities, such as jumps in curvature or other non-smooth differential quantities, by incorporating higher-order partial differential equations in place of Poisson’s equation. In addition, {\sharpnet} can be generalized to represent $C^{-1}$-continuous functions, that is, functions that are discontinuous at user-specified locations. This extension would significantly broaden its applicability, especially for tasks involving abrupt transitions, including image vectorization, medical image segmentation, and scientific simulations with shocks or material interfaces. The generalization can be realized by relaxing the continuity condition in Equation~\eqref{eqn:continuity} and applying Dirichlet boundary conditions that assign distinct function values on both sides of the feature set $M$. We identify this direction as an important avenue for future work.

\begin{acks}
We thank the anonymous reviewers for their constructive feedback.
We thank Xin Pan for assisting with the editing of part of the video demonstration.
We thank Christopher T. Howlett, from The Noun Project, for creating the bat shape used in Figure~\ref{fig:teaser}(\subref{fig:teaser_bat}), licensed under a \href{https://creativecommons.org/licenses/by/3.0/deed.en}{Creative Commons Attribution 3.0 Unported} License.
This research was partially supported by the National Key R\&D Program of China (2023YFB3002901), the Research Projects of ISCAS (ISCAS-JCMS-202303, ISCAS-ZD-202401, ISCAS-JCZD-202402, and ISCAS-JCMS-202403), and the Ministry of Education, Singapore, under its Academic Research Fund Grant RT19/22.
\end{acks}

\bibliographystyle{ACM-Reference-Format}
\bibliography{reference}


\begin{thebibliography}{46}


\ifx \showCODEN    \undefined \def \showCODEN     #1{\unskip}     \fi
\ifx \showISBNx    \undefined \def \showISBNx     #1{\unskip}     \fi
\ifx \showISBNxiii \undefined \def \showISBNxiii  #1{\unskip}     \fi
\ifx \showISSN     \undefined \def \showISSN      #1{\unskip}     \fi
\ifx \showLCCN     \undefined \def \showLCCN      #1{\unskip}     \fi
\ifx \shownote     \undefined \def \shownote      #1{#1}          \fi
\ifx \showarticletitle \undefined \def \showarticletitle #1{#1}   \fi
\ifx \showURL      \undefined \def \showURL       {\relax}        \fi
\providecommand\bibfield[2]{#2}
\providecommand\bibinfo[2]{#2}
\providecommand\natexlab[1]{#1}
\providecommand\showeprint[2][]{arXiv:#2}

\bibitem[Belhe et~al\mbox{.}(2023)]%
        {Belhe2023}
\bibfield{author}{\bibinfo{person}{Yash Belhe}, \bibinfo{person}{Micha\"{e}l Gharbi}, \bibinfo{person}{Matthew Fisher}, \bibinfo{person}{Iliyan Georgiev}, \bibinfo{person}{Ravi Ramamoorthi}, {and} \bibinfo{person}{Tzu-Mao Li}.} \bibinfo{year}{2023}\natexlab{}.
\newblock \showarticletitle{Discontinuity-Aware {2D} Neural Fields}.
\newblock \bibinfo{journal}{\emph{ACM Transactions on Graphics}} \bibinfo{volume}{42}, \bibinfo{number}{6}, Article \bibinfo{articleno}{217} (\bibinfo{date}{Dec.} \bibinfo{year}{2023}), \bibinfo{numpages}{11}~pages.
\newblock
\showISSN{0730-0301}
\href{https://doi.org/10.1145/3618379}{doi:\nolinkurl{10.1145/3618379}}


\bibitem[Chang et~al\mbox{.}(2025)]%
        {Chang2025LiftingWN}
\bibfield{author}{\bibinfo{person}{Yue Chang}, \bibinfo{person}{Mengfei Liu}, \bibinfo{person}{Zhecheng Wang}, \bibinfo{person}{Peter~Yichen Chen}, {and} \bibinfo{person}{Eitan Grinspun}.} \bibinfo{year}{2025}\natexlab{}.
\newblock \showarticletitle{Lifting the Winding Number: Precise Discontinuities in Neural Fields for Physics Simulation}. In \bibinfo{booktitle}{\emph{Proceedings of the Special Interest Group on Computer Graphics and Interactive Techniques Conference Papers}} \emph{(\bibinfo{series}{SIGGRAPH Conference Papers '25})}. \bibinfo{publisher}{Association for Computing Machinery}, \bibinfo{address}{New York, NY, USA}, Article \bibinfo{articleno}{25}, \bibinfo{numpages}{11}~pages.
\newblock
\showISBNx{9798400715402}
\href{https://doi.org/10.1145/3721238.3730597}{doi:\nolinkurl{10.1145/3721238.3730597}}


\bibitem[Chen et~al\mbox{.}(2025)]%
        {Chen2025FRCSG}
\bibfield{author}{\bibinfo{person}{Jiaxi Chen}, \bibinfo{person}{Zeyu Shen}, \bibinfo{person}{Mingyang Zhao}, \bibinfo{person}{Xiaohong Jia}, \bibinfo{person}{Dong-Ming Yan}, {and} \bibinfo{person}{Wencheng Wang}.} \bibinfo{year}{2025}\natexlab{}.
\newblock \showarticletitle{{FR-CSG}: Fast and Reliable Modeling for Constructive Solid Geometry}.
\newblock \bibinfo{journal}{\emph{IEEE Transactions on Visualization and Computer Graphics}} \bibinfo{volume}{31}, \bibinfo{number}{9} (\bibinfo{year}{2025}), \bibinfo{pages}{5869--5883}.
\newblock
\href{https://doi.org/10.1109/TVCG.2024.3481278}{doi:\nolinkurl{10.1109/TVCG.2024.3481278}}


\bibitem[Costabel(1987)]%
        {Costabel1987}
\bibfield{author}{\bibinfo{person}{Martin Costabel}.} \bibinfo{year}{1987}\natexlab{}.
\newblock \showarticletitle{Principles of boundary element methods}.
\newblock \bibinfo{journal}{\emph{Computer Physics Reports}} \bibinfo{volume}{6}, \bibinfo{number}{1} (\bibinfo{year}{1987}), \bibinfo{pages}{243--274}.
\newblock
\showISSN{0167-7977}
\href{https://doi.org/10.1016/0167-7977(87)90014-1}{doi:\nolinkurl{10.1016/0167-7977(87)90014-1}}


\bibitem[Cybenko(1989)]%
        {cybenko1989approximation}
\bibfield{author}{\bibinfo{person}{George Cybenko}.} \bibinfo{year}{1989}\natexlab{}.
\newblock \showarticletitle{Approximation by superpositions of a sigmoidal function}.
\newblock \bibinfo{journal}{\emph{Mathematics of Control, Signals and Systems}} \bibinfo{volume}{2}, \bibinfo{number}{4} (\bibinfo{year}{1989}), \bibinfo{pages}{303--314}.
\newblock
\showISSN{1435-568X}
\href{https://doi.org/10.1007/BF02551274}{doi:\nolinkurl{10.1007/BF02551274}}


\bibitem[Dong et~al\mbox{.}(2024)]%
        {Dong2024NeurCADRecon}
\bibfield{author}{\bibinfo{person}{Qiujie Dong}, \bibinfo{person}{Rui Xu}, \bibinfo{person}{Pengfei Wang}, \bibinfo{person}{Shuangmin Chen}, \bibinfo{person}{Shiqing Xin}, \bibinfo{person}{Xiaohong Jia}, \bibinfo{person}{Wenping Wang}, {and} \bibinfo{person}{Changhe Tu}.} \bibinfo{year}{2024}\natexlab{}.
\newblock \showarticletitle{{NeurCADRecon}: Neural Representation for Reconstructing CAD Surfaces by Enforcing Zero Gaussian Curvature}.
\newblock \bibinfo{journal}{\emph{ACM Transactions on Graphics}} \bibinfo{volume}{43}, \bibinfo{number}{4}, Article \bibinfo{articleno}{51} (\bibinfo{date}{July} \bibinfo{year}{2024}), \bibinfo{numpages}{17}~pages.
\newblock
\showISSN{0730-0301}
\href{https://doi.org/10.1145/3658171}{doi:\nolinkurl{10.1145/3658171}}


\bibitem[Du et~al\mbox{.}(2018)]%
        {du2018inversecsg}
\bibfield{author}{\bibinfo{person}{Tao Du}, \bibinfo{person}{Jeevana~Priya Inala}, \bibinfo{person}{Yewen Pu}, \bibinfo{person}{Andrew Spielberg}, \bibinfo{person}{Adriana Schulz}, \bibinfo{person}{Daniela Rus}, \bibinfo{person}{Armando Solar-Lezama}, {and} \bibinfo{person}{Wojciech Matusik}.} \bibinfo{year}{2018}\natexlab{}.
\newblock \showarticletitle{{InverseCSG}: automatic conversion of {3D} models to {CSG} trees}.
\newblock \bibinfo{journal}{\emph{ACM Transactions on Graphics}} \bibinfo{volume}{37}, \bibinfo{number}{6}, Article \bibinfo{articleno}{213} (\bibinfo{date}{Dec.} \bibinfo{year}{2018}), \bibinfo{numpages}{16}~pages.
\newblock
\showISSN{0730-0301}
\href{https://doi.org/10.1145/3272127.3275006}{doi:\nolinkurl{10.1145/3272127.3275006}}


\bibitem[Dupont et~al\mbox{.}(2025)]%
        {dupont2025transcad}
\bibfield{author}{\bibinfo{person}{Elona Dupont}, \bibinfo{person}{Kseniya Cherenkova}, \bibinfo{person}{Dimitrios Mallis}, \bibinfo{person}{Gleb Gusev}, \bibinfo{person}{Anis Kacem}, {and} \bibinfo{person}{Djamila Aouada}.} \bibinfo{year}{2025}\natexlab{}.
\newblock \showarticletitle{{TransCAD}: A Hierarchical Transformer for {CAD} Sequence Inference from Point Clouds}. In \bibinfo{booktitle}{\emph{Computer Vision -- ECCV 2024}}. \bibinfo{publisher}{Springer Nature Switzerland}, \bibinfo{address}{Cham}, \bibinfo{pages}{19--36}.
\newblock
\href{https://doi.org/10.1007/978-3-031-73030-6_2}{doi:\nolinkurl{10.1007/978-3-031-73030-6_2}}


\bibitem[Erler et~al\mbox{.}(2020)]%
        {Erler2020}
\bibfield{author}{\bibinfo{person}{Philipp Erler}, \bibinfo{person}{Paul Guerrero}, \bibinfo{person}{Stefan Ohrhallinger}, \bibinfo{person}{Niloy~J. Mitra}, {and} \bibinfo{person}{Michael Wimmer}.} \bibinfo{year}{2020}\natexlab{}.
\newblock \showarticletitle{Points2Surf Learning Implicit Surfaces from Point Clouds}. In \bibinfo{booktitle}{\emph{Computer Vision -- ECCV 2020}}. \bibinfo{publisher}{Springer International Publishing}, \bibinfo{address}{Cham}, \bibinfo{pages}{108--124}.
\newblock
\showISBNx{978-3-030-58558-7}
\href{https://doi.org/10.1007/978-3-030-58558-7_7}{doi:\nolinkurl{10.1007/978-3-030-58558-7_7}}


\bibitem[Guo et~al\mbox{.}(2022)]%
        {Guo2022NH-Rep}
\bibfield{author}{\bibinfo{person}{Hao-Xiang Guo}, \bibinfo{person}{Yang Liu}, \bibinfo{person}{Hao Pan}, {and} \bibinfo{person}{Baining Guo}.} \bibinfo{year}{2022}\natexlab{}.
\newblock \showarticletitle{Implicit Conversion of Manifold {B}-Rep Solids by Neural Halfspace Representation}.
\newblock \bibinfo{journal}{\emph{ACM Transactions on Graphics}} \bibinfo{volume}{41}, \bibinfo{number}{6}, Article \bibinfo{articleno}{276} (\bibinfo{date}{Nov.} \bibinfo{year}{2022}), \bibinfo{numpages}{15}~pages.
\newblock
\showISSN{0730-0301}
\href{https://doi.org/10.1145/3550454.3555502}{doi:\nolinkurl{10.1145/3550454.3555502}}


\bibitem[Hornik(1991)]%
        {hornik1991approximation}
\bibfield{author}{\bibinfo{person}{Kurt Hornik}.} \bibinfo{year}{1991}\natexlab{}.
\newblock \showarticletitle{Approximation capabilities of multilayer feedforward networks}.
\newblock \bibinfo{journal}{\emph{Neural Networks}} \bibinfo{volume}{4}, \bibinfo{number}{2} (\bibinfo{year}{1991}), \bibinfo{pages}{251--257}.
\newblock
\showISSN{0893-6080}
\href{https://doi.org/10.1016/0893-6080(91)90009-T}{doi:\nolinkurl{10.1016/0893-6080(91)90009-T}}


\bibitem[Imaizumi and Fukumizu(2019)]%
        {Imaizumi2019}
\bibfield{author}{\bibinfo{person}{Masaaki Imaizumi} {and} \bibinfo{person}{Kenji Fukumizu}.} \bibinfo{year}{2019}\natexlab{}.
\newblock \showarticletitle{Deep Neural Networks Learn Non-Smooth Functions Effectively}. In \bibinfo{booktitle}{\emph{Proceedings of the Twenty-Second International Conference on Artificial Intelligence and Statistics}} \emph{(\bibinfo{series}{Proceedings of Machine Learning Research}, Vol.~\bibinfo{volume}{89})}. \bibinfo{publisher}{PMLR}, \bibinfo{pages}{869--878}.
\newblock


\bibitem[Ismailov(2023)]%
        {Ismailov2023}
\bibfield{author}{\bibinfo{person}{Vugar~E. Ismailov}.} \bibinfo{year}{2023}\natexlab{}.
\newblock \showarticletitle{A three layer neural network can represent any multivariate function}.
\newblock \bibinfo{journal}{\emph{J. Math. Anal. Appl.}} \bibinfo{volume}{523}, \bibinfo{number}{1} (\bibinfo{year}{2023}), \bibinfo{pages}{127096}.
\newblock
\showISSN{0022-247X}
\href{https://doi.org/10.1016/j.jmaa.2023.127096}{doi:\nolinkurl{10.1016/j.jmaa.2023.127096}}


\bibitem[Ju et~al\mbox{.}(2002)]%
        {Ju2022DC}
\bibfield{author}{\bibinfo{person}{Tao Ju}, \bibinfo{person}{Frank Losasso}, \bibinfo{person}{Scott Schaefer}, {and} \bibinfo{person}{Joe Warren}.} \bibinfo{year}{2002}\natexlab{}.
\newblock \showarticletitle{Dual contouring of hermite data}.
\newblock \bibinfo{journal}{\emph{ACM Transactions on Graphics}} \bibinfo{volume}{21}, \bibinfo{number}{3} (\bibinfo{date}{July} \bibinfo{year}{2002}), \bibinfo{pages}{339--346}.
\newblock
\showISSN{0730-0301}
\href{https://doi.org/10.1145/566654.566586}{doi:\nolinkurl{10.1145/566654.566586}}


\bibitem[Kingma and Ba(2015)]%
        {Kingma2015Adam}
\bibfield{author}{\bibinfo{person}{Diederik~P. Kingma} {and} \bibinfo{person}{Jimmy Ba}.} \bibinfo{year}{2015}\natexlab{}.
\newblock \showarticletitle{Adam: A Method for Stochastic Optimization}. In \bibinfo{booktitle}{\emph{3rd International Conference on Learning Representations, {ICLR} 2015, San Diego, CA, USA, May 7-9, 2015, Conference Track Proceedings}}.
\newblock


\bibitem[Koch et~al\mbox{.}(2019)]%
        {Koch2019ABC}
\bibfield{author}{\bibinfo{person}{Sebastian Koch}, \bibinfo{person}{Albert Matveev}, \bibinfo{person}{Zhongshi Jiang}, \bibinfo{person}{Francis Williams}, \bibinfo{person}{Alexey Artemov}, \bibinfo{person}{Evgeny Burnaev}, \bibinfo{person}{Marc Alexa}, \bibinfo{person}{Denis Zorin}, {and} \bibinfo{person}{Daniele Panozzo}.} \bibinfo{year}{2019}\natexlab{}.
\newblock \showarticletitle{ABC: A Big CAD Model Dataset For Geometric Deep Learning}. In \bibinfo{booktitle}{\emph{2019 IEEE/CVF Conference on Computer Vision and Pattern Recognition (CVPR)}}. \bibinfo{pages}{9593--9603}.
\newblock
\href{https://doi.org/10.1109/CVPR.2019.00983}{doi:\nolinkurl{10.1109/CVPR.2019.00983}}


\bibitem[Li et~al\mbox{.}(2023)]%
        {Li2023}
\bibfield{author}{\bibinfo{person}{Qing Li}, \bibinfo{person}{Huifang Feng}, \bibinfo{person}{Kanle Shi}, \bibinfo{person}{Yi Fang}, \bibinfo{person}{Yu-Shen Liu}, {and} \bibinfo{person}{Zhizhong Han}.} \bibinfo{year}{2023}\natexlab{}.
\newblock \showarticletitle{Neural Gradient Learning and Optimization for Oriented Point Normal Estimation}. In \bibinfo{booktitle}{\emph{SIGGRAPH Asia 2023 Conference Papers}} (Sydney, NSW, Australia) \emph{(\bibinfo{series}{SA '23})}. \bibinfo{publisher}{Association for Computing Machinery}, \bibinfo{address}{New York, NY, USA}, Article \bibinfo{articleno}{122}, \bibinfo{numpages}{9}~pages.
\newblock
\showISBNx{9798400703157}
\href{https://doi.org/10.1145/3610548.3618253}{doi:\nolinkurl{10.1145/3610548.3618253}}


\bibitem[Li et~al\mbox{.}(2025)]%
        {li2025deep}
\bibfield{author}{\bibinfo{person}{Yuanqi Li}, \bibinfo{person}{Hongshen Wang}, \bibinfo{person}{Yansong Liu}, \bibinfo{person}{Jingcheng Huang}, \bibinfo{person}{Shun Liu}, \bibinfo{person}{Chenyu Huang}, \bibinfo{person}{Jianwei Guo}, \bibinfo{person}{Jie Guo}, {and} \bibinfo{person}{Yanwen Guo}.} \bibinfo{year}{2025}\natexlab{}.
\newblock \showarticletitle{Deep Point Cloud Edge Reconstruction via Surface Patch Segmentation}.
\newblock \bibinfo{journal}{\emph{IEEE Transactions on Visualization and Computer Graphics}} \bibinfo{volume}{31}, \bibinfo{number}{10} (\bibinfo{year}{2025}), \bibinfo{pages}{7463--7477}.
\newblock
\href{https://doi.org/10.1109/TVCG.2025.3547411}{doi:\nolinkurl{10.1109/TVCG.2025.3547411}}


\bibitem[Lin et~al\mbox{.}(2025)]%
        {Lin2025PatchGrid}
\bibfield{author}{\bibinfo{person}{Guying Lin}, \bibinfo{person}{Lei Yang}, \bibinfo{person}{Congyi Zhang}, \bibinfo{person}{Hao Pan}, \bibinfo{person}{Yuhan Ping}, \bibinfo{person}{Guodong Wei}, \bibinfo{person}{Taku Komura}, \bibinfo{person}{John Keyser}, {and} \bibinfo{person}{Wenping Wang}.} \bibinfo{year}{2025}\natexlab{}.
\newblock \showarticletitle{{Patch-Grid}: An Efficient and Feature-Preserving Neural Implicit Surface Representation}.
\newblock \bibinfo{journal}{\emph{ACM Transactions on Graphics}} \bibinfo{volume}{44}, \bibinfo{number}{2}, Article \bibinfo{articleno}{16} (\bibinfo{date}{April} \bibinfo{year}{2025}), \bibinfo{numpages}{21}~pages.
\newblock
\showISSN{0730-0301}
\href{https://doi.org/10.1145/3727142}{doi:\nolinkurl{10.1145/3727142}}


\bibitem[Liu et~al\mbox{.}(2025b)]%
        {Liu2025}
\bibfield{author}{\bibinfo{person}{Chenxi Liu}, \bibinfo{person}{Siqi Wang}, \bibinfo{person}{Matthew Fisher}, \bibinfo{person}{Deepali Aneja}, {and} \bibinfo{person}{Alec Jacobson}.} \bibinfo{year}{2025}\natexlab{b}.
\newblock \showarticletitle{{2D} Neural Fields with Learned Discontinuities}.
\newblock \bibinfo{journal}{\emph{Computer Graphics Forum}} (\bibinfo{year}{2025}), \bibinfo{pages}{e70023}.
\newblock
\href{https://doi.org/10.1111/cgf.70023}{doi:\nolinkurl{10.1111/cgf.70023}}


\bibitem[Liu et~al\mbox{.}(2025a)]%
        {Liu2025PGD}
\bibfield{author}{\bibinfo{person}{Mengfei Liu}, \bibinfo{person}{Yue Chang}, \bibinfo{person}{Zhecheng Wang}, \bibinfo{person}{Peter~Yichen Chen}, {and} \bibinfo{person}{Eitan Grinspun}.} \bibinfo{year}{2025}\natexlab{a}.
\newblock \showarticletitle{Precise Gradient Discontinuities in Neural Fields for Subspace Physics}. In \bibinfo{booktitle}{\emph{Proceedings of the SIGGRAPH Asia 2025 Conference Papers}} \emph{(\bibinfo{series}{SA Conference Papers '25})}. \bibinfo{publisher}{Association for Computing Machinery}, \bibinfo{address}{New York, NY, USA}, Article \bibinfo{articleno}{26}, \bibinfo{numpages}{11}~pages.
\newblock
\showISBNx{9798400721373}
\href{https://doi.org/10.1145/3757377.3763810}{doi:\nolinkurl{10.1145/3757377.3763810}}


\bibitem[Liu et~al\mbox{.}(2024a)]%
        {Liu2024SplitandFit}
\bibfield{author}{\bibinfo{person}{Yilin Liu}, \bibinfo{person}{Jiale Chen}, \bibinfo{person}{Shanshan Pan}, \bibinfo{person}{Daniel Cohen-Or}, \bibinfo{person}{Hao Zhang}, {and} \bibinfo{person}{Hui Huang}.} \bibinfo{year}{2024}\natexlab{a}.
\newblock \showarticletitle{Split-and-Fit: Learning B-Reps via Structure-Aware Voronoi Partitioning}.
\newblock \bibinfo{journal}{\emph{ACM Transactions on Graphics}} \bibinfo{volume}{43}, \bibinfo{number}{4}, Article \bibinfo{articleno}{108} (\bibinfo{date}{July} \bibinfo{year}{2024}), \bibinfo{numpages}{13}~pages.
\newblock
\showISSN{0730-0301}
\href{https://doi.org/10.1145/3658155}{doi:\nolinkurl{10.1145/3658155}}


\bibitem[Liu et~al\mbox{.}(2024b)]%
        {liu2024point2cad}
\bibfield{author}{\bibinfo{person}{Yujia Liu}, \bibinfo{person}{Anton Obukhov}, \bibinfo{person}{Jan~Dirk Wegner}, {and} \bibinfo{person}{Konrad Schindler}.} \bibinfo{year}{2024}\natexlab{b}.
\newblock \showarticletitle{{Point2CAD}: Reverse Engineering {CAD} Models from {3D} Point Clouds}. In \bibinfo{booktitle}{\emph{2024 IEEE/CVF Conference on Computer Vision and Pattern Recognition (CVPR)}}. \bibinfo{pages}{3763--3772}.
\newblock
\href{https://doi.org/10.1109/CVPR52733.2024.00361}{doi:\nolinkurl{10.1109/CVPR52733.2024.00361}}


\bibitem[Liu et~al\mbox{.}(2025c)]%
        {Liu2025KAN}
\bibfield{author}{\bibinfo{person}{Ziming Liu}, \bibinfo{person}{Yixuan Wang}, \bibinfo{person}{Sachin Vaidya}, \bibinfo{person}{Fabian Ruehle}, \bibinfo{person}{James Halverson}, \bibinfo{person}{Marin Soljacic}, \bibinfo{person}{Thomas~Y. Hou}, {and} \bibinfo{person}{Max Tegmark}.} \bibinfo{year}{2025}\natexlab{c}.
\newblock \showarticletitle{{KAN}: Kolmogorov{\textendash}Arnold Networks}. In \bibinfo{booktitle}{\emph{The Thirteenth International Conference on Learning Representations}}.
\newblock


\bibitem[Luo et~al\mbox{.}(2026)]%
        {Luo2026TopoMesh}
\bibfield{author}{\bibinfo{person}{Guan Luo}, \bibinfo{person}{Xiu Li}, \bibinfo{person}{Rui Chen}, \bibinfo{person}{Xuanyu Yi}, \bibinfo{person}{Jing Lin}, \bibinfo{person}{Chia-Hao Chen}, \bibinfo{person}{Jiahang Liu}, \bibinfo{person}{Song-Hai Zhang}, {and} \bibinfo{person}{Jianfeng Zhang}.} \bibinfo{year}{2026}\natexlab{}.
\newblock \bibinfo{title}{{TopoMesh}: High-Fidelity Mesh Autoencoding via Topological Unification}.
\newblock
\showeprint[arxiv]{2603.24278}~[cs.CV]
\urldef\tempurl%
\url{https://arxiv.org/abs/2603.24278}
\showURL{%
\tempurl}


\bibitem[Ma et~al\mbox{.}(2021)]%
        {NeuralPull}
\bibfield{author}{\bibinfo{person}{Baorui Ma}, \bibinfo{person}{Zhizhong Han}, \bibinfo{person}{Yu-Shen Liu}, {and} \bibinfo{person}{Matthias Zwicker}.} \bibinfo{year}{2021}\natexlab{}.
\newblock \showarticletitle{Neural-Pull: Learning Signed Distance Function from Point Clouds by Learning to Pull Space onto Surface}. In \bibinfo{booktitle}{\emph{Proceedings of the 38th International Conference on Machine Learning}} \emph{(\bibinfo{series}{Proceedings of Machine Learning Research}, Vol.~\bibinfo{volume}{139})}. \bibinfo{publisher}{PMLR}, \bibinfo{pages}{7246--7257}.
\newblock


\bibitem[Mescheder et~al\mbox{.}(2019)]%
        {Mescheder2019}
\bibfield{author}{\bibinfo{person}{Lars Mescheder}, \bibinfo{person}{Michael Oechsle}, \bibinfo{person}{Michael Niemeyer}, \bibinfo{person}{Sebastian Nowozin}, {and} \bibinfo{person}{Andreas Geiger}.} \bibinfo{year}{2019}\natexlab{}.
\newblock \showarticletitle{Occupancy Networks: Learning {3D} Reconstruction in Function Space}. In \bibinfo{booktitle}{\emph{2019 IEEE/CVF Conference on Computer Vision and Pattern Recognition (CVPR)}}. \bibinfo{address}{Los Alamitos, CA, USA}, \bibinfo{pages}{4455--4465}.
\newblock
\href{https://doi.org/10.1109/CVPR.2019.00459}{doi:\nolinkurl{10.1109/CVPR.2019.00459}}


\bibitem[Mildenhall et~al\mbox{.}(2020)]%
        {NeRF}
\bibfield{author}{\bibinfo{person}{Ben Mildenhall}, \bibinfo{person}{Pratul~P. Srinivasan}, \bibinfo{person}{Matthew Tancik}, \bibinfo{person}{Jonathan~T. Barron}, \bibinfo{person}{Ravi Ramamoorthi}, {and} \bibinfo{person}{Ren Ng}.} \bibinfo{year}{2020}\natexlab{}.
\newblock \showarticletitle{{NeRF}: Representing Scenes as Neural Radiance Fields for View Synthesis}. In \bibinfo{booktitle}{\emph{Computer Vision -- ECCV 2020}}. \bibinfo{publisher}{Springer International Publishing}, \bibinfo{address}{Cham}, \bibinfo{pages}{405--421}.
\newblock
\showISBNx{978-3-030-58452-8}
\href{https://doi.org/10.1007/978-3-030-58452-8_24}{doi:\nolinkurl{10.1007/978-3-030-58452-8_24}}


\bibitem[M\"{u}ller et~al\mbox{.}(2022)]%
        {Muller2022INGP}
\bibfield{author}{\bibinfo{person}{Thomas M\"{u}ller}, \bibinfo{person}{Alex Evans}, \bibinfo{person}{Christoph Schied}, {and} \bibinfo{person}{Alexander Keller}.} \bibinfo{year}{2022}\natexlab{}.
\newblock \showarticletitle{Instant neural graphics primitives with a multiresolution hash encoding}.
\newblock \bibinfo{journal}{\emph{ACM Transactions on Graphics}} \bibinfo{volume}{41}, \bibinfo{number}{4}, Article \bibinfo{articleno}{102} (\bibinfo{date}{July} \bibinfo{year}{2022}), \bibinfo{numpages}{15}~pages.
\newblock
\showISSN{0730-0301}
\href{https://doi.org/10.1145/3528223.3530127}{doi:\nolinkurl{10.1145/3528223.3530127}}


\bibitem[Pang et~al\mbox{.}(2023)]%
        {Pang2023}
\bibfield{author}{\bibinfo{person}{Bo Pang}, \bibinfo{person}{Zhongtian Zheng}, \bibinfo{person}{Guoping Wang}, {and} \bibinfo{person}{Peng-Shuai Wang}.} \bibinfo{year}{2023}\natexlab{}.
\newblock \showarticletitle{Learning the Geodesic Embedding with Graph Neural Networks}.
\newblock \bibinfo{journal}{\emph{ACM Transactions on Graphics}} \bibinfo{volume}{42}, \bibinfo{number}{6}, Article \bibinfo{articleno}{236} (\bibinfo{date}{Dec.} \bibinfo{year}{2023}), \bibinfo{numpages}{12}~pages.
\newblock
\showISSN{0730-0301}
\href{https://doi.org/10.1145/3618317}{doi:\nolinkurl{10.1145/3618317}}


\bibitem[Park et~al\mbox{.}(2019)]%
        {DeepSDF}
\bibfield{author}{\bibinfo{person}{Jeong~Joon Park}, \bibinfo{person}{Peter Florence}, \bibinfo{person}{Julian Straub}, \bibinfo{person}{Richard Newcombe}, {and} \bibinfo{person}{Steven Lovegrove}.} \bibinfo{year}{2019}\natexlab{}.
\newblock \showarticletitle{{DeepSDF}: Learning Continuous Signed Distance Functions for Shape Representation}. In \bibinfo{booktitle}{\emph{2019 IEEE/CVF Conference on Computer Vision and Pattern Recognition (CVPR)}}. \bibinfo{pages}{165--174}.
\newblock
\href{https://doi.org/10.1109/CVPR.2019.00025}{doi:\nolinkurl{10.1109/CVPR.2019.00025}}


\bibitem[Rockafellar and Wets(1998)]%
        {rockafellar1998variational}
\bibfield{author}{\bibinfo{person}{R.~Tyrrell Rockafellar} {and} \bibinfo{person}{Roger J-B. Wets}.} \bibinfo{year}{1998}\natexlab{}.
\newblock \bibinfo{booktitle}{\emph{Variational Analysis}}. \bibinfo{series}{Grundlehren der mathematischen Wissenschaften}, Vol.~\bibinfo{volume}{317}.
\newblock \bibinfo{publisher}{Springer Berlin, Heidelberg}.
\newblock
\href{https://doi.org/10.1007/978-3-642-02431-3}{doi:\nolinkurl{10.1007/978-3-642-02431-3}}


\bibitem[Sharma et~al\mbox{.}(2022)]%
        {Sharma2022Neural}
\bibfield{author}{\bibinfo{person}{Gopal Sharma}, \bibinfo{person}{Rishabh Goyal}, \bibinfo{person}{Difan Liu}, \bibinfo{person}{Evangelos Kalogerakis}, {and} \bibinfo{person}{Subhransu Maji}.} \bibinfo{year}{2022}\natexlab{}.
\newblock \showarticletitle{Neural Shape Parsers for Constructive Solid Geometry}.
\newblock \bibinfo{journal}{\emph{IEEE Transactions on Pattern Analysis and Machine Intelligence}} \bibinfo{volume}{44}, \bibinfo{number}{5} (\bibinfo{year}{2022}), \bibinfo{pages}{2628--2640}.
\newblock
\href{https://doi.org/10.1109/TPAMI.2020.3044749}{doi:\nolinkurl{10.1109/TPAMI.2020.3044749}}


\bibitem[Shen et~al\mbox{.}(2025)]%
        {shen2025mesh2brep}
\bibfield{author}{\bibinfo{person}{Zeyu Shen}, \bibinfo{person}{Mingyang Zhao}, \bibinfo{person}{Dong-Ming Yan}, {and} \bibinfo{person}{Wencheng Wang}.} \bibinfo{year}{2025}\natexlab{}.
\newblock \showarticletitle{{Mesh2Brep}: B-Rep Reconstruction via Robust Primitive Fitting and Intersection-Aware Constraints}.
\newblock \bibinfo{journal}{\emph{IEEE Transactions on Visualization and Computer Graphics}} \bibinfo{volume}{31}, \bibinfo{number}{10} (\bibinfo{year}{2025}), \bibinfo{pages}{6661--6676}.
\newblock
\href{https://doi.org/10.1109/TVCG.2025.3525844}{doi:\nolinkurl{10.1109/TVCG.2025.3525844}}


\bibitem[Sitzmann et~al\mbox{.}(2020)]%
        {sitzmann2020siren}
\bibfield{author}{\bibinfo{person}{Vincent Sitzmann}, \bibinfo{person}{Julien~N.P. Martel}, \bibinfo{person}{Alexander~W. Bergman}, \bibinfo{person}{David~B. Lindell}, {and} \bibinfo{person}{Gordon Wetzstein}.} \bibinfo{year}{2020}\natexlab{}.
\newblock \showarticletitle{Implicit Neural Representations with Periodic Activation Functions}. In \bibinfo{booktitle}{\emph{Advances in Neural Information Processing Systems}}, Vol.~\bibinfo{volume}{33}. \bibinfo{publisher}{Curran Associates, Inc.}, \bibinfo{pages}{7462--7473}.
\newblock


\bibitem[Sorkine(2006)]%
        {Sorkine2006}
\bibfield{author}{\bibinfo{person}{Olga Sorkine}.} \bibinfo{year}{2006}\natexlab{}.
\newblock \showarticletitle{Differential Representations for Mesh Processing}.
\newblock \bibinfo{journal}{\emph{Computer Graphics Forum}} \bibinfo{volume}{25}, \bibinfo{number}{4} (\bibinfo{year}{2006}), \bibinfo{pages}{789--807}.
\newblock
\href{https://doi.org/10.1111/j.1467-8659.2006.00999.x}{doi:\nolinkurl{10.1111/j.1467-8659.2006.00999.x}}


\bibitem[Takikawa et~al\mbox{.}(2021)]%
        {Takikawa2021NGLOD}
\bibfield{author}{\bibinfo{person}{Towaki Takikawa}, \bibinfo{person}{Joey Litalien}, \bibinfo{person}{Kangxue Yin}, \bibinfo{person}{Karsten Kreis}, \bibinfo{person}{Charles Loop}, \bibinfo{person}{Derek Nowrouzezahrai}, \bibinfo{person}{Alec Jacobson}, \bibinfo{person}{Morgan McGuire}, {and} \bibinfo{person}{Sanja Fidler}.} \bibinfo{year}{2021}\natexlab{}.
\newblock \showarticletitle{Neural Geometric Level of Detail: Real-time Rendering with Implicit 3D Shapes}. In \bibinfo{booktitle}{\emph{2021 IEEE/CVF Conference on Computer Vision and Pattern Recognition (CVPR)}}. \bibinfo{pages}{11353--11362}.
\newblock
\href{https://doi.org/10.1109/CVPR46437.2021.01120}{doi:\nolinkurl{10.1109/CVPR46437.2021.01120}}


\bibitem[Usama et~al\mbox{.}(2026)]%
        {usama2026nurbgen}
\bibfield{author}{\bibinfo{person}{Muhammad Usama}, \bibinfo{person}{Mohammad~Sadil Khan}, \bibinfo{person}{Didier Stricker}, {and} \bibinfo{person}{Muhammad~Zeshan Afzal}.} \bibinfo{year}{2026}\natexlab{}.
\newblock \showarticletitle{{NURBGen}: High-Fidelity Text-to-{CAD} Generation through {LLM}-Driven {NURBS} Modeling}. In \bibinfo{booktitle}{\emph{Proceedings of the AAAI Conference on Artificial Intelligence}}, Vol.~\bibinfo{volume}{40}. \bibinfo{pages}{9603--9611}.
\newblock
\href{https://doi.org/10.1609/aaai.v40i12.37922}{doi:\nolinkurl{10.1609/aaai.v40i12.37922}}


\bibitem[Wang et~al\mbox{.}(2025)]%
        {Wang2025NeuVAS}
\bibfield{author}{\bibinfo{person}{Pengfei Wang}, \bibinfo{person}{Qiujie Dong}, \bibinfo{person}{Fangtian Liang}, \bibinfo{person}{Hao Pan}, \bibinfo{person}{Lei Yang}, \bibinfo{person}{Congyi Zhang}, \bibinfo{person}{Guying Lin}, \bibinfo{person}{Caiming Zhang}, \bibinfo{person}{Yuanfeng Zhou}, \bibinfo{person}{Changhe Tu}, \bibinfo{person}{Shiqing Xin}, \bibinfo{person}{Alla Sheffer}, \bibinfo{person}{Xin Li}, {and} \bibinfo{person}{Wenping Wang}.} \bibinfo{year}{2025}\natexlab{}.
\newblock \showarticletitle{NeuVAS: Neural Implicit Surfaces for Variational Shape Modeling}.
\newblock \bibinfo{journal}{\emph{ACM Transactions on Graphics}} \bibinfo{volume}{44}, \bibinfo{number}{6}, Article \bibinfo{articleno}{186} (\bibinfo{date}{Dec.} \bibinfo{year}{2025}), \bibinfo{numpages}{14}~pages.
\newblock
\showISSN{0730-0301}
\href{https://doi.org/10.1145/3763331}{doi:\nolinkurl{10.1145/3763331}}


\bibitem[Wang et~al\mbox{.}(2021)]%
        {NeuS}
\bibfield{author}{\bibinfo{person}{Peng Wang}, \bibinfo{person}{Lingjie Liu}, \bibinfo{person}{Yuan Liu}, \bibinfo{person}{Christian Theobalt}, \bibinfo{person}{Taku Komura}, {and} \bibinfo{person}{Wenping Wang}.} \bibinfo{year}{2021}\natexlab{}.
\newblock \showarticletitle{NeuS: Learning Neural Implicit Surfaces by Volume Rendering for Multi-view Reconstruction}. In \bibinfo{booktitle}{\emph{Advances in Neural Information Processing Systems}}, Vol.~\bibinfo{volume}{34}. \bibinfo{publisher}{Curran Associates, Inc.}, \bibinfo{pages}{27171--27183}.
\newblock


\bibitem[Wang et~al\mbox{.}(2022)]%
        {Wang2022}
\bibfield{author}{\bibinfo{person}{Yifan Wang}, \bibinfo{person}{Lukas Rahmann}, {and} \bibinfo{person}{Olga Sorkine{-}Hornung}.} \bibinfo{year}{2022}\natexlab{}.
\newblock \showarticletitle{Geometry-Consistent Neural Shape Representation with Implicit Displacement Fields}. In \bibinfo{booktitle}{\emph{The Tenth International Conference on Learning Representations, {ICLR} 2022, Virtual Event, April 25-29, 2022}}. \bibinfo{publisher}{OpenReview.net}.
\newblock


\bibitem[Wu et~al\mbox{.}(2018)]%
        {Wu2018Constructing}
\bibfield{author}{\bibinfo{person}{Qiaoyun Wu}, \bibinfo{person}{Kai Xu}, {and} \bibinfo{person}{Jun Wang}.} \bibinfo{year}{2018}\natexlab{}.
\newblock \showarticletitle{Constructing {3D} {CSG} Models from {3D} Raw Point Clouds}.
\newblock \bibinfo{journal}{\emph{Computer Graphics Forum}} \bibinfo{volume}{37}, \bibinfo{number}{5} (\bibinfo{year}{2018}), \bibinfo{pages}{221--232}.
\newblock
\href{https://doi.org/10.1111/cgf.13504}{doi:\nolinkurl{10.1111/cgf.13504}}


\bibitem[Xu et~al\mbox{.}(2022)]%
        {Xu2022REEPS}
\bibfield{author}{\bibinfo{person}{Rui Xu}, \bibinfo{person}{Zixiong Wang}, \bibinfo{person}{Zhiyang Dou}, \bibinfo{person}{Chen Zong}, \bibinfo{person}{Shiqing Xin}, \bibinfo{person}{Mingyan Jiang}, \bibinfo{person}{Tao Ju}, {and} \bibinfo{person}{Changhe Tu}.} \bibinfo{year}{2022}\natexlab{}.
\newblock \showarticletitle{RFEPS: Reconstructing Feature-Line Equipped Polygonal Surface}.
\newblock \bibinfo{journal}{\emph{ACM Transactions on Graphics}} \bibinfo{volume}{41}, \bibinfo{number}{6}, Article \bibinfo{articleno}{228} (\bibinfo{date}{Nov.} \bibinfo{year}{2022}), \bibinfo{numpages}{15}~pages.
\newblock
\showISSN{0730-0301}
\href{https://doi.org/10.1145/3550454.3555443}{doi:\nolinkurl{10.1145/3550454.3555443}}


\bibitem[Yu et~al\mbox{.}(2023)]%
        {Yu2023D2CSG}
\bibfield{author}{\bibinfo{person}{Fenggen Yu}, \bibinfo{person}{Qimin Chen}, \bibinfo{person}{Maham Tanveer}, \bibinfo{person}{Ali Mahdavi~Amiri}, {and} \bibinfo{person}{Hao Zhang}.} \bibinfo{year}{2023}\natexlab{}.
\newblock \showarticletitle{{D\textsuperscript{2}CSG}: Unsupervised Learning of Compact {CSG} Trees with Dual Complements and Dropouts}. In \bibinfo{booktitle}{\emph{Advances in Neural Information Processing Systems}}, Vol.~\bibinfo{volume}{36}. \bibinfo{publisher}{Curran Associates, Inc.}, \bibinfo{pages}{22807--22819}.
\newblock
\href{https://doi.org/10.52202/075280-0989}{doi:\nolinkurl{10.52202/075280-0989}}


\bibitem[Yu et~al\mbox{.}(2022)]%
        {Yu2022CAPRI-Net}
\bibfield{author}{\bibinfo{person}{Fenggen Yu}, \bibinfo{person}{Zhiqin Chen}, \bibinfo{person}{Manyi Li}, \bibinfo{person}{Aditya Sanghi}, \bibinfo{person}{Hooman Shayani}, \bibinfo{person}{Ali Mahdavi-Amiri}, {and} \bibinfo{person}{Hao Zhang}.} \bibinfo{year}{2022}\natexlab{}.
\newblock \showarticletitle{CAPRI-Net: Learning Compact CAD Shapes with Adaptive Primitive Assembly}. In \bibinfo{booktitle}{\emph{2022 IEEE/CVF Conference on Computer Vision and Pattern Recognition (CVPR)}}. \bibinfo{pages}{11758--11768}.
\newblock
\href{https://doi.org/10.1109/CVPR52688.2022.01147}{doi:\nolinkurl{10.1109/CVPR52688.2022.01147}}


\bibitem[Zhu et~al\mbox{.}(2023)]%
        {Zhu2023NerVE}
\bibfield{author}{\bibinfo{person}{Xiangyu Zhu}, \bibinfo{person}{Dong Du}, \bibinfo{person}{Weikai Chen}, \bibinfo{person}{Zhiyou Zhao}, \bibinfo{person}{Yinyu Nie}, {and} \bibinfo{person}{Xiaoguang Han}.} \bibinfo{year}{2023}\natexlab{}.
\newblock \showarticletitle{{NerVE}: Neural Volumetric Edges for Parametric Curve Extraction from Point Cloud}. In \bibinfo{booktitle}{\emph{2023 IEEE/CVF Conference on Computer Vision and Pattern Recognition (CVPR)}}. \bibinfo{pages}{13601--13610}.
\newblock
\href{https://doi.org/10.1109/CVPR52729.2023.01307}{doi:\nolinkurl{10.1109/CVPR52729.2023.01307}}


\end{thebibliography}

\appendix

\section{Feature Surface Construction from Meshes}
\label{sec:feature_surface_mesh}

\begin{figure}[!htbp]
\centering
\includegraphics[width=\linewidth]{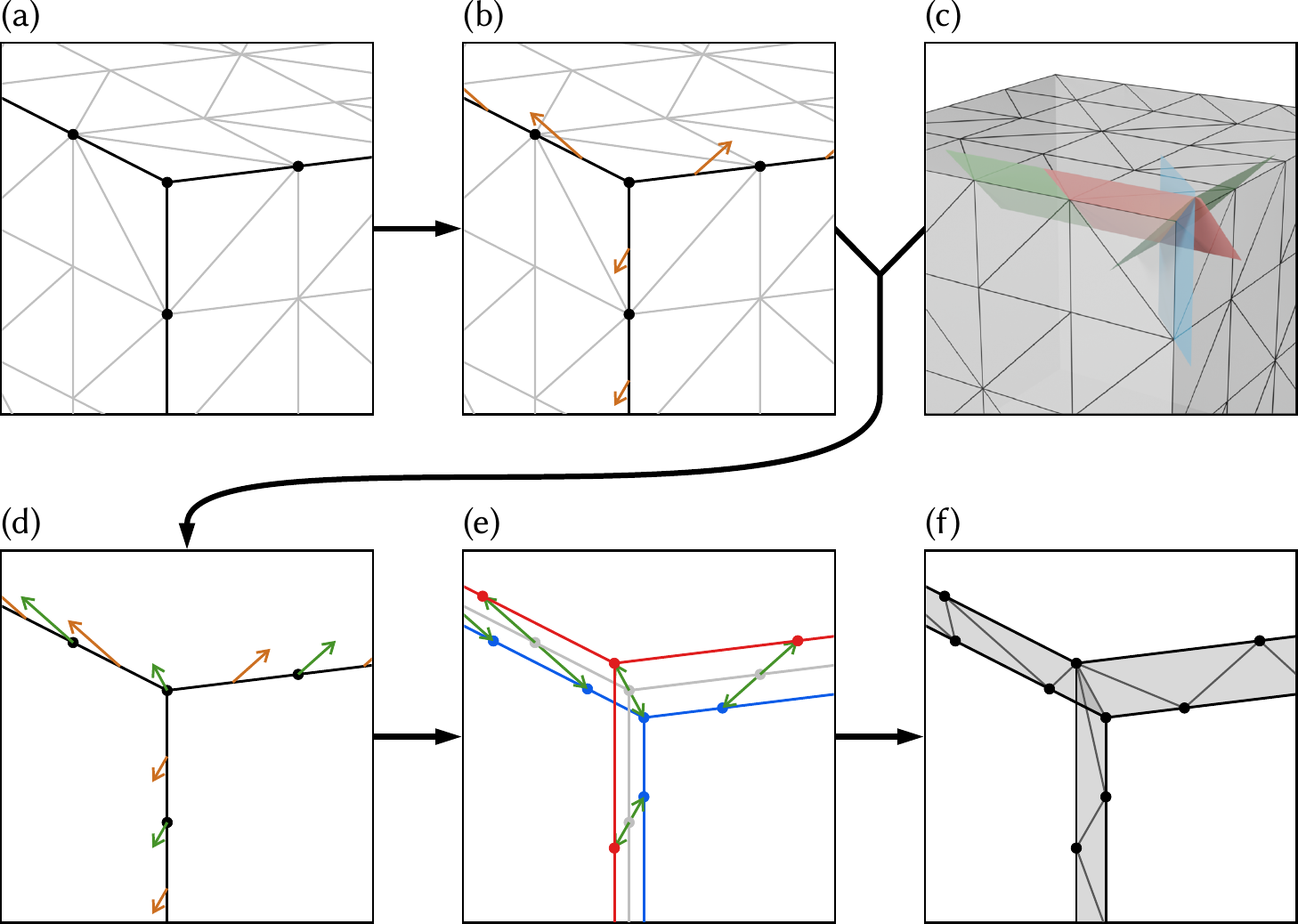}
\caption{For 3D tasks, the \(C^0\) feature surface \(M\) of {\sharpnet} can be constructed either from a mesh or directly from a point cloud.
(a) Sharp edges identified from the input mesh. 
(b) Guiding directions computed for each sharp edge. 
(c) Local bisector planes associated with the sharp edges. 
(d) Guiding directions propagated from edges to all vertices, chosen to lie as closely as possible within the local bisector planes of the incident edges.
(e) Each feature vertex is offset bidirectionally along its guiding direction to generate the feature surface \(\partial M\). 
(f) The strip-like triangular mesh \(M\).}
\label{fig:gen_c0_feature}
\end{figure}

Given a CAD triangular mesh as input, sharp edges are identified by evaluating the dihedral angle between each pair of triangles that share an edge. As illustrated in Figure~\ref{fig:gen_c0_feature}(a), we obtain the sharp curve set, where the dihedral angles are smaller than a user specified threshold. The sharp curve set is represented as an undirected graph \(\mathcal{S}\) composed of all sharp edges together with their incident vertices.
Next, we construct the feature surface \(M\) by expanding \(\mathcal{S}\) along certain guiding directions to form a strip‑like surface \(M\) that approximates the medial surface near the sharp edges. As illustrated in Figure~\ref{fig:gen_c0_feature}(b), for each sharp edge \(e_i \in \mathcal{S}\), we compute the angular bisector plane of its two adjacent triangle faces, which serves as an approximation of the local medial surface. For every edge \(e_i \in \mathcal{S}\), we then compute a guiding direction \(g_{e_i}\) of a unit vector, defined as a vector orthogonal to \(e_i\) and lying within its bisector plane, pointing to the model exterior. 

As illustrated in Figure~\ref{fig:gen_c0_feature}(c,d), these edge‑based guiding directions are subsequently propagated to all vertices \(v_j\) of \(\mathcal{S}\) to obtain vertex guiding directions \(g_{v_j}\). For a vertex \(v\), let \(\mathbf{E}_v\) denote the set of all edges incident to \(v\). When \(\norm{\mathbf{E}_v} = 1\), i.e., when the degree satisfies \(\deg(v)=1\), the vertex \(v\) is an end vertex, and we directly assign its guiding direction \(g_v\) to be identical to the guiding direction \(g_e\) of its unique incident edge \(e\). When \(\norm{\mathbf{E}_v} > 1\), we require \(g_v\) to align as closely as possible with the bisector planes of all edges in \(\mathbf{E}_v\), meaning that the dot product between \(g_v\) and the normal of each bisector plane should be as close to zero as possible. Although the mean of the edge guiding directions \(\Sigma_{e \in \mathbf{E}_v} g_e/\norm{\mathbf{E}_v}\) provides a reasonable approximation, it is not exact. Ideally, as shown in Figure~\ref{fig:gen_c0_feature}(c), the bisector planes of all incident edges should intersect along a single line, and we take the direction of this line as \(g_v\).

Taking these considerations into account, we determine the guiding direction \(g_v\) for each vertex \(v\) by minimizing the following energy:
\begin{gather*}
\mathbf{E}_v \;=\; \condset{e_i}{e_i \in \mathbb{S},e_i \ni v}\nonumber \\
g_v \;=\; \argmin_{g \in \mathbb{S}^2} \; \sum_{e_i \in \mathbf{E}_v} \abs*{%
    g \cdot \left(g_{e_i} \times \frac{\vec{e_i}}{\norm{\vec{e_i}}}\right)%
} + \lambda \cdot \norm*{%
    g - \frac{\sum_{e_i \in \mathbb{E}_v}g_{e_i}}{\norm{\mathbf{E}_v}}%
},
\label{eqn:c0_gen}
\end{gather*}
where \(\mathbb{S}^2\) denotes the unit sphere. The first term encourages \(g_v\) to lie within the bisector planes of the incident edges, while the second prevents \(g_v\) from deviating too far from the mean of the edge directions \(\set{g_{e_i}}\). This improves robustness and also handles degree-one vertices correctly. 

Given a width parameter \(w\), we then offset each vertex \(v\in\mathcal{S}\) by \(\pm w\) along its guiding direction $g_v$. This bidirectional displacement converts the polyline feature set \(\mathcal{S}\) into a strip-like surface \(M\), as shown in Figure~\ref{fig:gen_c0_feature}(e,f). By construction, \(M\) remains close to the original mesh and provides a localized representation of the region where the distance field exhibits a directional-derivative jump.

\section{Offset Field}
\label{sec:offset}

\begin{figure}[!htbp]
\centering
\includegraphics[width=.7\linewidth]{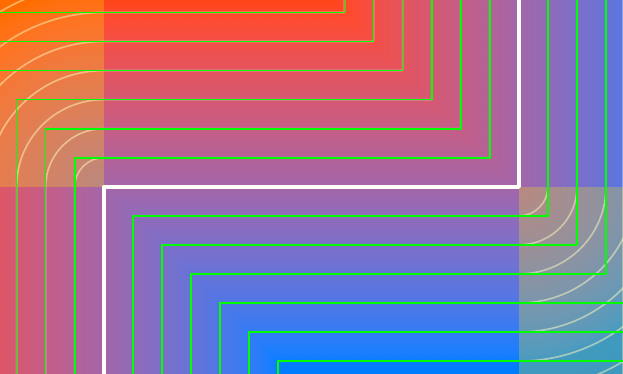}
\caption{A 2D illustration contrasting the signed offset distance with the standard signed distance. Inside the normal cone of a sharp vertex (shown in yellow), the offset isolines continue smoothly from those outside the cone and meet along the angular bisector, thereby maintaining the non-differentiable structure at the sharp feature. By comparison, the signed distance isolines yield rounded contours within the normal cone, which fails to reflect the underlying sharp geometry.}
\label{fig:offset}
\end{figure}
\citet{Luo2026TopoMesh} present a $L_\infty$ distance for sharp feature preservation. Likewise, we present a stricter definition on the continuous surface.
For any point \(p\), let the set of its closest points on the surface \(S\) be

\begin{equation*}
\mathrm{proj}_S(p) = \argmin_{x \in S}\,\norm{p - x}.
\end{equation*}

We define pseudo signed offset distance as

\begin{equation*}
\Psi^*_S(p) =%
\lim_{\epsilon \to 0} \; \mathrm{sign} \left(\overrightarrow{p - q\mathstrut}\cdot \overrightarrow{n_S(q)\mathstrut} \right) \cdot \max_{\substack{q \in S , \norm{q-p'} \le \epsilon \\ p' \in \mathrm{proj}_S(p)}} \abs*{ \overrightarrow{p - q\mathstrut}\cdot \overrightarrow{n_S(q)\mathstrut} }
\,,
\end{equation*}
where \(n_S(q)\) denotes the surface normal at \(q\). 

With the pseudo signed offset distance \(\Psi^*_S(p)\) defined above, we introduce the pseudo signed offset surface at distance \(d\) as
\begin{equation*}
\Gamma^*_d(S) =%
\condset{\, p \in \mathbb{R}^3}{\Psi^*_S(p) = d\,}.
\end{equation*}

We further define the signed offset surface \(\Gamma_d(S)\) as the limit of repeatedly applying infinitesimal pseudo-offsets:

\begin{equation*}
    \Gamma_d(S) = \lim_{n \to \infty} \left( \Gamma_{d/n}^* \circ \Gamma_{d/n}^* \circ \dotsb \circ \Gamma_{d/n}^* \right) (S)
\end{equation*}

For smooth surfaces, the signed offset surface coincides exactly with the signed distance surface. However, when the surface contains sharp features, the two constructions differ precisely within the normal cones~\cite{rockafellar1998variational} of those features. Inside these cones, the signed distance surface exhibits rounded behavior, whereas the signed offset surface preserves the sharp geometry, as shown in Figure~\ref{fig:offset}.

\section{Feature Surface Initialization from Points}
\label{sec:feature_surface_points}

\begin{figure}[!htbp]
\centering
\begin{subcaptionblock}{.47\linewidth}
    \includegraphics[width=\linewidth,page=1]{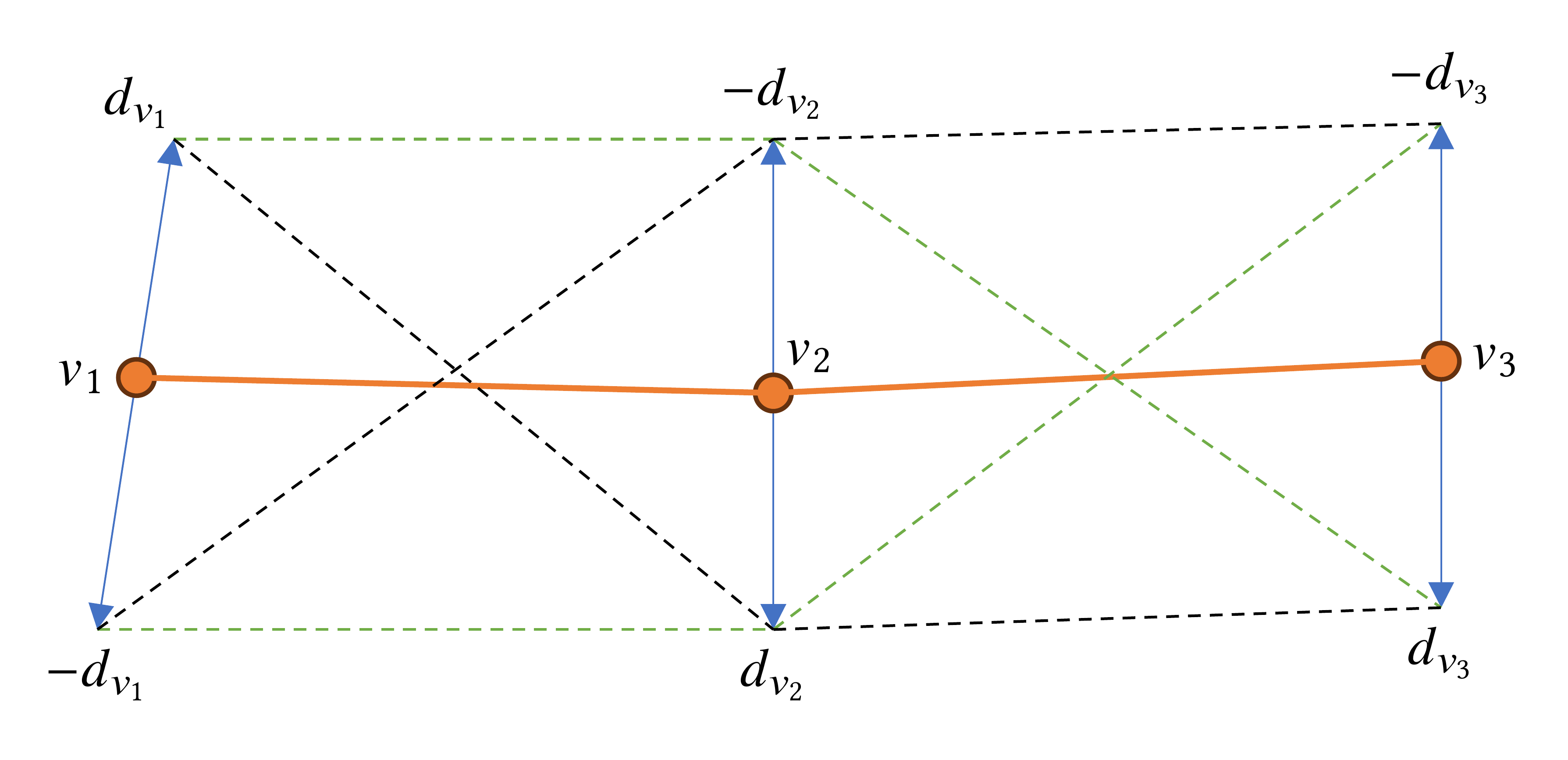}
    \caption{}
    \label{fig:parallel_cross_a}
\end{subcaptionblock}
\begin{subcaptionblock}{.47\linewidth}
    \includegraphics[width=\linewidth,page=2]{img/CAD/parallel_cross.pdf}
    \caption{}
    \label{fig:parallel_cross_b}
\end{subcaptionblock}
\caption{%
    Feature-surface generation from point-based offsets.
    (\subref{fig:parallel_cross_a})~For each $d_{v_i}$, we create offsets in both directions, $d_{v_i}$ and $-d_{v_i}$. We then evaluate the two possible connection patterns: $\set{(d_{v_i}, d_{v_j}), (-d_{v_i}, -d_{v_j})}$ (black dashed lines) and $\set{(-d_{v_i}, d_{v_j}), (d_{v_i}, -d_{v_j})}$ (green dashed lines).
    (\subref{fig:parallel_cross_b})~We choose the configuration whose paired connections are more nearly parallel in order to construct the feature surface. The added diagonals triangulate the resulting rectangle.%
}
\label{fig:parallel_cross}
\Description{A demonstration on the selection of subdivision edges.}
\end{figure}

Given a point cloud as input, since the dihedral angles cannot be computed directly, we adopt NerVE~\cite{Zhu2023NerVE} to initialize the sharp curve set \(\mathcal{S}\) from the point cloud itself. Similarly, the angular bisector plane required for extracting the feature surface \(M\) cannot be evaluated from the point cloud. Instead, we estimate a coarse guiding direction \(d_v\) for each vertex \(v\in \mathcal{S}\) by forming a vector from the centroid of \(v\)'s 32 nearest neighbors in the point cloud to \(v\). Finally, we offset all \(v\in S\) in both directions of \(d_v\) and \(-d_v\) to generate the stripe-like feature surface \(M\). As shown in Figure~\ref{fig:parallel_cross}, because \(d_v\) cannot be given a consistent orientation, some \(d_{v_i}\) may point in the opposite direction. To prevent rectangles from folding, we evaluate both connection patterns (the black and green dashed segment pairs) and select the one that is closer to being parallel.

\end{document}